\documentclass[11pt]{article}

\usepackage[final]{acl}

\usepackage{times}
\usepackage{latexsym}
\usepackage{subcaption}
\usepackage{tabularx}
\usepackage{adjustbox}
\usepackage{graphicx}
\usepackage{amsmath}
\usepackage{titlesec}
\usepackage{array}
\usepackage[T1]{fontenc}

\usepackage[utf8]{inputenc}

\usepackage{microtype}

\usepackage{inconsolata}
\usepackage{longtable}
\usepackage{booktabs}

\usepackage{graphicx}

\title{Exploration of Perceptual Speech Features for Clinical Decision-Support in Mental Health Care}

\author{
  \textbf{Vassilis Lyberatos\textsuperscript{1,2}},
  \textbf{Edmund G. Dervakos\textsuperscript{2}},
  \textbf{Eleni Adamidi\textsuperscript{2}}\\
  \textbf{Athanasios Voulodimos\textsuperscript{1}},
  \textbf{Giorgos Stamou\textsuperscript{1}}
\\[0.5em]
  \textsuperscript{1}National Technical University of Athens, Athens, Greece \\
  \textsuperscript{2}PsychNow \\
  \texttt{vaslyb@ails.ece.ntua.gr} \\
  \texttt{eddie@psychnow.com}
}
\begin{document}
\pagenumbering{gobble}
\pagestyle{empty}
\thispagestyle{empty}
\maketitle
\thispagestyle{empty}
\begin{abstract}

Speech and language technologies offer valuable opportunities for supporting mental health assessment through objective and interpretable cues. We present a systematic feature-based analysis framework leveraging perceptually grounded acoustic and linguistic characteristics, including prosody, vocal quality, semantic coherence, syntactic structure, and sarcasm. Using statistical analysis and interpretable machine learning (XGBoost with SHAP and LIME), we examine associations between speech features and validated symptom measures of depression, anxiety, and ADHD.  Evaluated on both controlled benchmark datasets (StressID, DAIC-WOZ, Androids, EATD) and a real-world clinical dataset, the framework reveals stable and consistent relationships between symptom severity and vocal irregularities (e.g., shimmer, jitter), lexical-syntactic patterns, and affective tone. An ablation study conducted across all datasets further identifies the most informative feature groups. This work explores a transparent and clinically interpretable approach to speech-based mental health analysis.

\end{abstract}

\section{Introduction}
\label{sec:intro}

Mental health disorders represent a major global health burden, affecting nearly 970 million people worldwide in 2019 and accounting for approximately $16\%$ of global years lived with disability (YLDs), which makes them one of the leading causes of disability worldwide \cite{vos2020global}. Traditional screening and intake approaches rely heavily on subjective evaluations, including clinical interviews and self-reported symptoms, which are time-consuming and vulnerable to clinician bias, recall bias, and stigma-driven underreporting. In this context, there is a pressing need not for diagnostic tools in the narrow sense, but for clinically supportive technologies that provide objective and interpretable cues without enforcing rigid categorical decisions \cite{berisha2024responsible}. Particularly in the early stages of assessment, tools that function as perceptual and behavioral aids can help clinicians detect patterns, guide conversation, and observe phenotypes while avoiding premature labeling and stigma reinforcement \cite{kotov2017hierarchical}.

Speech and language provide a rich, non-invasive source of information for understanding mental health. As natural modes of human expression, they encode cognitive, emotional, and neurological states through acoustic properties (e.g., pitch, prosody, intensity) and linguistic structure (e.g., lexical richness, syntactic complexity) \cite{cummins2015review,low2020speech}. A growing body of research shows that psychiatric disorders are reflected in distinctive speech and language patterns, including alterations in articulation, semantic coherence, prosody, and syntax, particularly in conditions such as depression, anxiety, schizophrenia, and cognitive impairment \cite{alhanai2018detecting,arevian2020clinical}.

Recent advances in speech and language technologies have enabled automated analysis for a range of mental health applications. Speech collected in naturalistic settings, including mobile environments, has been used to identify depression, anxiety, insomnia, and fatigue with encouraging accuracy and robustness across populations \cite{berisha2024automated}. However, many state-of-the-art models function as “black boxes,” which limits interpretability and undermines clinician trust. There is growing demand, motivated by both ethical and regulatory considerations, for explainable and interpretable AI systems in healthcare, particularly in high-stakes domains such as mental health assessment \cite{ng2024tutorial,holzinger2019causability,doshi2017towards}.

Perceptually motivated acoustic and linguistic features, such as prosodic, spectral, and lexical-syntactic markers, are inherently interpretable and grounded in well-established clinical phenomena \cite{tasnim2023depac,voleti2019review,jiao2017interpretable,tu2017interpretable,premananth2025multimodal,neumann2025multimodal}. Models that rely on these features are typically more transparent and accessible to clinicians. Such features can be obtained through standard signal processing, extracted using auxiliary deep neural networks \cite{jiao2017interpretable,tu2017interpretable}, or learned directly in end-to-end architectures \cite{leschly2025exploration,korzekwa2019interpretable,xu2023dysarthria}. When combined with interpretable classifiers or post-hoc explanation methods such as attribution analyses, perceptual feature pipelines offer clear insight into which speech characteristics drive model predictions. This approach improves accountability and supports clinical decision-making by providing human-understandable evidence that aligns with expert knowledge. Consequently, perceptual feature-based explainable AI frameworks are a promising strategy for bridging complex machine learning models and practical mental health assessment \cite{ntalampiras2025interpretable,menne2024voice}.

In this work, we explore the development of transparent and clinically meaningful speech and language analysis methods for mental health assessment. We present a systematic, interpretable framework that combines perceptually grounded acoustic and linguistic features with feature-based modeling and explainable AI techniques \cite{guidotti2018survey}. To evaluate the robustness and generalizability of the proposed approach, we conduct experiments on five speech datasets spanning controlled laboratory stress elicitation, semi-structured clinical interviews, multilingual public depression corpora, and real-world digital mental health assessments. Across datasets, the framework is applied consistently to binary classification tasks involving stress, depression, anxiety, and attention-related difficulties. In addition, we perform statistical analyses and ablation studies on feature aggregation and representation strategies to assess the stability and interpretability of the extracted speech markers. By examining speech characteristics across diverse recording conditions, languages, and clinical contexts, this work explores a clinically grounded approach that links perceptual feature design with explainable modeling, contributing to more transparent and interpretable speech-based technologies for mental health.

\section{Methodology}

Our guiding principle in designing the experimental methodology was to prioritize the extraction of clinically interpretable features by leveraging reliable tools. In line with prior work on clinical applications of voice analysis~\cite{jiao2017interpretable,berisha2024automated,article}, we adopted feature extraction strategies that draw upon multiple disciplines, including natural language processing, signal processing, and auxiliary deep neural models. As a next step, we applied traditional statistical and machine learning analyses, further enhanced with post-hoc explainable AI (XAI) techniques~\cite{guidotti2018survey}, in order to improve the interpretability of our results and analysis.

\subsection{Feature Extraction}\label{subsec:feature}

Clinical psychopathology can be detected from speech based not only on what is said, but also on how it is said~\cite{aloshban2022you}. In our approach, we incorporate both modalities, audio and language, into the analysis. Accordingly, our feature extraction methods can be divided into two categories: acoustic features and linguistic features. A list of all extracted features is provided in Table~\ref{tab:features_mathsafe} in Appendix~\ref{sec:feature}.

\subsubsection{Acoustic Features}

For the audio analysis, we used Parselmouth~\cite{parselmouth}, which enables Praat ’s~\cite {praat} validated acoustic analyses directly within python. Acoustic features were extracted from voiced segments and included pitch statistics, intensity, jitter, shimmer, harmonic-to-noise ratio (HNR), zero-crossing rate (ZCR), pauses, phonation and articulation rates, rhythm (pairwise variability index), and speech entropy. Prior to analysis, waveforms were converted to mono, resampled to 16 kHz, and normalized in amplitude to ensure comparability across recordings. Beyond low-level acoustics, we extracted higher-level paralinguistic and linguistic representations using pretrained neural models. Emotion-related features were obtained with a HuBERT-based speech emotion recognition model~\footnote{\url{https://huggingface.co/superb/hubert-base-superb-er}}~\cite{yang2021superb}, fine-tuned on the IEMOCAP corpus~\cite{busso2008iemocap}. The extracted acoustic descriptors are organized into three interpretable groups: prosodic/fluency features (e.g., pitch and intensity statistics, pause measures, phonation and articulation rates, and rhythm/variability indices), voice quality features (e.g., jitter, shimmer, and harmonic-to-noise ratio), and psycholinguistic (auxiliary emotion and sarcasm estimates).

These representations have also been studied in prior clinical work. Prosodic reductions such as flatter pitch range/variability and increased pausing are commonly associated with monotone speech, depression, and blunted affect, while elevated variability may reflect agitation or manic states~\cite{alpert2001reflections,low2020automated}. Voice quality perturbations have been linked to psychopathology in prior work, with shimmer in particular reported as a correlate of depression severity in some settings~\cite{ettore2023digital,honig2014automatic}. Finally, sarcasm and related pragmatic indicators have been associated with increased risk of anxiety, stress, and depression~\cite{dionigi2023understanding,gross2014emotion,pope1970anxiety}.

\subsubsection{Linguistic Features}

Linguistic features were extracted from transcripts using spaCy~\cite{honnibal2020spacy} and Stanza~\cite{qi2020stanza}, providing tokenization, POS tagging, lemmatization, dependency parsing, and constituency trees. From these annotations, we derived lexical indices (type-token ratio, MATTR, Brunet’s index, Honore’s statistic, lemma diversity, and morphological richness), syntactic measures (mean sentence length, clause ratio, syntactic and constituency depth, and passive voice ratio), and graph-based discourse metrics (connectivity, loops, density, diameter, and path statistics). Psycholinguistic information was obtained using VADER sentiment analysis~\cite{hutto2014vader}, capturing positive, neutral, and negative valence. This set of descriptors captures lexical diversity, morphosyntactic complexity, discourse organization, and affective tone. To further capture semantic information, pretrained neural models were integrated, with sentence-level embeddings obtained using Sentence-BERT~\footnote{\url{https://huggingface.co/sentence-transformers/paraphrase-MiniLM-L6-v2}}~\cite{reimers2019sentence}, enabling estimation of discourse coherence, cohesion, and repetition patterns. The resulting linguistic descriptors are organized into four interpretable groups: lexical features (e.g., word/sentence counts, lexical diversity indices such as TTR/MATTR/Brunet/Honor\'e, content--function and pronoun ratios, and morphology/POS diversity), syntactic features (e.g., sentence/clause lengths, dependency and constituency depth, passive voice usage, and graph-based measures of discourse structure and repetition), semantic features (e.g., first-/second-order coherence and repetition/cohesion measures), and psycholinguistic features capturing affective tone and pragmatics, including sentiment and sarcasm. 

These groupings are motivated by prior clinical findings: reduced lexical richness and shifts in tense/pronoun usage have been linked to schizophrenia, dementia, and depression, and pronoun patterns may reflect self-focus or social withdrawal~\cite{compton2023lexical,pennebaker2003psychological}. Reduced syntactic complexity is commonly associated with cognitive impairment and depression, while increased repetition and disfluency-like patterns have been reported in ADHD~\cite{sung2020syntactic,engelhardt2011language}. At the semantic level, reduced coherence has been associated with disorganized thought and psychosis-related phenomena and has also been observed in ADHD and manic states~\cite{corcoran2018prediction,engelhardt2011language}. Finally, affective language indicators (e.g., elevated negative sentiment and reduced positive emotion) are associated with mood disorders, while pragmatic cues such as sarcasm can correlate with elevated risk and may track agitation/psychosis-related expression in some settings~\cite{sonnenschein2018linguistic,dionigi2023understanding}.

\subsubsection{Sarcasm}

Sarcasm is a subtle communicative cue that conveys implicit emotional and attitudinal meaning beyond literal word content. Since sarcasm has been correlated with depressive symptoms~\cite{dionigi2023understanding}, it provides a relevant paralinguistic marker for our study. To capture this dimension, we trained a multimodal sarcasm detection model on the MUStARD dataset~\cite{mustard}, achieving an accuracy of approximately 70\%. The model integrates linguistic and acoustic information by combining BERT\footnote{\url{https://huggingface.co/google-bert/bert-base-uncased}}~\cite{bertbase} for text and Wav2Vec2\footnote{\url{https://huggingface.co/facebook/wav2vec2-base-960h}}~\cite{baevski2020wav2vec} for audio, with frozen encoders whose pooled embeddings are projected into a shared space, concatenated, and passed through a feedforward classifier. Once trained, this auxiliary model was applied to our datasets to infer sarcasm probabilities, which were then included as additional features alongside acoustic, linguistic, and emotional representations to enrich our analysis. This procedure yielded a total of 82 scalar, interpretable features.

\subsection{Analytical Framework}\label{subsec:ana}

The analytical framework was designed to explore statistical and structural relationships between psychopathology and speech features in an exploratory manner. We combined inferential statistics with interpretable machine learning and explainability techniques, an integration that, to our knowledge, remains relatively uncommon in this area, to examine associations across datasets.

Group-level comparisons were first conducted using independent samples t-tests to assess significant differences in perceptual and paralinguistic features between participant subgroups defined by clinical thresholds on PHQ-9, GAD-7, and ASRS scores. To control for multiple comparisons, resulting p-values were adjusted using the false discovery rate (FDR) correction according to the Benjamini-Hochberg procedure. These tests provided an initial understanding of feature distributions and effect directions underlying symptom-related variation.

To capture higher-order and nonlinear dependencies, we employed XGBoost~\cite{chen2016xgboost} classifiers as analytical models linking voice features to mental health categories. XGBoost was selected due to its strong performance on tabular data and its compatibility with transparent, feature-level interpretability. The internal structure of these models was examined through feature importance rankings and post-hoc explainability analyses using SHAP~\cite{lundberg2017unified} and LIME~\cite{ribeiro2016should}. LIME explanations were aggregated across all instances to derive cumulative patterns of feature influence, enabling a global interpretation of localized explanations. Complementarily, SHAP summary plots were employed to visualize the overall distribution, magnitude, and direction of feature effects. Together, these analyses indicated correlations and interaction patterns between prosodic, spectral, and temporal features and psychopathological dimensions.

In addition, we conducted an ablation study to identify the most informative groups of features, using the feature groupings defined in the Section~\ref{subsec:feature}. By systematically training models with individual feature groups isolated, we assessed the relative contribution of acoustic, prosodic, spectral, and linguistic feature sets to the observed associations with psychopathology. This analysis provided further insight into which categories of features drive the explanatory power of the framework.

Overall, this framework follows an explanatory modeling approach in which both statistical tests and interpretable machine learning are used to identify and validate relationships between acoustic patterns and mental health indicators, providing analytical insight rather than purely predictive outcomes.

\section{Experiments}
We conducted experiments on five speech datasets spanning controlled laboratory conditions, semi-structured clinical interviews, multilingual public corpora, and real-world digital mental health assessments. Across all datasets, we applied the proposed feature extraction pipeline to extract acoustic and linguistic descriptors from speech and examined their relationship to clinically meaningful labels and screening instruments. Depending on the dataset, the experimental task was formulated as binary classification, including stress recognition and the detection of depression, anxiety, and attention-related difficulties. Statistical analysis and explainable machine learning methods were used to support robust evaluation and interpretable feature attribution.

\subsection{Datasets}

As a primary benchmark, we used the \textsc{StressID} dataset~\cite{chaptoukaev2023stressid}, a multimodal corpus designed for stress identification. \textsc{StressID} includes synchronized audio, facial video, and physiological recordings (ECG, EDA, respiration) from 65 participants performing 11 stress-inducing and neutral tasks. The dataset contains over 39 hours of annotated data, with self-reported measures of stress, relaxation, arousal, and valence. In this study, we used only the audio modality and formulated a binary classification task distinguishing stressed from non-stressed speech.

To assess depression detection in structured clinical interviews, we used the \textsc{DAIC-WOZ} corpus~\cite{gratch2014distress}, a subset of the Distress Analysis Interview Corpus. It consists of semi-structured interviews conducted by a virtual agent using a Wizard-of-Oz framework, and includes synchronized audio, video, transcripts, and validated PHQ-8 depression scores. We extracted participant speech only and performed binary depression classification based on standard PHQ-8 thresholds.

We further evaluated our approach on the \textsc{Androids} corpus~\cite{tao2023androids}, a clinically validated Italian-language dataset comprising speech recordings from 118 participants, including individuals diagnosed with depression and healthy controls. The dataset contains both read and spontaneous speech collected in real clinical environments, with expert diagnostic labels. In our experiments, we used the audio recordings to perform binary depression classification.

To examine cross-linguistic generalization, we employed the \textsc{EATD} corpus~\cite{shen2022automatic}, the first publicly available Chinese-language dataset for depression detection. \textsc{EATD} includes audio recordings and transcripts from 162 participants who completed standardized self-report assessments using the Self-Rating Depression Scale (SDS). Participants responded to emotionally eliciting prompts, and we used the speech recordings to define a binary depression classification task based on SDS score thresholds.

Finally, we used a proprietary real-world dataset collected through a digital mental health platform, comprising approximately 200 participants who completed a 30-minute self-report questionnaire and provided spoken responses, as part of psychiatric intake. The dataset includes audio recordings and standardized clinical screening scores for depression (PHQ-9), anxiety (GAD-7), and attention-deficit/hyperactivity disorder (ASRS). We analyzed speech features in relation to these clinical scores to evaluate the generalizability of our approach in real-world conditions. The dataset is balanced across diagnostic categories and participant sex. All participants provided informed consent, and all data were de-identified prior to analysis.

\subsection{Results}\label{sec:res}

Results are reported across five datasets (\textsc{StressID}, \textsc{DAIC-WOZ}, \textsc{Androids}, \textsc{EATD}, and \textsc{Real}), spanning multiple clinical conditions, languages, and recording contexts. All experiments were formulated as binary classification tasks using dataset-specific labels or clinically validated questionnaire cutoffs. 

The analyses focus on predictive performance using interpretable models, feature-group ablation to assess the contribution of predefined acoustic and linguistic representations, group-level statistical differences, and feature-level interpretability using SHAP, LIME, and partial dependence analysis. Additional results are provided in the Appendix~\ref{sec:additional}.

\subsubsection{Feature Group Ablation}

\begin{figure}[t]
    \centering
    \includegraphics[width=0.75\linewidth]{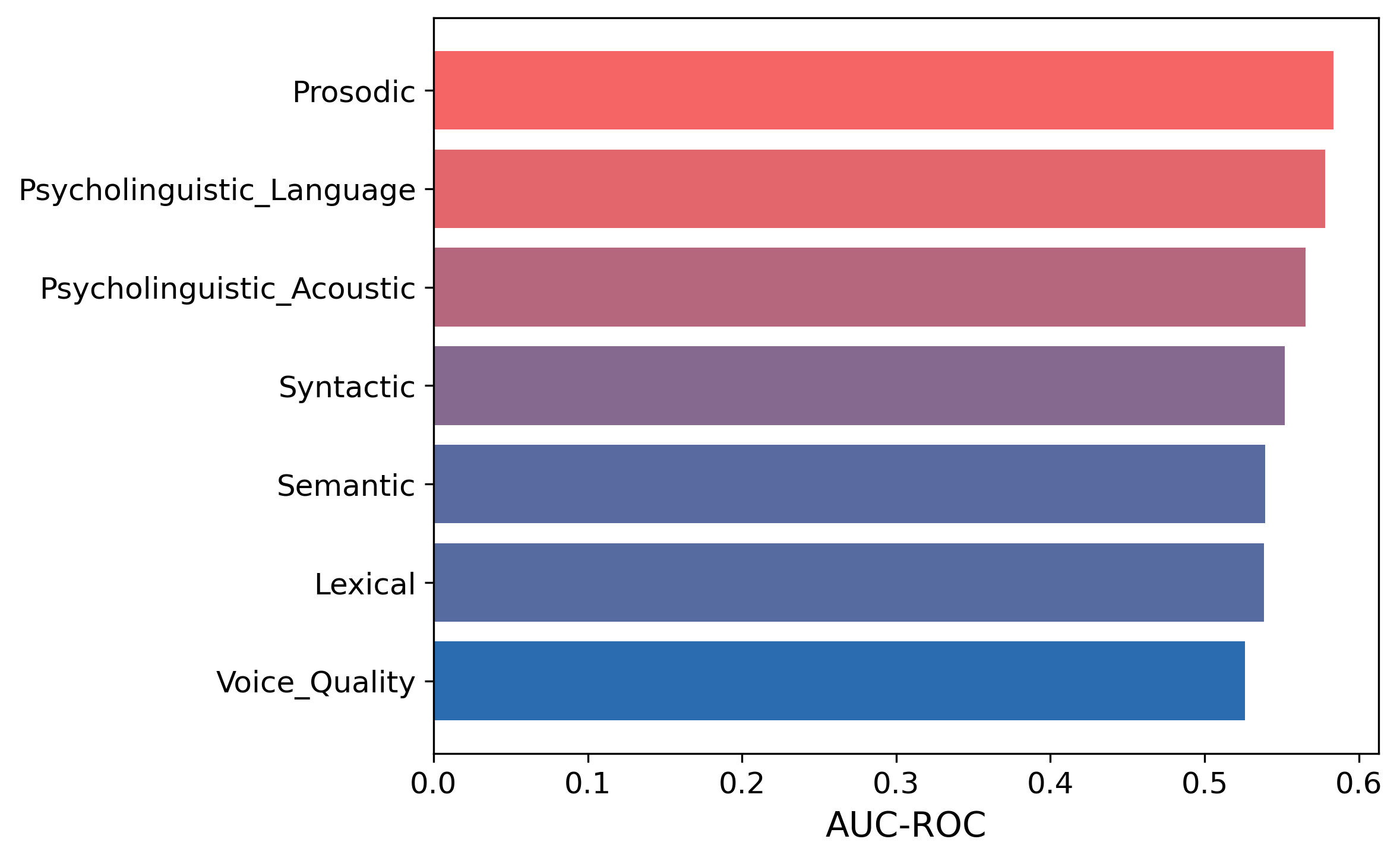}
    \caption{Average AUC-ROC across datasets for XGBoost models trained with only one feature group at a time.}
    \label{fig:avg_group_performance_auc}
\end{figure}

An ablation study evaluated every predefined feature groups within each dataset. For each combination, a summary AUC-ROC was computed from cross-validation. Figure~\ref{fig:avg_group_performance_auc} reports the mean single-group AUC-ROC across datasets for each feature group in isolation. Prosodic features show the highest standalone average performance, followed by psycholinguistic language and acoustic groups. Syntactic, semantic, and lexical groups are intermediate, while voice-quality features are weakest alone. This cross-dataset summary suggests that no single group is sufficient, motivating the use of complementary feature combinations.

\subsubsection{StressID}

In the \textsc{StressID} dataset, we used the labels corresponding to the stress and non-stress conditions, with the dataset being balanced. In the original study~\cite{chaptoukaev2023stressid}, which employed a large pretrained Wav2Vec model followed by a logistic regression classifier, an average accuracy of 0.66 (variance = 0.03) and an F1-score of 0.70 (variance = 0.02) were reported. In our approach, using perceptual features combined with an XGBoost classifier, we achieved a higher average accuracy of 0.70 (variance = 0.01) and an F1-score of 0.81 (variance = 0.01). These results are based on 10 random runs, consistent with the evaluation procedure described in the original paper.

\begin{figure*}[t!]
    \centering
    \begin{subfigure}[b]{0.3\linewidth}
        \includegraphics[width=\linewidth]{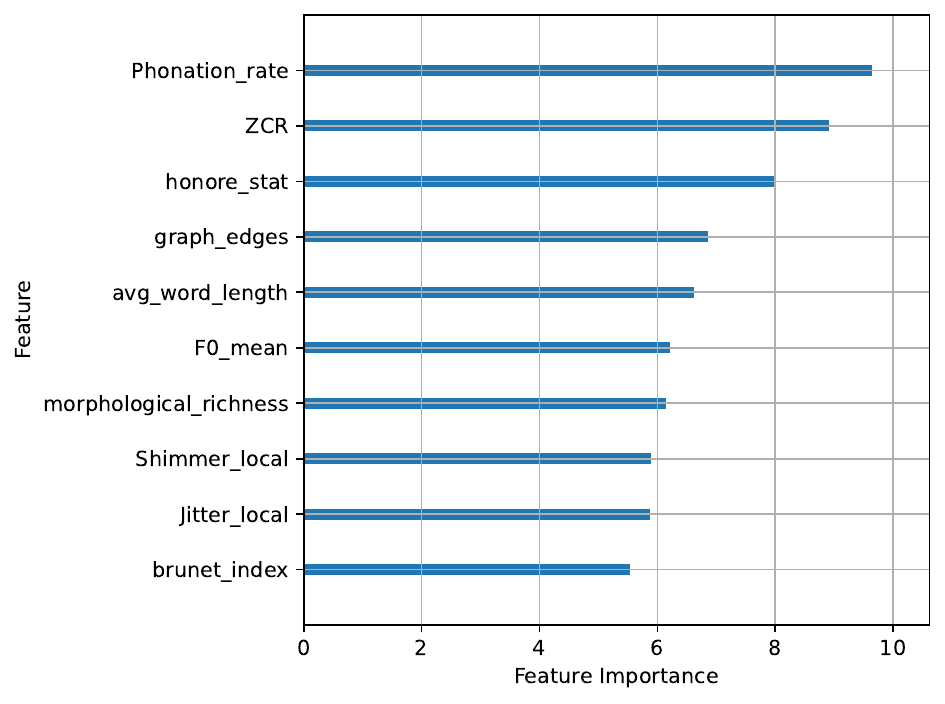}
        \caption{Stress -- XGBoost}
        \label{fig:stressid_xgb}
    \end{subfigure}
    \hfill
    \begin{subfigure}[b]{0.3\linewidth}
        \includegraphics[width=\linewidth]{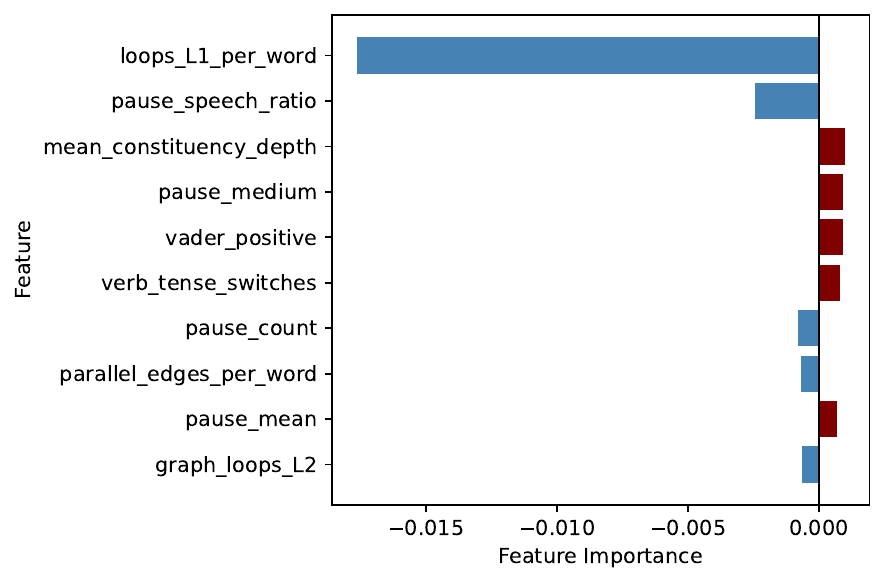}
        \caption{Stress -- LIME}
        \label{fig:stressid_lime}
    \end{subfigure}
    \hfill
    \begin{subfigure}[b]{0.3\linewidth}
        \includegraphics[width=\linewidth]{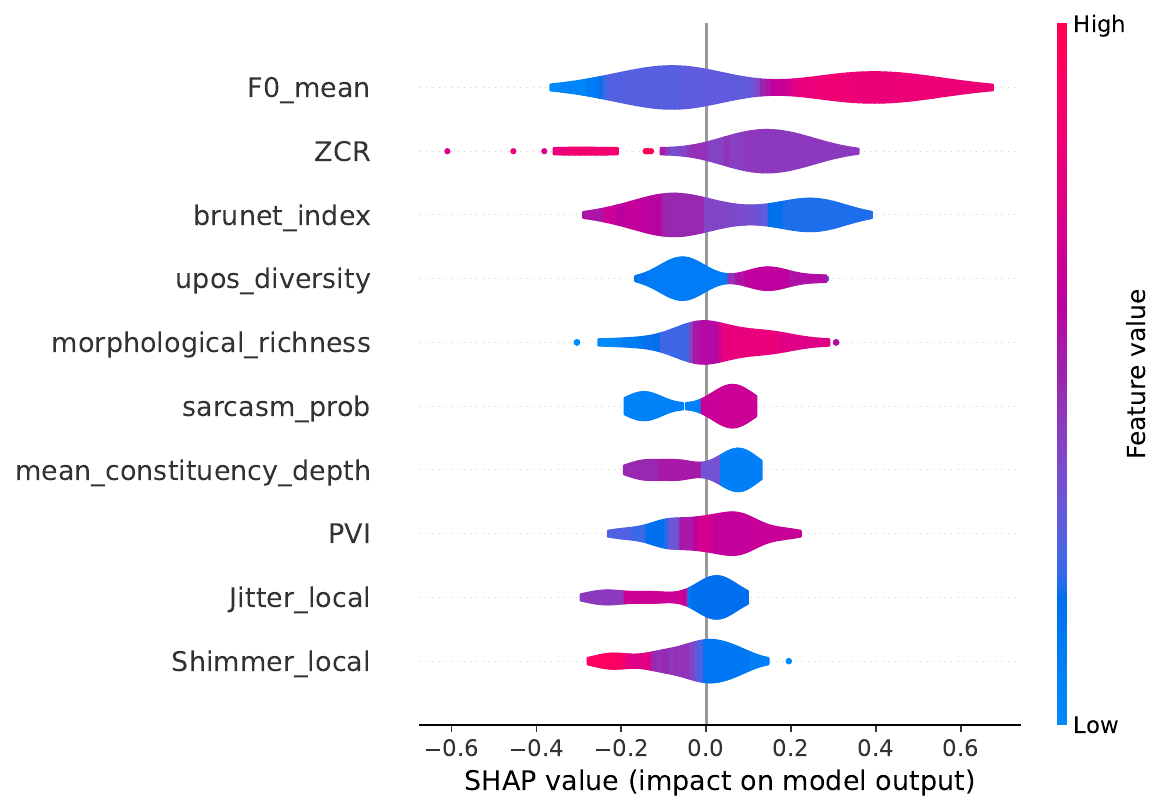}
        \caption{Stress -- SHAP}
        \label{fig:stressid_shap}
    \end{subfigure}

    \caption{
        Top predictive features for the \textsc{StressID} dataset derived from acoustic and linguistic descriptors.
    }
    \label{fig:stressid_feature_comparison}
\end{figure*}

Table~\ref{tab:stress_features} presents the features that differ significantly between the two groups after FDR correction, highlighting associations with emotional expression, speech quality, and the duration of voiced segments. To analyze the XGBoost model results described in Section~\ref{subsec:ana}, we visualized feature importance based on gain, plotted SHAP values, and examined the aggregated local explanations obtained from LIME. Figure~\ref{fig:stressid_feature_comparison} illustrates the top 10 most important features. Features related to syntax, lexical properties, and voice quality consistently emerged as the most influential across interpretability methods, with voice quality features showing the strongest overall contribution.

\begin{table}[t]
\centering
\footnotesize
\setlength{\tabcolsep}{2.5pt} 
\renewcommand{\arraystretch}{1.1}
\begin{tabular}{lccc}
\toprule
\textbf{Feature} & \textbf{Non-Stressed} & \textbf{Stressed} & \textbf{$p$-value} \\
\midrule
Shimmer\_local       &  0.343 & -0.142 & 1.27$\times$10$^{-5}$ \\
Jitter\_local        &  0.368 & -0.152 & 4.14$\times$10$^{-4}$ \\
emotion\_hap         & -0.238 &  0.098 & 5.42$\times$10$^{-3}$ \\
emotion\_sad         &  0.220 & -0.091 & 1.08$\times$10$^{-2}$ \\
loops\_L1\_per\_word & -0.156 &  0.065 & 1.64$\times$10$^{-2}$ \\
vader\_positive      &  0.211 & -0.087 & 2.54$\times$10$^{-2}$ \\
vader\_compound      &  0.178 & -0.073 & 2.95$\times$10$^{-2}$ \\
Number=Plur          &  0.186 & -0.077 & 3.80$\times$10$^{-2}$ \\
pause\_count         & -0.147 &  0.061 & 4.10$\times$10$^{-2}$ \\
pause\_short         & -0.147 &  0.061 & 4.12$\times$10$^{-2}$ \\
F0\_mean             & -0.151 &  0.063 & 4.29$\times$10$^{-2}$ \\
\bottomrule
\end{tabular}
\caption{
Mean feature values for Non-Stressed and Stressed groups (\textsc{StressID}), 
with corresponding $p$-values from two-sample $t$-tests ($p<0.05$).
}
\label{tab:stress_features}
\end{table}

\subsubsection{\textsc{DAIC-WOZ}}

Binary depression classification was performed using participant speech, with subject-level acoustic and linguistic features aggregated using robust statistics and modeled with XGBoost. Classification performance was moderate (accuracy = 0.66, F1-score = 0.56, AUC-ROC = 0.63), compared with an LSTM model with an F1-score of 0.64~\cite{arioz2022scoping}. Statistical analysis indicated that depressed participants exhibited higher pause frequency and larger pause-to-speech ratios, reflecting reduced fluency. Linguistic differences were limited and did not remain significant after false discovery rate correction. Feature importance analyses using XGBoost, SHAP, and LIME were consistent, identifying pausing behavior, reduced pitch variability, lower vocal intensity, and increased voice quality instability as the most influential predictors. 

\subsubsection{\textsc{Androids}}

Depression classification was conducted using participant-level acoustic and linguistic features aggregated across speech tasks and modeled with XGBoost. The model achieved strong performance (accuracy = 75.6\%, F1-score = 77.1\%, AUC-ROC = 87.6\%), compared with the LSTM model reported in~\cite{tao2023androids}, which achieved an F1-score of 0.83. Statistical analysis revealed significant group differences in emotional expression, vocal intensity, pause behavior, and discourse structure. After false discovery rate correction, sadness-related emotion, negative sentiment, reduced intensity, and lower semantic coherence remained significant. Interpretability analyses consistently highlighted emotional polarity, voice quality measures (jitter and shimmer), vocal energy, and discourse coherence as dominant predictors. 

\subsubsection{\textsc{EATD}}

Depression classification was performed using multimodal acoustic and linguistic features and an XGBoost classifier. Model performance was variable (accuracy = 82.1\%, F1-score = 53.9\%, AUC-ROC = 73.4\%), compared with a GRU model reported in~\cite{shen2022automatic}, which achieved an F1-score of 0.71. Statistical group-level differences did not remain significant after false discovery rate correction. Despite this, feature importance analyses consistently identified reduced prosodic variability, lower articulation rate, and sadness-related emotional cues as relevant predictors. 

\subsubsection{Real dataset}

For the \textsc{Real} dataset, we converted the PHQ-9, GAD-7, and ASRS scores into binary classification tasks using clinically established thresholds: a PHQ-9 score of 15 or higher was considered indicative of at least moderate depressive symptoms, a GAD-7 score of 10 or higher indicated at least moderate anxiety symptoms, and an ASRS score of 13 or higher was considered indicative of clinically relevant ADHD symptoms. This binarization aimed to reduce variability in interpretation of self-reported questionnaire scores and to enhance consistency across participants~\cite{van2016validation}.  Using an XGBoost classifier, we achieved AUC-ROC of 0.67 (variance = 0.05) for ASRS, 0.63 (variance = 0.03) for PHQ-9, and 0.59 (variance = 0.02) for GAD-7 under 4-fold, subject-independent cross-validation. All features were first aggregated to the subject level by taking the median across each participant’s available audio files, and folds were created on subjects (speaker-disjoint), eliminating any train/test leakage.

\begin{figure*}[t!]
    \centering
    
    \begin{subfigure}[b]{0.3\linewidth}
        \includegraphics[width=\linewidth]{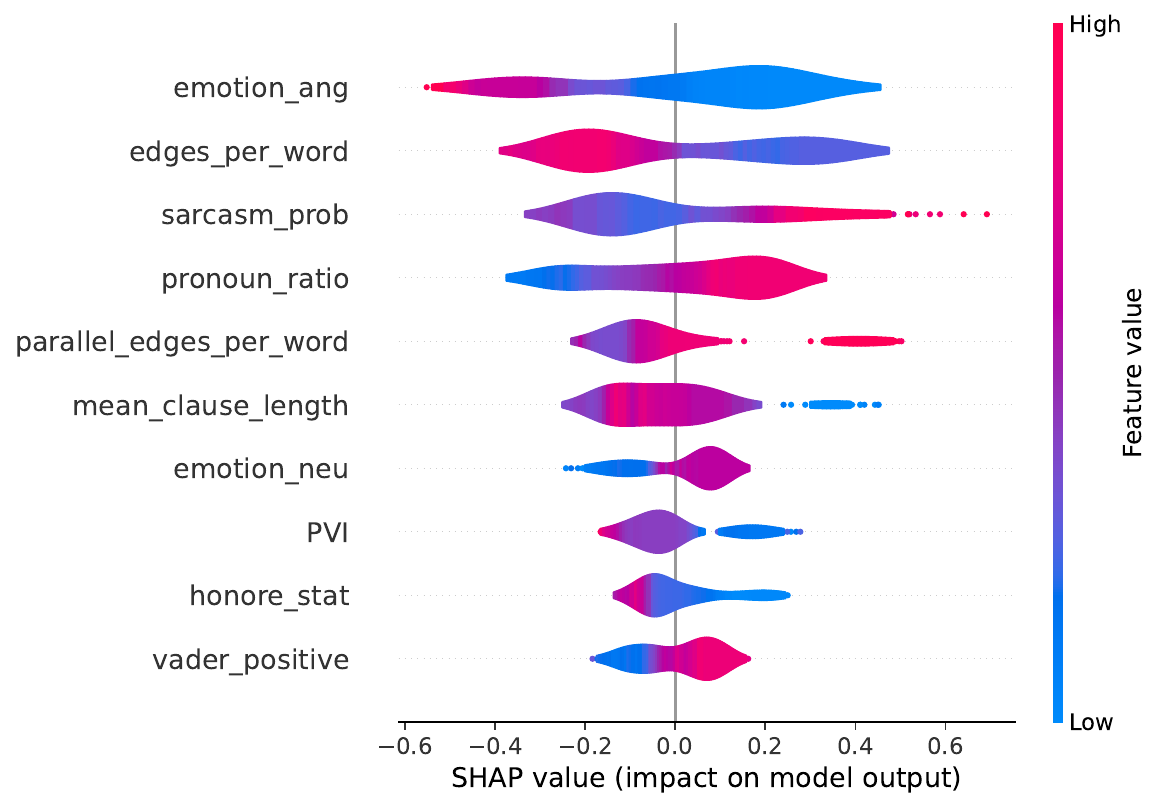}
        \caption{ASRS}
        \label{fig:asrs_shap}
    \end{subfigure}
    \begin{subfigure}[b]{0.3\linewidth}
        \includegraphics[width=\linewidth]{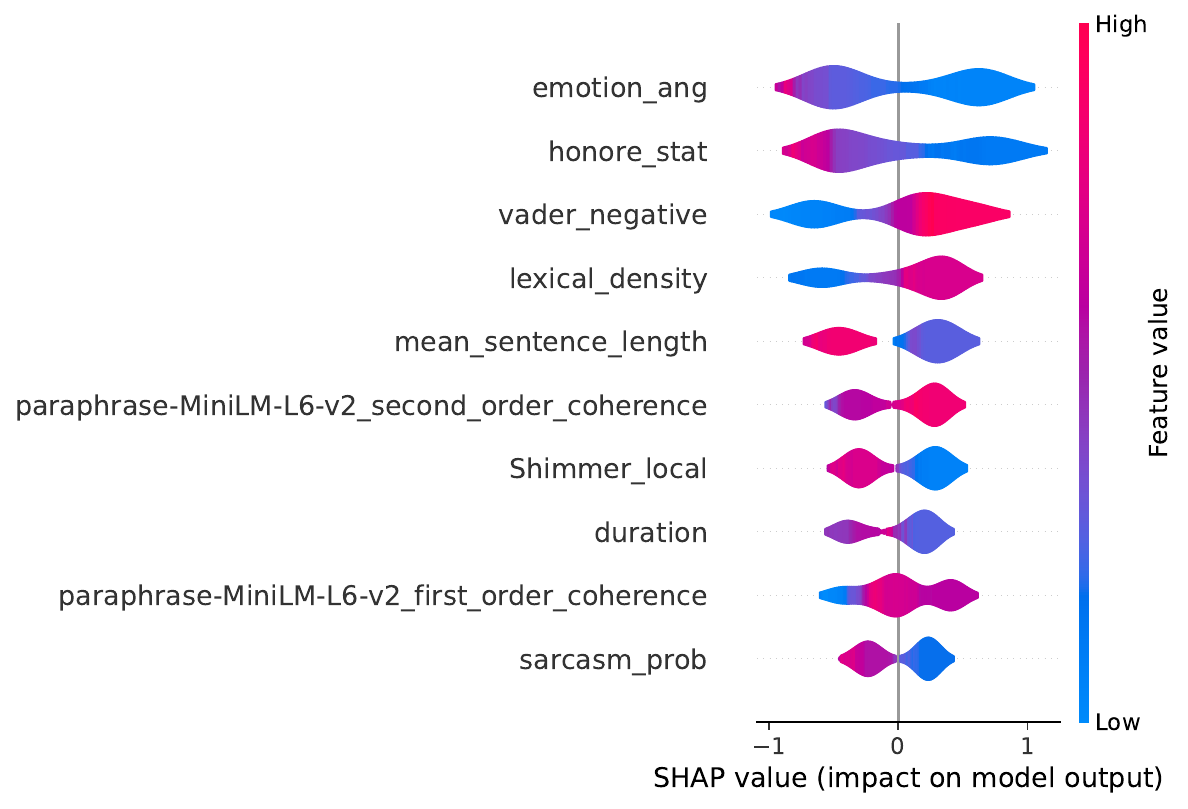}
        \caption{GAD-7}
        \label{fig:gad7_shap}
    \end{subfigure}
    \begin{subfigure}[b]{0.3\linewidth}
        \includegraphics[width=\linewidth]{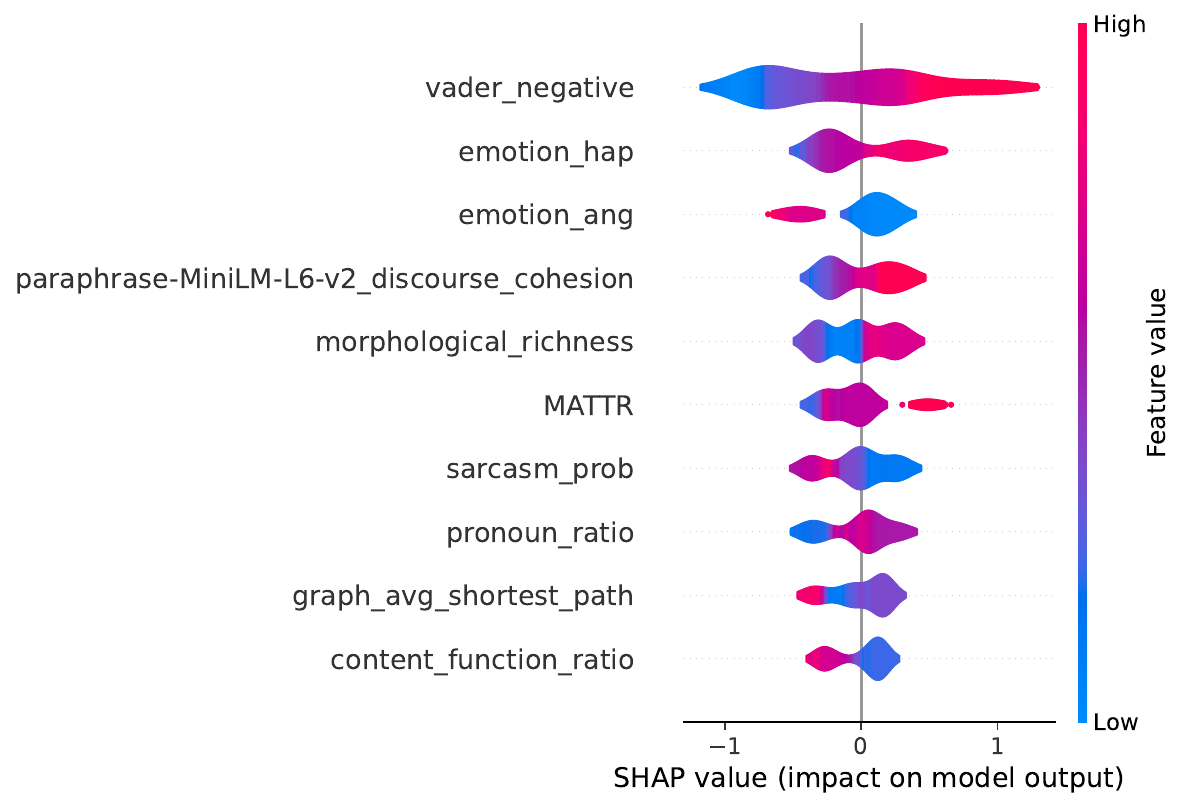}
        \caption{PHQ-9}
        \label{fig:phq9_shap}
    \end{subfigure}
    \caption{SHAP explanations for top predictive features from the \textsc{Real} dataset for ASRS, GAD-7, and PHQ-9.}    \label{fig:feature_comparison}
\end{figure*}

Table~\ref{tab:phq9_medians_pvalues} presents the features distinguishing the depressed and non-depressed groups. After applying FDR correction, the \textit{VADER\_negative} feature remained statistically significant, indicating higher levels of negative sentiment and altered emotional expression in the depressed group. To further validate these findings, we compared the results with those obtained from other methodologies, as shown in Figure~\ref{fig:feature_comparison}. This analysis confirmed that emotional, lexical, and syntactic features play a key role in differentiating between the two groups.

\begin{table}[!b]
\centering
\footnotesize
\setlength{\tabcolsep}{2.5pt} 
\renewcommand{\arraystretch}{1.1} 
\begin{tabular}{lccc}
\toprule
\textbf{Feature} & \textbf{Non-Dep} & \textbf{Dep} & \textbf{$p$-value} \\
\midrule
vader\_negative          & 0.030 & 0.050 & 1.22$\times$10$^{-4}$ \\
vader\_compound          & 0.652 & 0.497 & 4.98$\times$10$^{-3}$ \\
content\_function\_ratio & 1.333 & 1.273 & 9.93$\times$10$^{-3}$ \\
loops\_L1\_per\_word     & 0.003 & 0.001 & 1.26$\times$10$^{-2}$ \\
pause\_medium            & 0.295 & 0.102 & 1.91$\times$10$^{-2}$ \\
graph\_loops\_L1         & 0.302 & 0.148 & 1.77$\times$10$^{-2}$ \\
MATTR                    & 0.693 & 0.704 & 4.78$\times$10$^{-2}$ \\
emotion\_neu             & 0.198 & 0.169 & 4.26$\times$10$^{-2}$ \\
\bottomrule
\end{tabular}
\caption{
Mean feature values for Non-Depression and Depression groups (PHQ-9) from the \textsc{Real} dataset, 
with corresponding $p$-values from two-sample $t$-tests ($p<0.05$).
}
\label{tab:phq9_medians_pvalues}

\end{table}

Table~\ref{tab:asrs_medians_pvalues} presents the features that show statistically significant differences for the ASRS-based ADHD recognition; note that only the last four features did not remain significant after FDR correction. The most influential features appear to be those capturing repetition patterns, and the fluency of the speaker. As illustrated in Figure~\ref{fig:feature_comparison}, features related to graph-based representations, verb tense variation, emotional expression, and sarcasm detection also played a key role in distinguishing individuals with high ASRS scores, indicative of ADHD. Overall, across all analyses, the frequency of verb tense shifts and graph features capturing repetition consistently emerged as the most important indicators.

\begin{table}[!t]
\centering
\footnotesize
\setlength{\tabcolsep}{2pt} 
\renewcommand{\arraystretch}{1.05} 
\resizebox{\columnwidth}{!}{%
\begin{tabular}{lccc}
\toprule
\textbf{Feature} & \textbf{Non-ADHD} & \textbf{ADHD} & \textbf{$p$-value} \\
\midrule
Tense=Pres & 6.22 & 7.81 & 2.28$\times$10$^{-4}$ \\
graph\_repeated\_edges & 3.12 & 4.48 & 1.75$\times$10$^{-4}$ \\
graph\_diameter & 8.71 & 8.23 & 3.85$\times$10$^{-4}$ \\
edges\_per\_word & 0.93 & 0.92 & 7.15$\times$10$^{-4}$ \\
parallel\_edges\_per\_word & 0.04 & 0.05 & 1.09$\times$10$^{-3}$ \\
verb\_tense\_switches & 8.54 & 10.44 & 2.16$\times$10$^{-3}$ \\
graph\_loops\_L3 & 1.52 & 1.99 & 4.15$\times$10$^{-3}$ \\
max\_constituency\_depth & 13.88 & 14.83 & 5.49$\times$10$^{-3}$ \\
pronoun\_ratio & 0.15 & 0.16 & 5.43$\times$10$^{-3}$ \\
discourse\_cohesion & 0.09 & 0.10 & 7.75$\times$10$^{-3}$ \\
filler\_count & 116.45 & 136.47 & 7.74$\times$10$^{-3}$ \\
emotion\_ang & 0.06 & 0.04 & 8.68$\times$10$^{-3}$ \\
brunet\_index & 9.94 & 10.32 & 1.60$\times$10$^{-2}$ \\
clause\_ratio & 0.73 & 0.85 & 2.31$\times$10$^{-2}$ \\
honore\_stat & 1572.53 & 1522.98 & 4.42$\times$10$^{-2}$ \\
graph\_loops\_L2 & 0.75 & 0.95 & 4.52$\times$10$^{-2}$ \\
\bottomrule
\end{tabular}%
}

\caption{
Mean feature values for Non-ADHD and ADHD groups (ASRS) from the \textsc{Real} dataset, 
with corresponding $p$-values from two-sample $t$-tests ($p<0.05$). 
}
\label{tab:asrs_medians_pvalues}
\end{table}

Table~\ref{tab:gad7_means_pvalues} presents the most significant features identified by the t-tests for distinguishing individuals scoring above the cutoff for moderate anxiety in the GAD-7, with features associated with voice quality and emotional expression emerging as the most influential. However, after applying FDR correction, no features remained statistically significant. As shown in Figure~\ref{fig:feature_comparison}, further analysis revealed that graph-based, semantic, voice quality, emotional, and lexical features played key roles in discrimination. Across all methodologies, features related to voice quality and emotional characteristics consistently proved to be the most important indicators of anxiety.

\begin{table}[!b]
\centering
\footnotesize
\setlength{\tabcolsep}{2.5pt} 
\renewcommand{\arraystretch}{1.1}
\begin{tabular}{lccc}
\toprule
\textbf{Feature} & \textbf{Non-Anxiety} & \textbf{Anxiety} & \textbf{$p$-value} \\
\midrule
vader\_negative\ & 0.030 & 0.041 & 6.81$\times$10$^{-3}$ \\
Shimmer\_local\  & 0.111 & 0.103 & 2.17$\times$10$^{-2}$ \\
\bottomrule
\end{tabular}
\caption{
Mean feature values for Non-Anxiety and Anxiety groups (GAD-7), 
with corresponding $p$-values from two-sample $t$-tests ($p<0.05$).
}
\label{tab:gad7_means_pvalues}

\end{table}

\section{Discussion}
Section~\ref{sec:res} suggests two key points. First, predicting psychopathology from voice is a challenging task, as also reported in~\cite{berisha2024responsible}. Second, there is preliminary evidence that certain features could act as potential indicators across different conditions and clinical cases.

For anxiety and stress, Shimmer was the most prominent feature across both datasets. Shimmer measures cycle-to-cycle amplitude variation in successive glottal cycles, reflecting irregularities in vocal fold vibration. This finding is consistent with prior work~\cite{teferra2022acoustic} and supports the potential of Shimmer as an indicator for anxiety. In our analysis, higher Shimmer values were associated with increased anxiety, in line with~\citet{jones2011vocal}, although opposite trends have also been reported~\cite{basar2023relationship}, indicating the need for further validation before clinical use.

For the ADHD-related ASRS score, we found that the most important features were graph-based measures capturing repetition patterns~\cite{Mota2012}, as well as verb-tense-related features that quantify how frequently speakers shift between tenses. This aligns with previous literature~\cite{engelhardt2011language}, which reports that individuals with ADHD often exhibit higher levels of repetition and disfluency—patterns that our methodology appears capable of capturing as well.

For PHQ-9 based depression, we found that features capturing emotional content from both audio and text played an important role, along with the content-function ratio feature. The content-function ratio measures the balance between meaning-bearing words and grammatical words, indicating how information-dense or syntactically simple a text is. From this observation, we see that our system captures features reflecting both emotional expression (e.g., sadness) and syntactic richness. These findings are consistent with previous work showing that emotional and affective cues in both speech and language serve as reliable indicators of depressive states~\cite{cummins2015review}. Likewise, the relevance of the content-function ratio aligns with the linguistic style framework proposed by Pennebaker and King~\cite{pennebaker1999linguistic}, where the balance between content and function words reflects cognitive style and syntactic complexity often associated with mood and mental health variations.

These findings highlight the value of an integrative, feature-centered approach to speech-based psychopathology analysis. The work combines foundation-model-based representations with traditional acoustic and linguistic features, classical statistical analysis, and interpretable machine learning across conditions and datasets. We posit that this integrated and exploratory approach is particularly well suited to clinical and interdisciplinary contexts, where transparent indicators are preferred, and may facilitate interpretable insights and hypothesis generation.

\section{Conclusion}
\label{sec:conclusion}

We introduced an interpretable, perceptually grounded framework for exploring speech-based correlates of psychopathology across heterogeneous datasets, languages, and labeling schemes. Using a compact set of acoustic and linguistic descriptors and transparent models with complementary explanations, we examined feature patterns (prosody/fluency, affect, voice quality, and discourse markers) that appeared to be associated with stress, depression, anxiety, and ADHD-related screening status. Future work should further investigate these candidate indicators under stronger reference standards and in longitudinal, out-of-domain settings.

\section{Limitations}
Despite promising within-dataset results, applying speech-based models for psychopathology prediction in real-world settings remains difficult. As highlighted by Berisha and Liss~\cite{berisha2024responsible}, speech is affected by numerous confounding factors, such as  fatigue, and background noise, that obscure condition-specific indicators and reduce model robustness. Differences in recording devices, linguistic and cultural backgrounds, and labeling protocols further introduce domain bias. Although some cross-setting generalization was observed, performance remained sensitive to acoustic variability and contextual shifts. Moreover, short speech samples and static features may overlook temporal cues critical for capturing symptom dynamics. Advancing clinical applicability therefore requires adaptive preprocessing, domain-invariant features, and large datasets that reflect naturalistic conditions. Finally, because several datasets use questionnaire-based cutoffs (including PHQ-8/9, GAD-7, and ASRS), the supervision signal reflects widely used, but imperfect, measurement instruments, and some errors may reflect measurement/cutoff variability rather than model behavior alone. This does not change the goal of the study (interpretability and clinical plausibility), but it does bound how literally one should read classification metrics as reflecting “true” clinical status across datasets. Additionally, neural-based features such as sarcasm detection should be interpreted cautiously, as these models have imperfect accuracy and may reproduce biases present in their training data.

\section{Acknowledgment}
This work was supported by PsychNow. We thank the clinical and product teams at PsychNow for their support and feedback.

\bibliography{custom}
\clearpage
\appendix

\section{Demographic Data}

This section summarizes the demographic and clinical score distributions across the evaluated datasets. Where available, standardized psychometric instruments are used to characterize symptom severity and population variability.

\begin{figure*}[!h]
    \centering
    \begin{minipage}{0.45\textwidth}
        \centering
        \includegraphics[width=\linewidth]{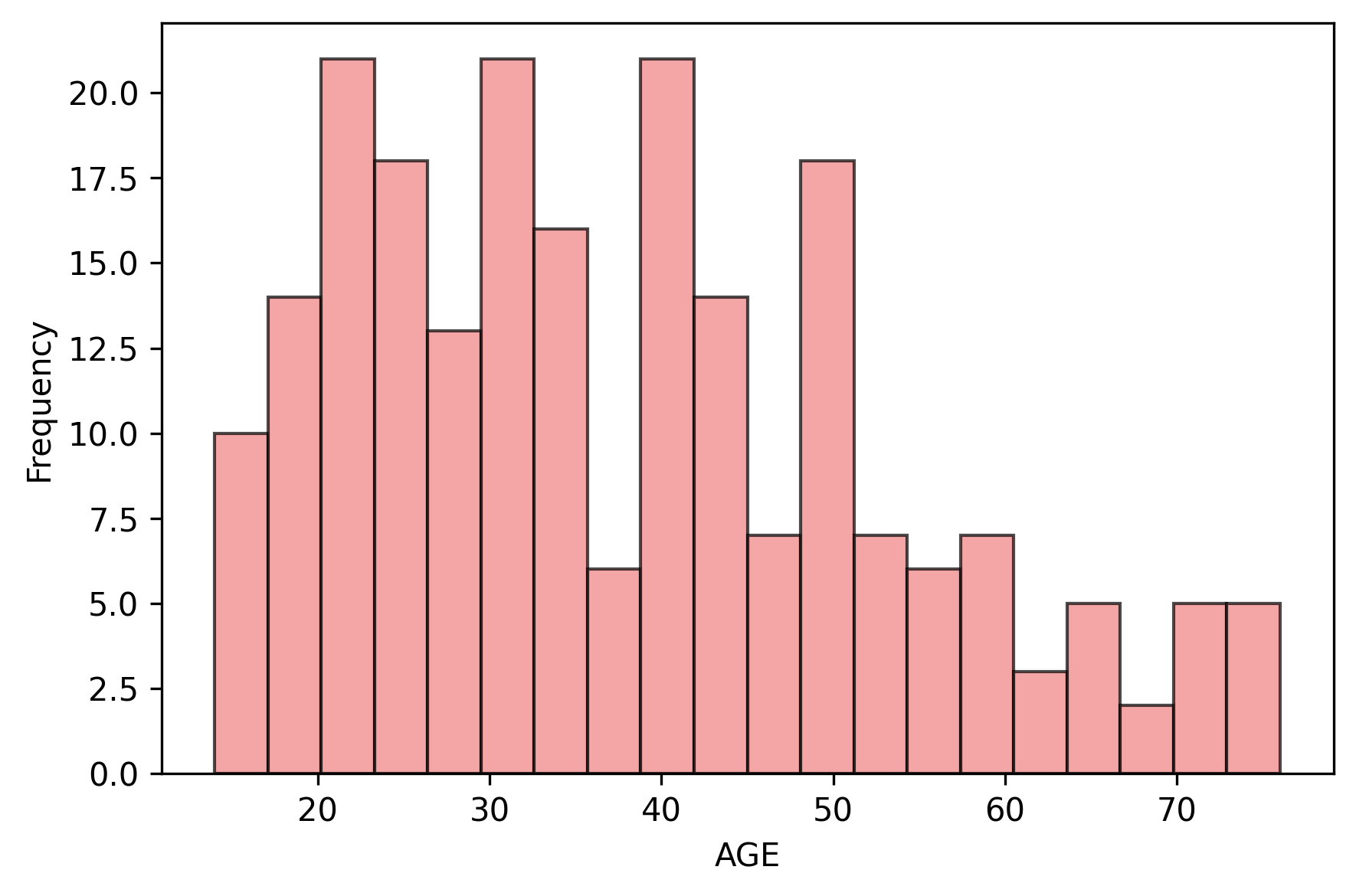}
        \caption*{Age}
    \end{minipage}
    \hfill
    \begin{minipage}{0.45\textwidth}
        \centering
        \includegraphics[width=\linewidth]{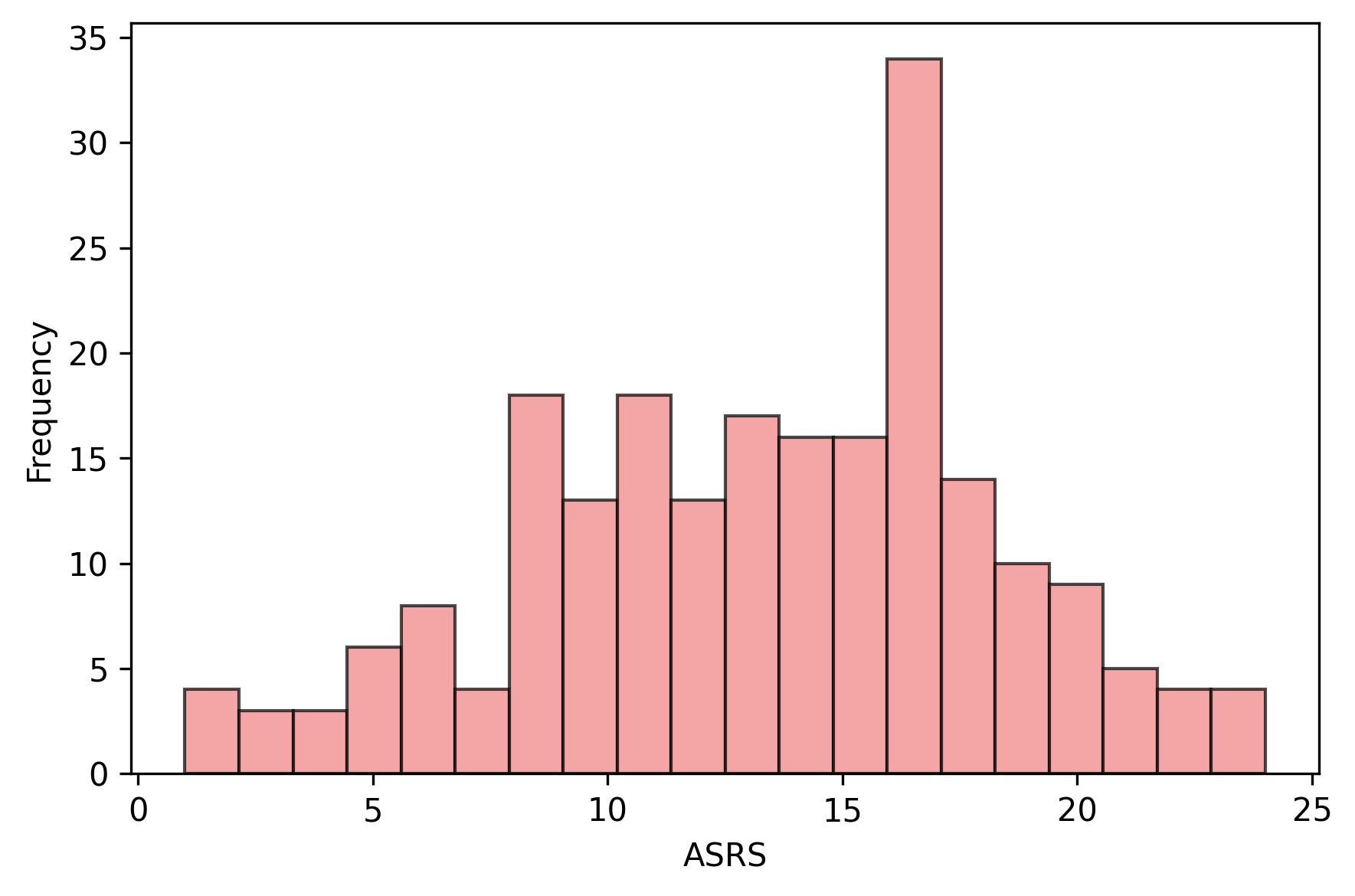}
        \caption*{ASRS}
    \end{minipage}

    \vspace{0.4cm}

    \begin{minipage}{0.45\textwidth}
        \centering
        \includegraphics[width=\linewidth]{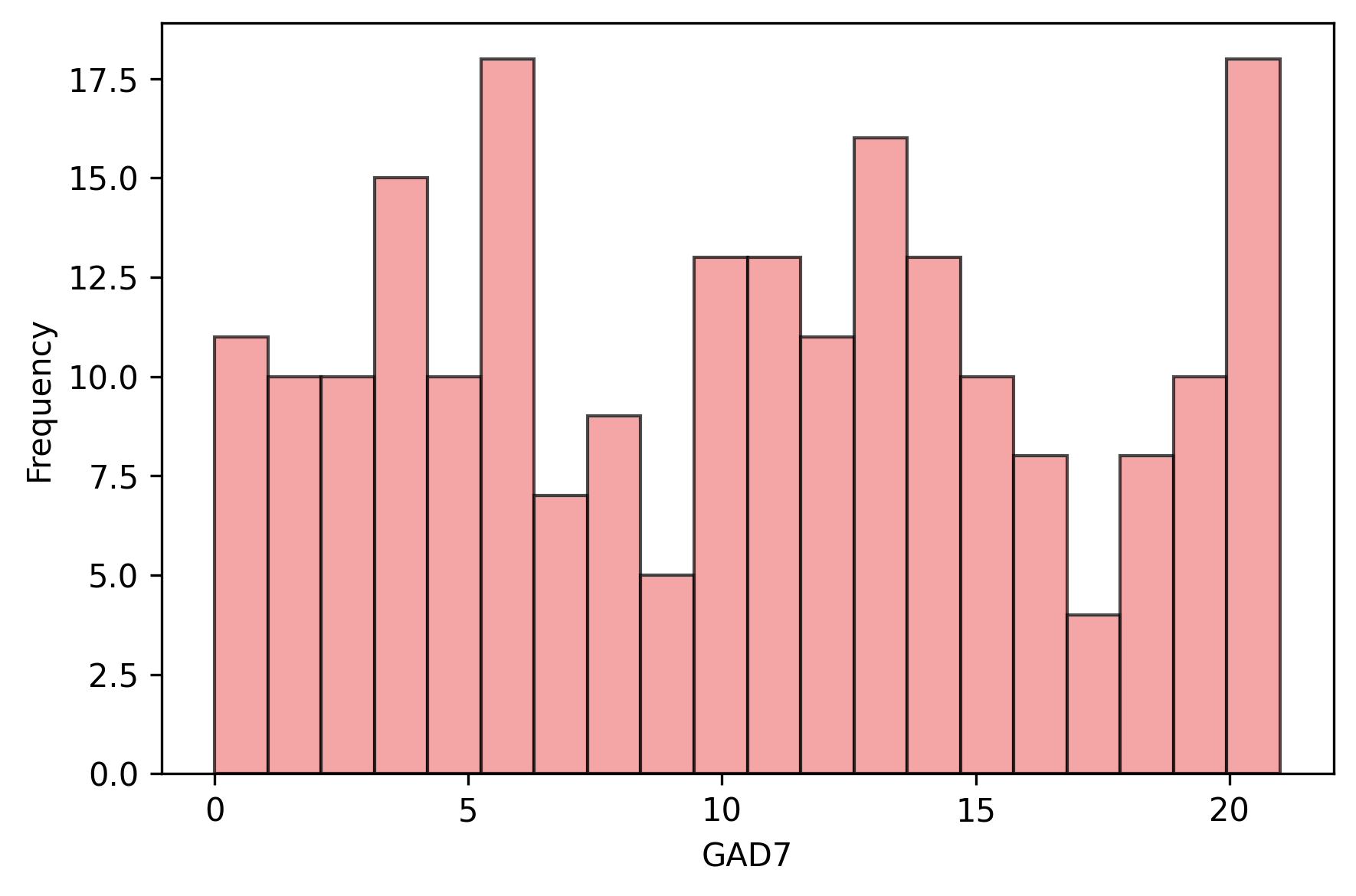}
        \caption*{GAD-7}
    \end{minipage}
    \hfill
    \begin{minipage}{0.45\textwidth}
        \centering
        \includegraphics[width=\linewidth]{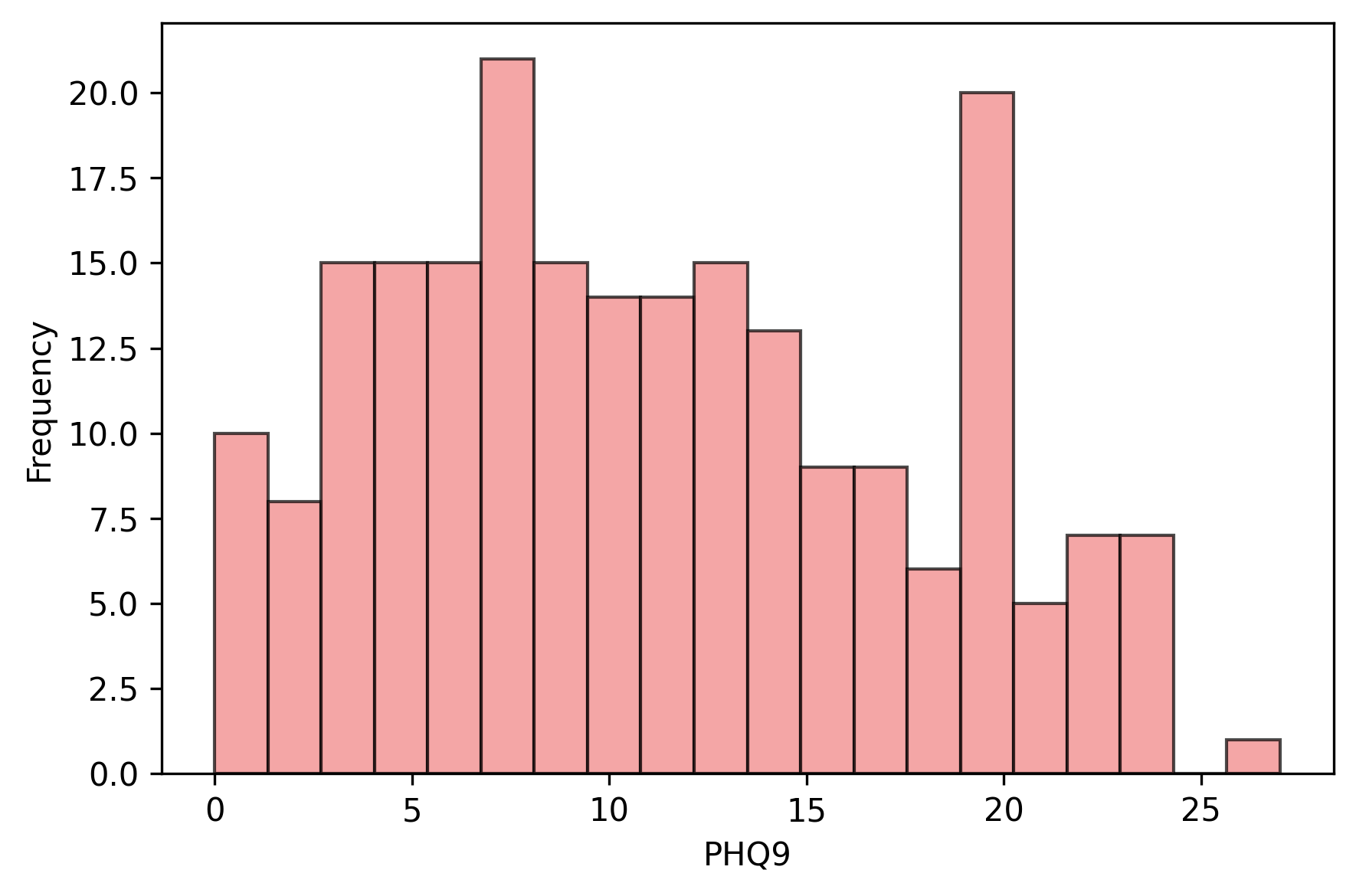}
        \caption*{PHQ-9}
    \end{minipage}

    \caption{Distributions of demographic and clinical variables in the \textsc{Real} dataset.}
    \label{fig:dataset_histograms_real}
\end{figure*}

For the \textsc{Real} dataset, Figure~\ref{fig:dataset_histograms_real} presents distributions of age and self-reported clinical scales. Participant ages span a broad range, with most individuals between their 20s and 50s. ASRS scores show moderate dispersion with a noticeable concentration at higher values, indicating a substantial proportion of participants exhibiting elevated ADHD-related traits. GAD-7 scores are broadly distributed, suggesting balanced representation across anxiety severity levels, while PHQ-9 scores span the full clinical range, with many participants in the moderate to severe depression categories.

\begin{figure}[!h]
    \centering
    \includegraphics[width=1\linewidth]{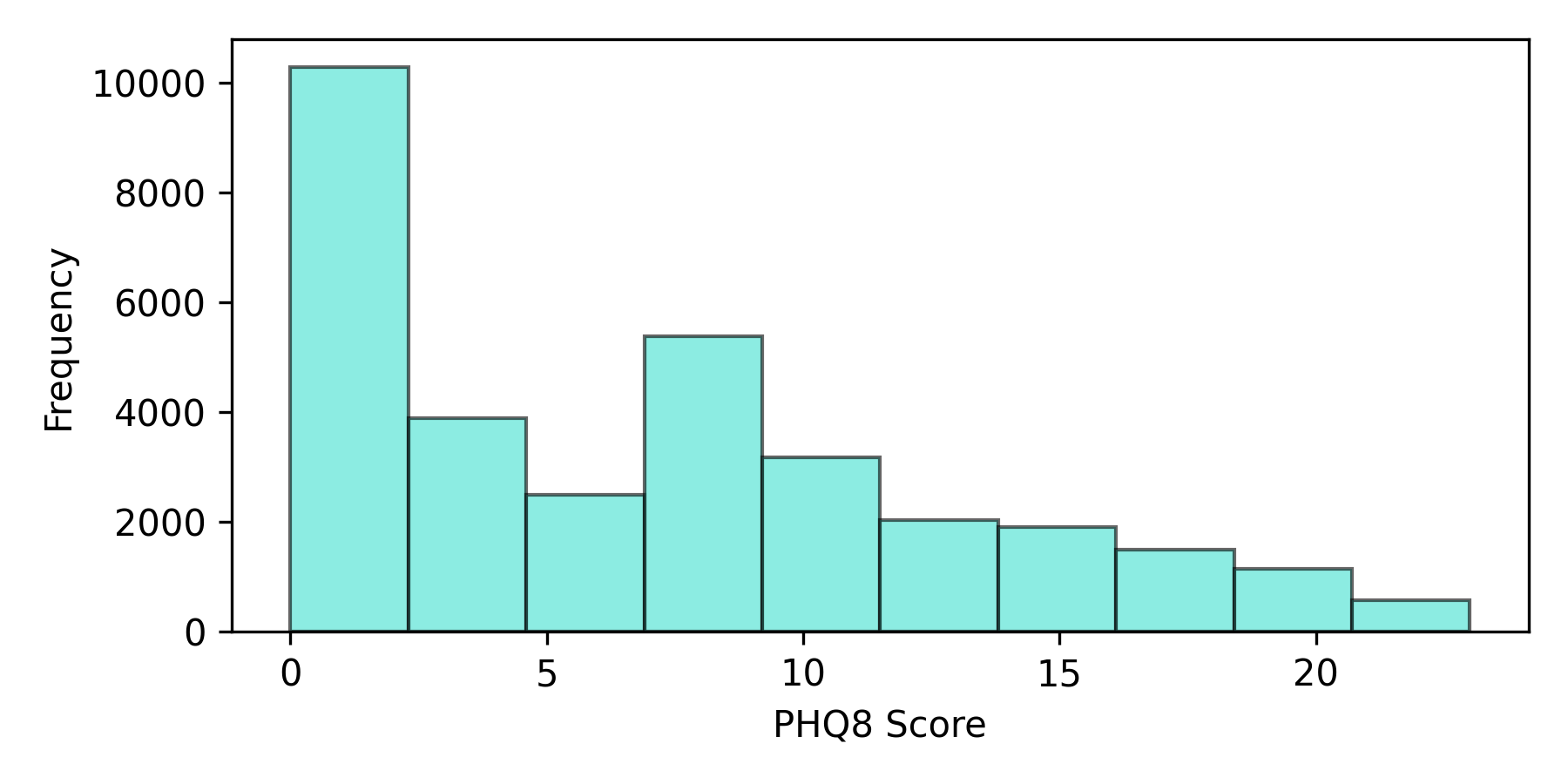}
    \caption{Distribution of PHQ-8 depression scores in the \textsc{DAIC-WOZ} dataset.}
    \label{fig:dataset_histograms_daic}
\end{figure}

For the \textsc{DAIC-WOZ} corpus, Figure~\ref{fig:dataset_histograms_daic} illustrates the distribution of PHQ-8 scores obtained during semi-structured clinical interviews. The distribution reflects substantial variability in depressive symptom severity, supporting its use as a benchmark dataset for depression detection from speech.

\begin{figure}[!h]
    \centering
    \includegraphics[width=1\linewidth]{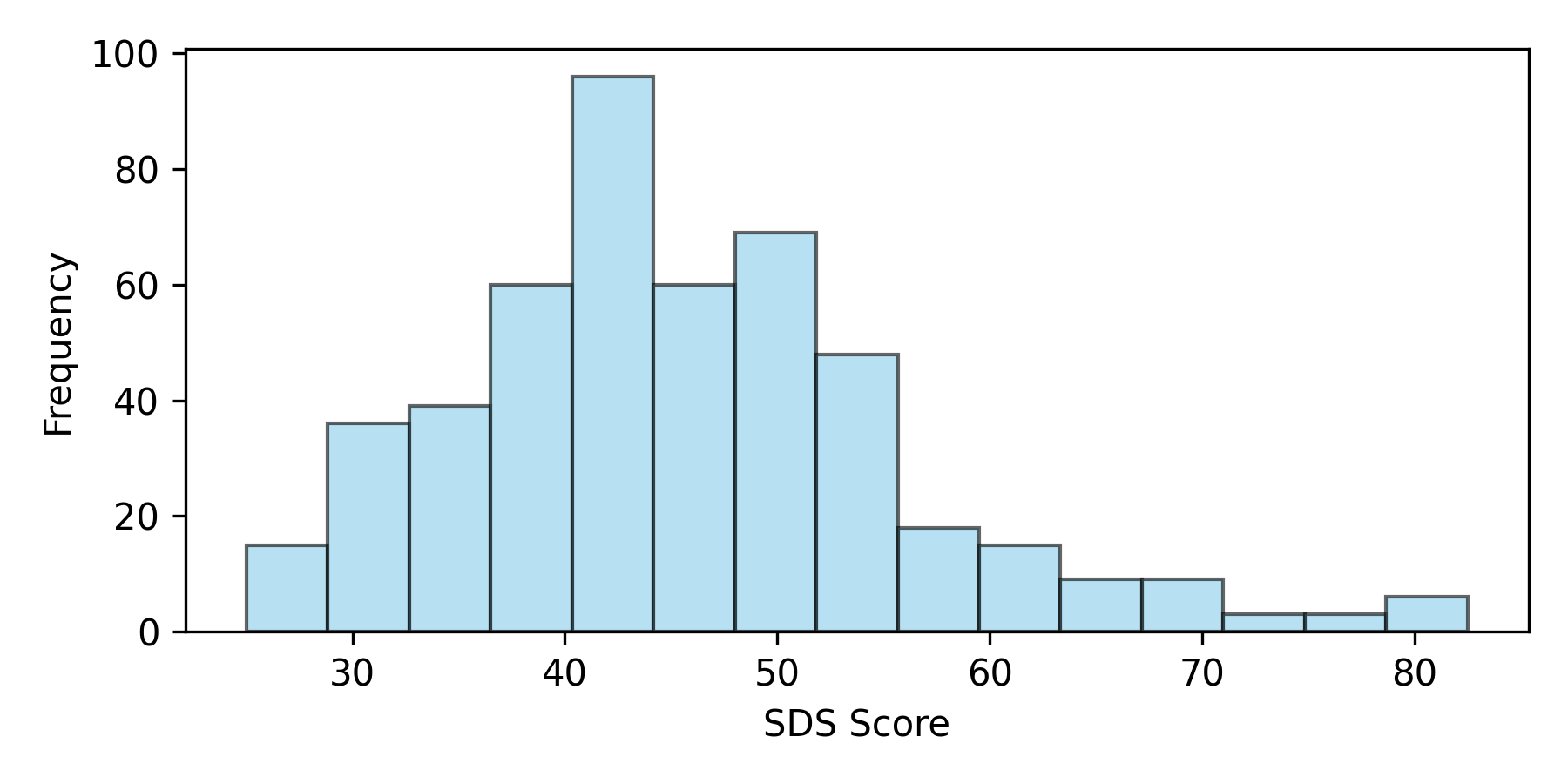}
    \caption{Distribution of SDS depression scores in the \textsc{EATD} dataset.}
    \label{fig:dataset_histograms_eatd}
\end{figure}

Figure~\ref{fig:dataset_histograms_eatd} shows the distribution of SDS scores for the \textsc{EATD} dataset. Most participants fall below the clinical threshold, with a smaller subset exhibiting elevated depression scores, consistent with the dataset’s class imbalance and student-based recruitment.

\section{Feature Distributions}

\begin{figure*}[t]
    \centering
    \setlength{\tabcolsep}{2pt}
    \renewcommand{\arraystretch}{0.2}
    \scriptsize

    \begin{tabular}{cccccccccc}
        \includegraphics[width=0.09\textwidth]{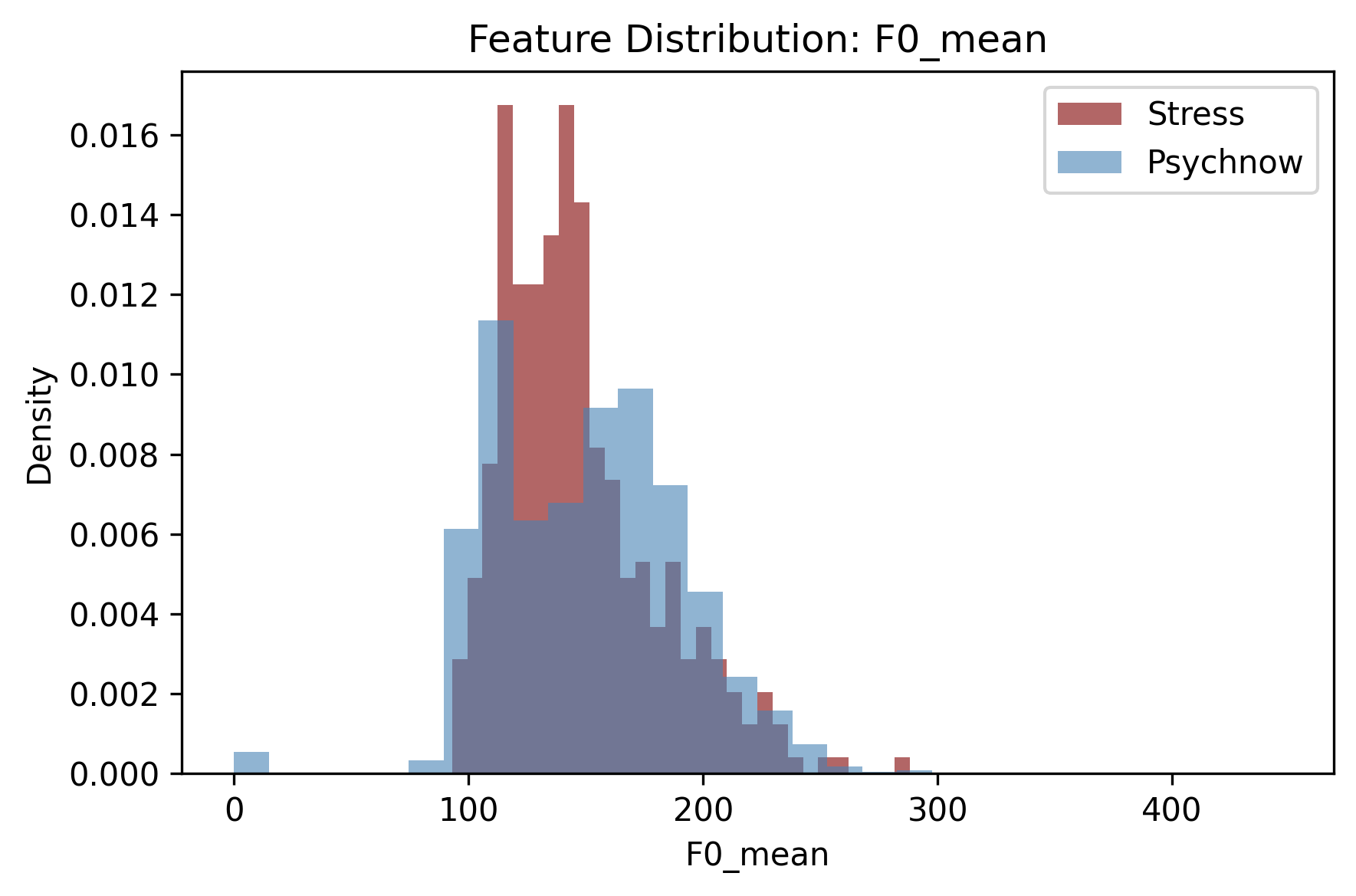} &
        \includegraphics[width=0.09\textwidth]{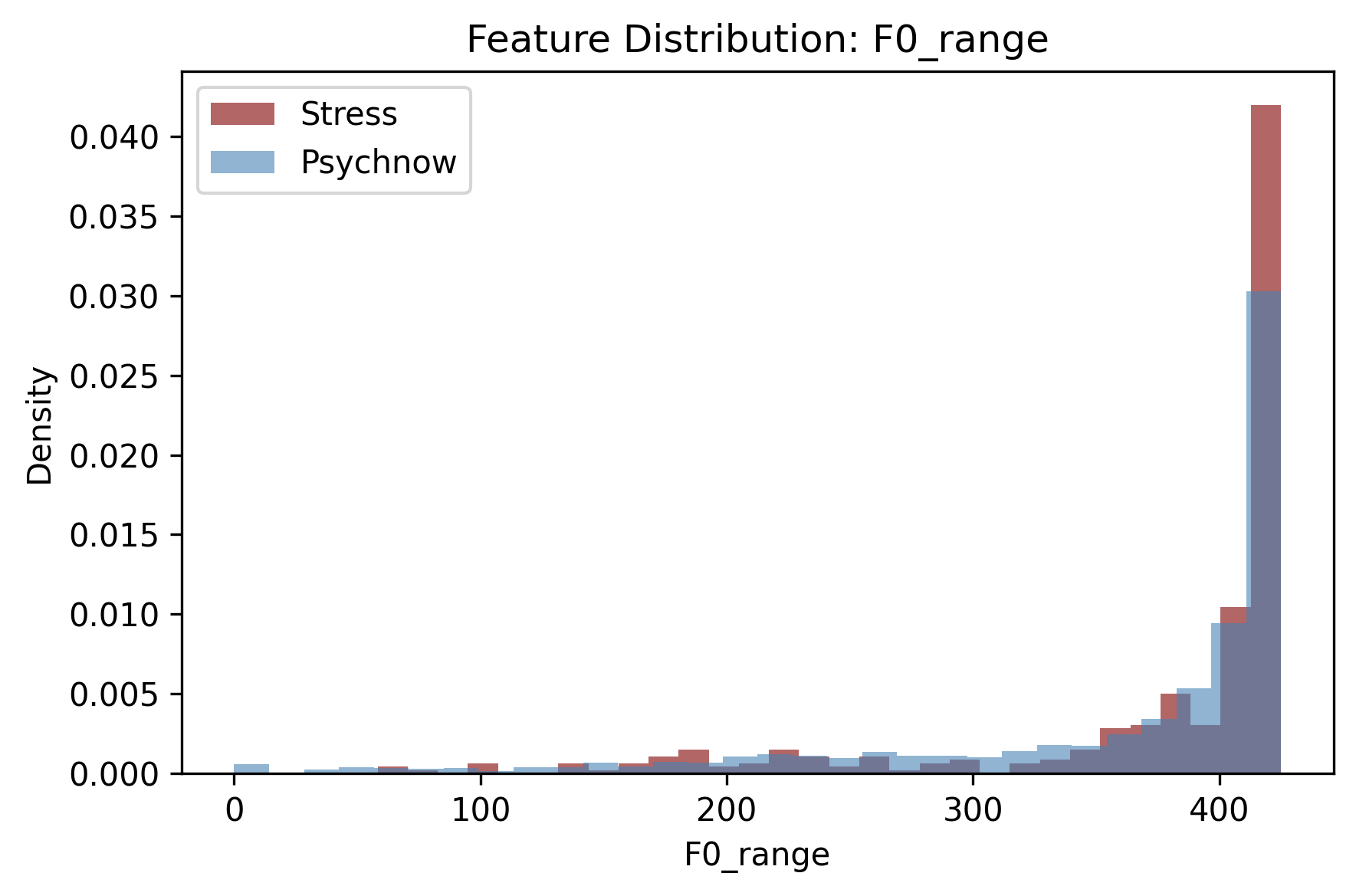} &
        \includegraphics[width=0.09\textwidth]{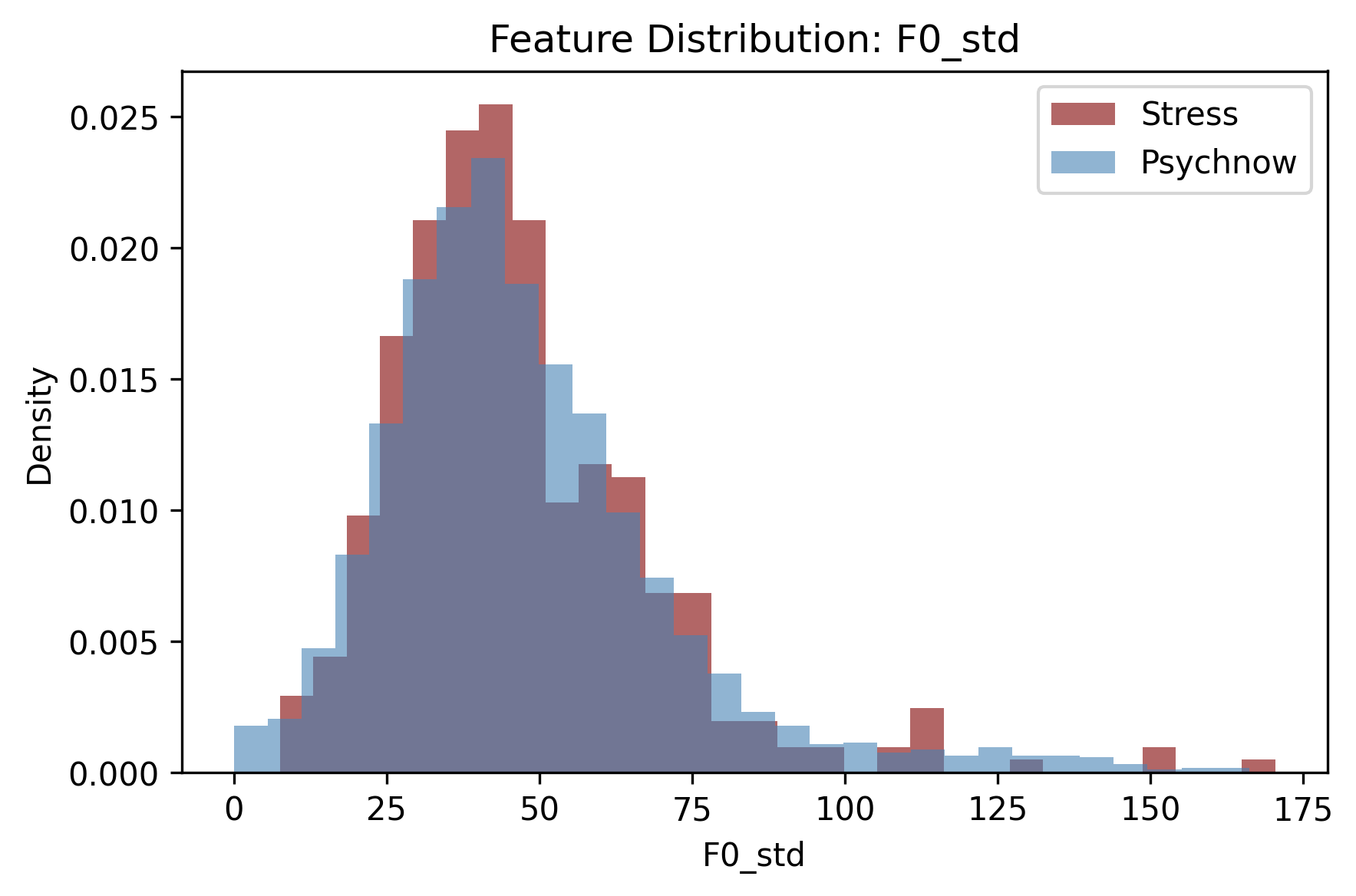} &
        \includegraphics[width=0.09\textwidth]{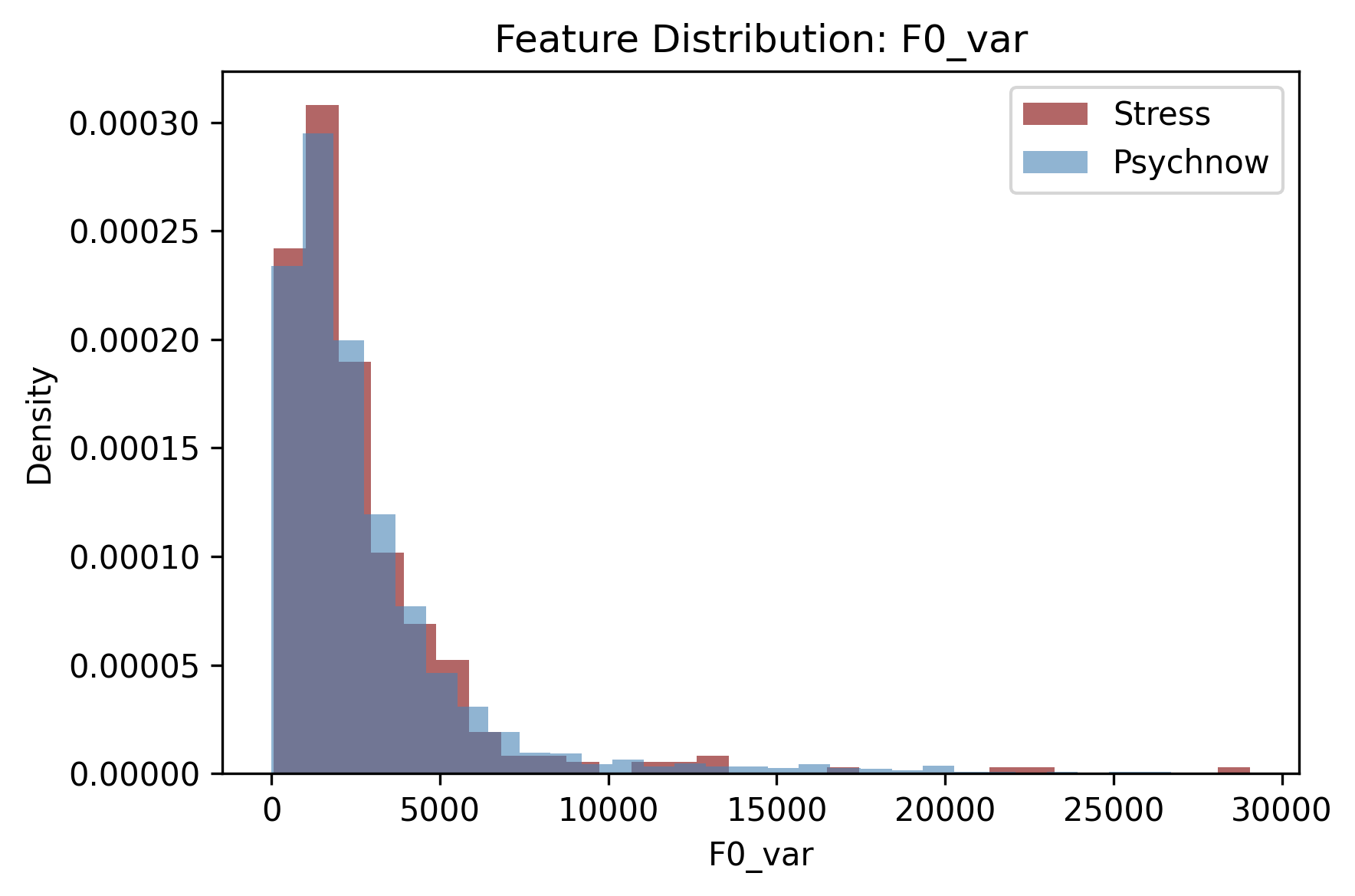} &
        \includegraphics[width=0.09\textwidth]{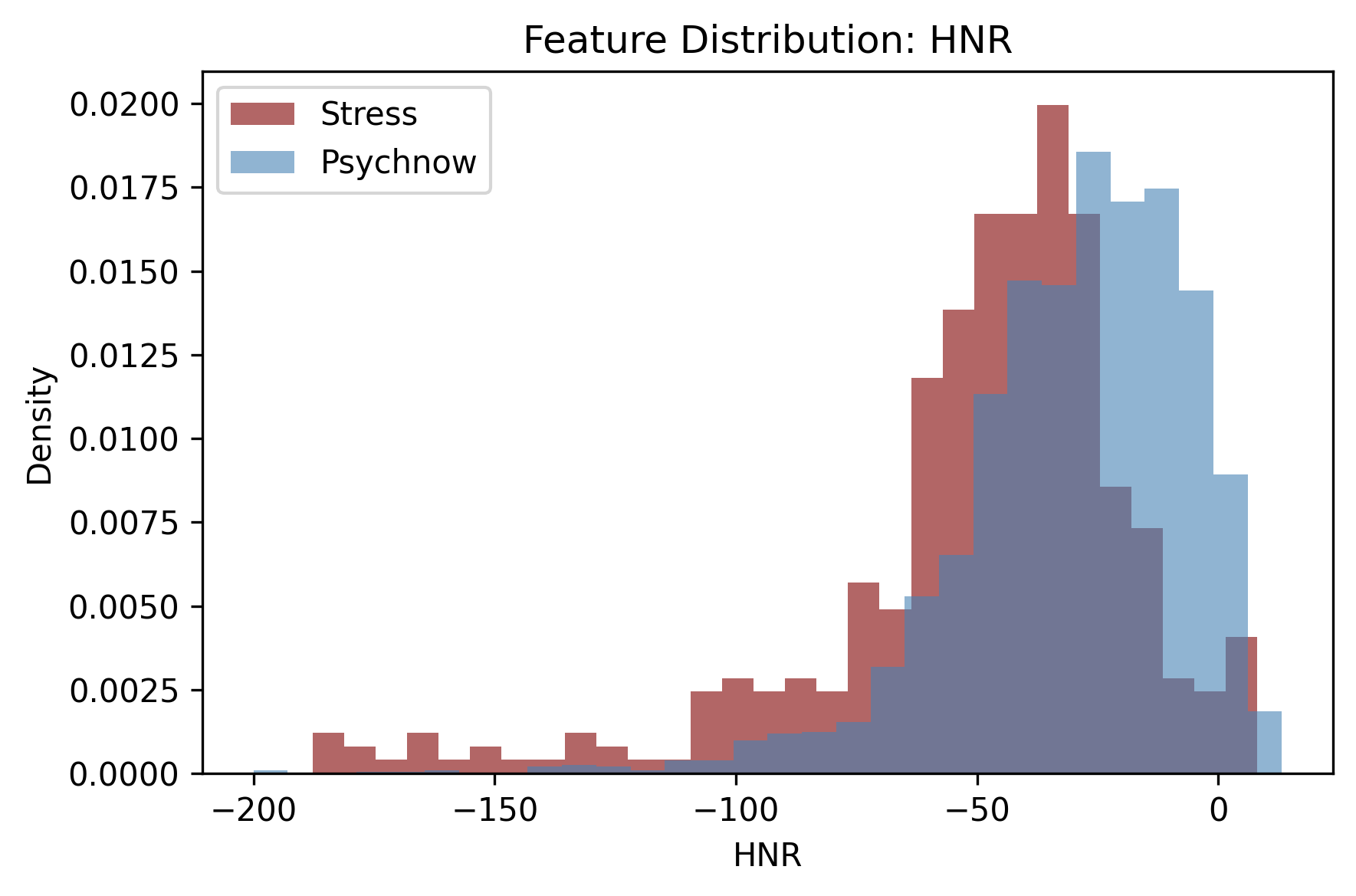} &
        \includegraphics[width=0.09\textwidth]{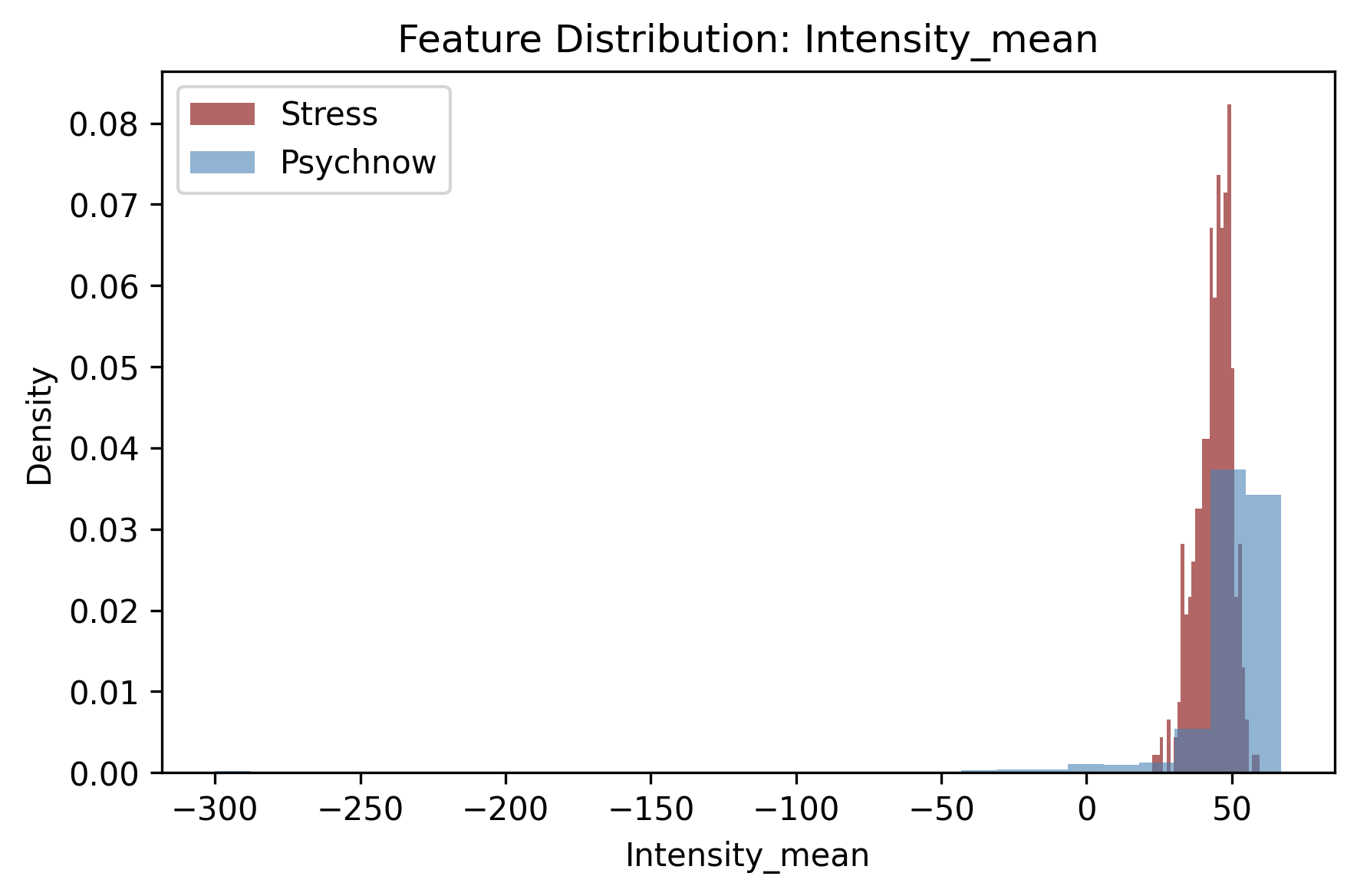} &
        \includegraphics[width=0.09\textwidth]{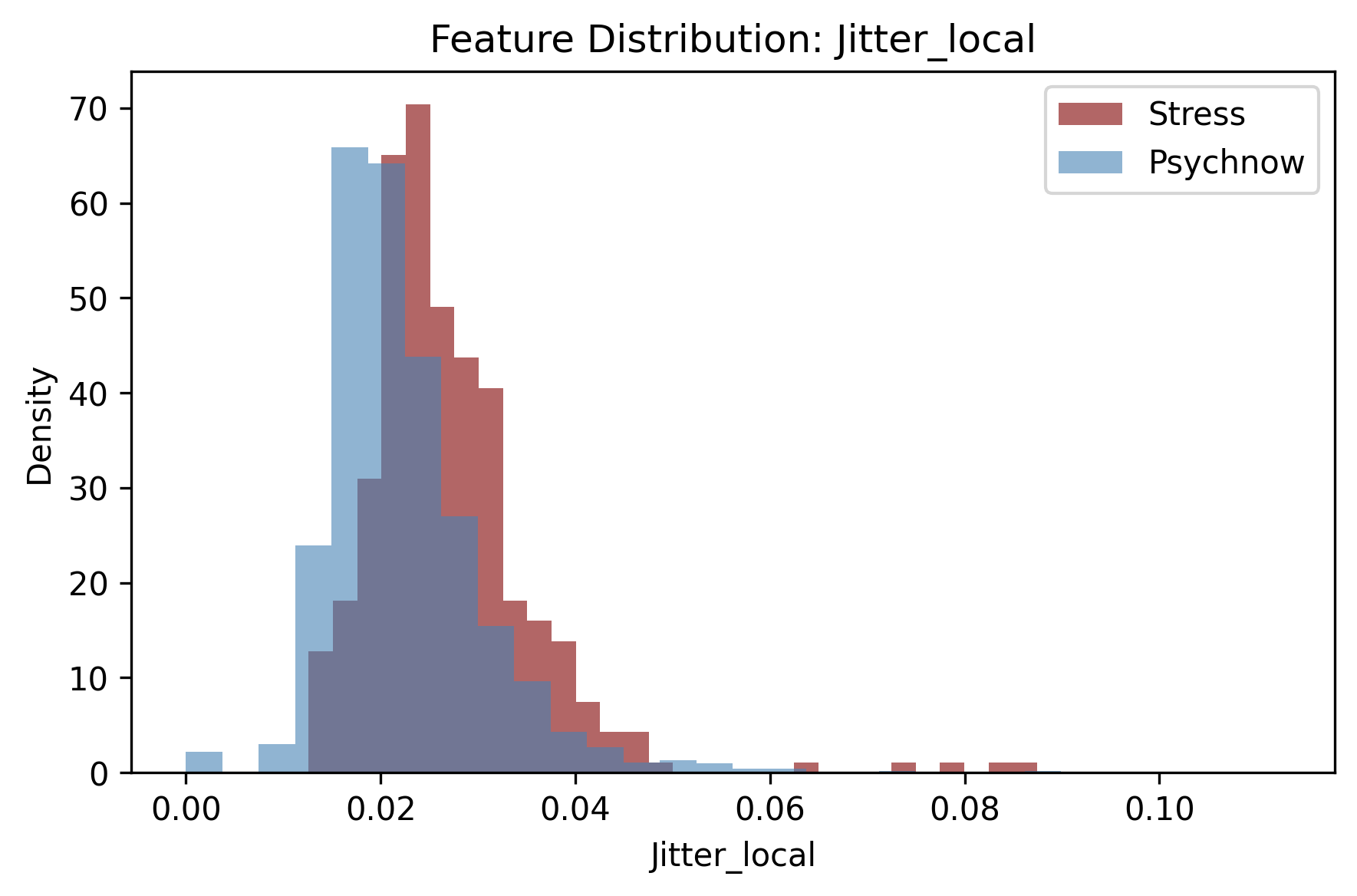} &
        \includegraphics[width=0.09\textwidth]{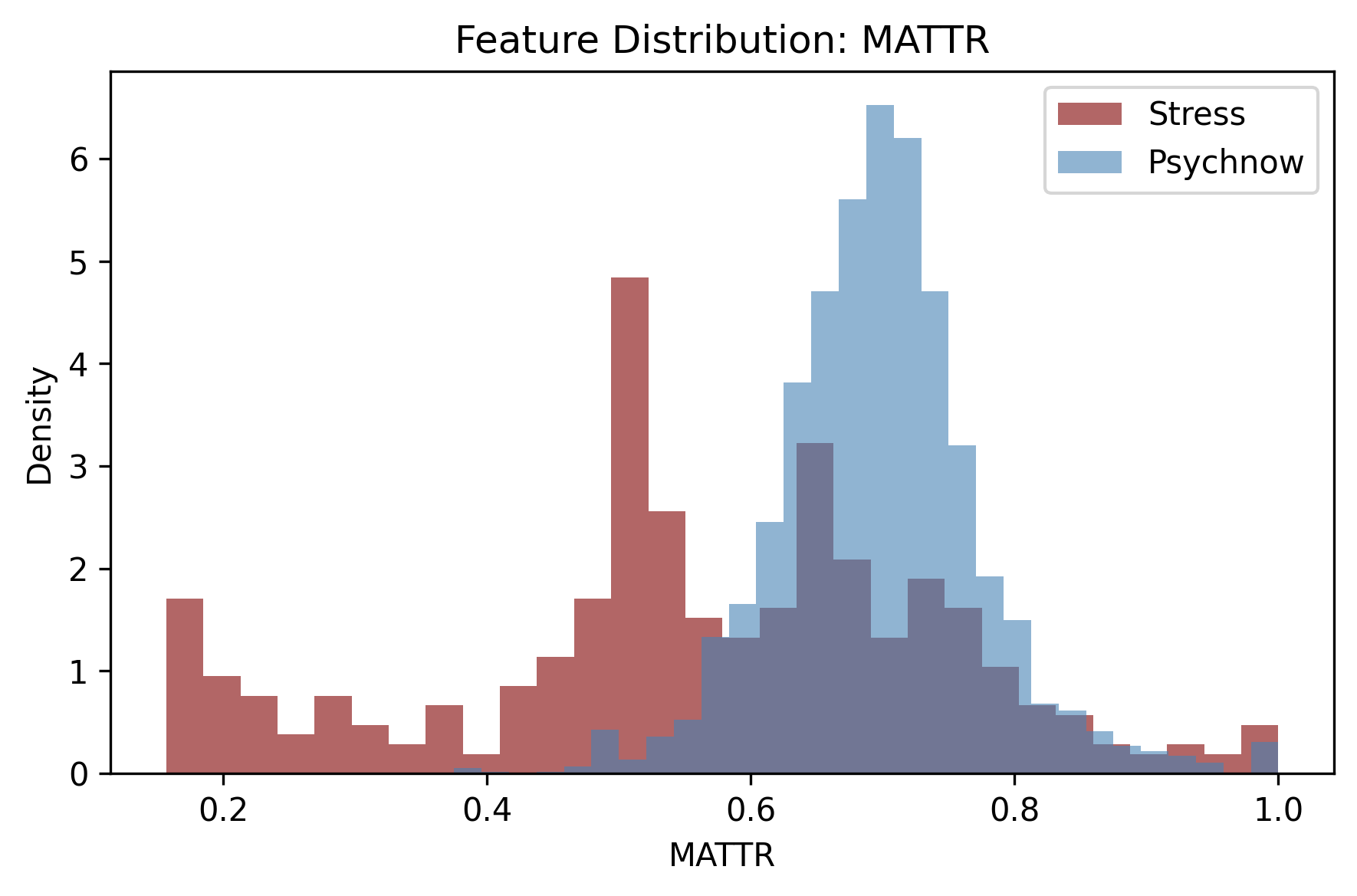} &
        \includegraphics[width=0.09\textwidth]{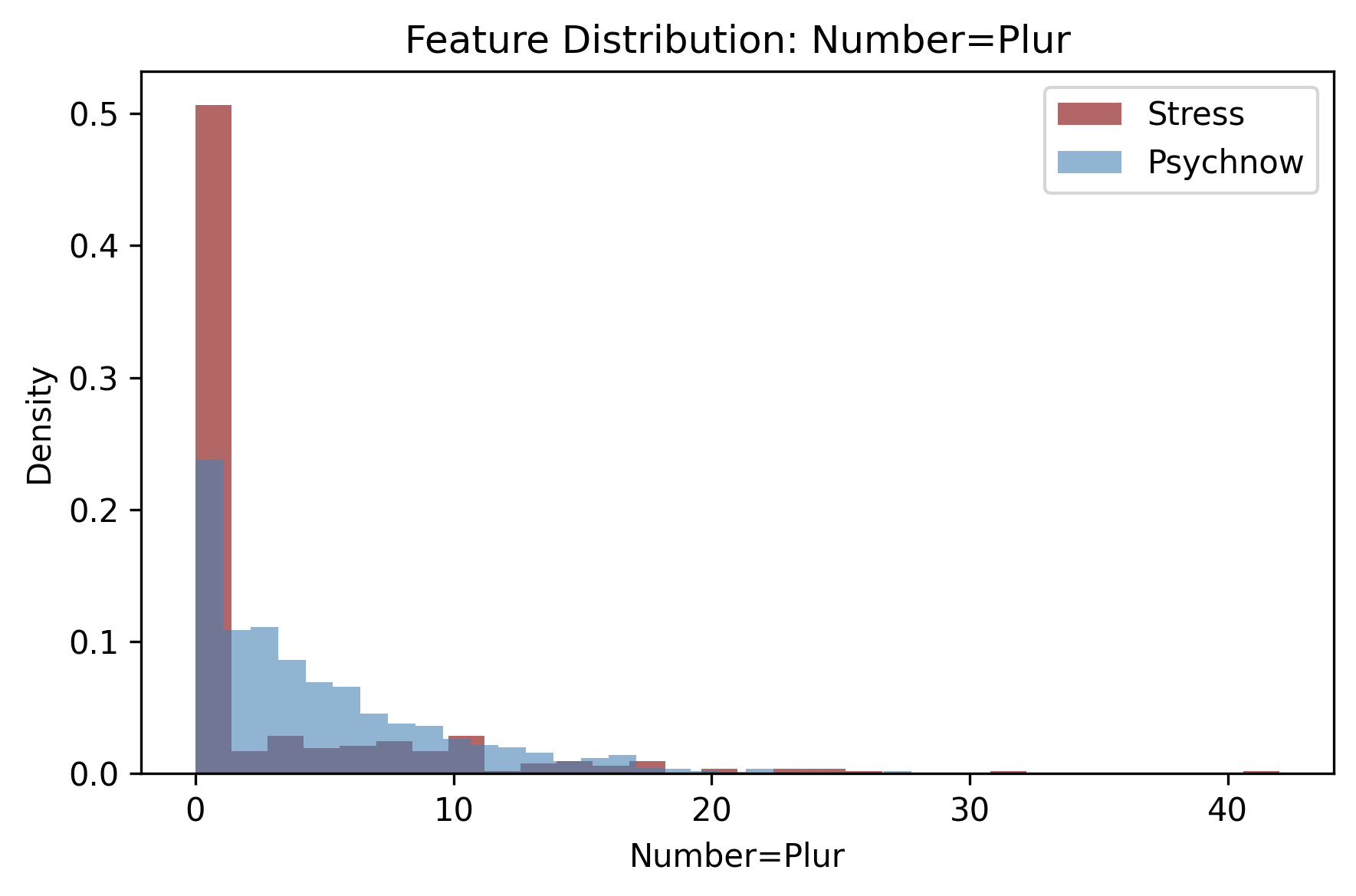} &
        \includegraphics[width=0.09\textwidth]{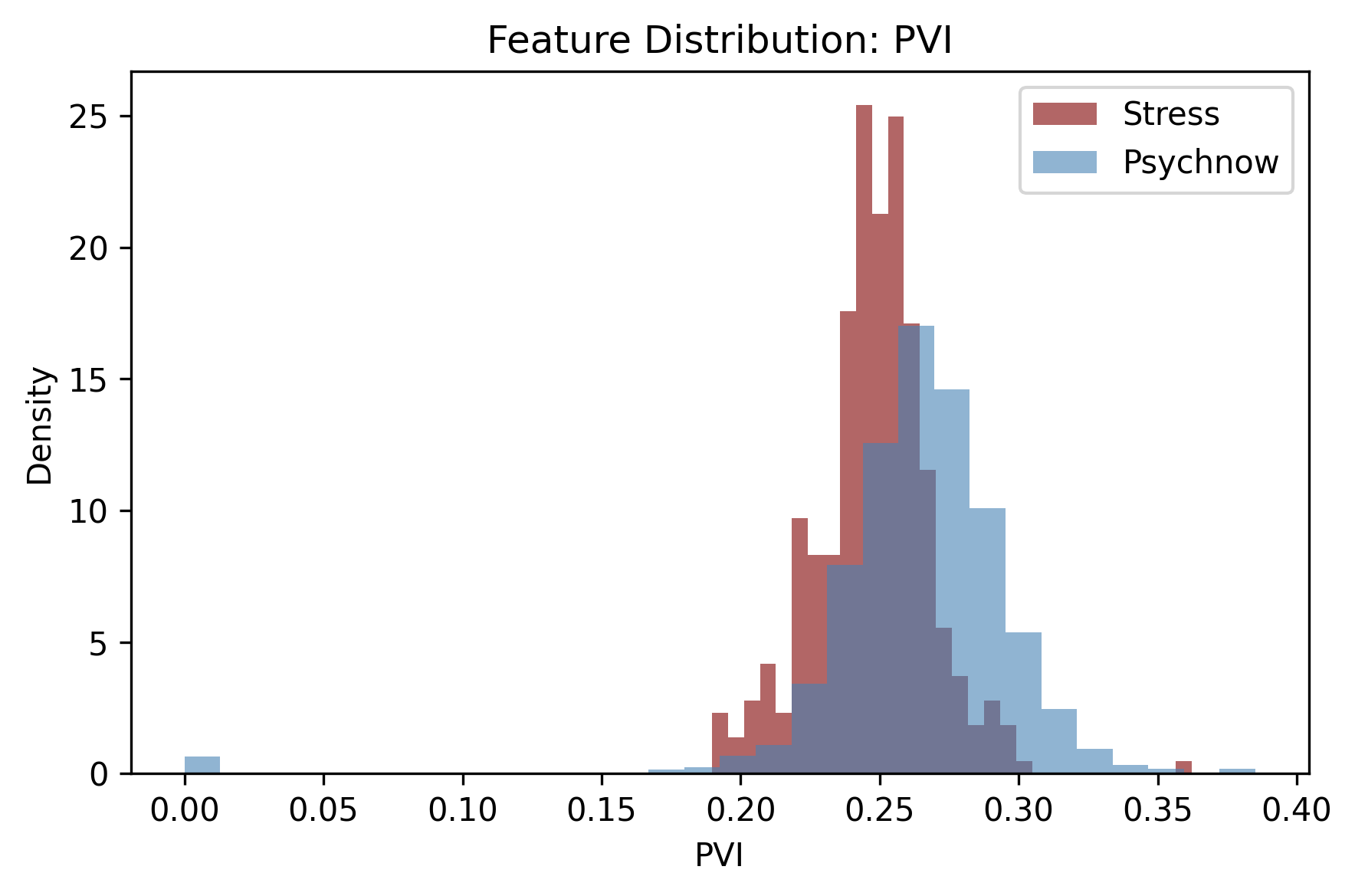} \\[-2pt]

        \includegraphics[width=0.09\textwidth]{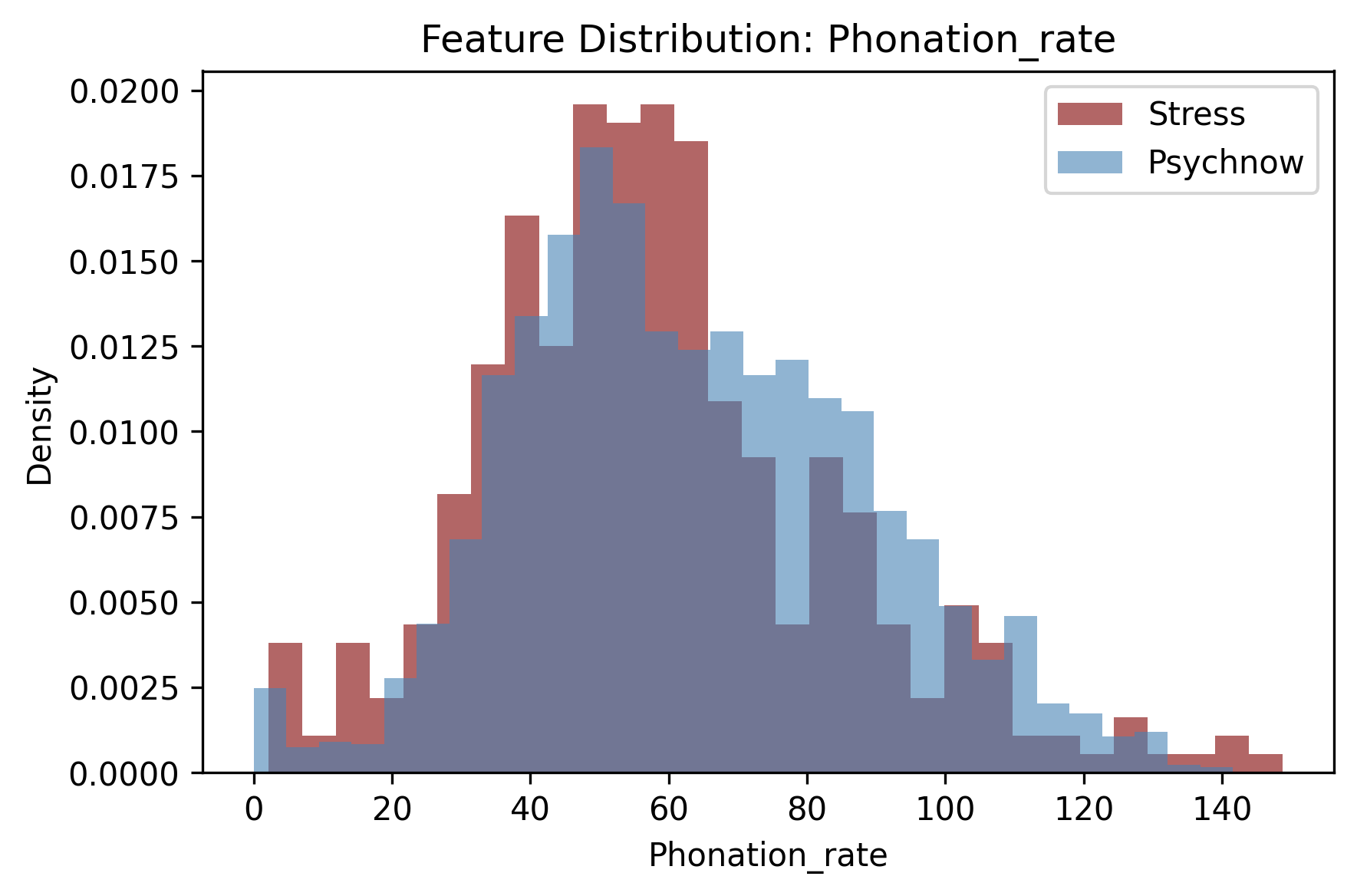} &
        \includegraphics[width=0.09\textwidth]{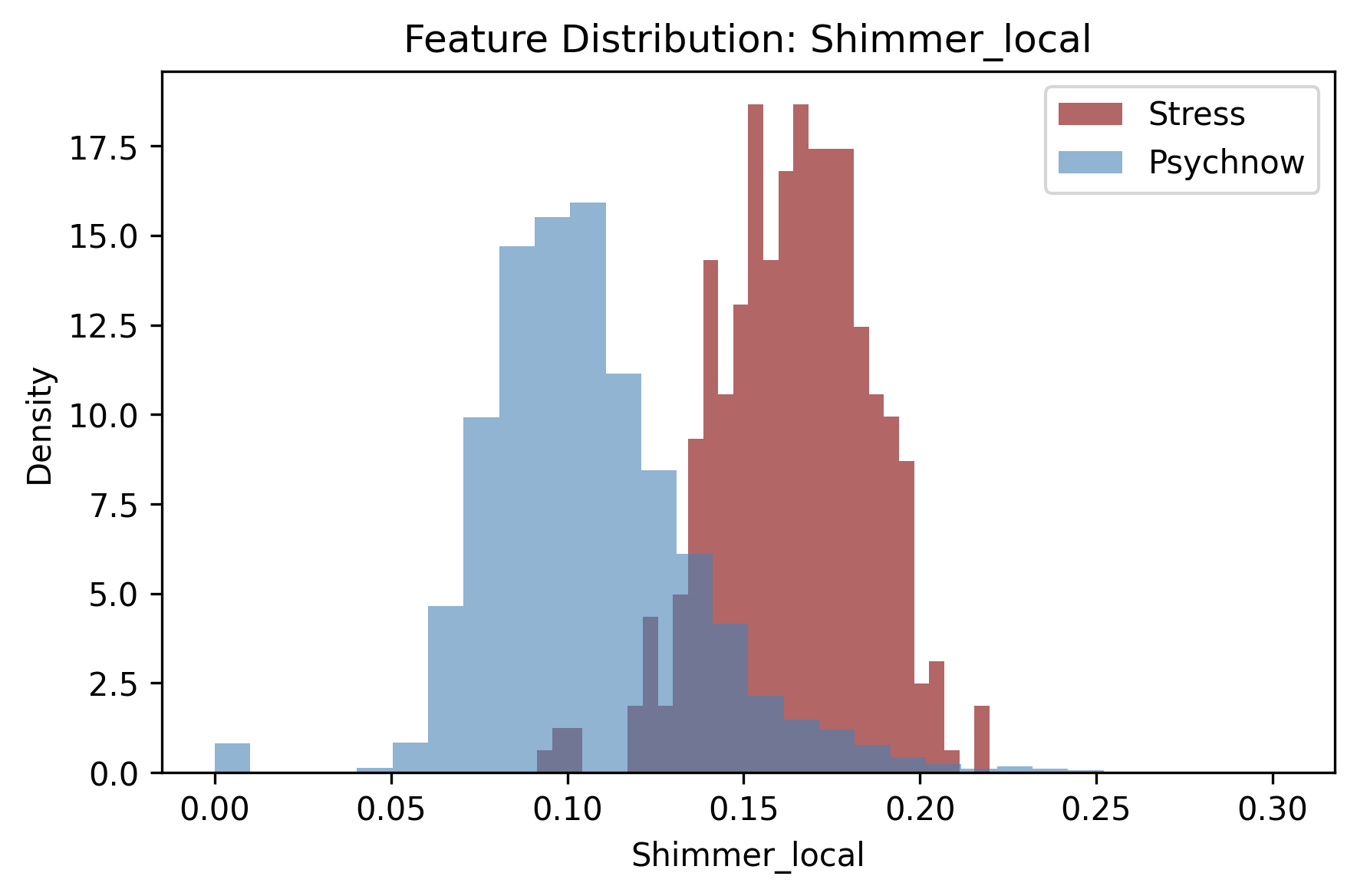} &
        \includegraphics[width=0.09\textwidth]{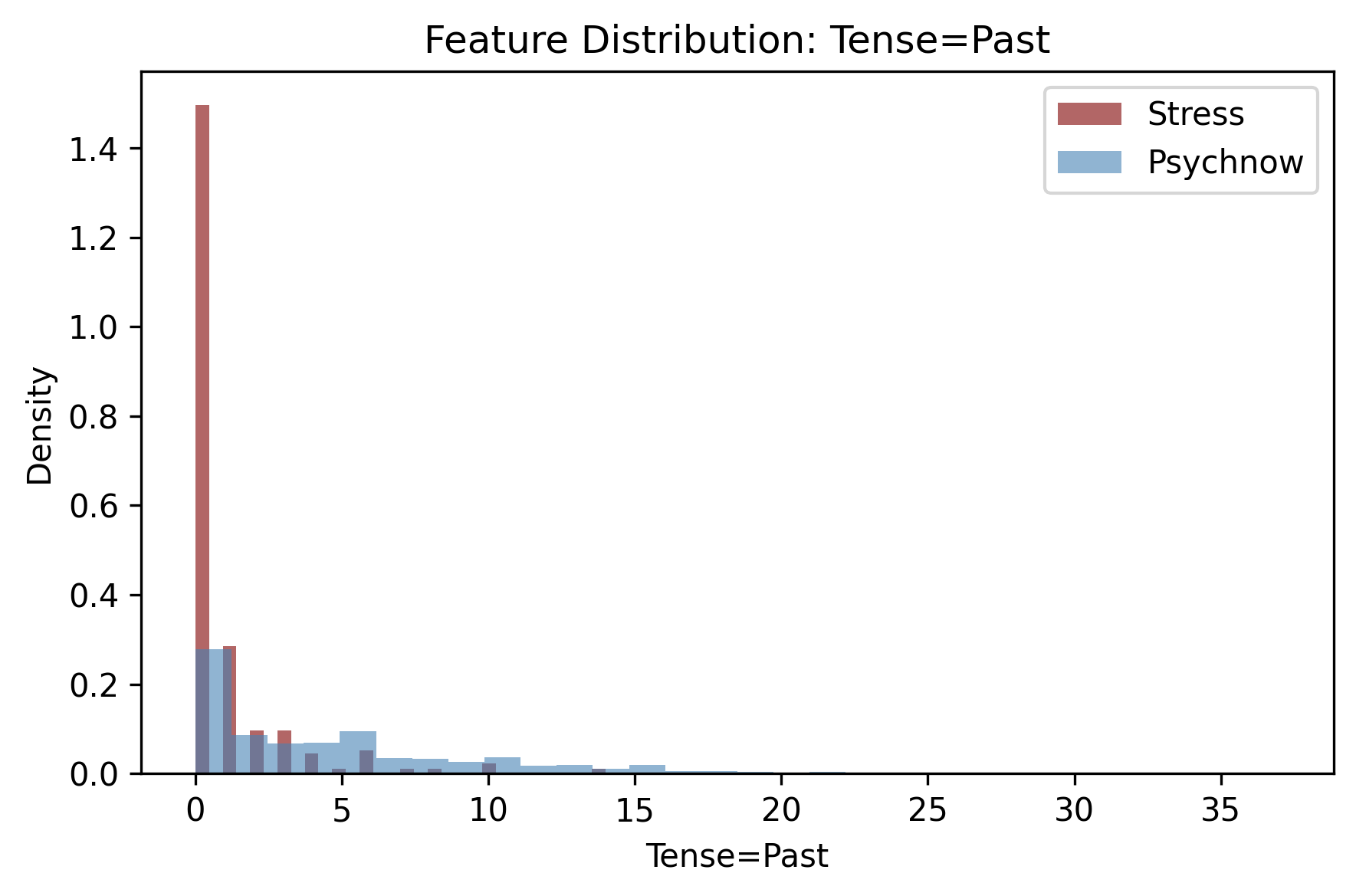} &
        \includegraphics[width=0.09\textwidth]{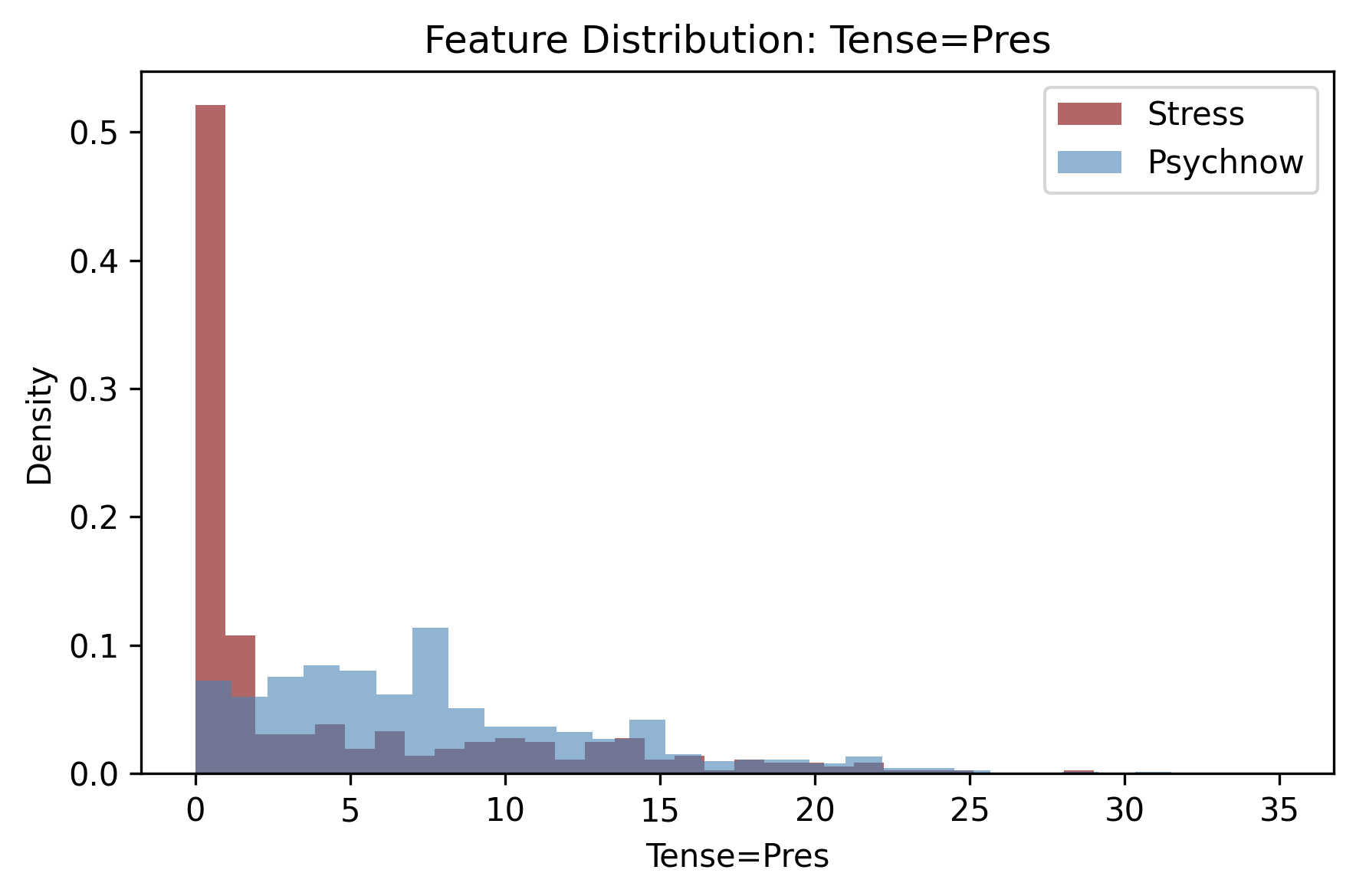} &
        \includegraphics[width=0.09\textwidth]{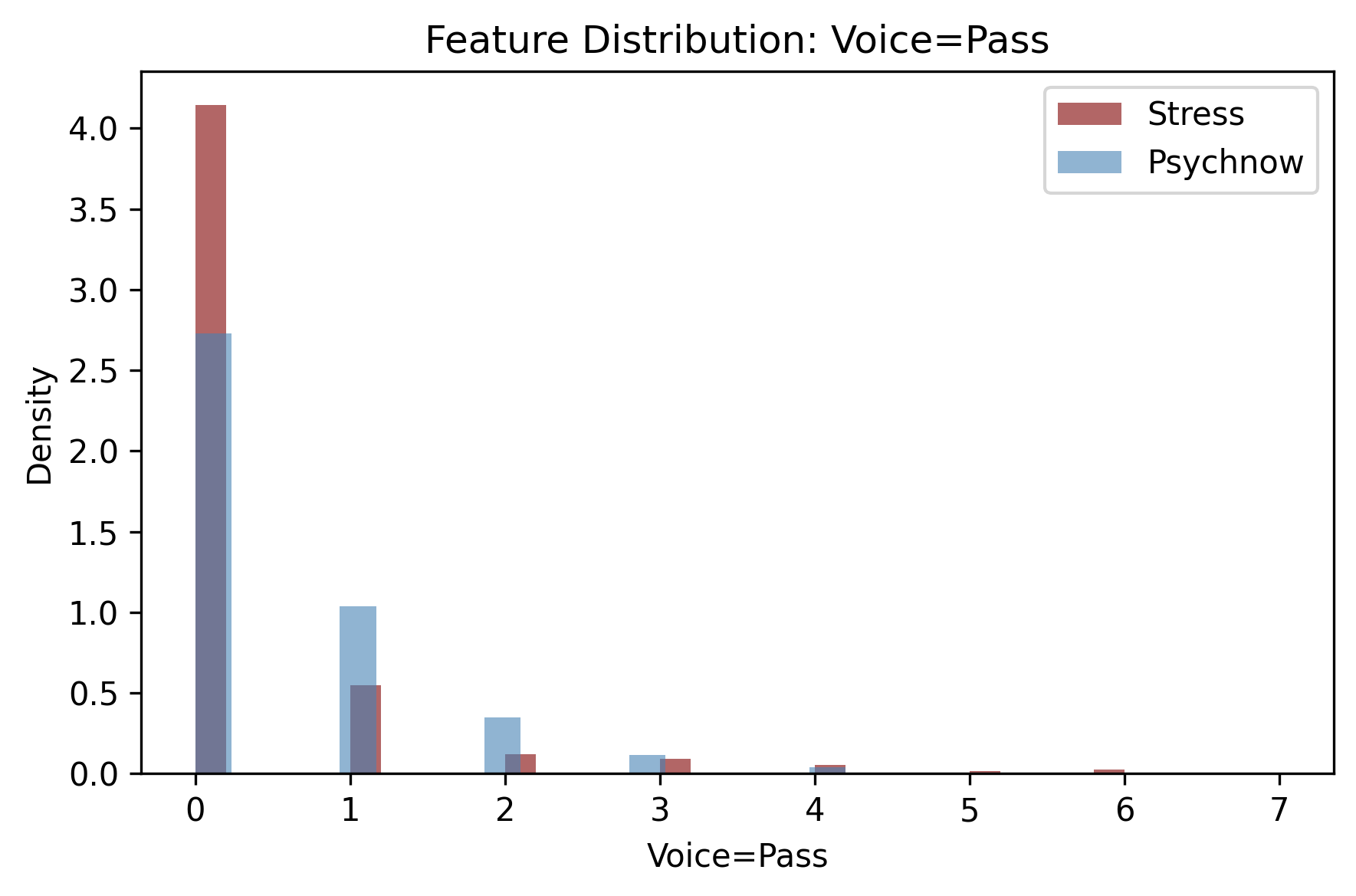} &
        \includegraphics[width=0.09\textwidth]{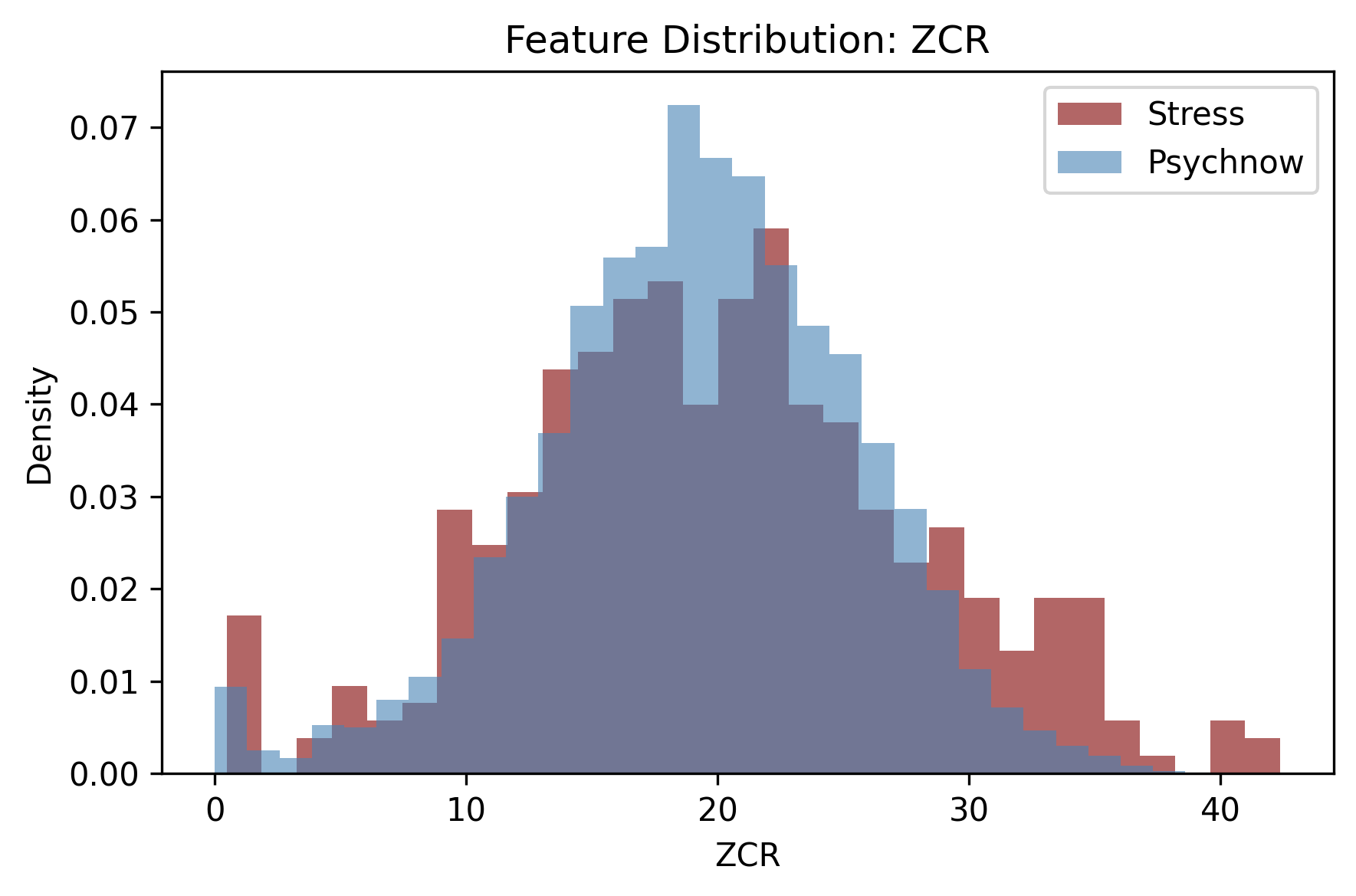} &
        \includegraphics[width=0.09\textwidth]{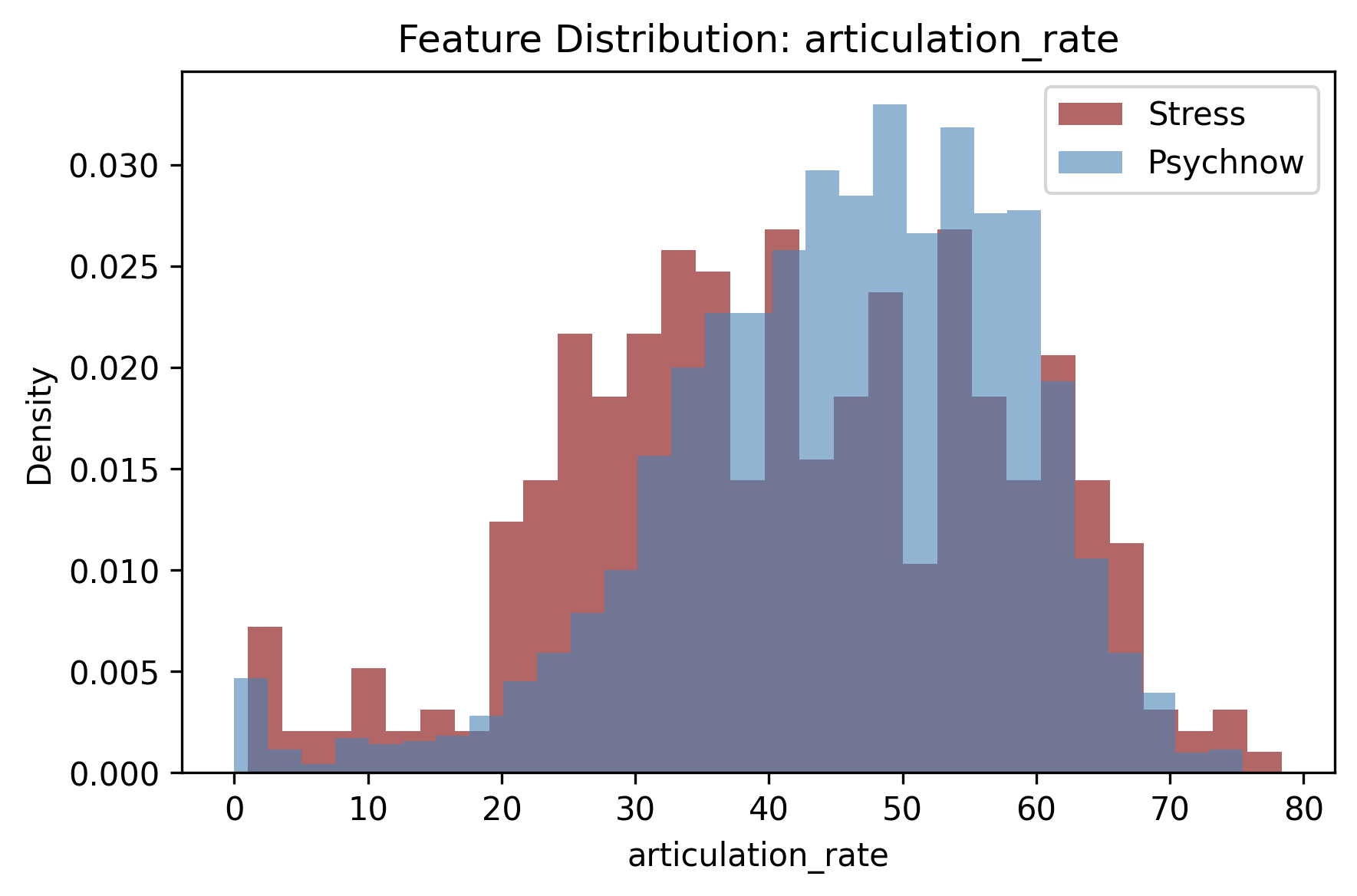} &
        \includegraphics[width=0.09\textwidth]{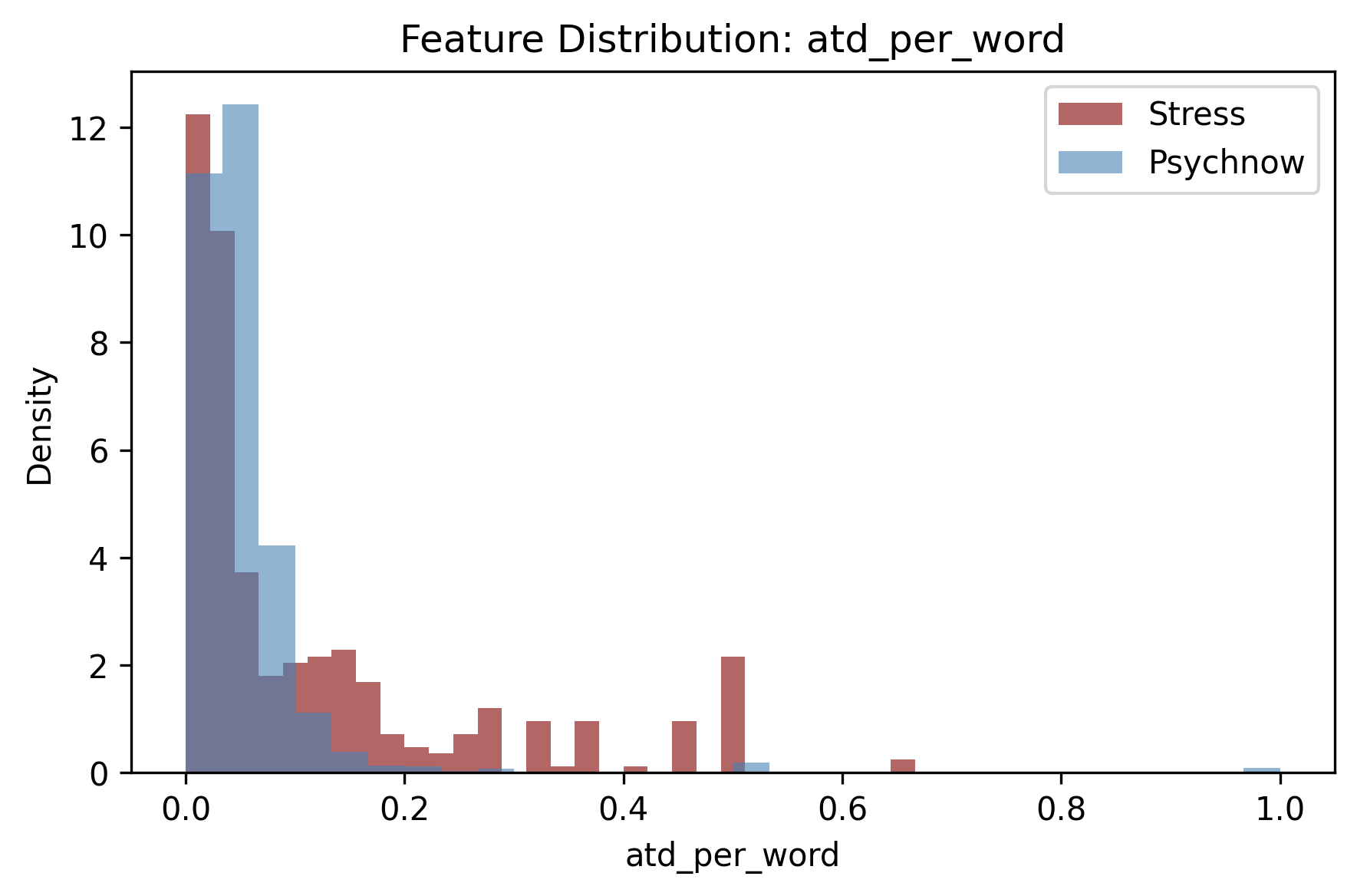} &
        \includegraphics[width=0.09\textwidth]{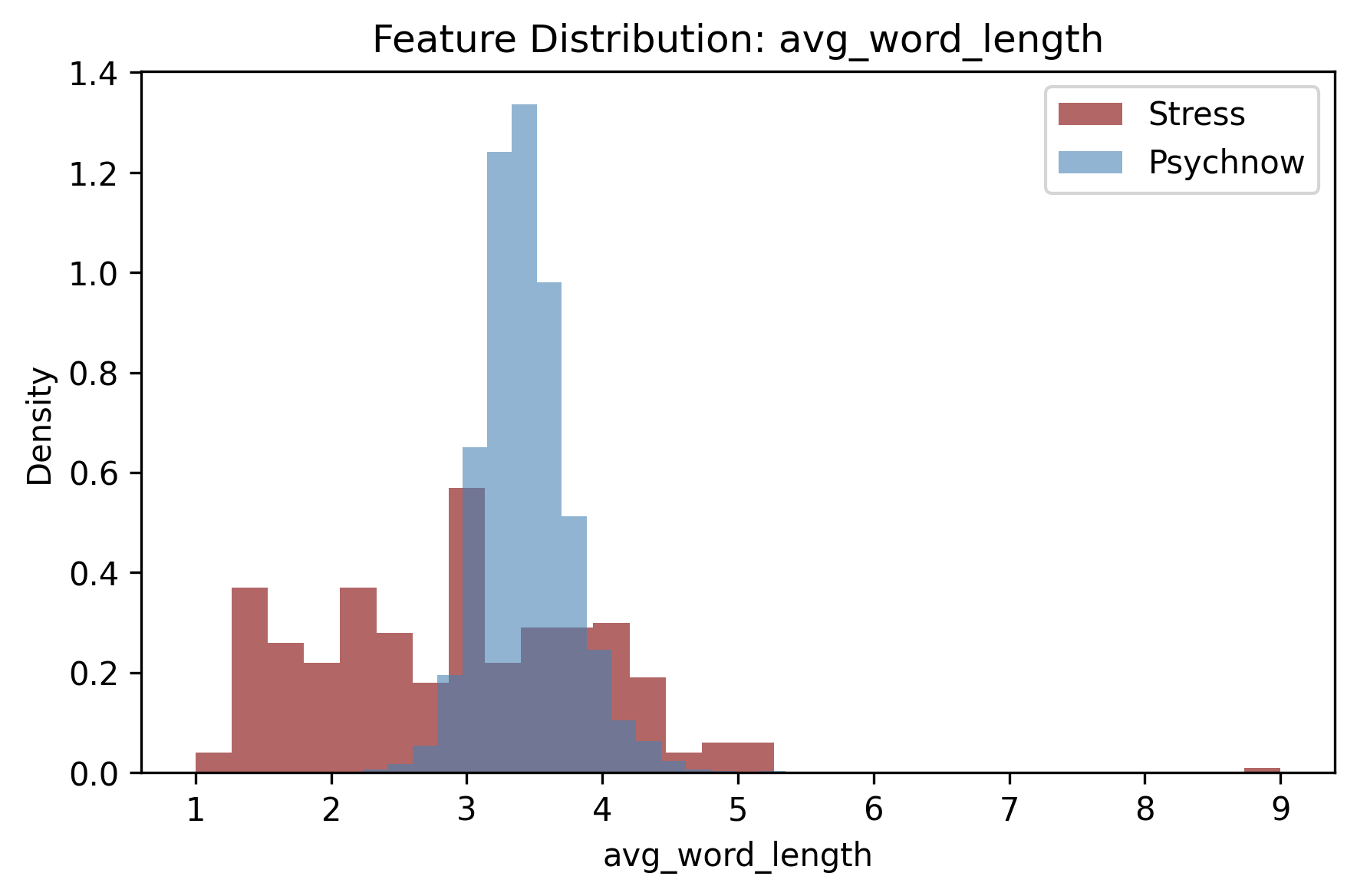} &
        \includegraphics[width=0.09\textwidth]{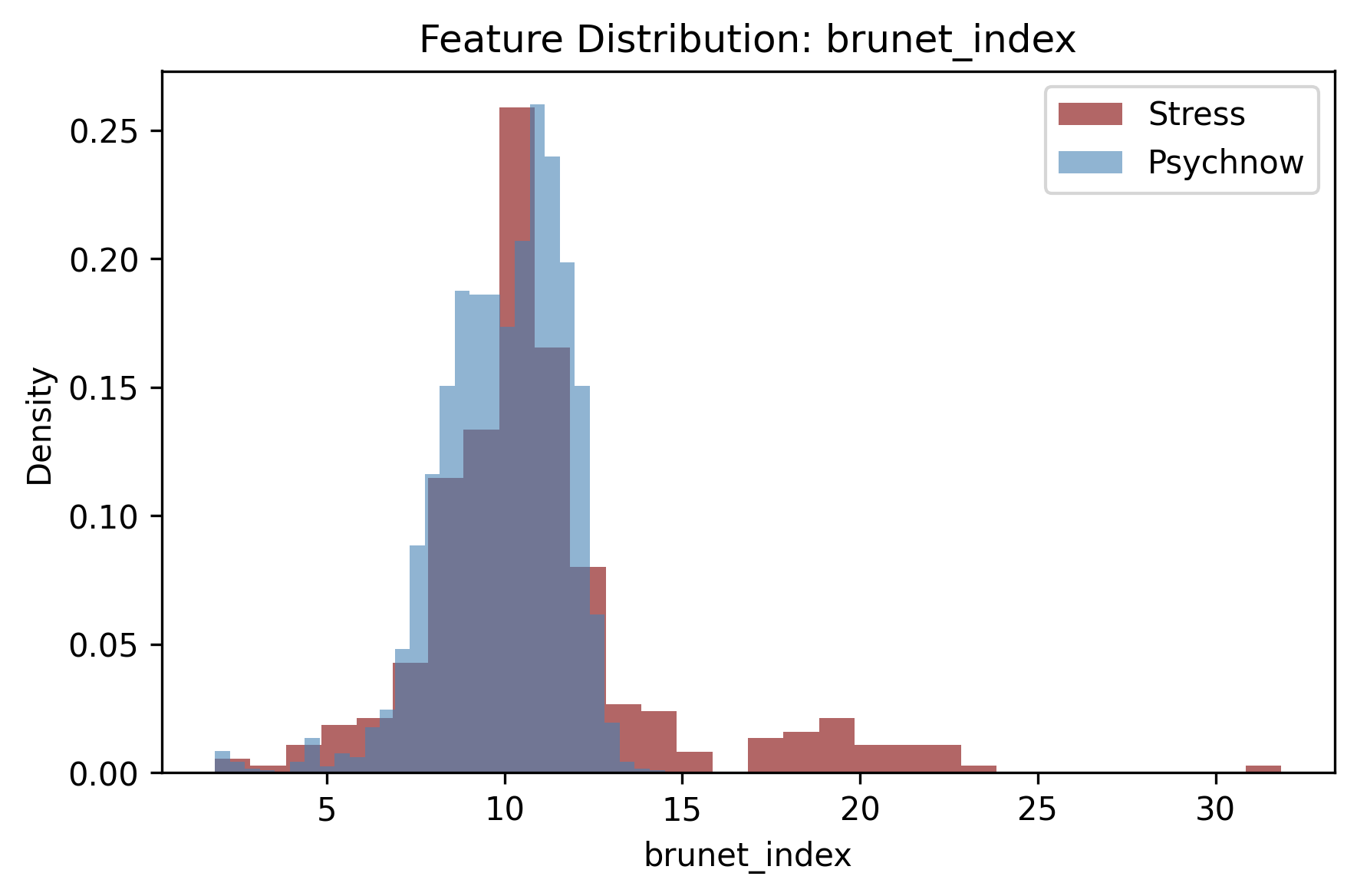} \\[-2pt]

        \includegraphics[width=0.09\textwidth]{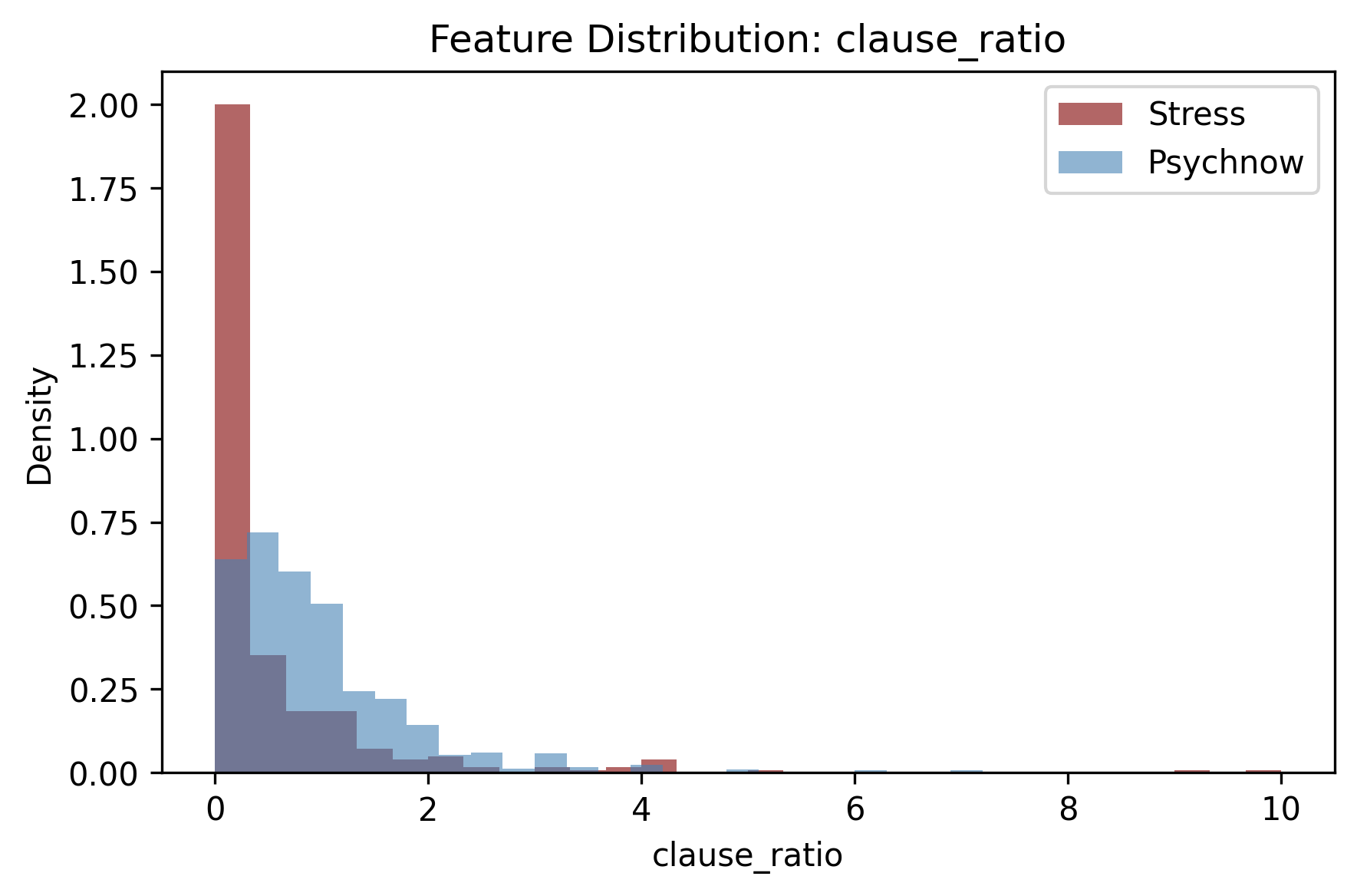} &
        \includegraphics[width=0.09\textwidth]{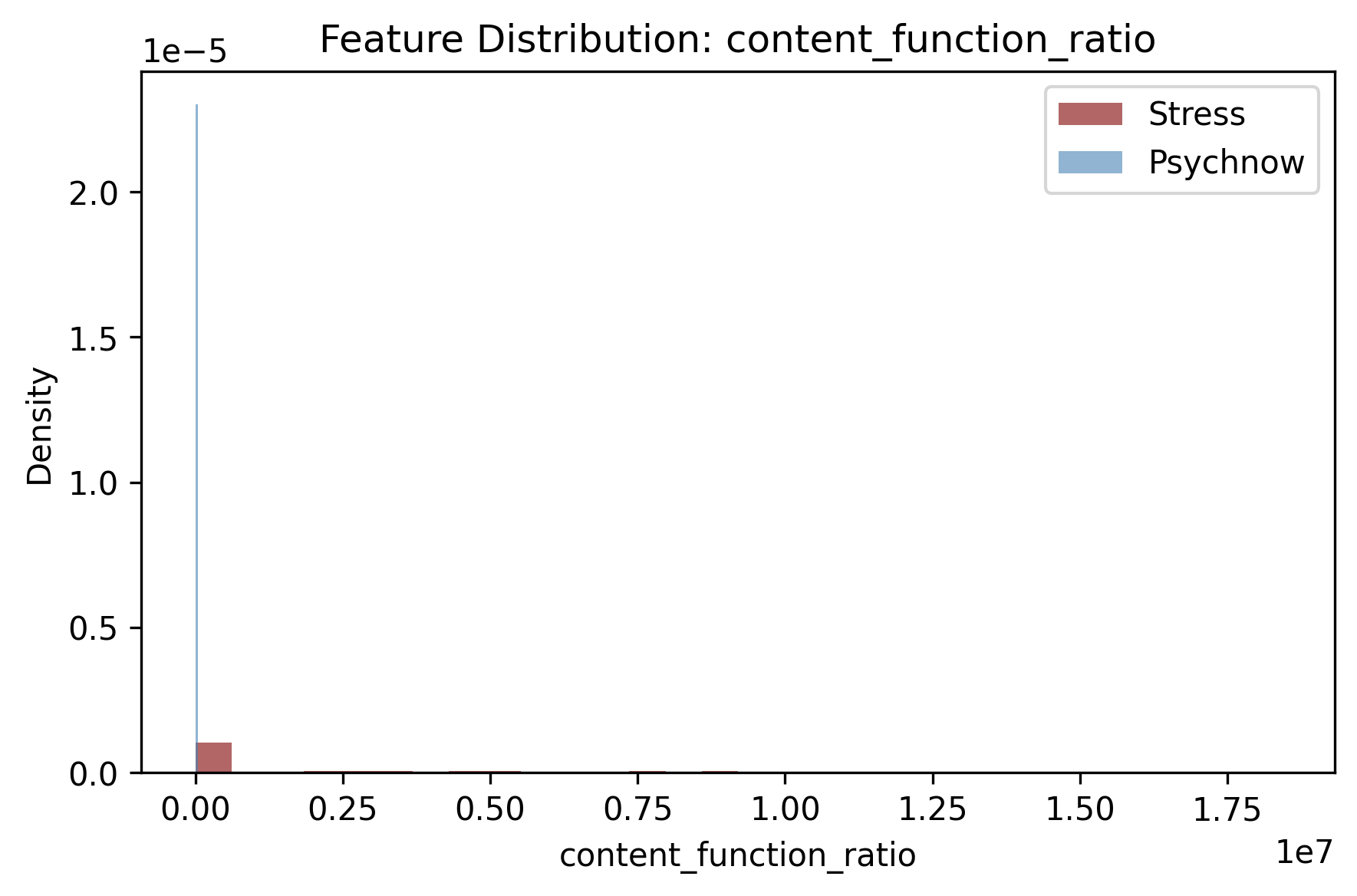} &
        \includegraphics[width=0.09\textwidth]{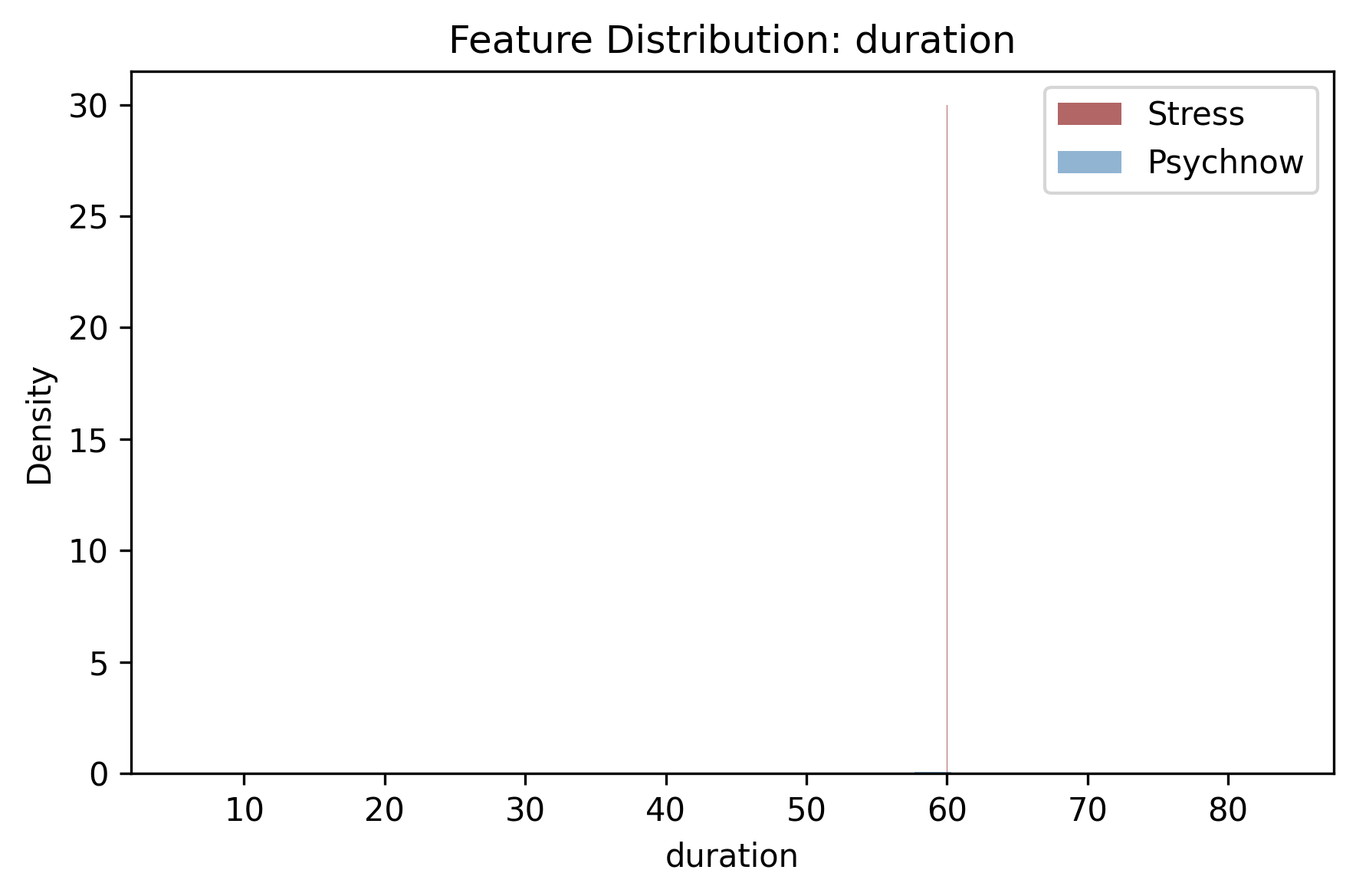} &
        \includegraphics[width=0.09\textwidth]{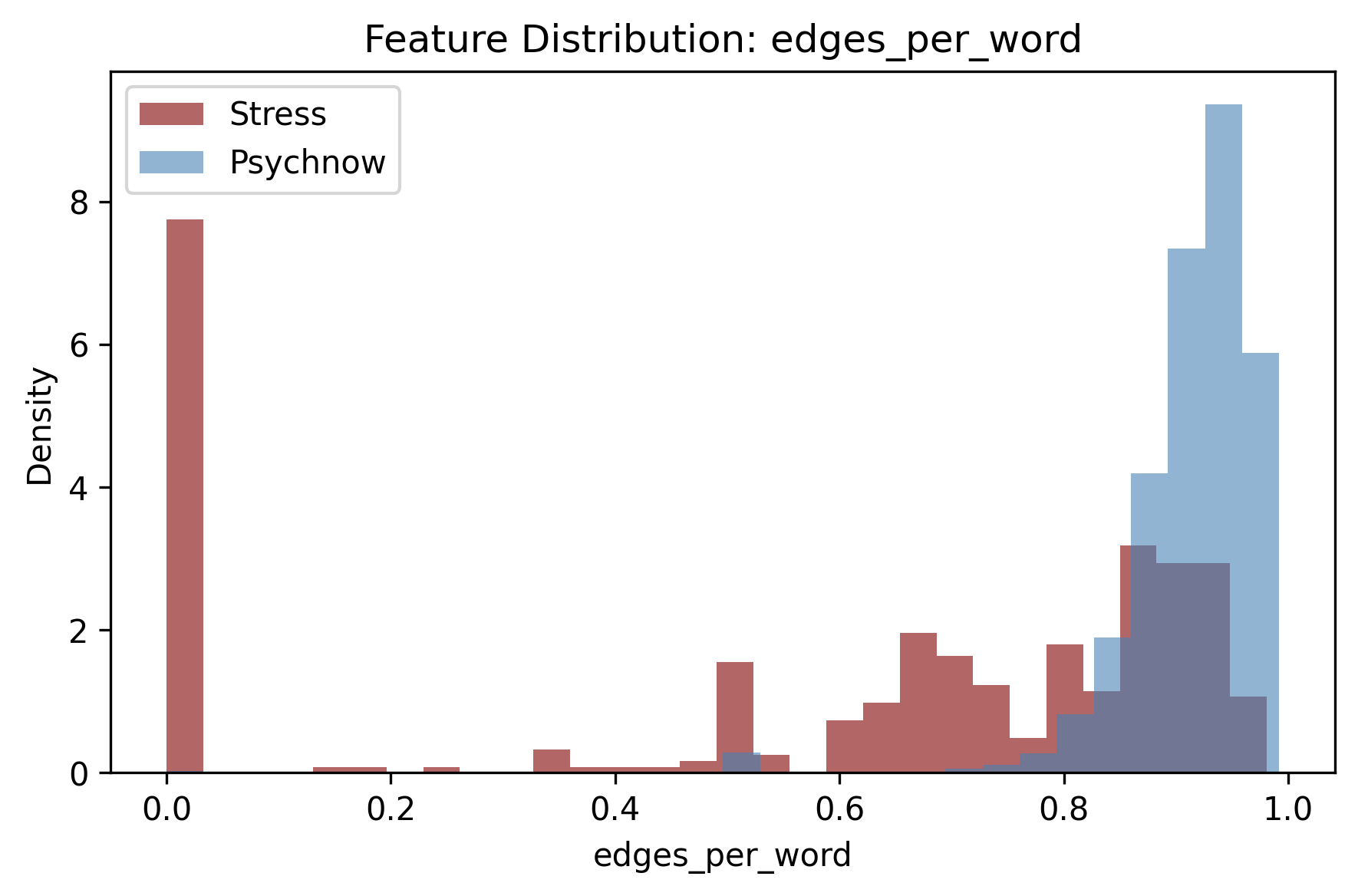} &
        \includegraphics[width=0.09\textwidth]{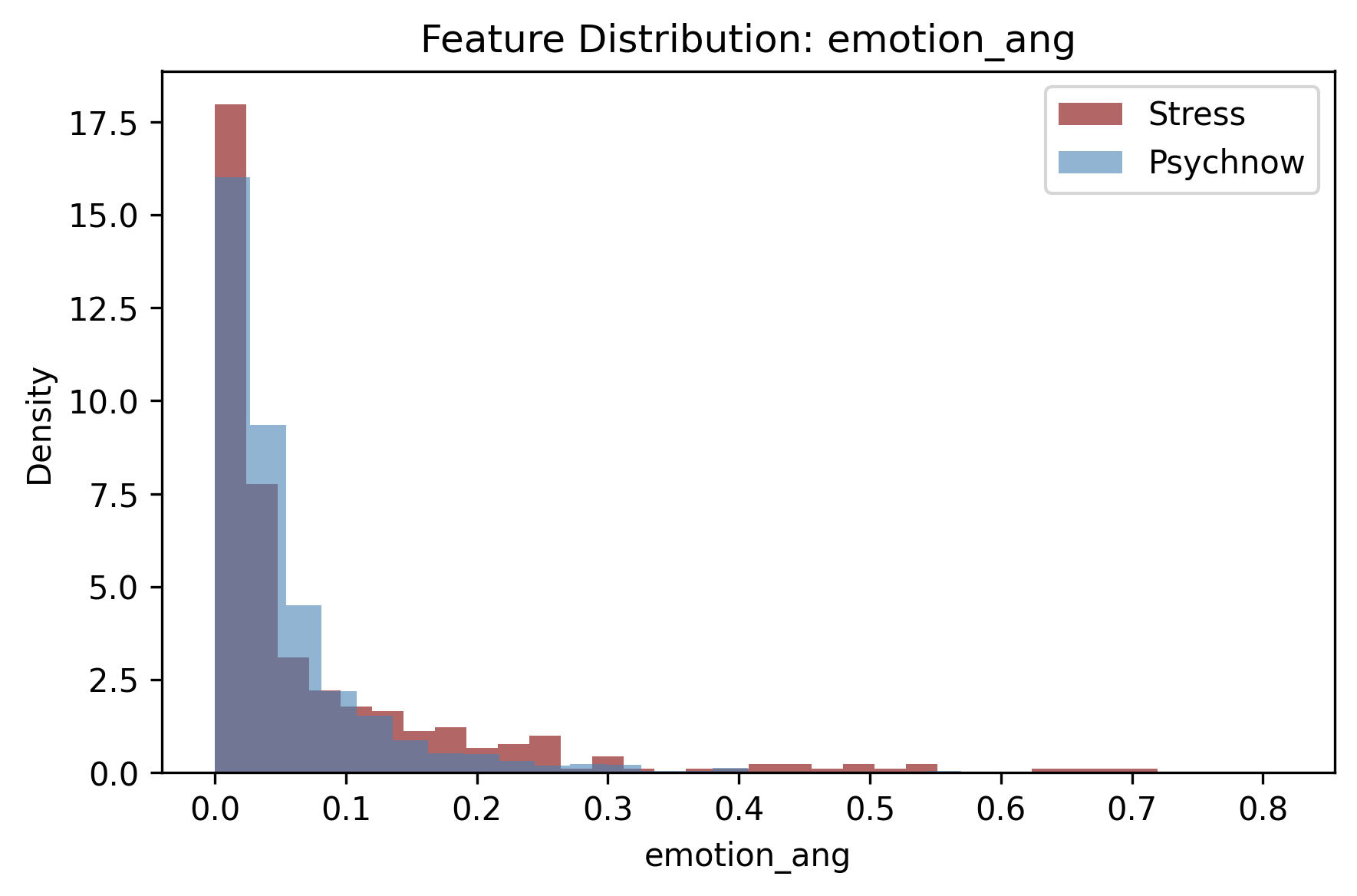} &
        \includegraphics[width=0.09\textwidth]{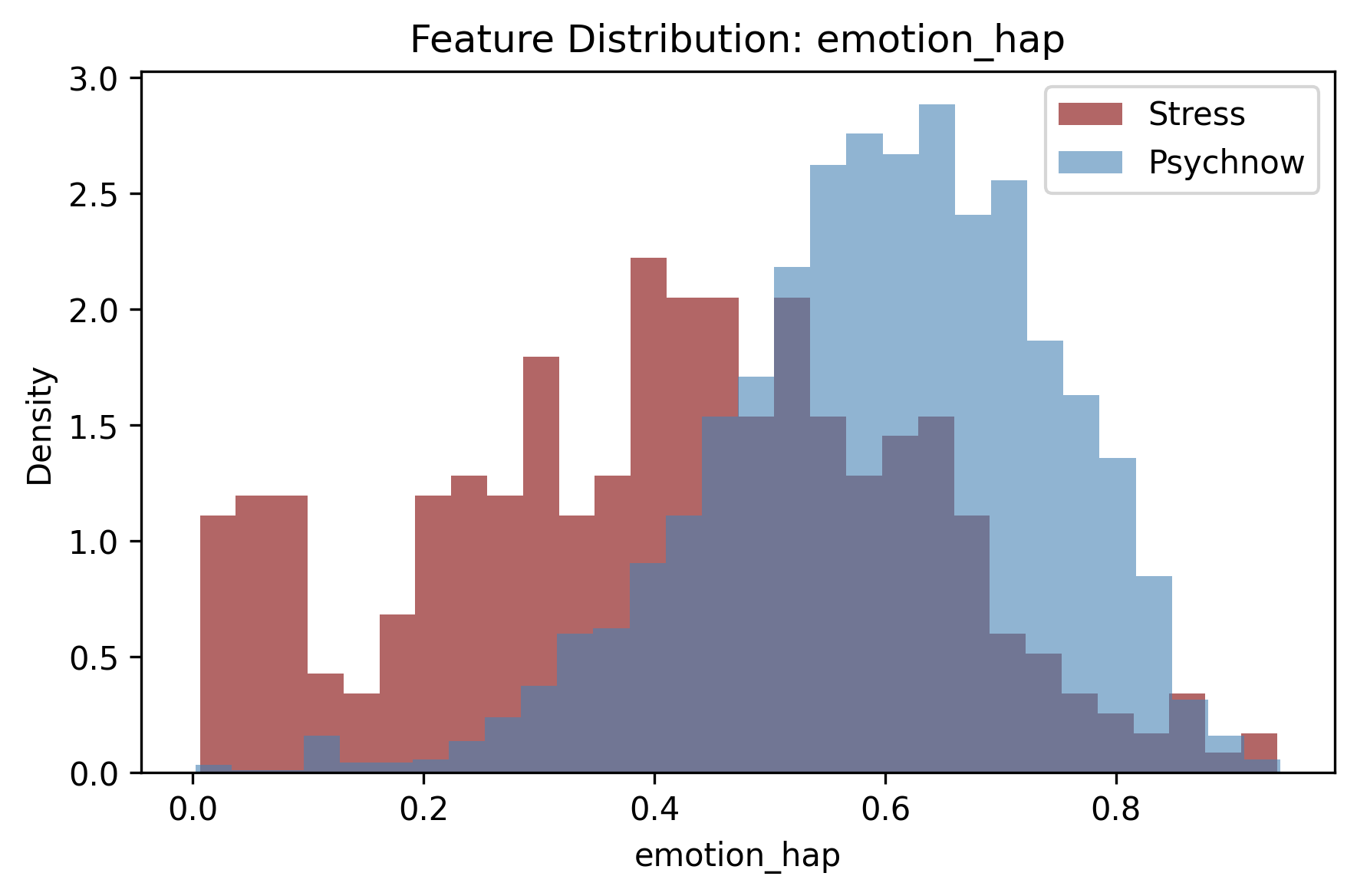} &
        \includegraphics[width=0.09\textwidth]{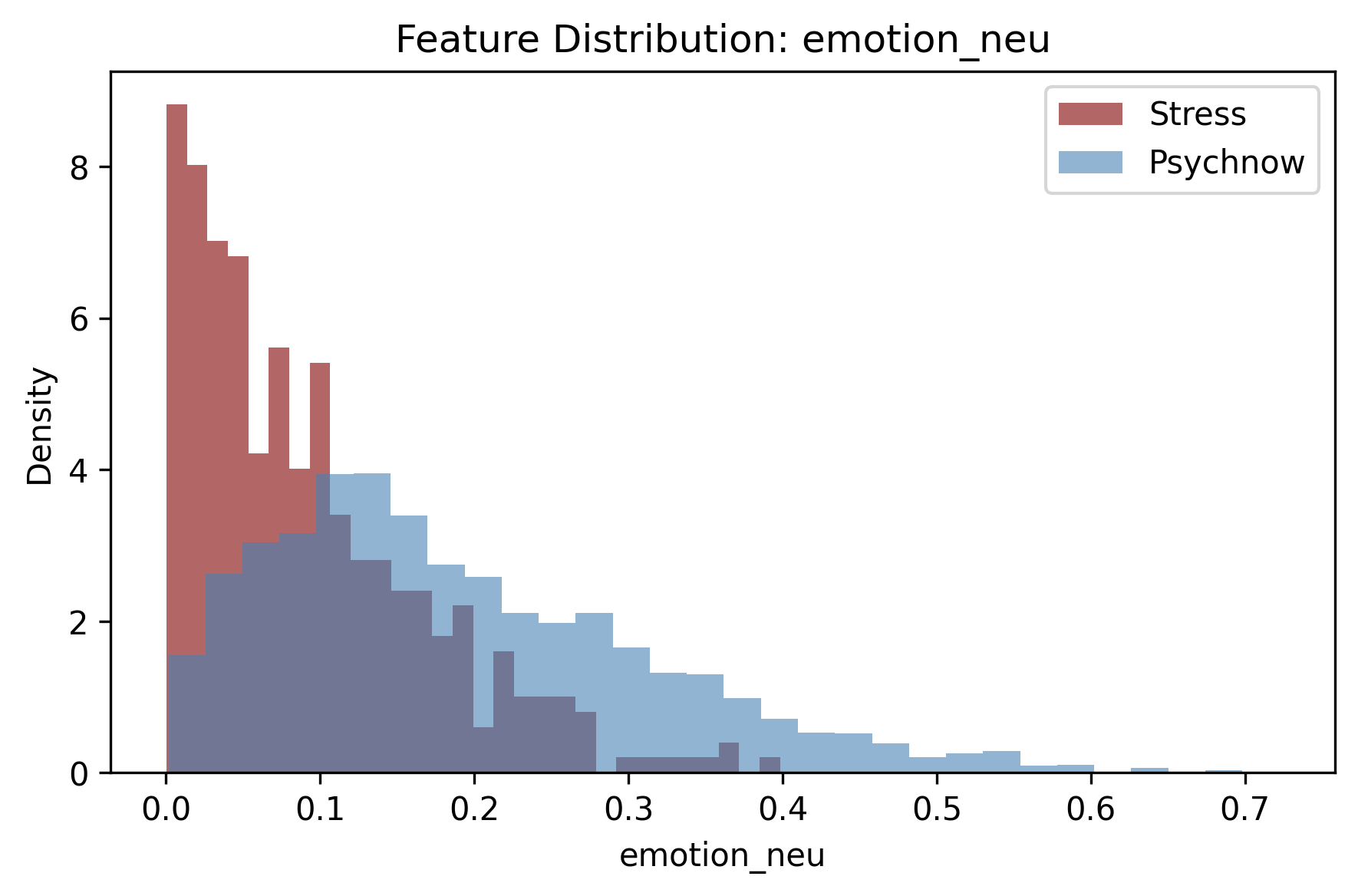} &
        \includegraphics[width=0.09\textwidth]{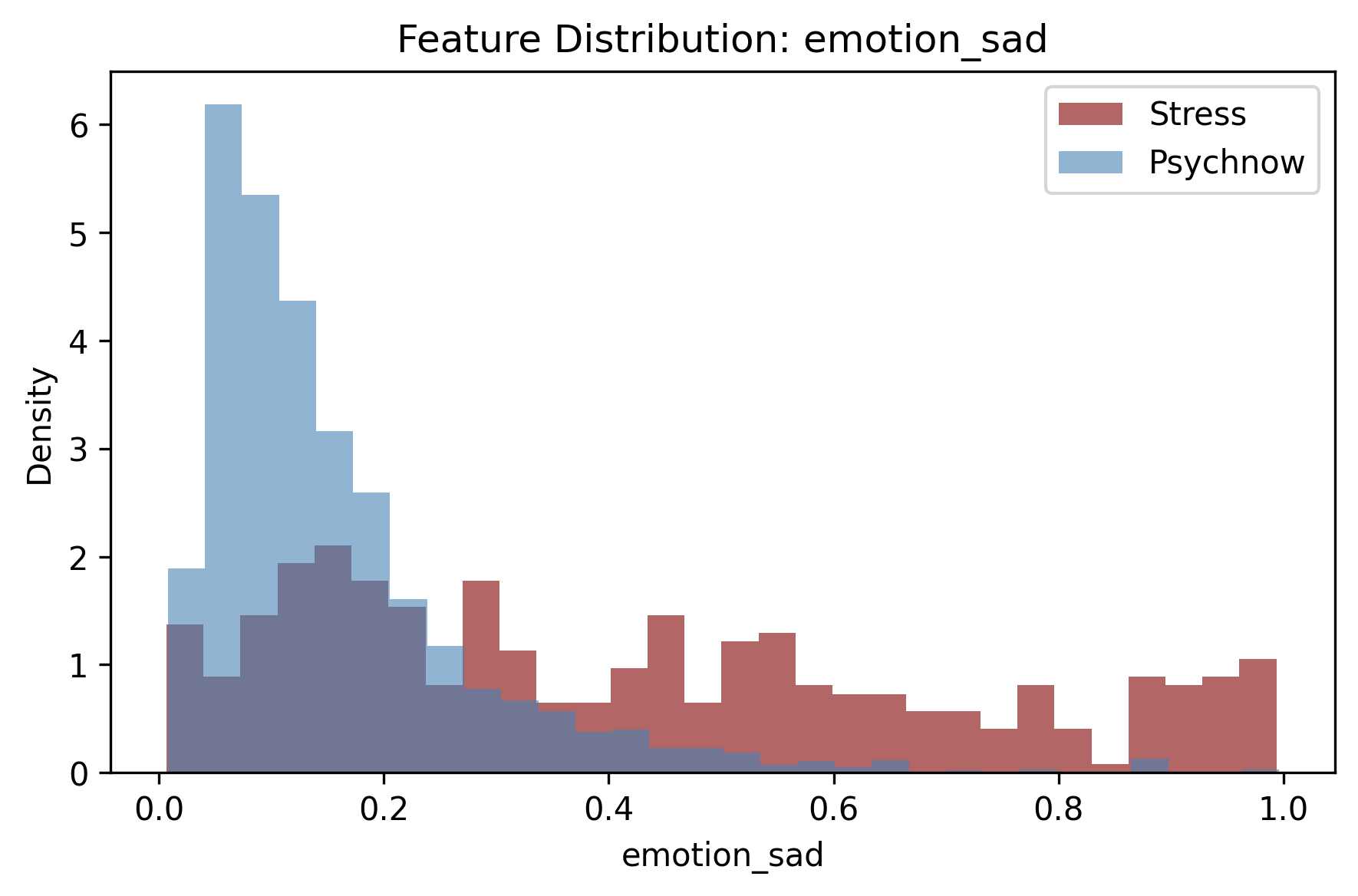} &
        \includegraphics[width=0.09\textwidth]{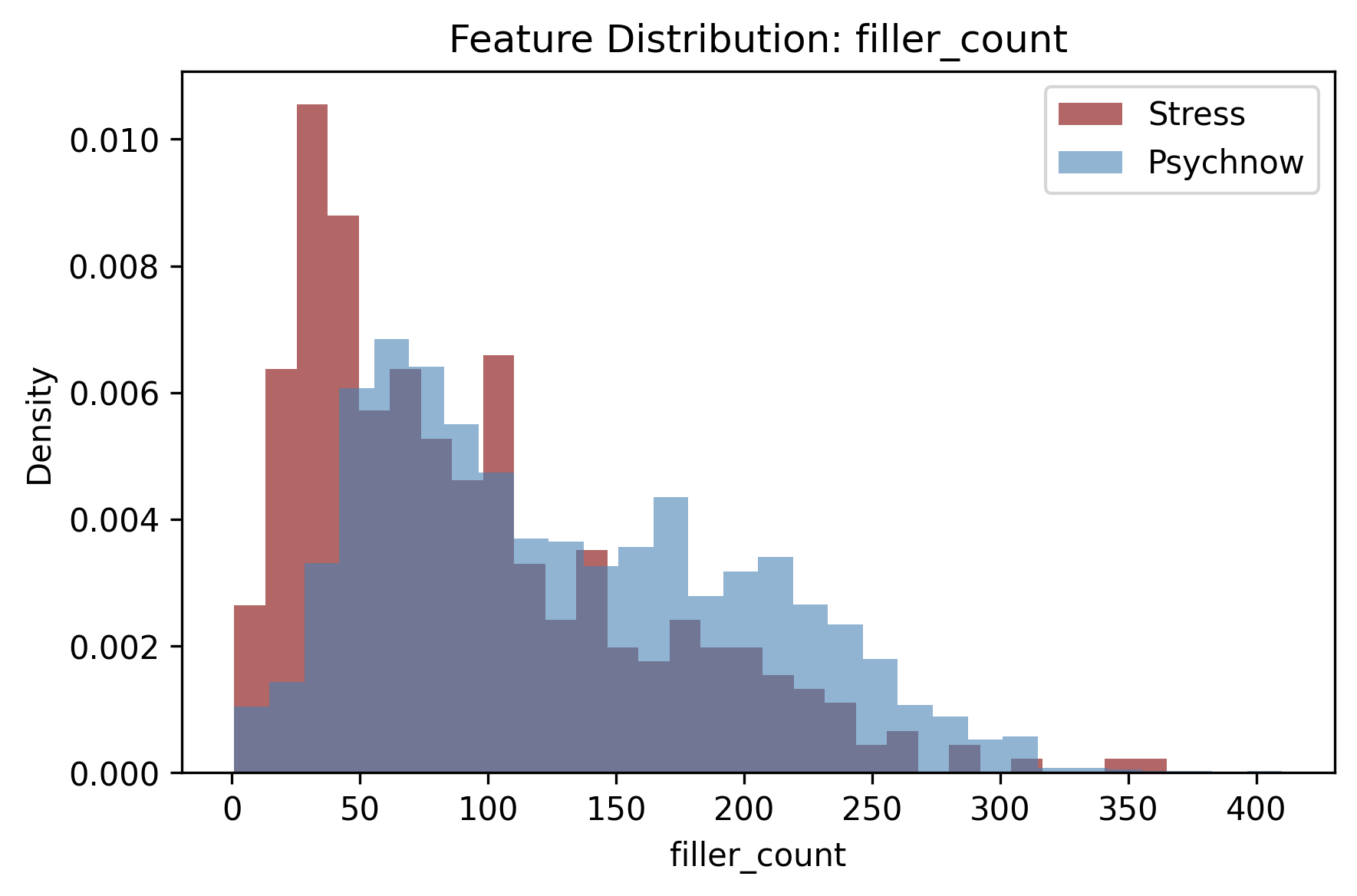} &
        \includegraphics[width=0.09\textwidth]{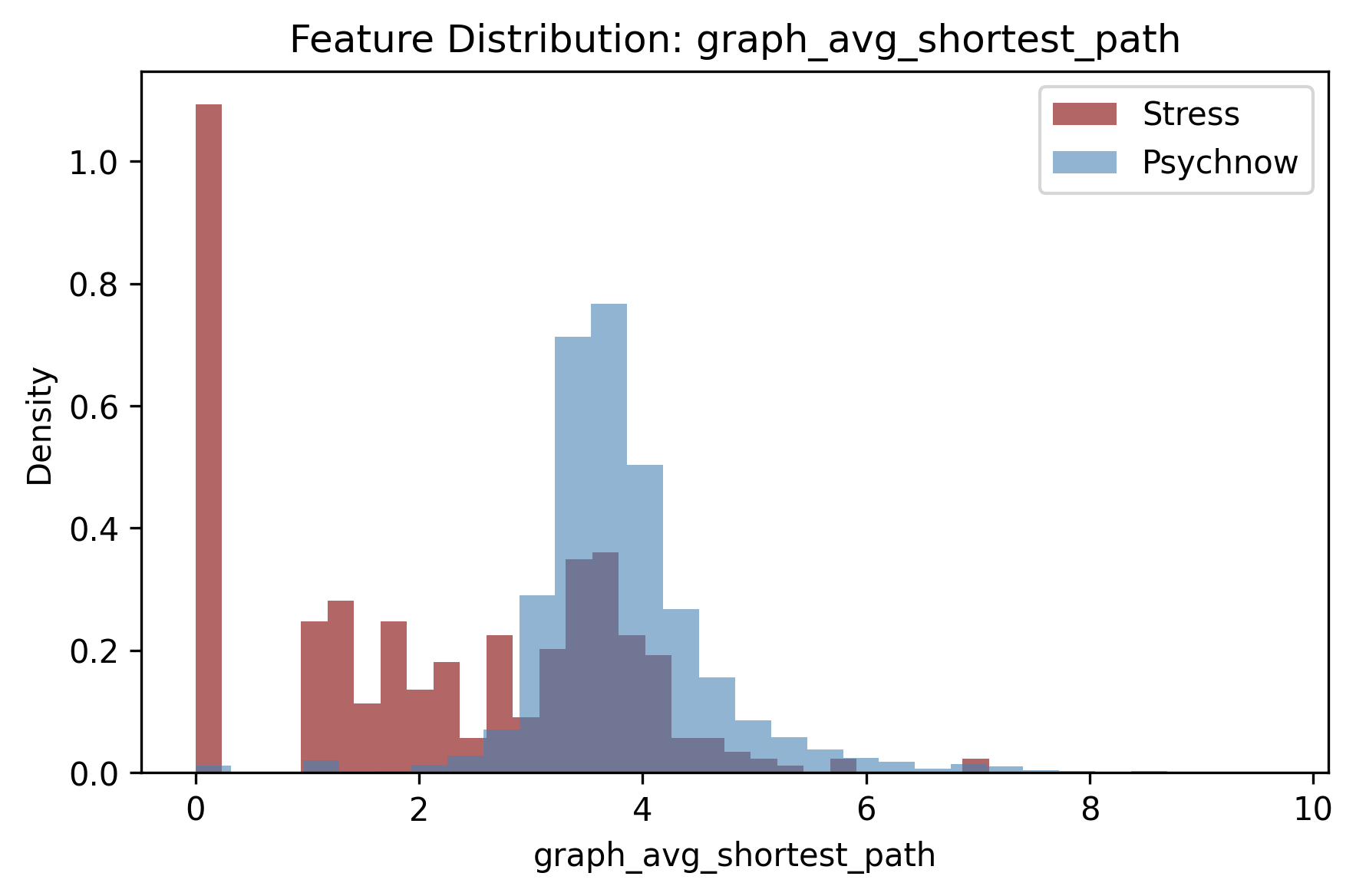} \\[-2pt]

        \includegraphics[width=0.09\textwidth]{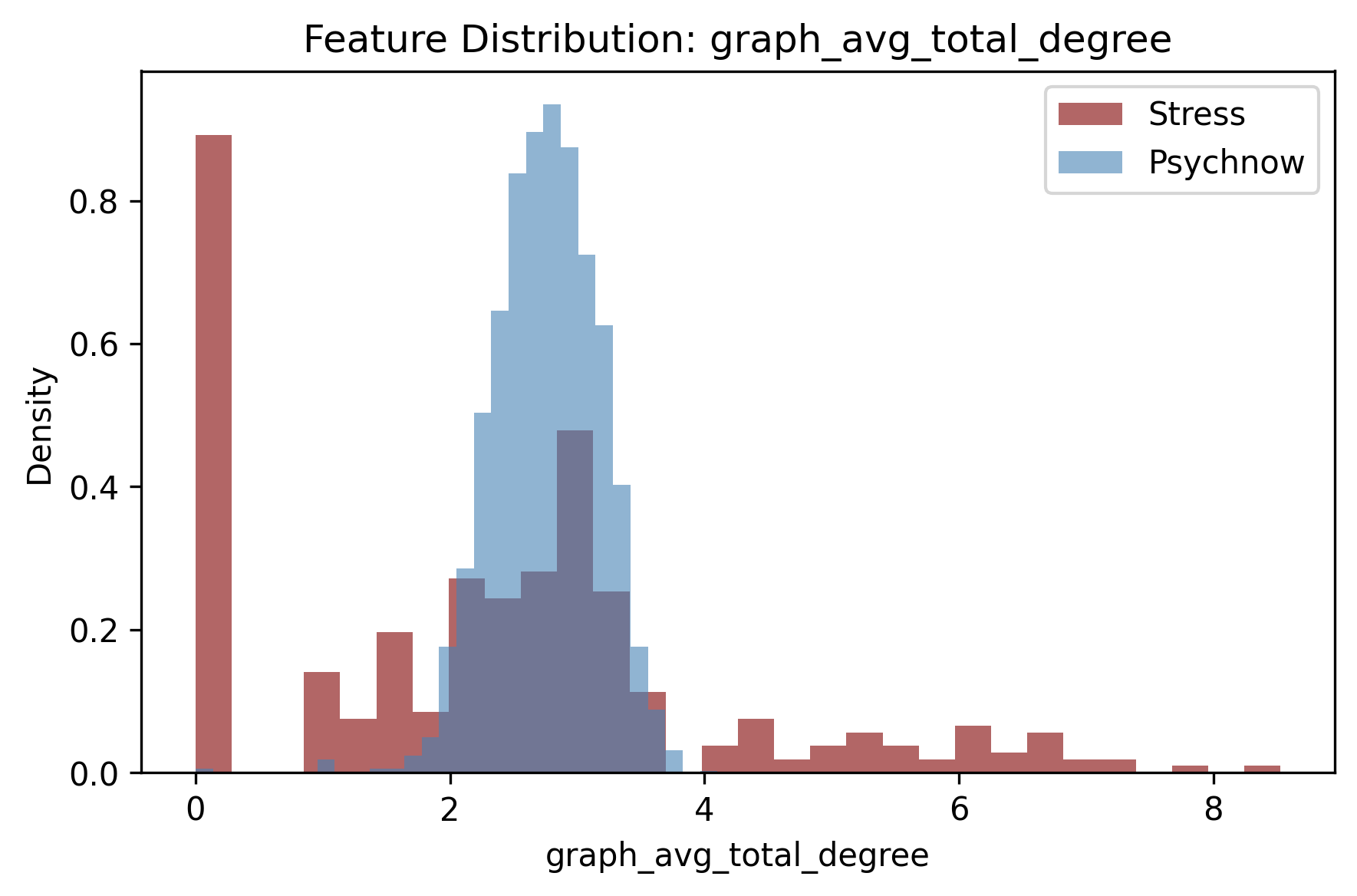} &
        \includegraphics[width=0.09\textwidth]{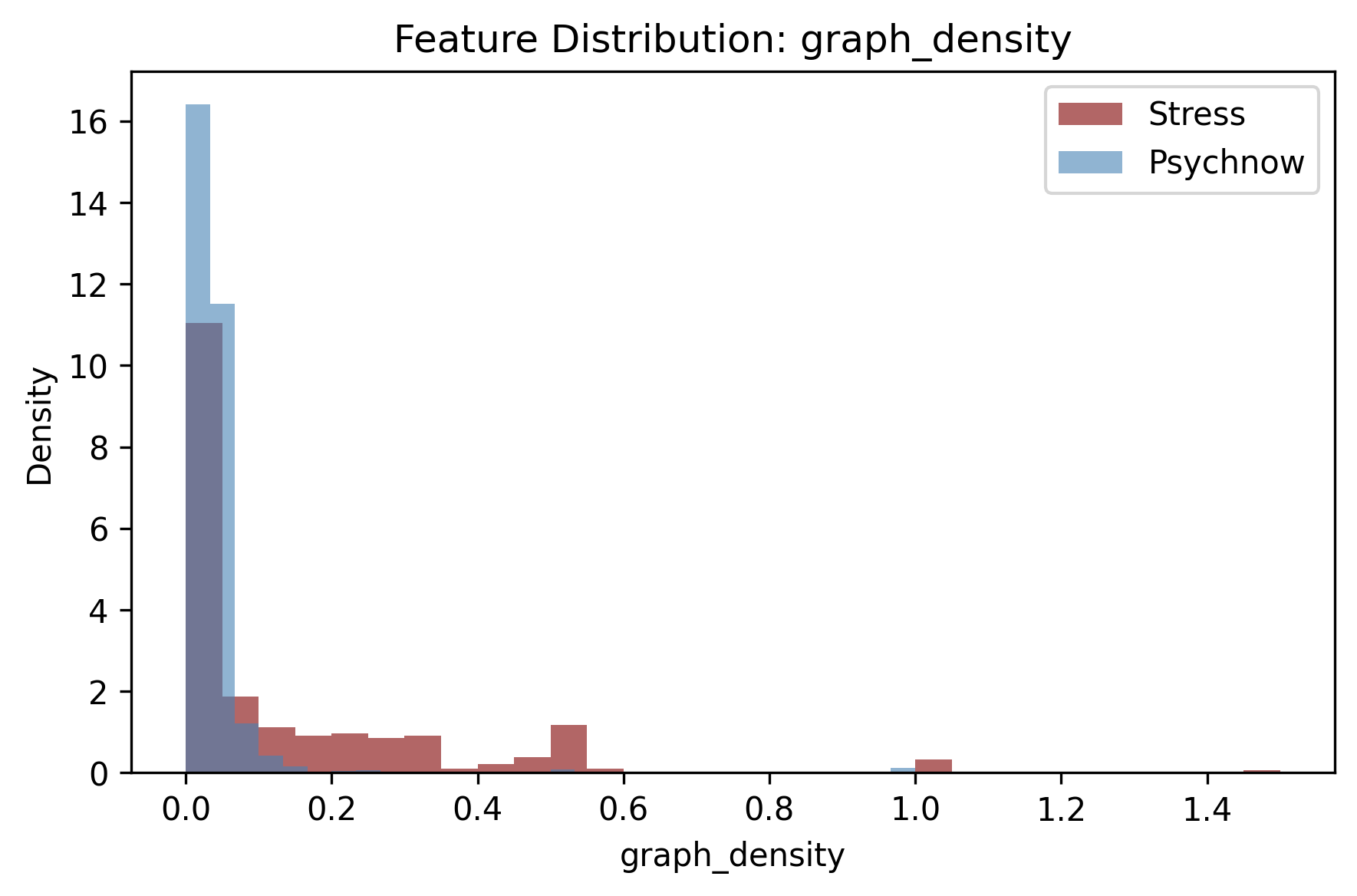} &
        \includegraphics[width=0.09\textwidth]{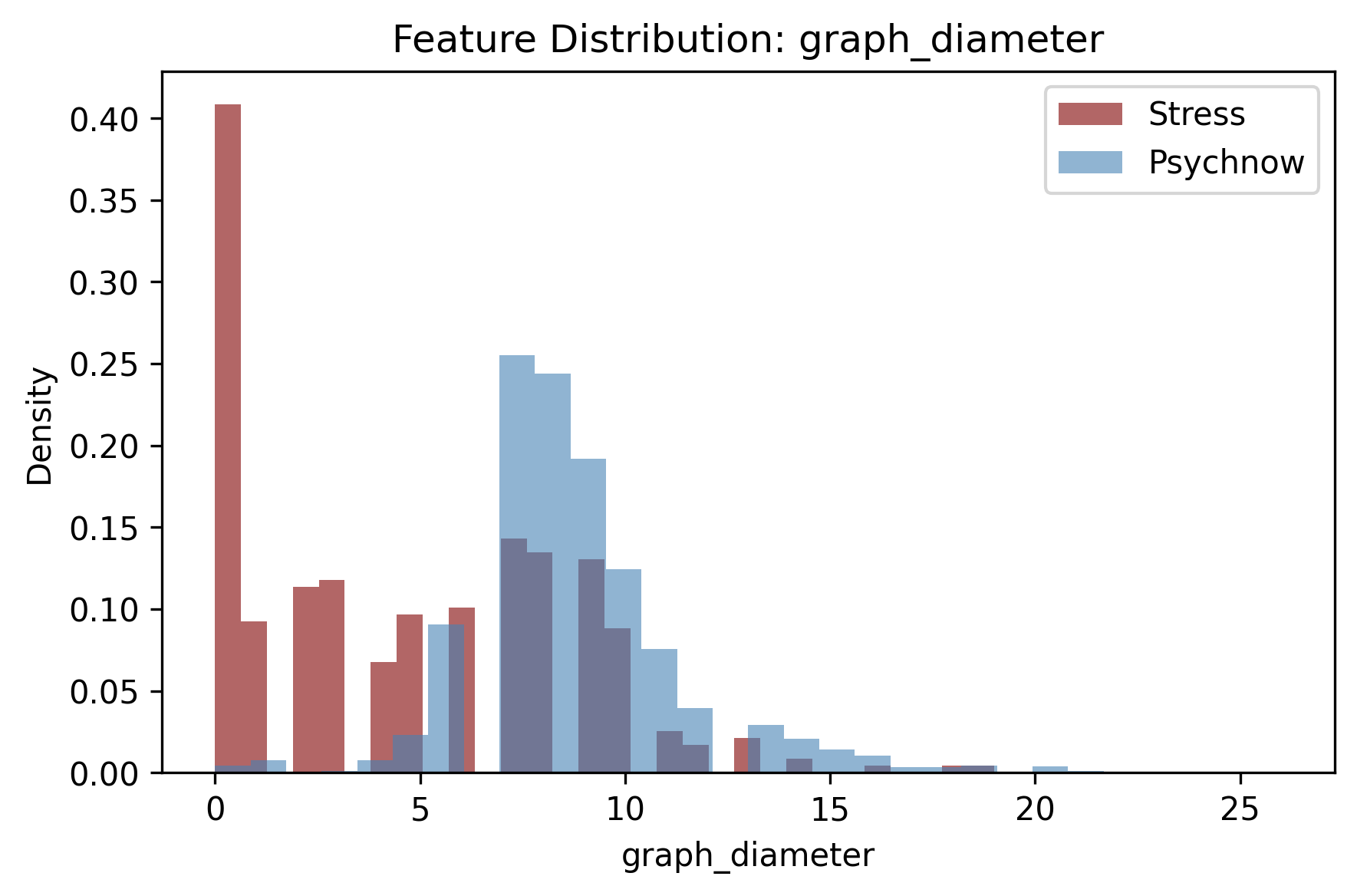} &
        \includegraphics[width=0.09\textwidth]{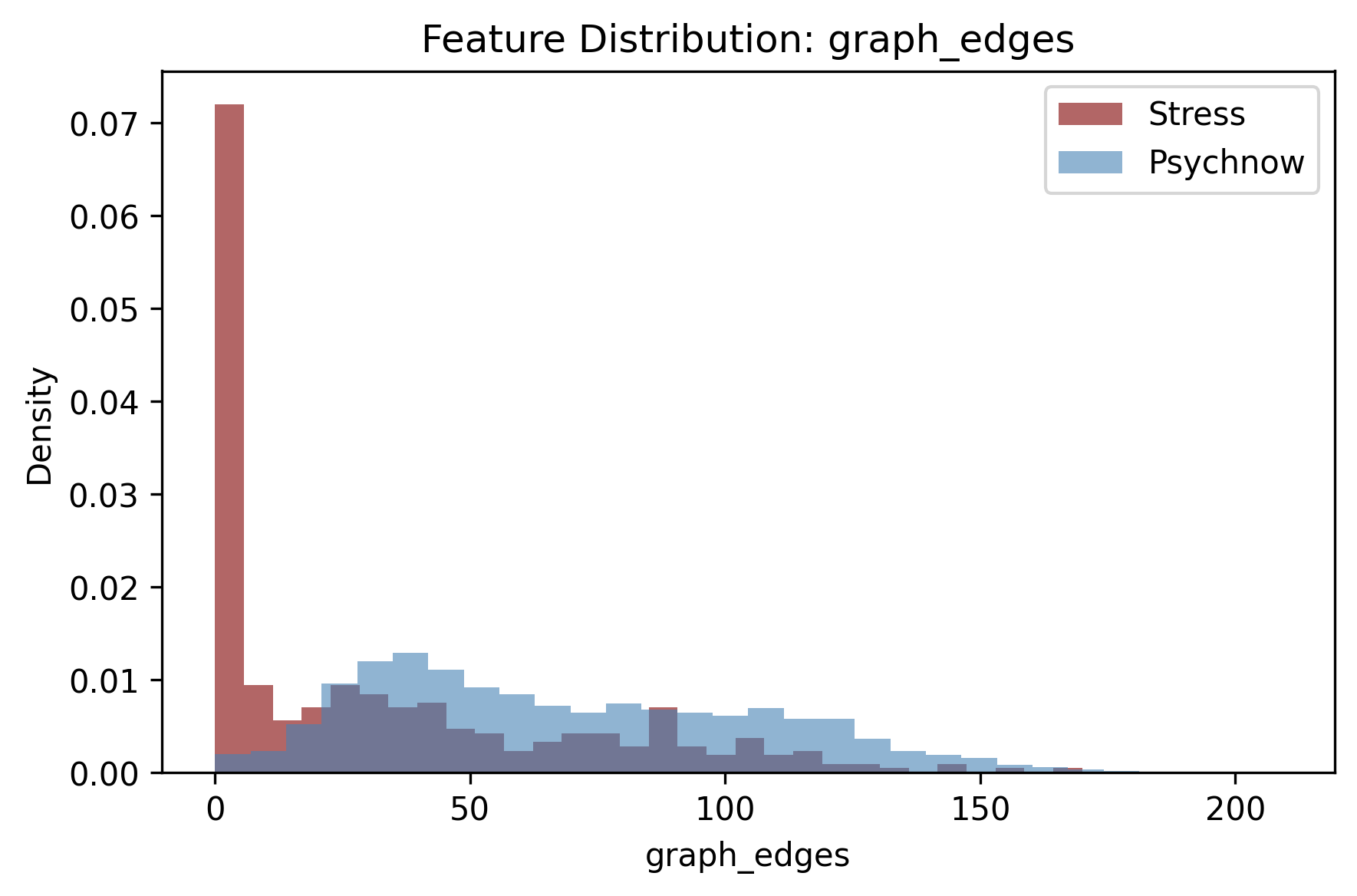} &
        \includegraphics[width=0.09\textwidth]{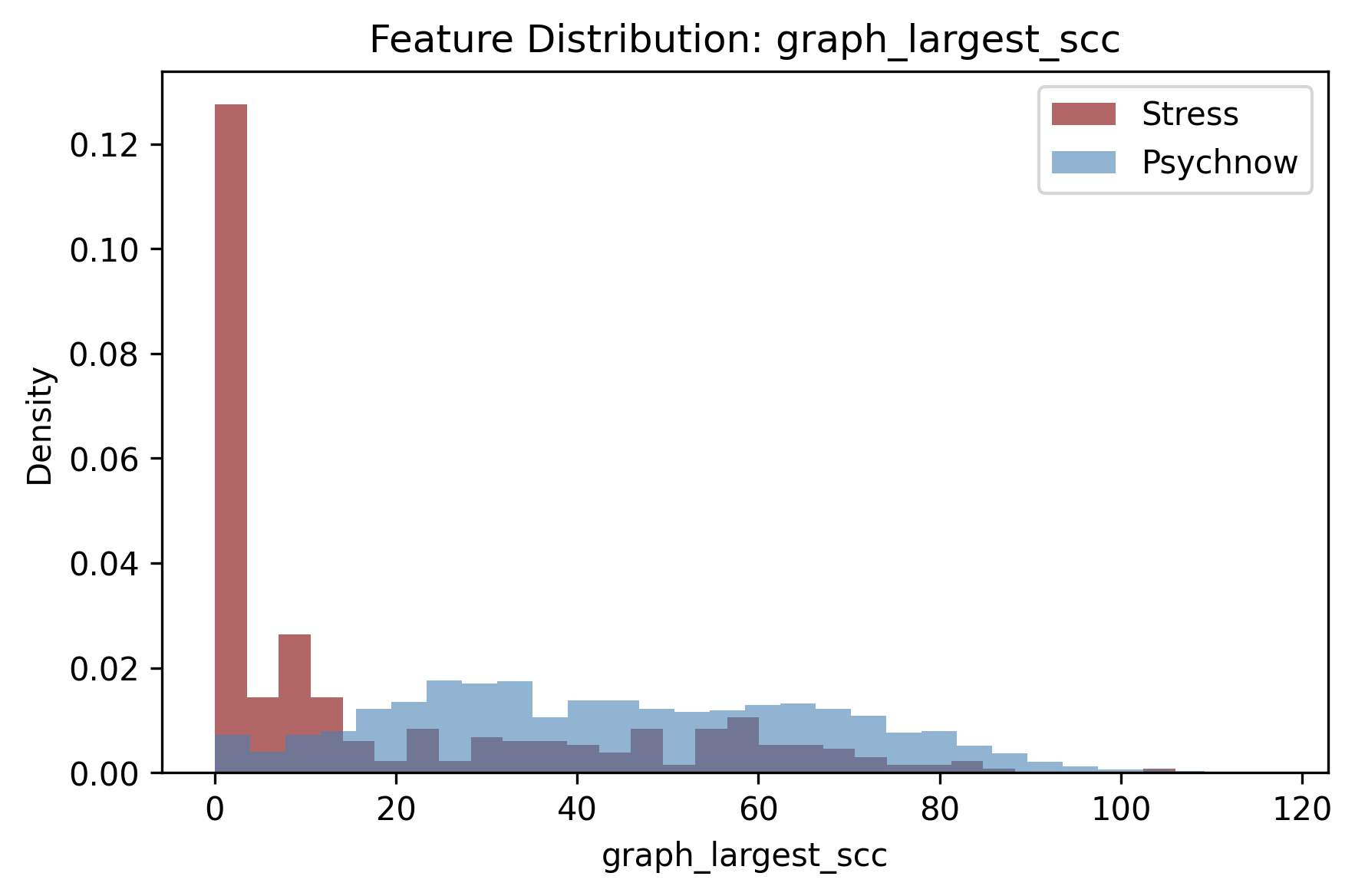} &
        \includegraphics[width=0.09\textwidth]{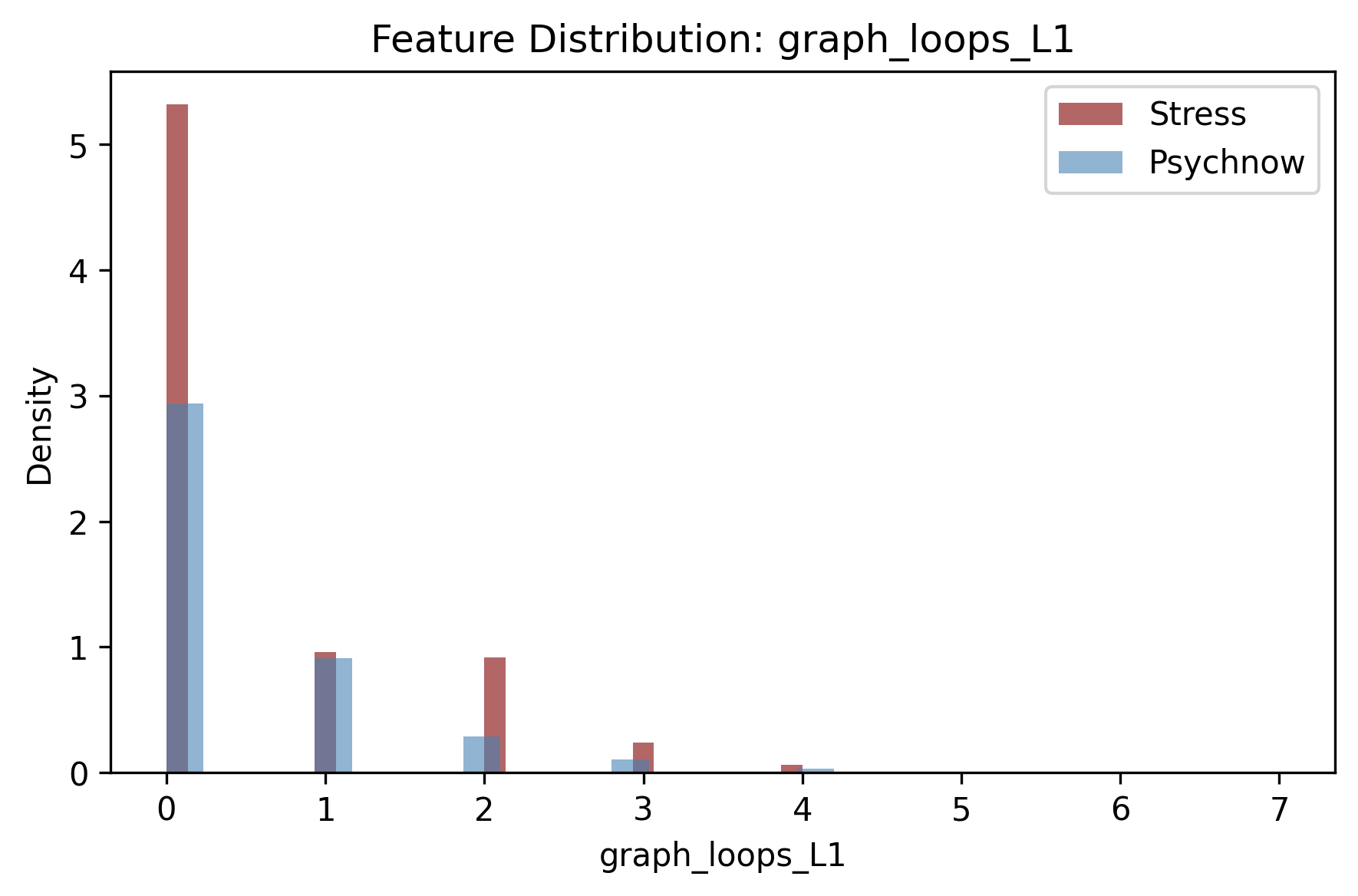} &
        \includegraphics[width=0.09\textwidth]{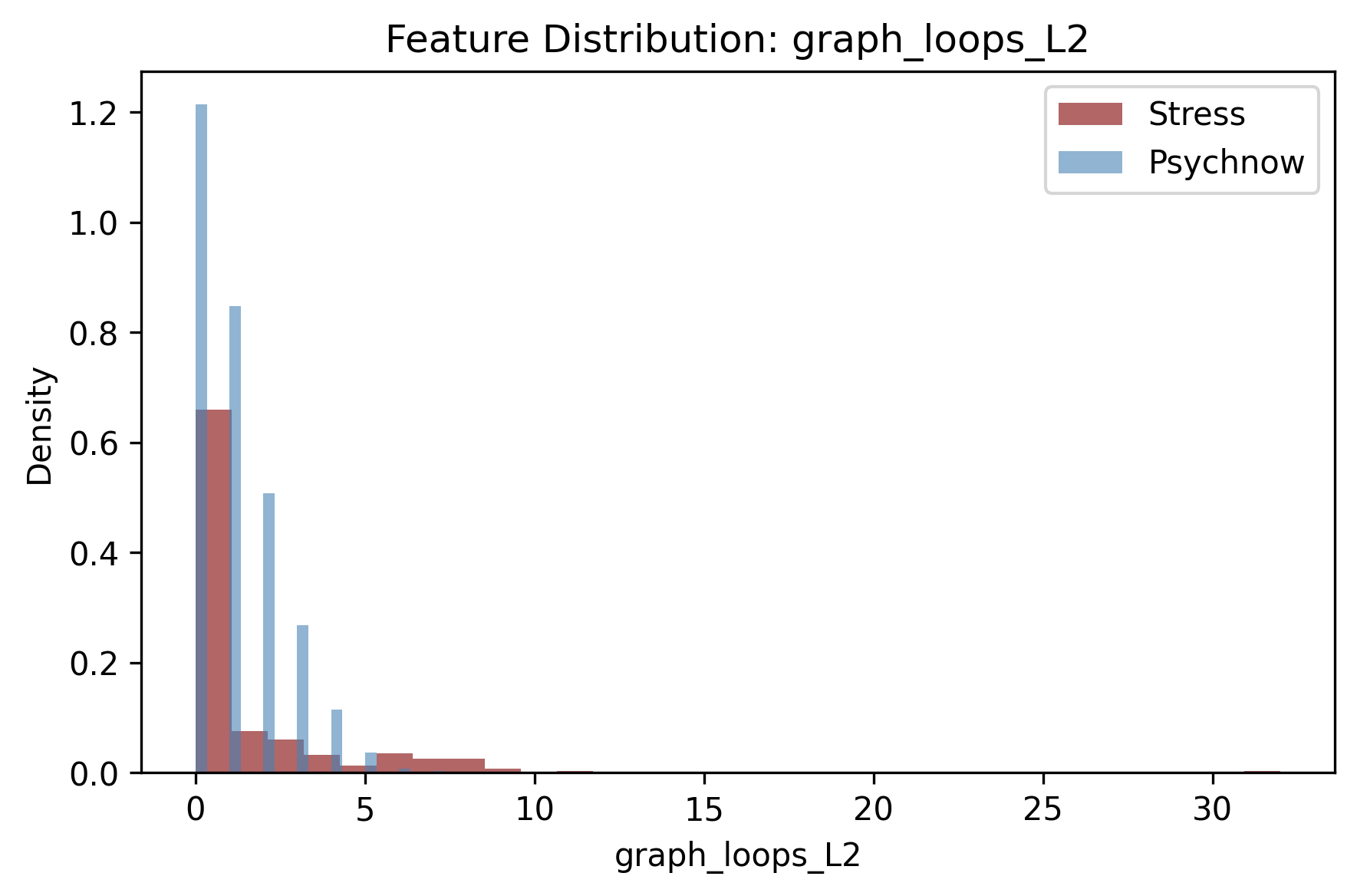} &
        \includegraphics[width=0.09\textwidth]{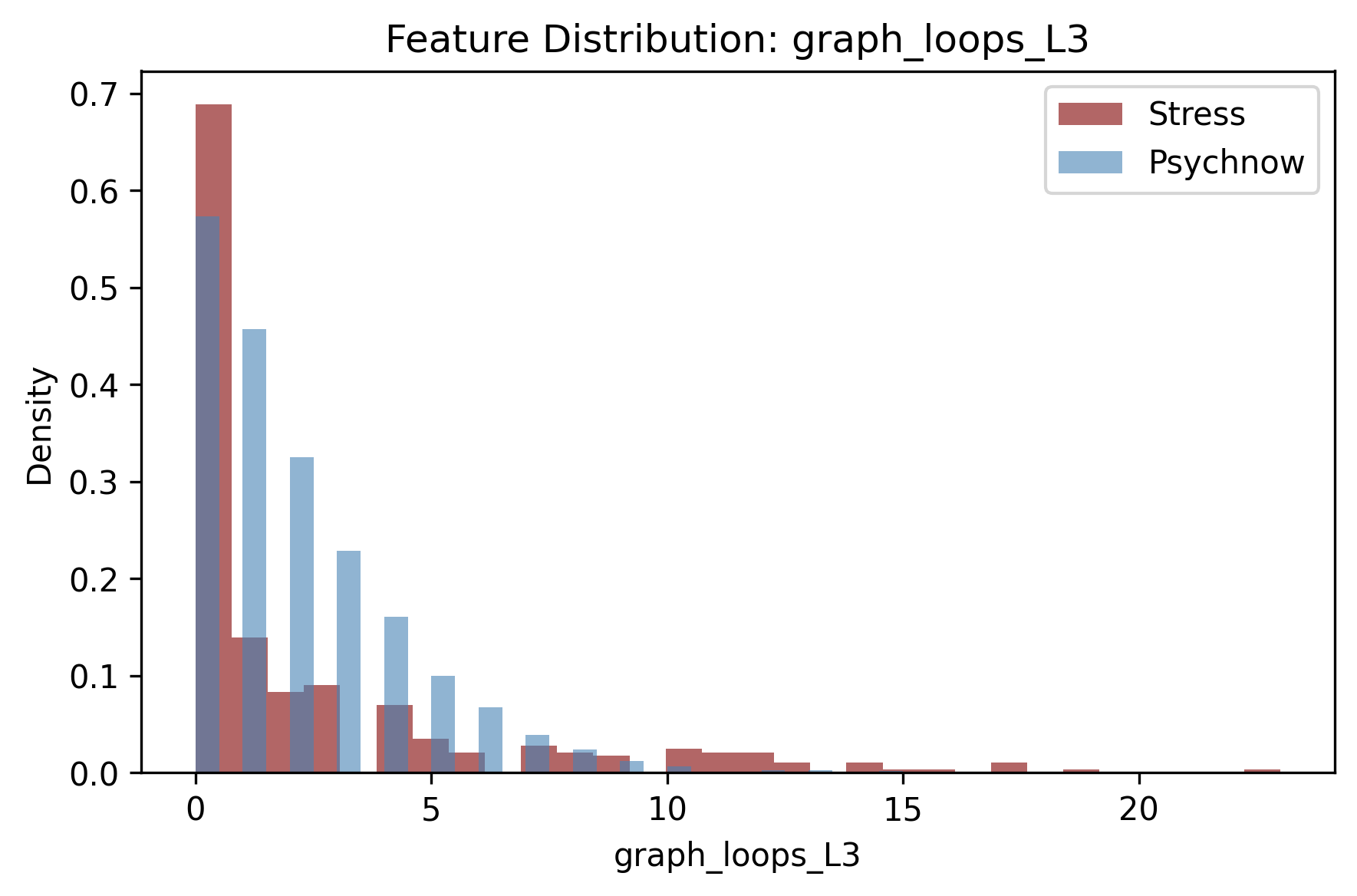} &
        \includegraphics[width=0.09\textwidth]{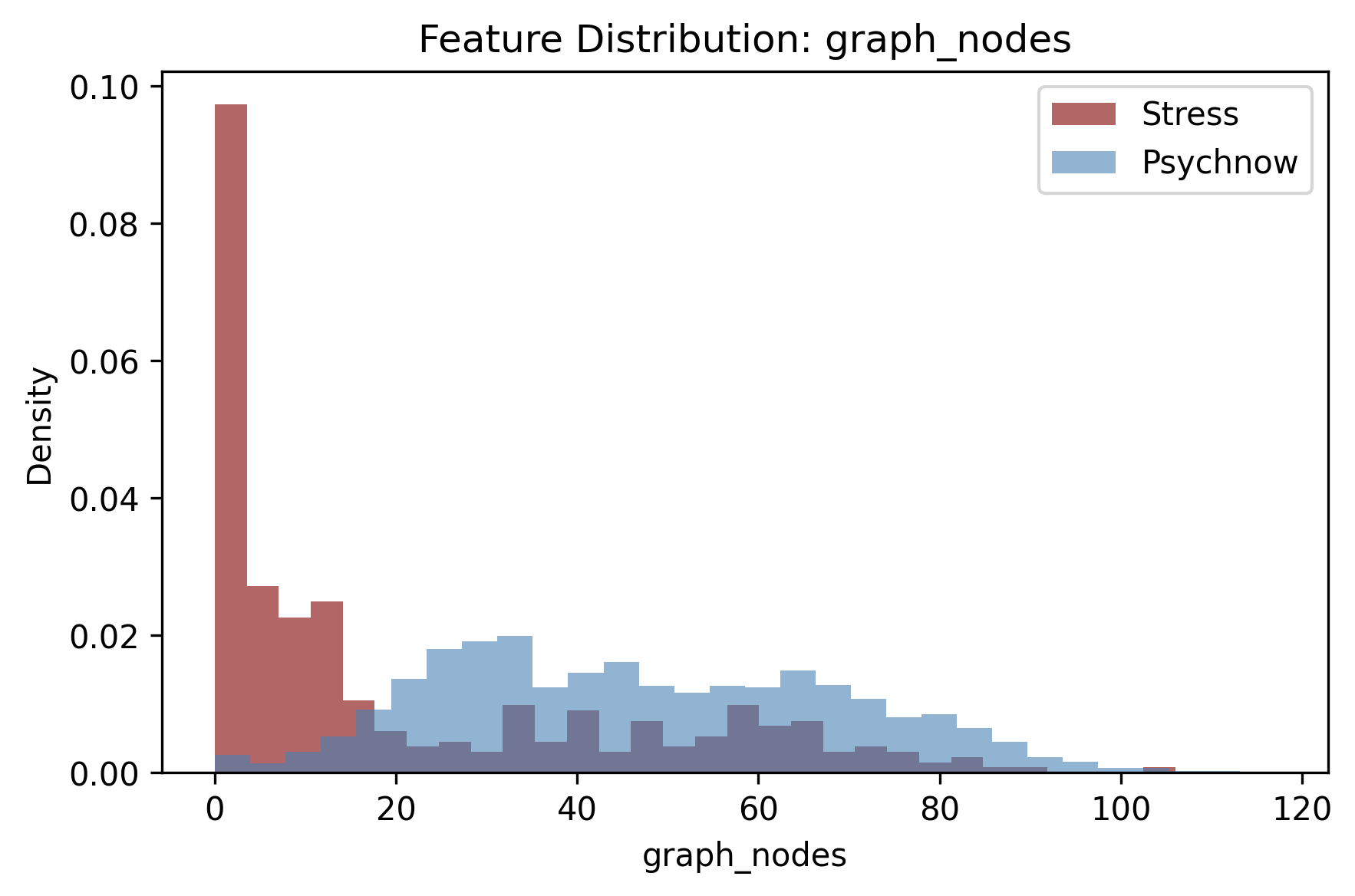} &
        \includegraphics[width=0.09\textwidth]{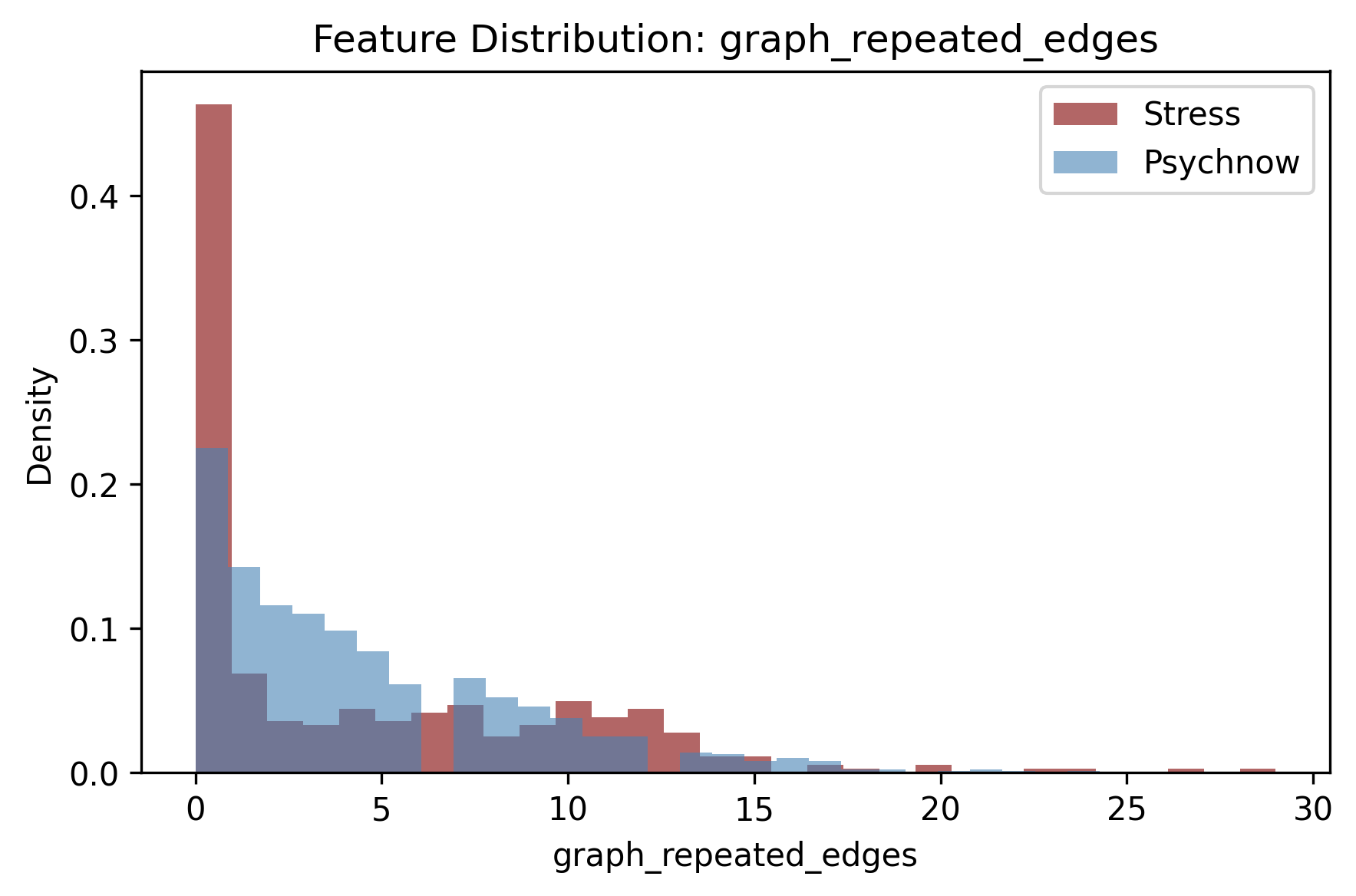} \\[-2pt]

        \includegraphics[width=0.09\textwidth]{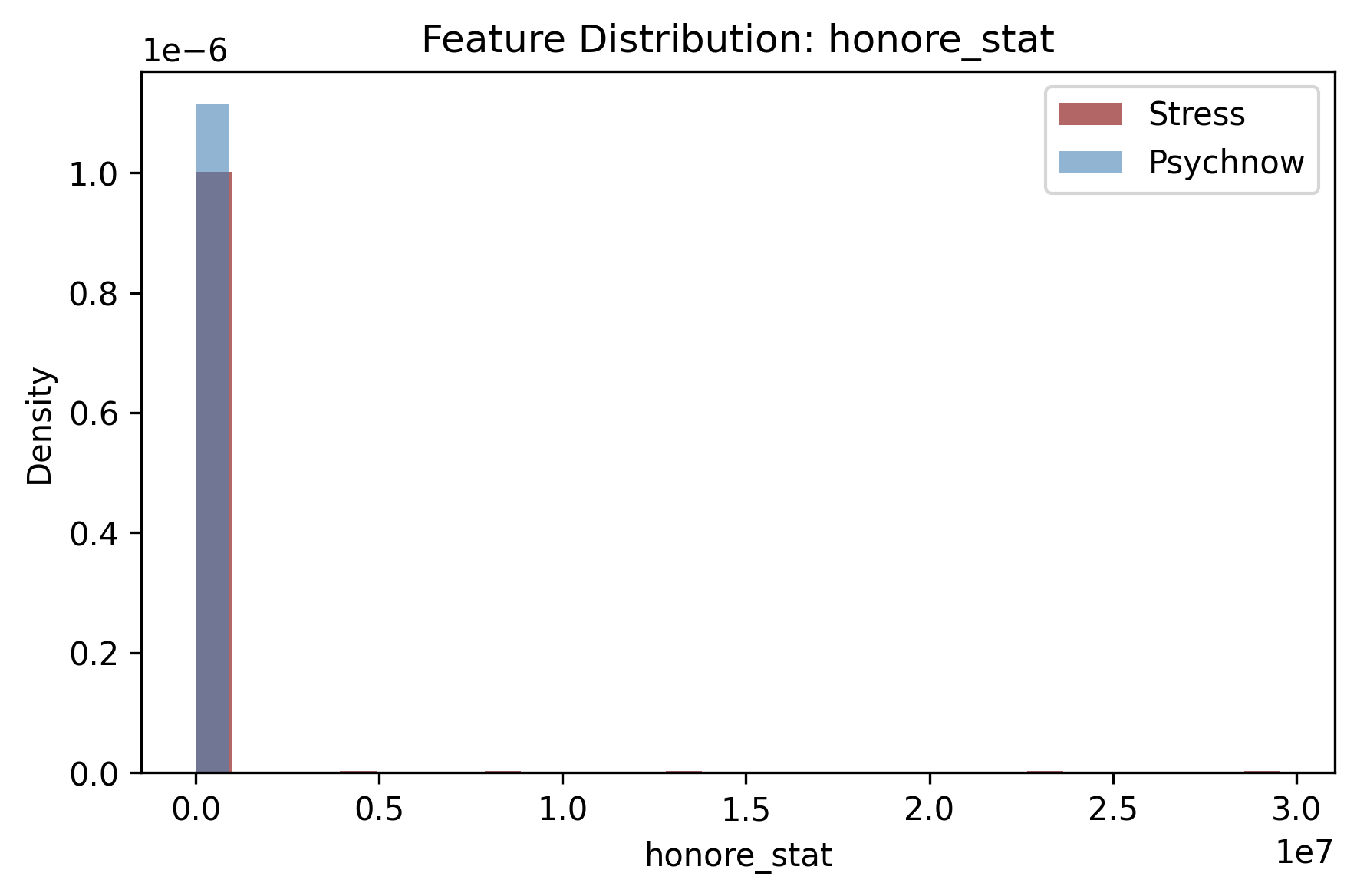} &
        \includegraphics[width=0.09\textwidth]{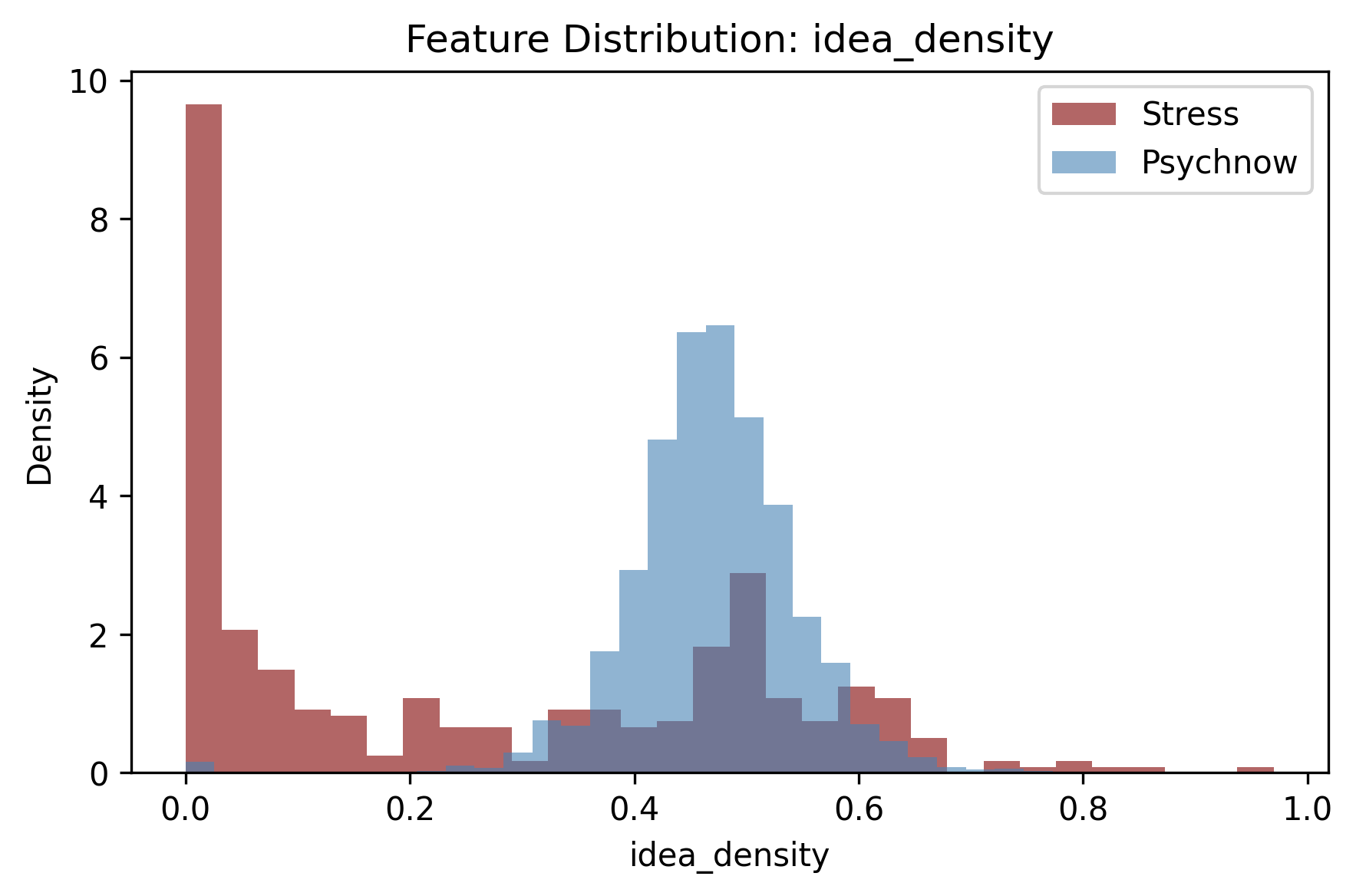} &
        \includegraphics[width=0.09\textwidth]{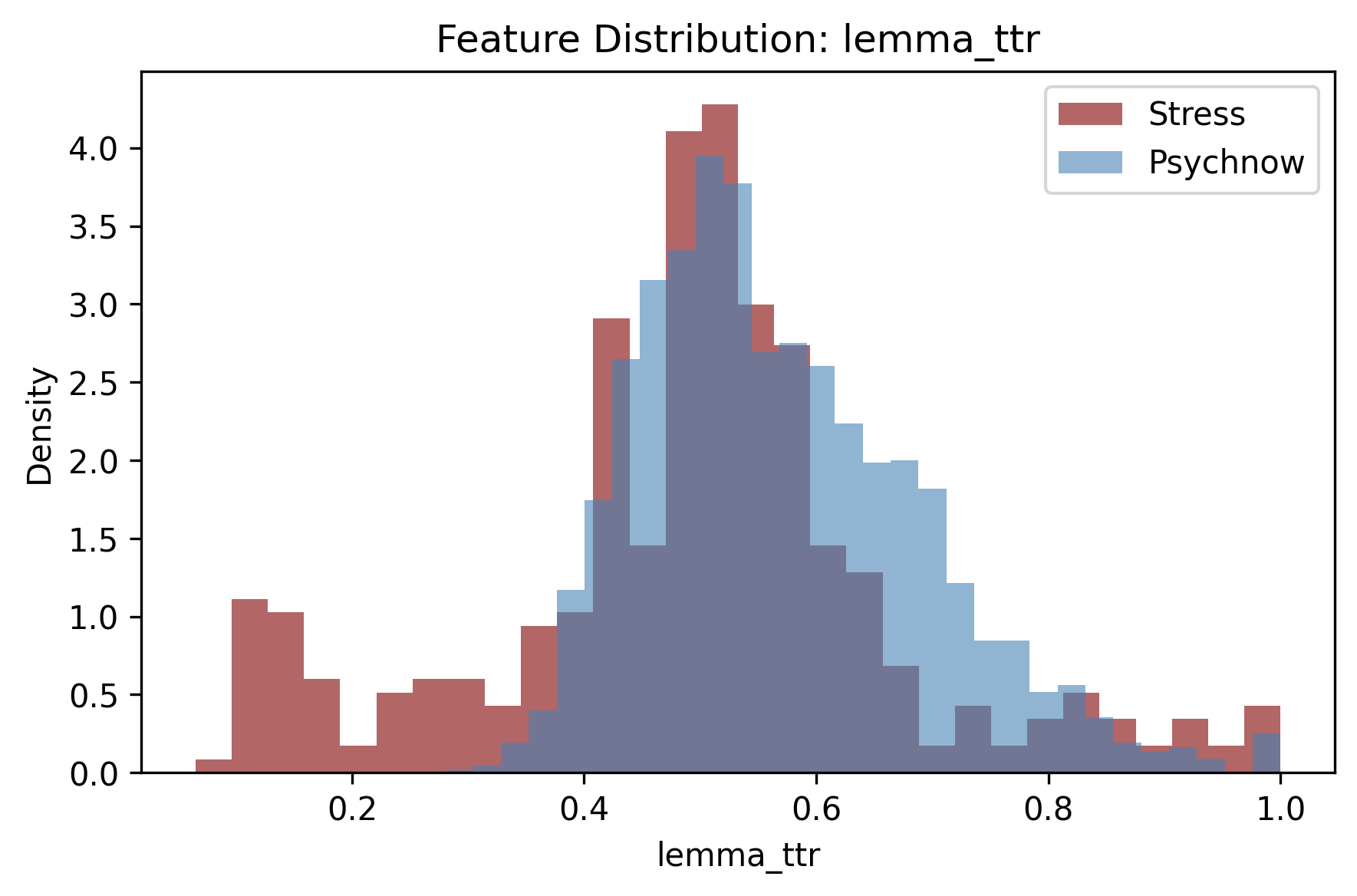} &
        \includegraphics[width=0.09\textwidth]{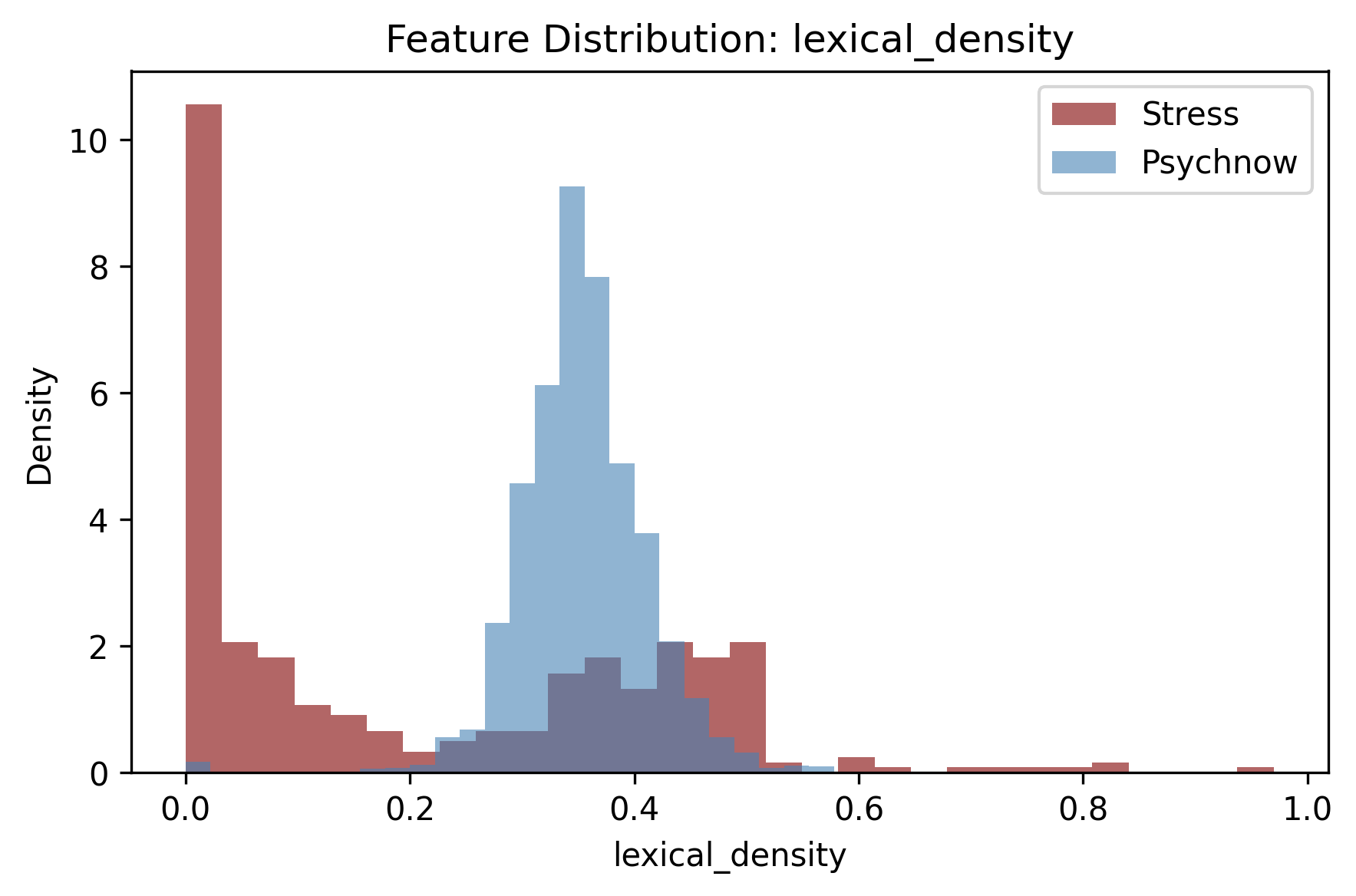} &
        \includegraphics[width=0.09\textwidth]{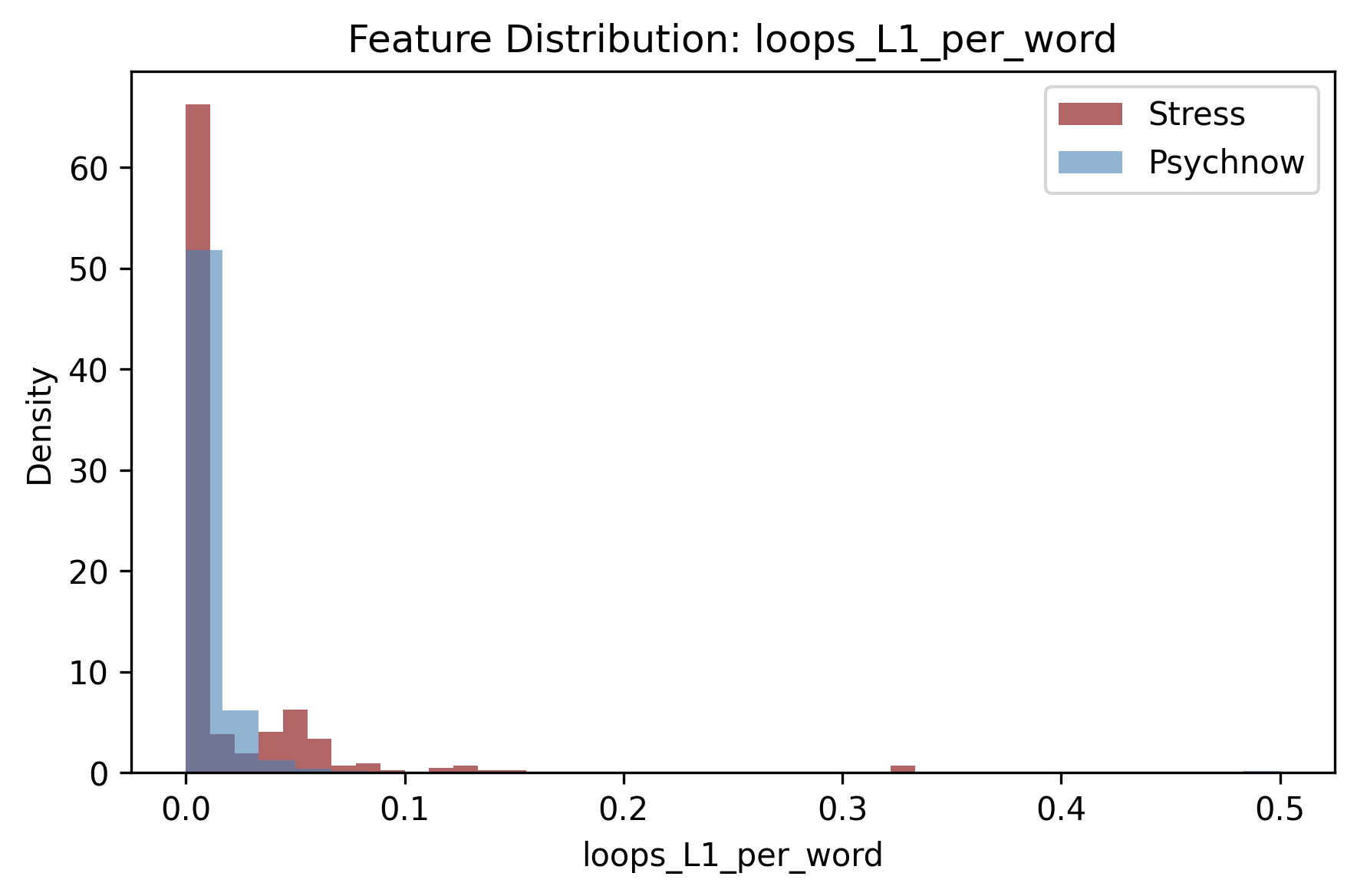} &
        \includegraphics[width=0.09\textwidth]{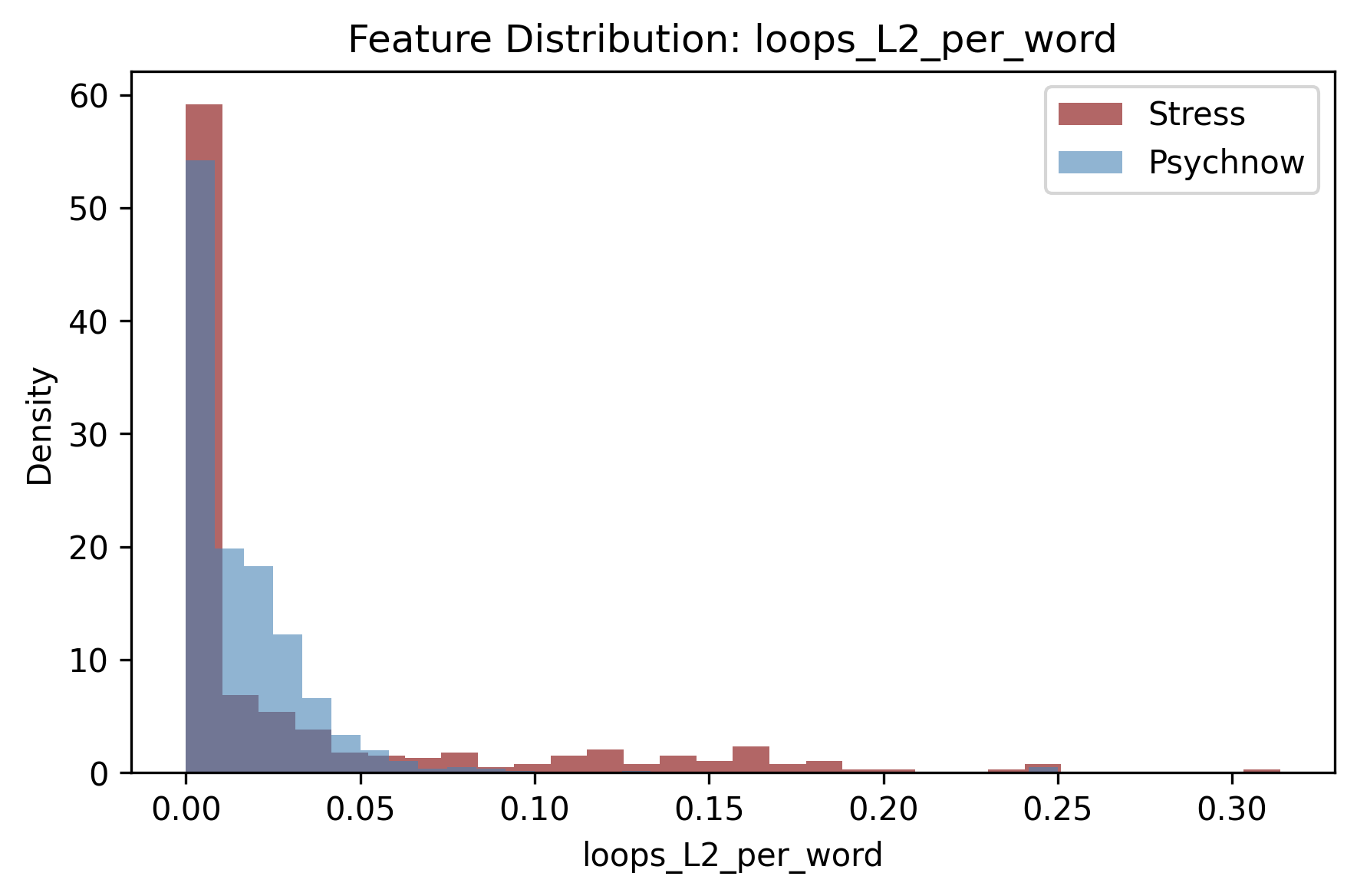} &
        \includegraphics[width=0.09\textwidth]{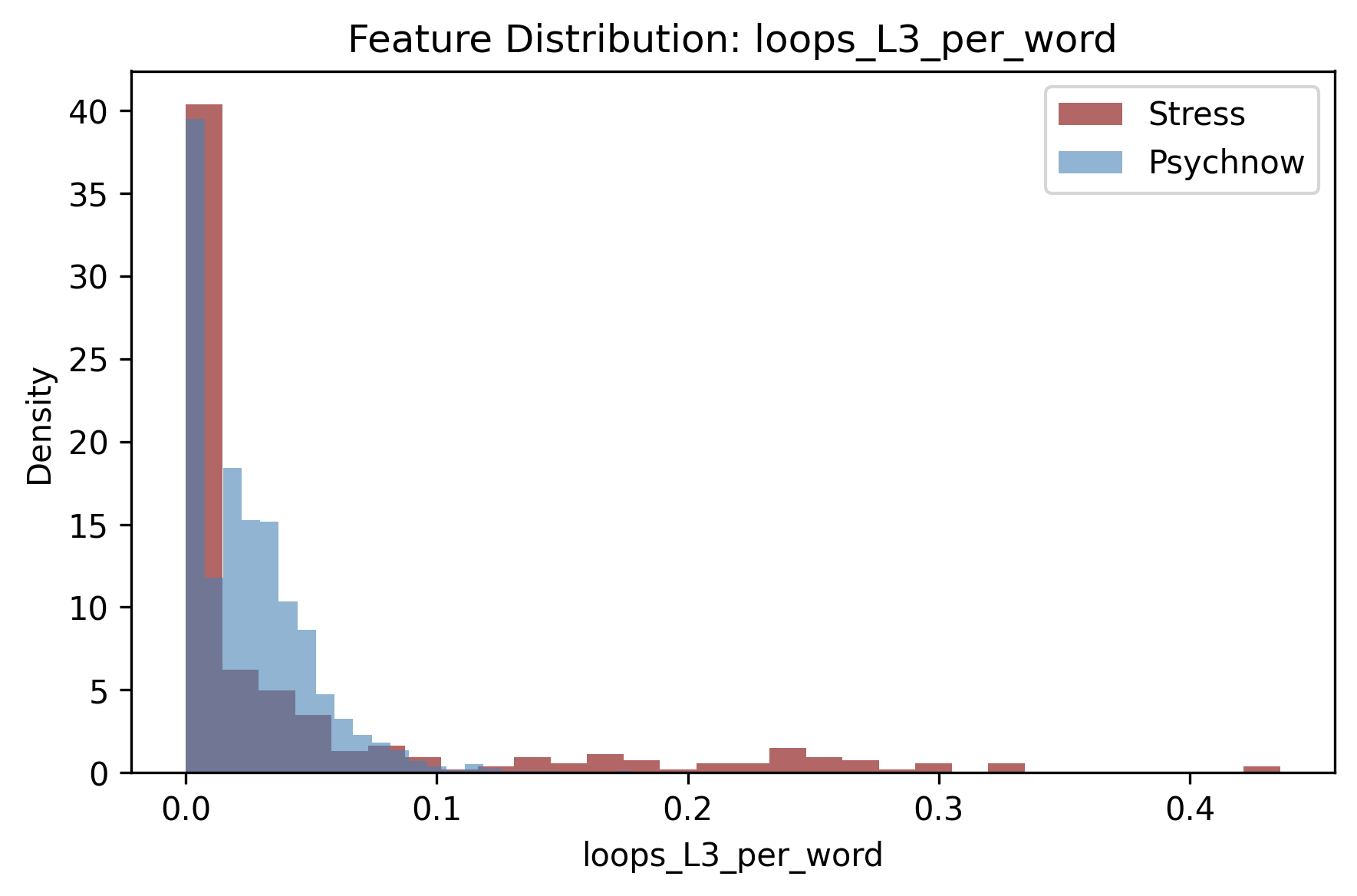} &
        \includegraphics[width=0.09\textwidth]{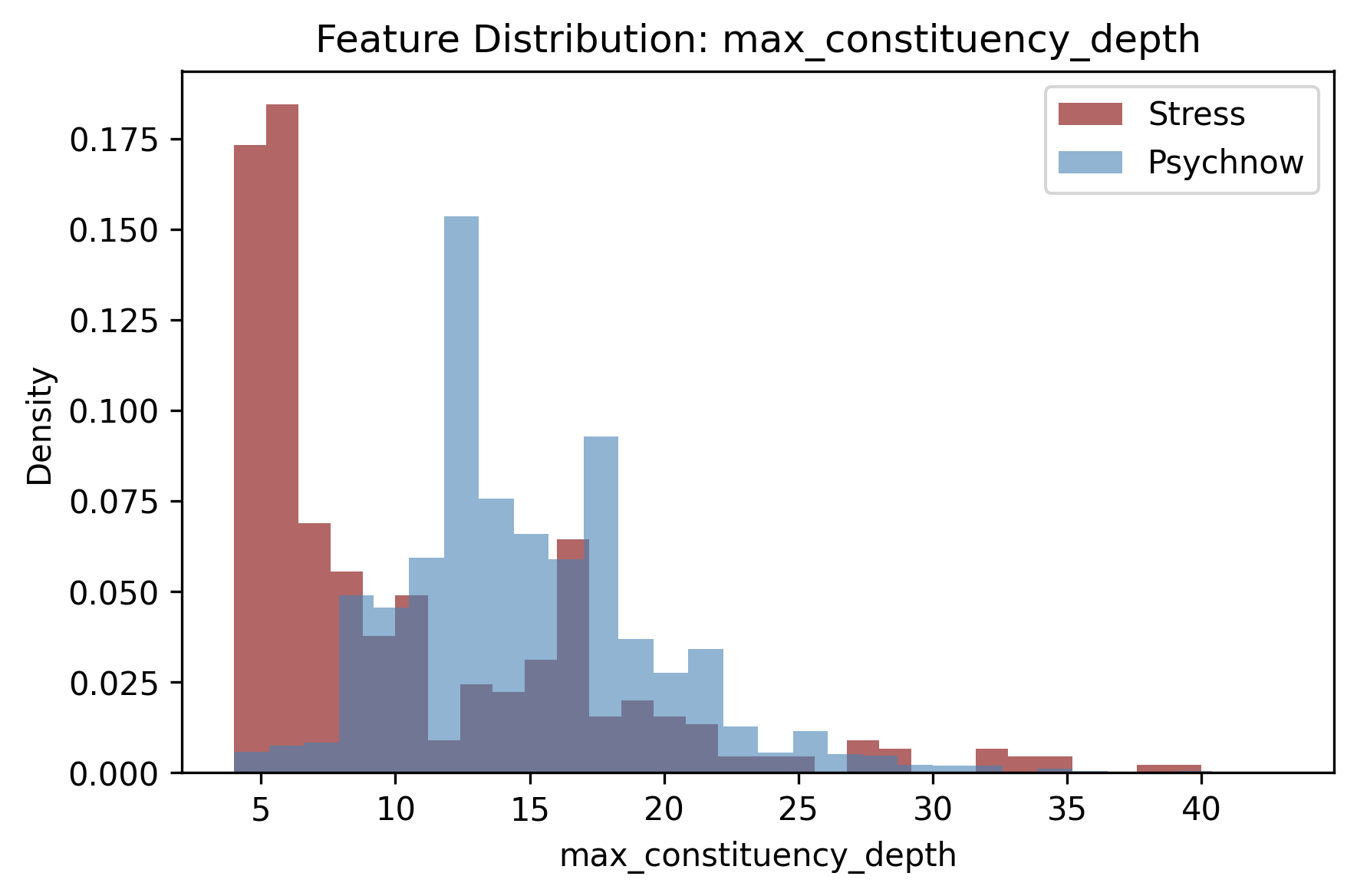} &
        \includegraphics[width=0.09\textwidth]{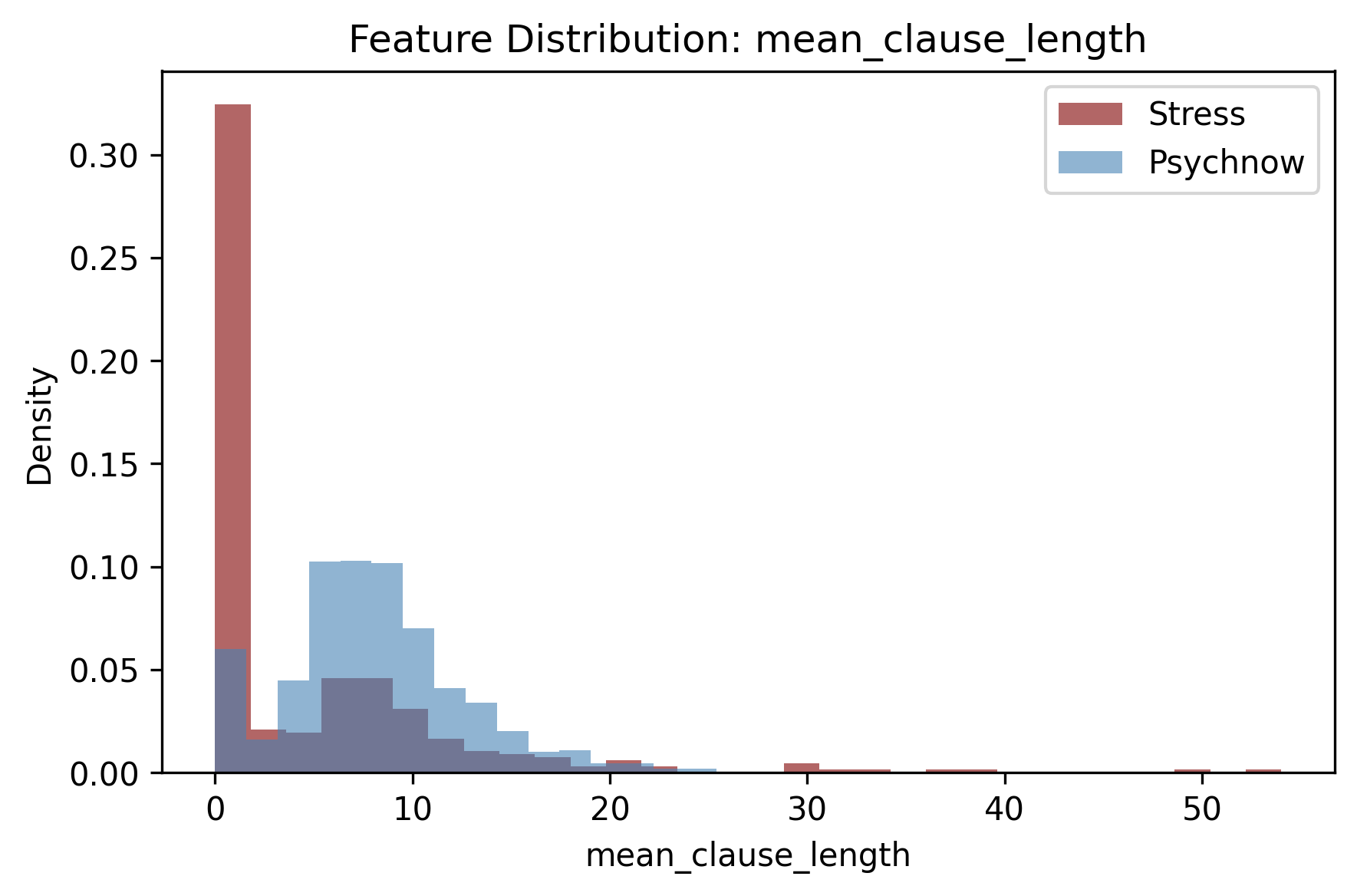} &
        \includegraphics[width=0.09\textwidth]{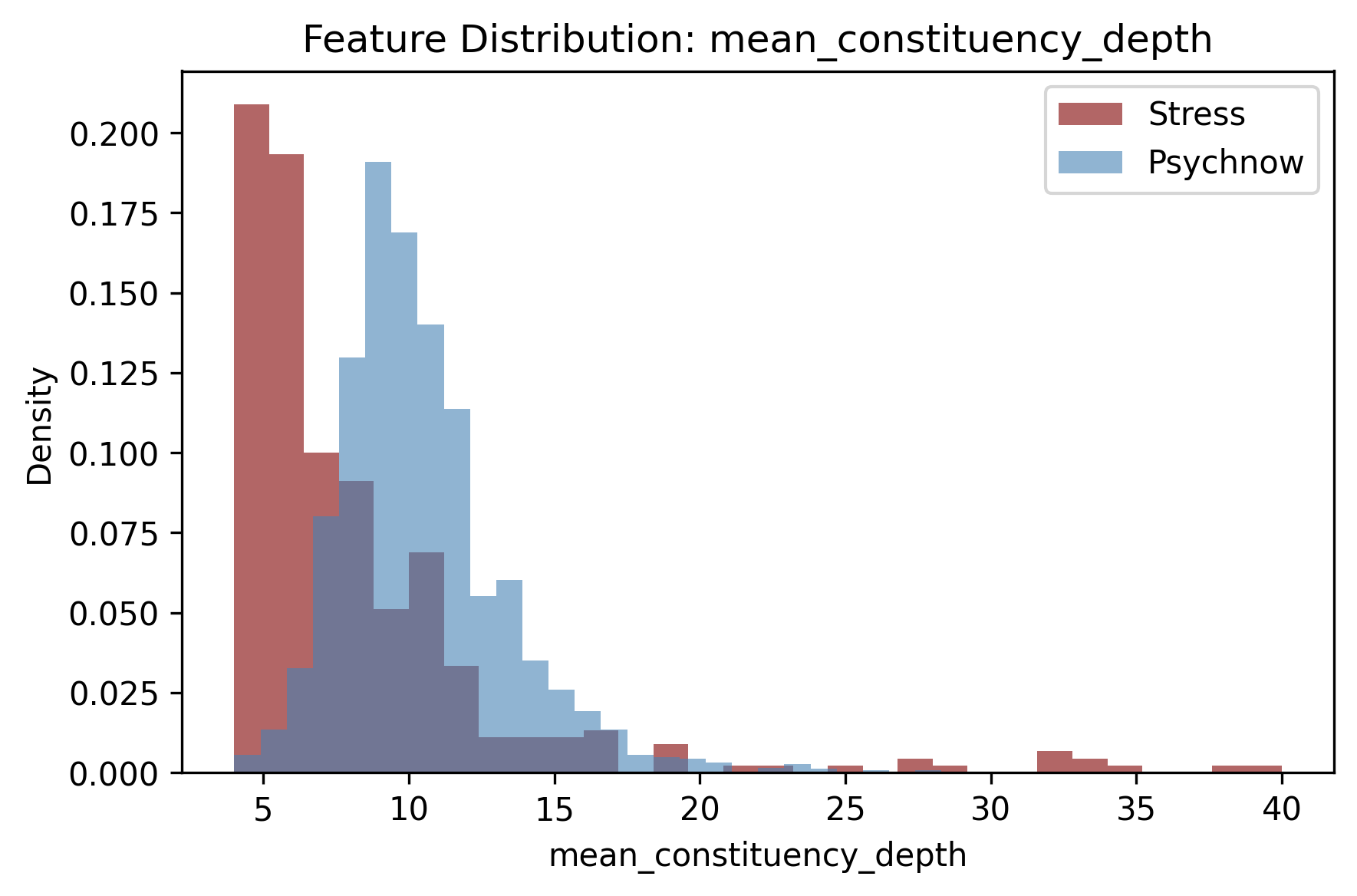} \\[-2pt]

        \includegraphics[width=0.09\textwidth]{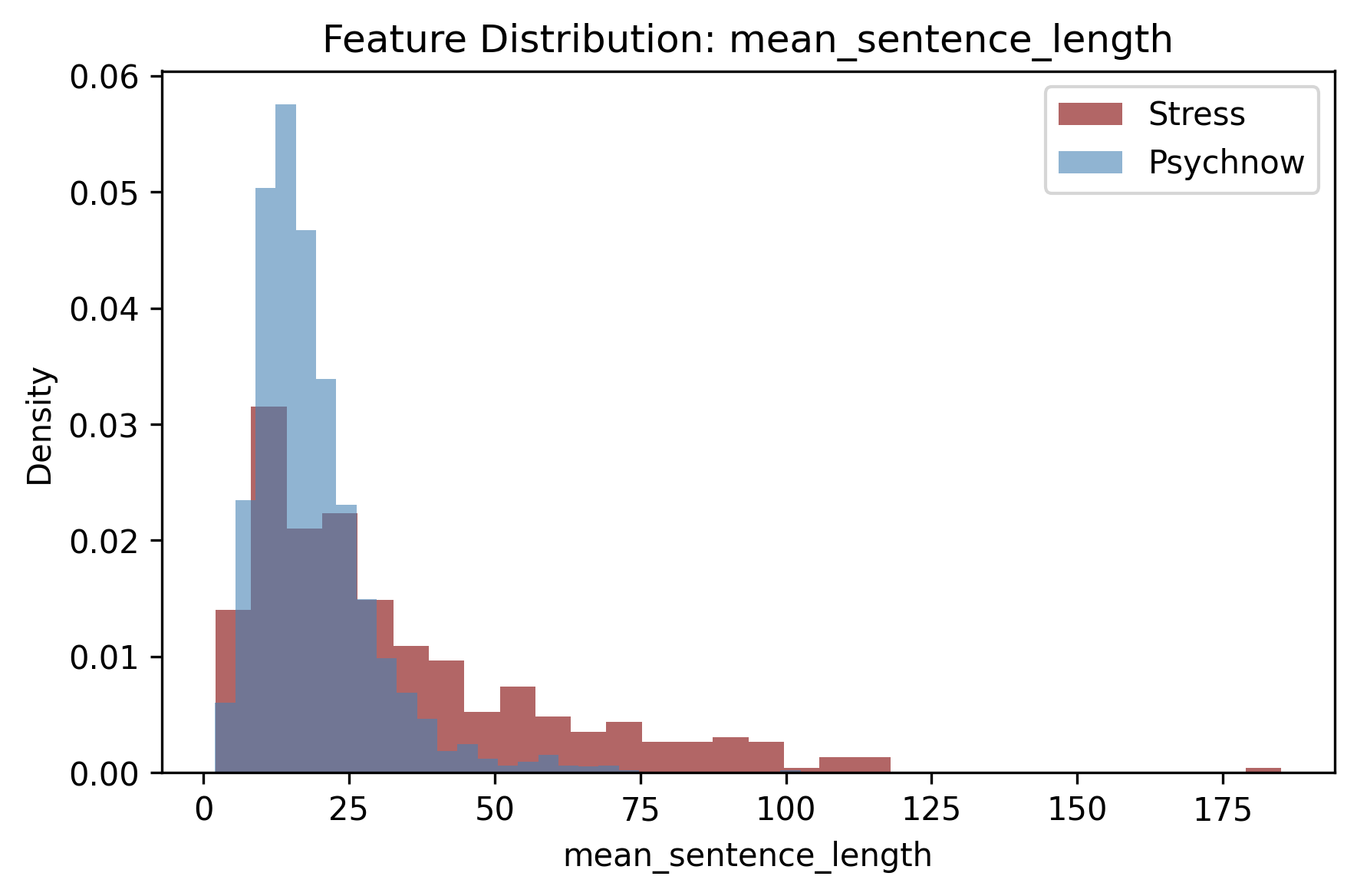} &
        \includegraphics[width=0.09\textwidth]{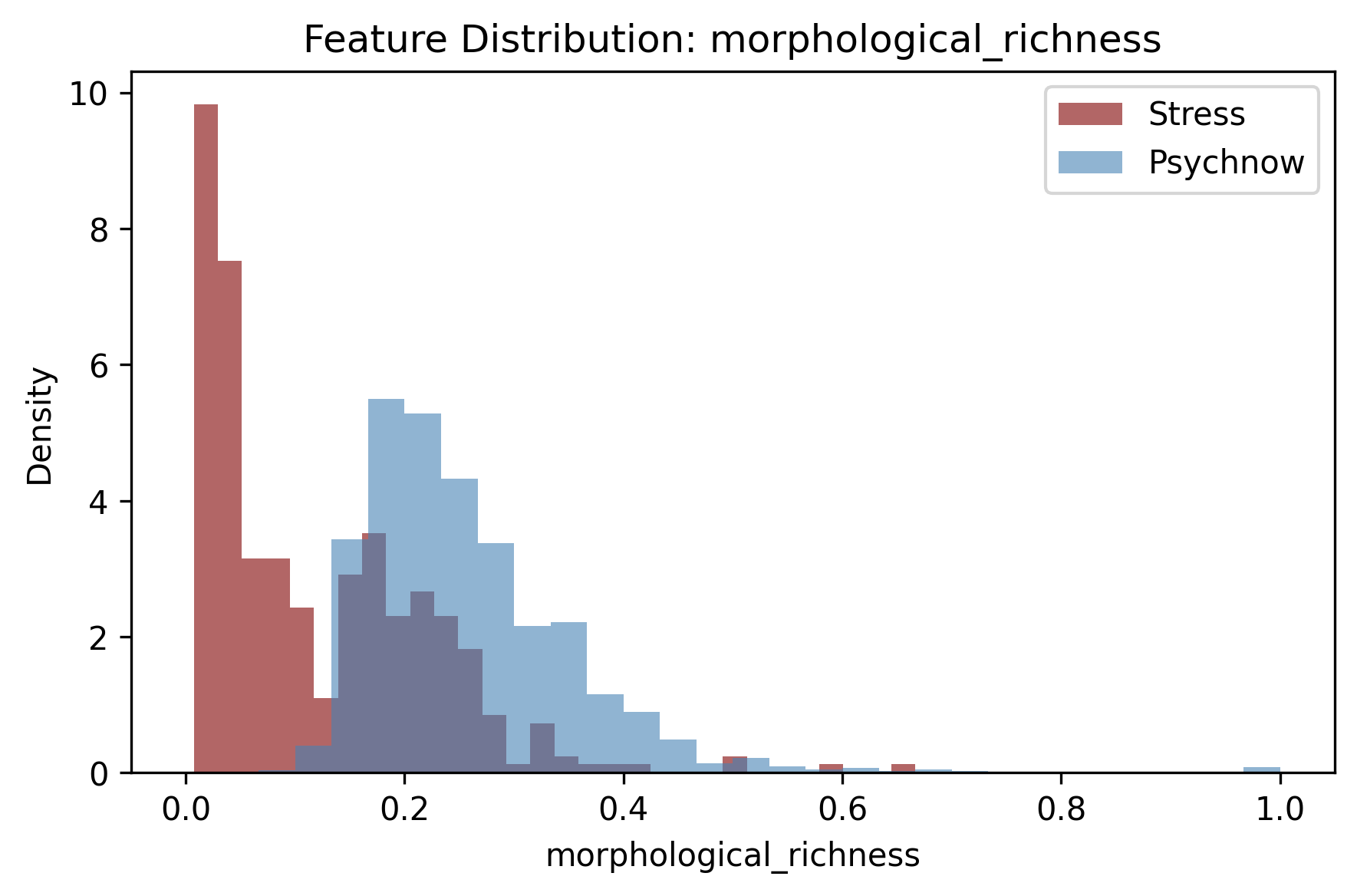} &
        \includegraphics[width=0.09\textwidth]{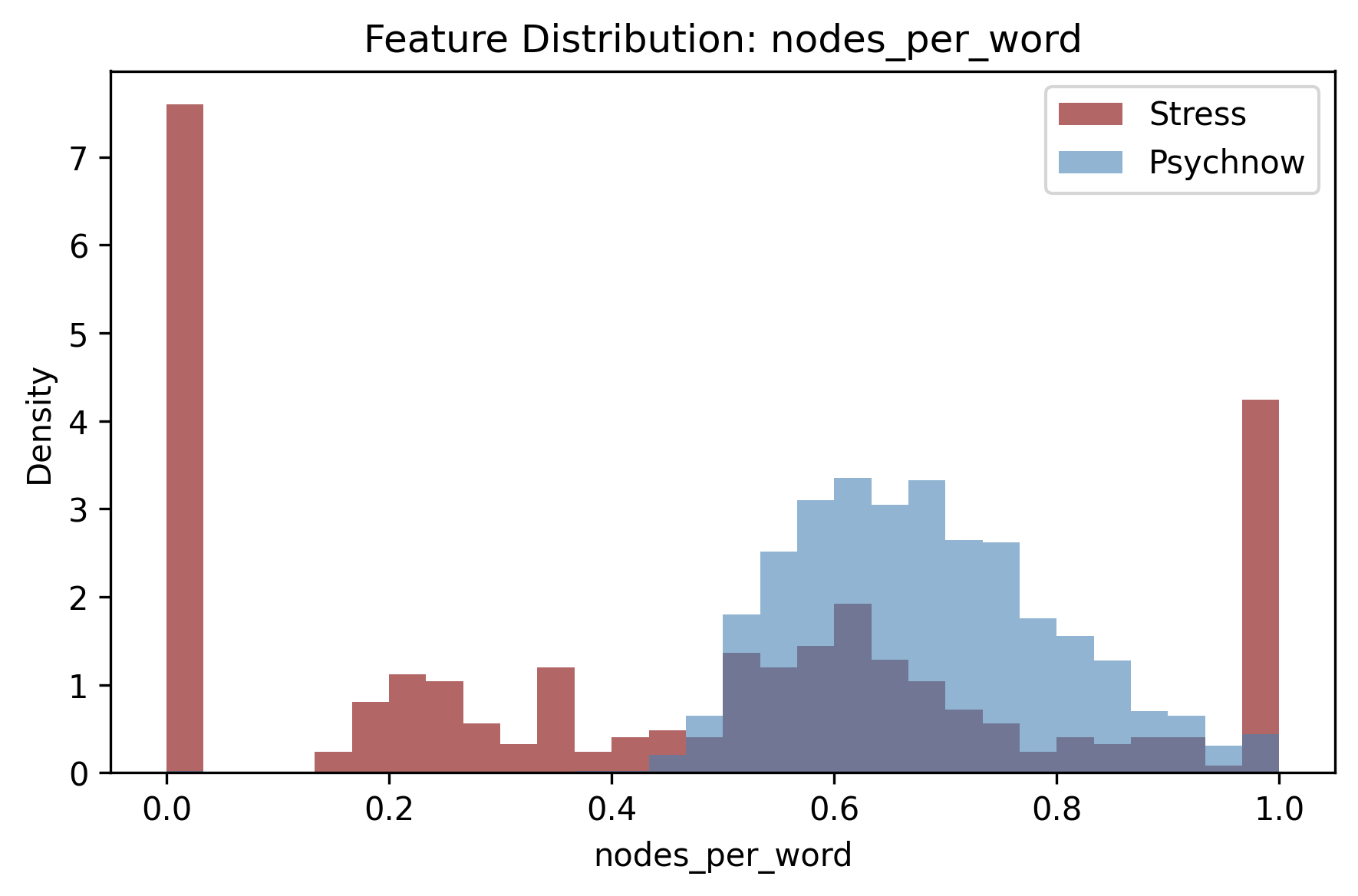} &
        \includegraphics[width=0.09\textwidth]{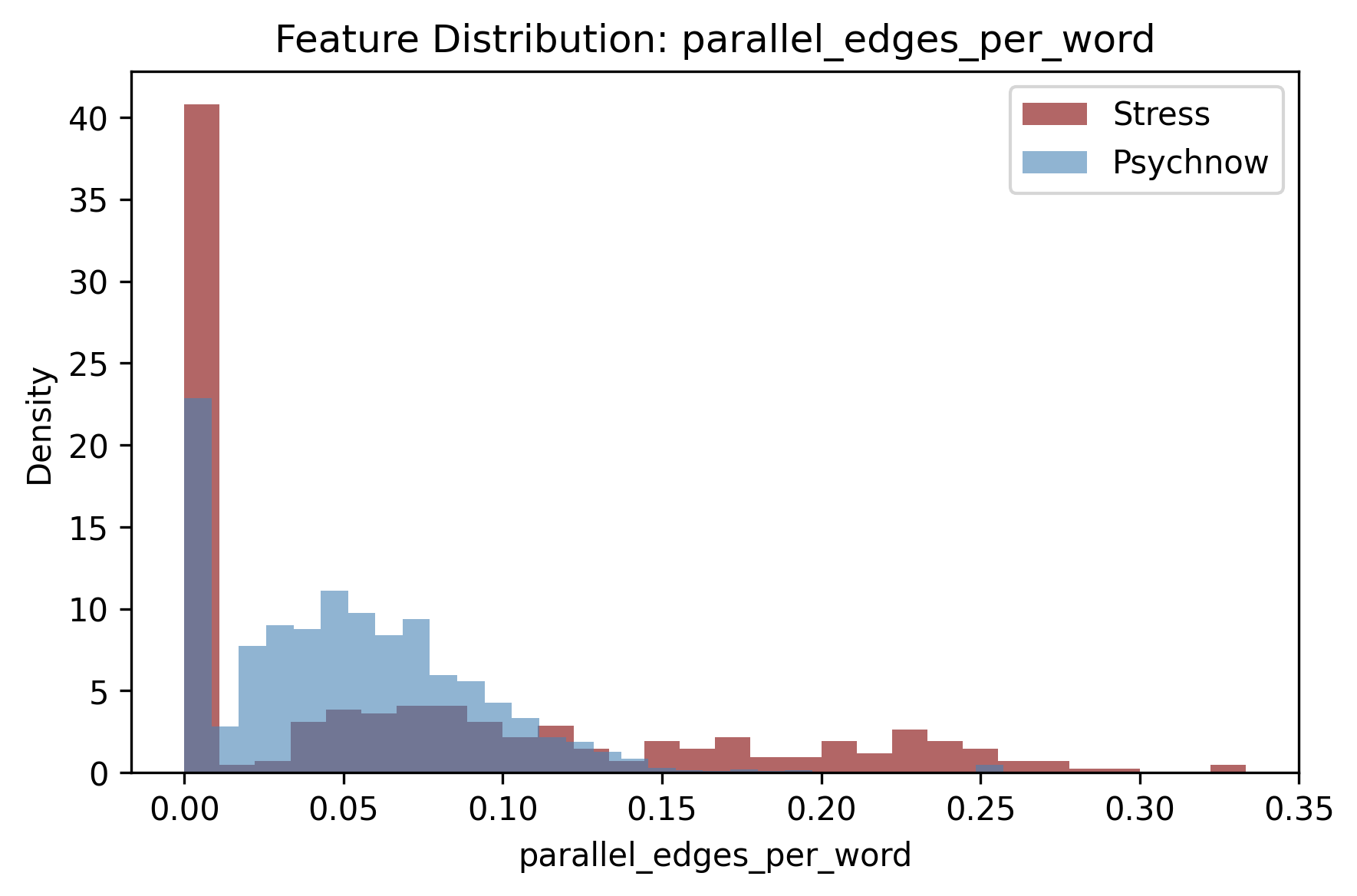} &
        \includegraphics[width=0.09\textwidth]{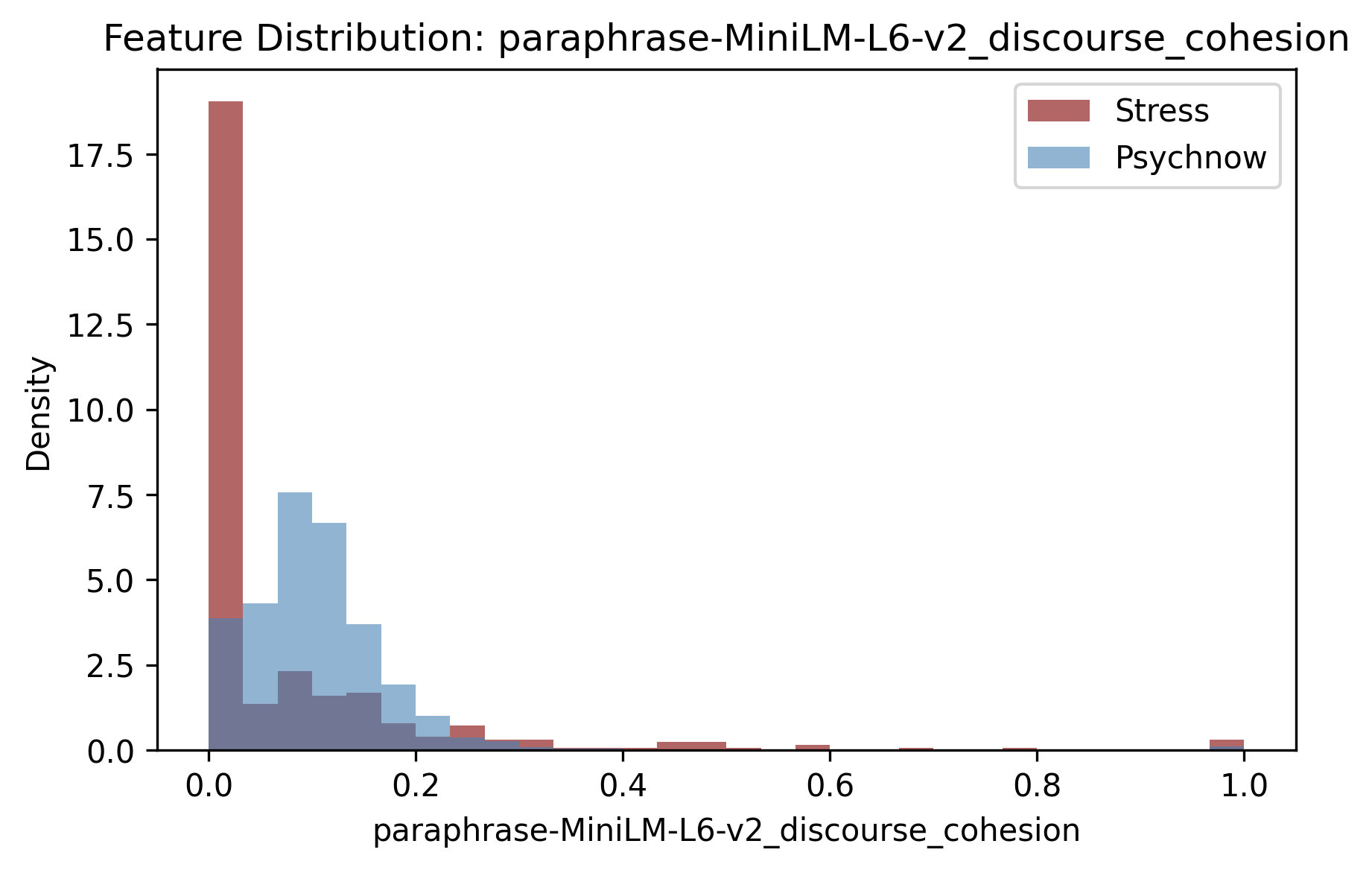} &
        \includegraphics[width=0.09\textwidth]{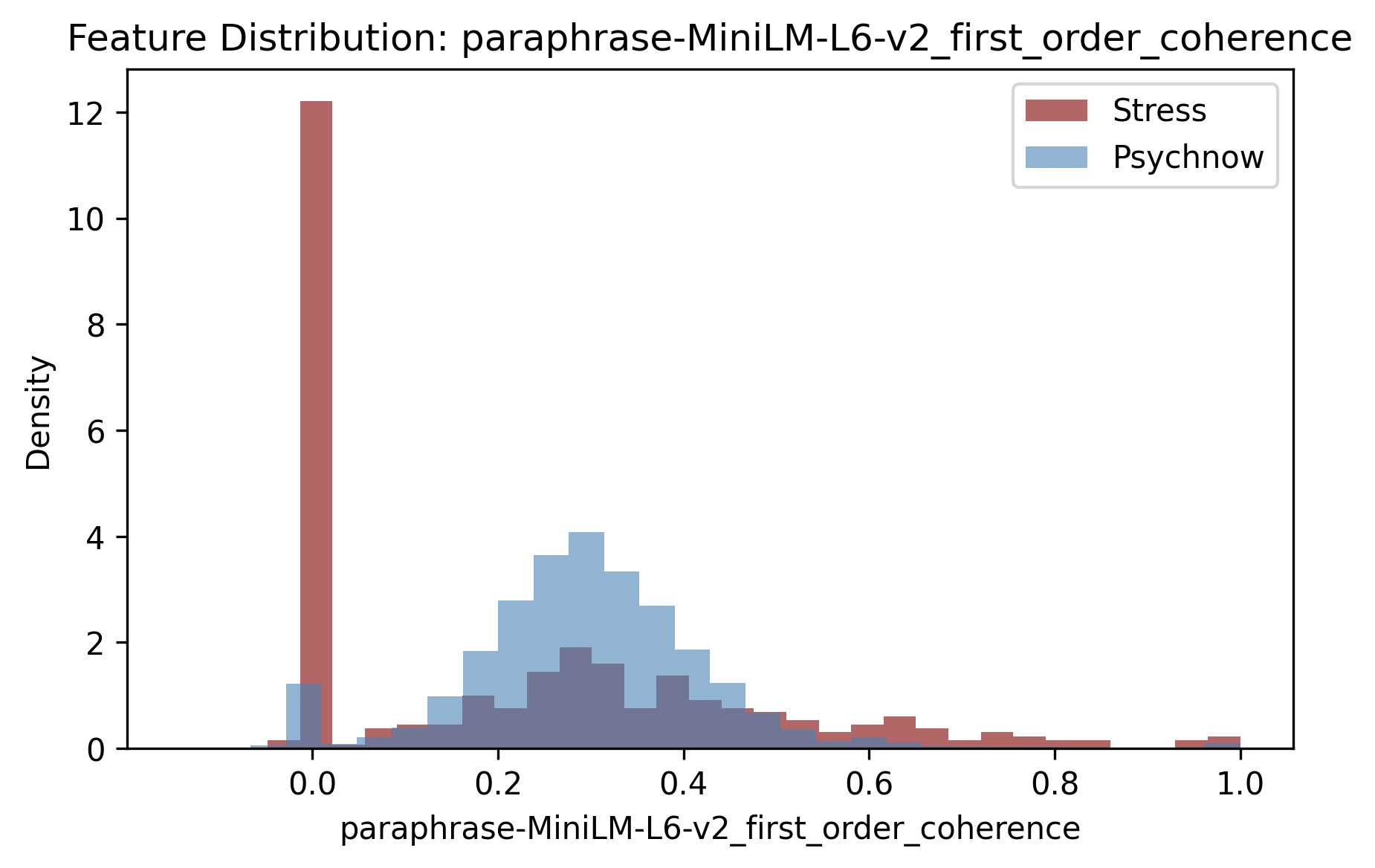} &
        \includegraphics[width=0.09\textwidth]{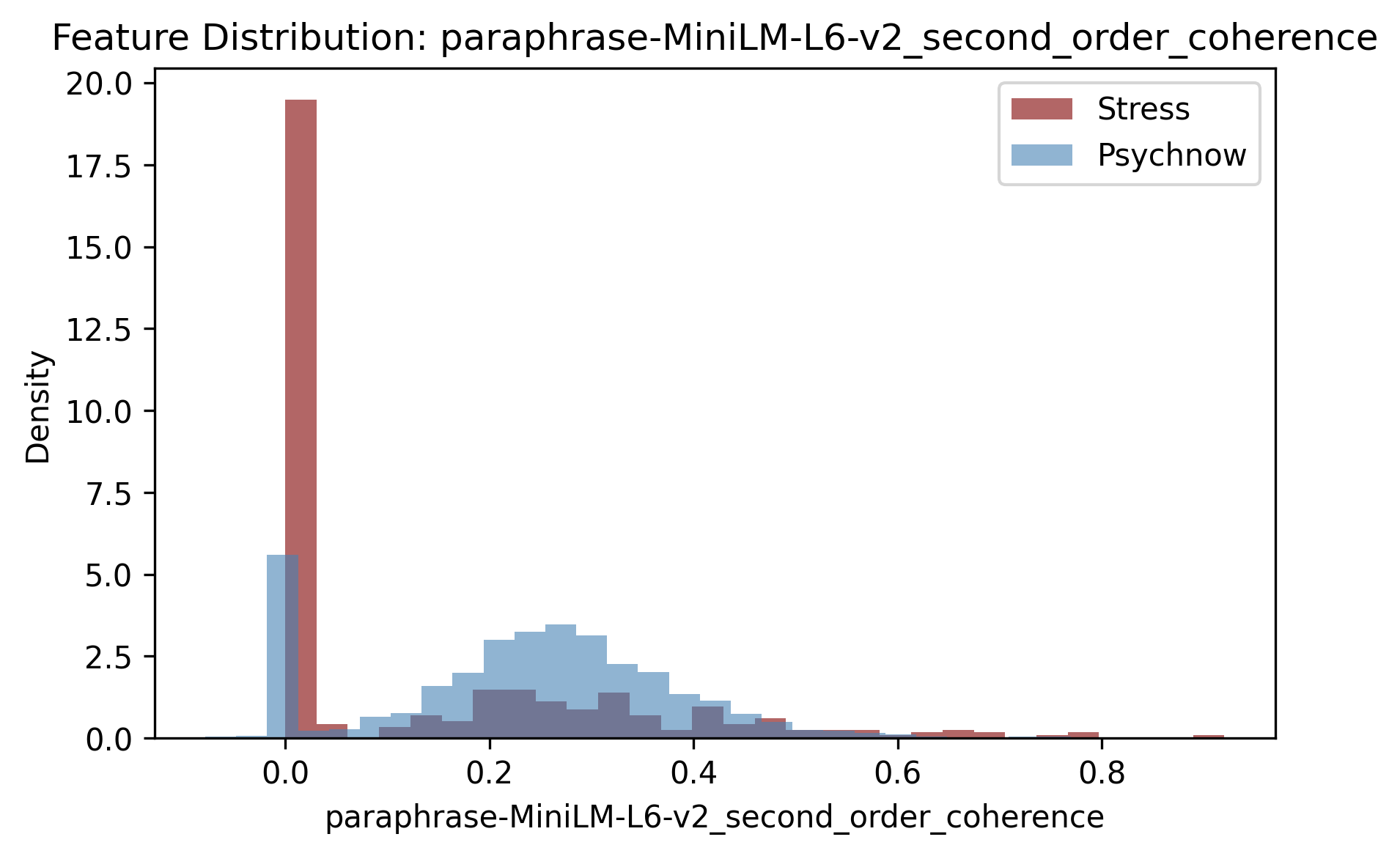} &
        \includegraphics[width=0.09\textwidth]{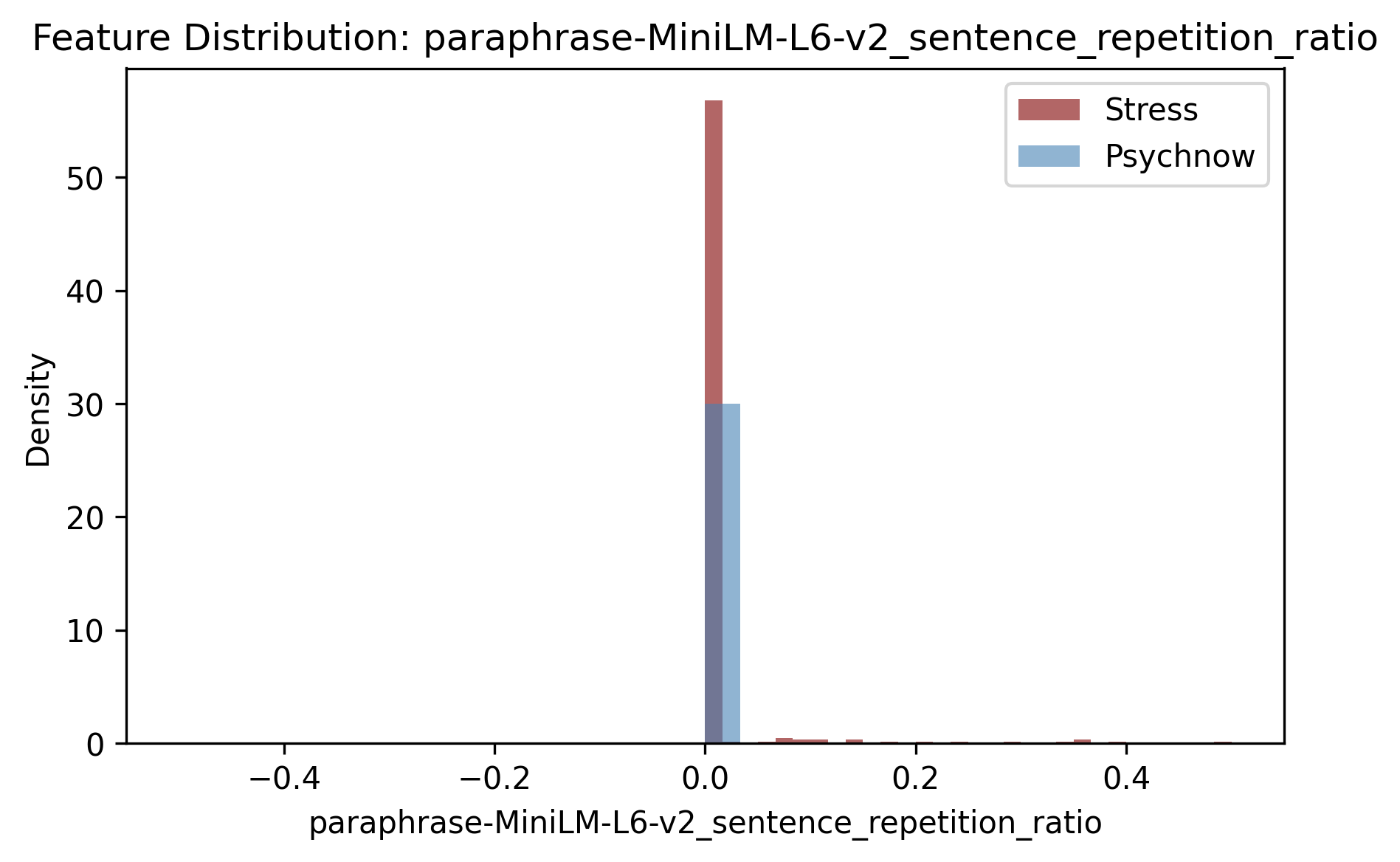} &
        \includegraphics[width=0.09\textwidth]{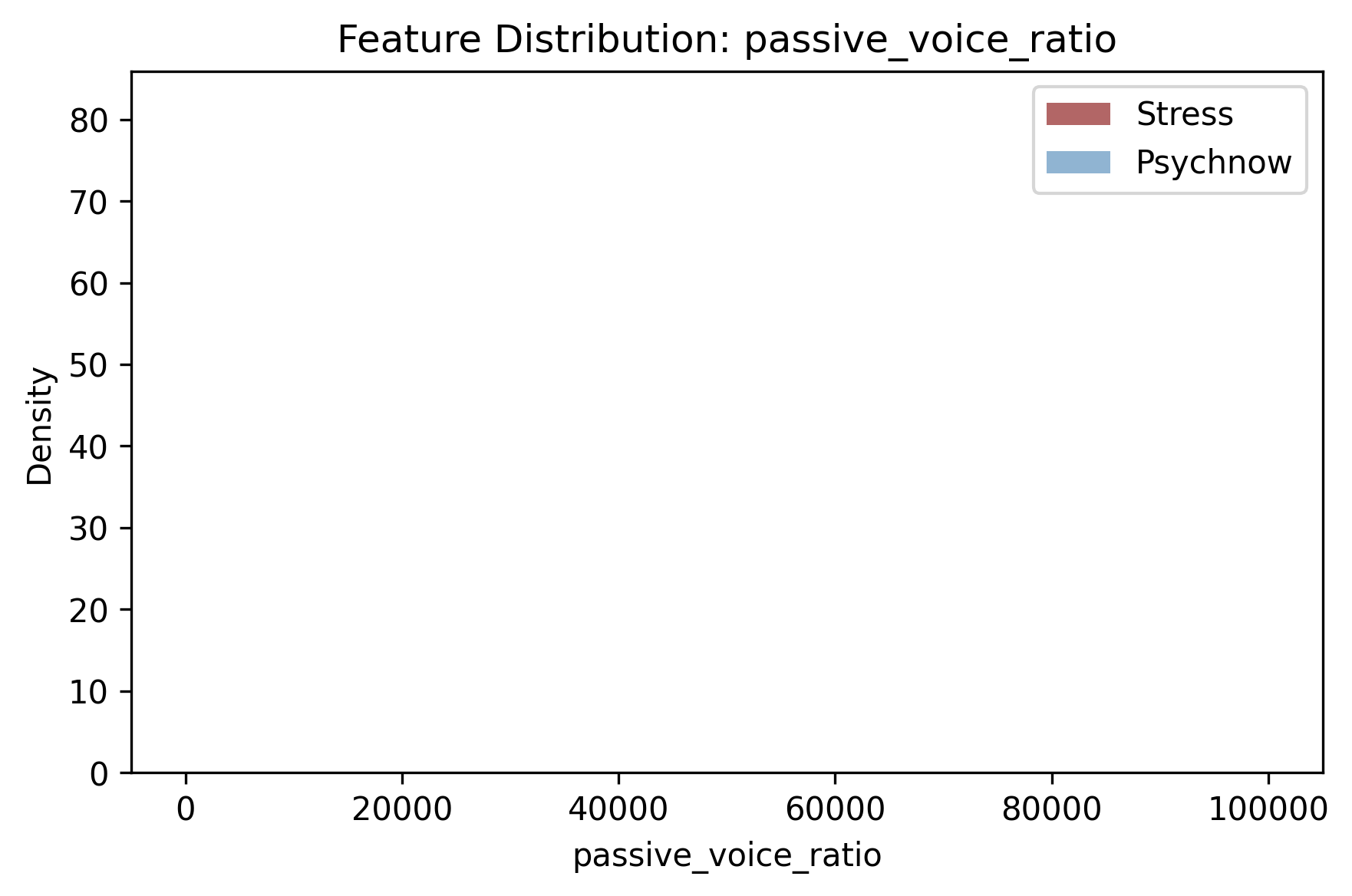} &
        \includegraphics[width=0.09\textwidth]{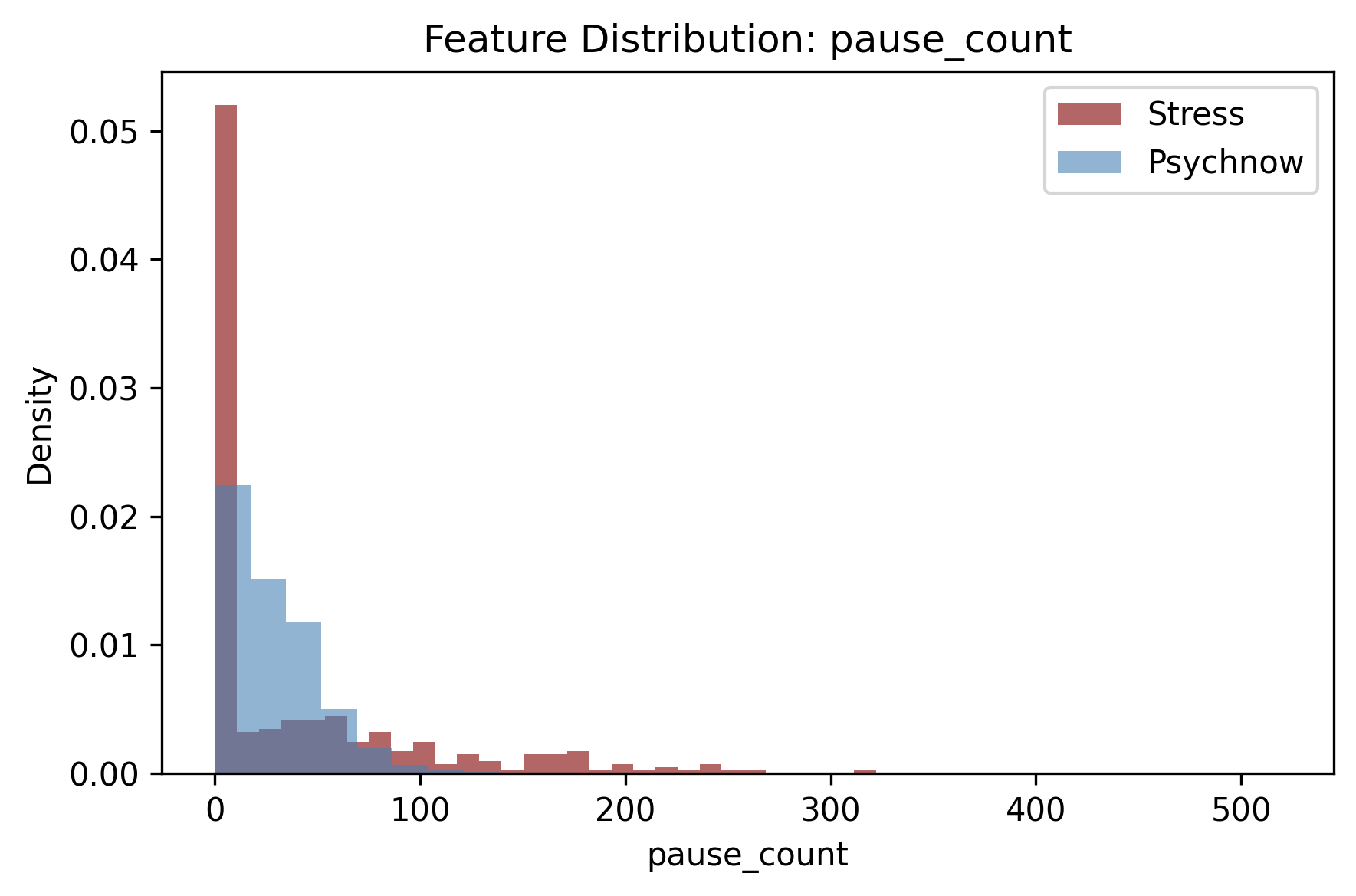} \\[-2pt]

        \includegraphics[width=0.09\textwidth]{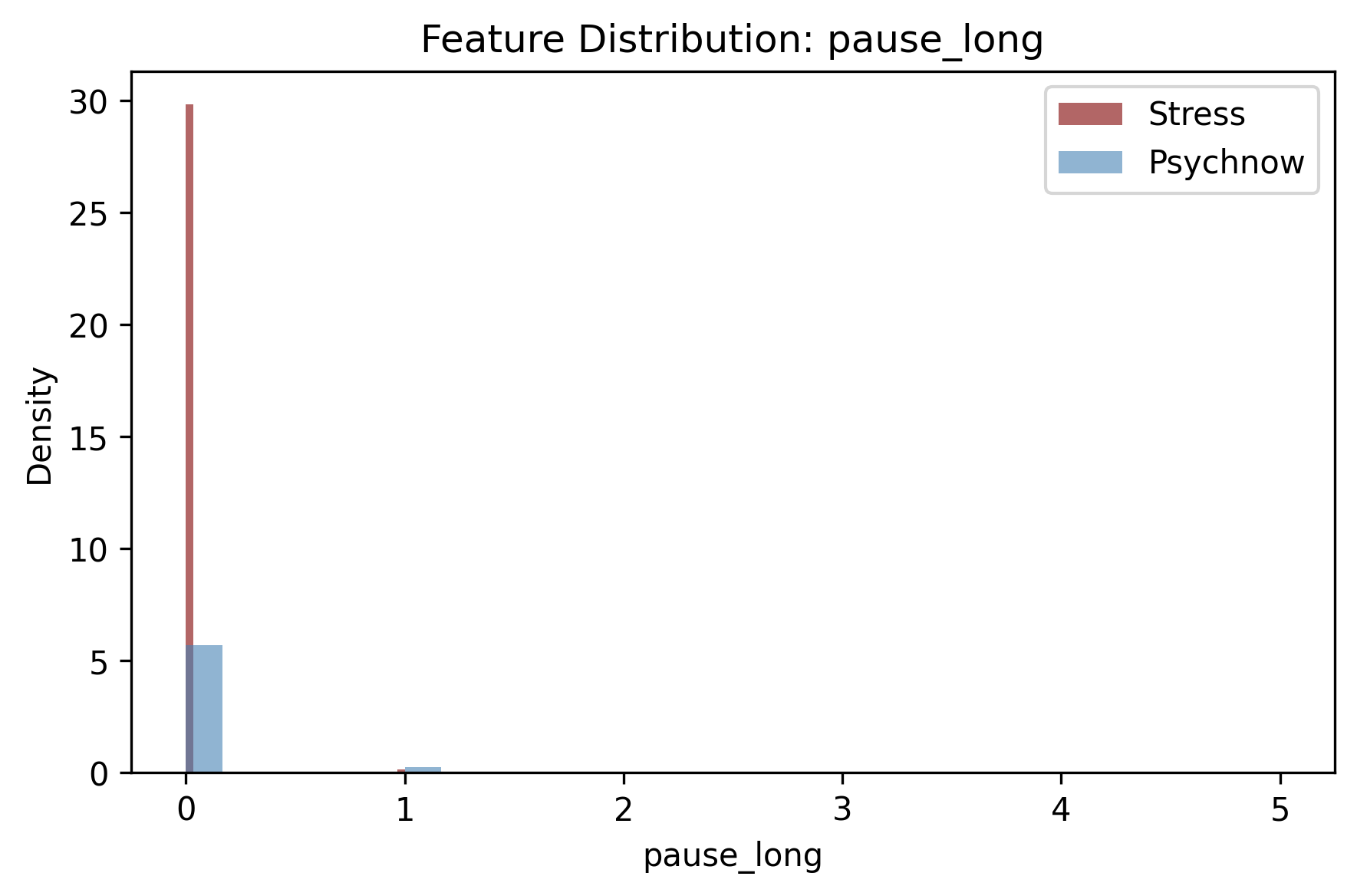} &
        \includegraphics[width=0.09\textwidth]{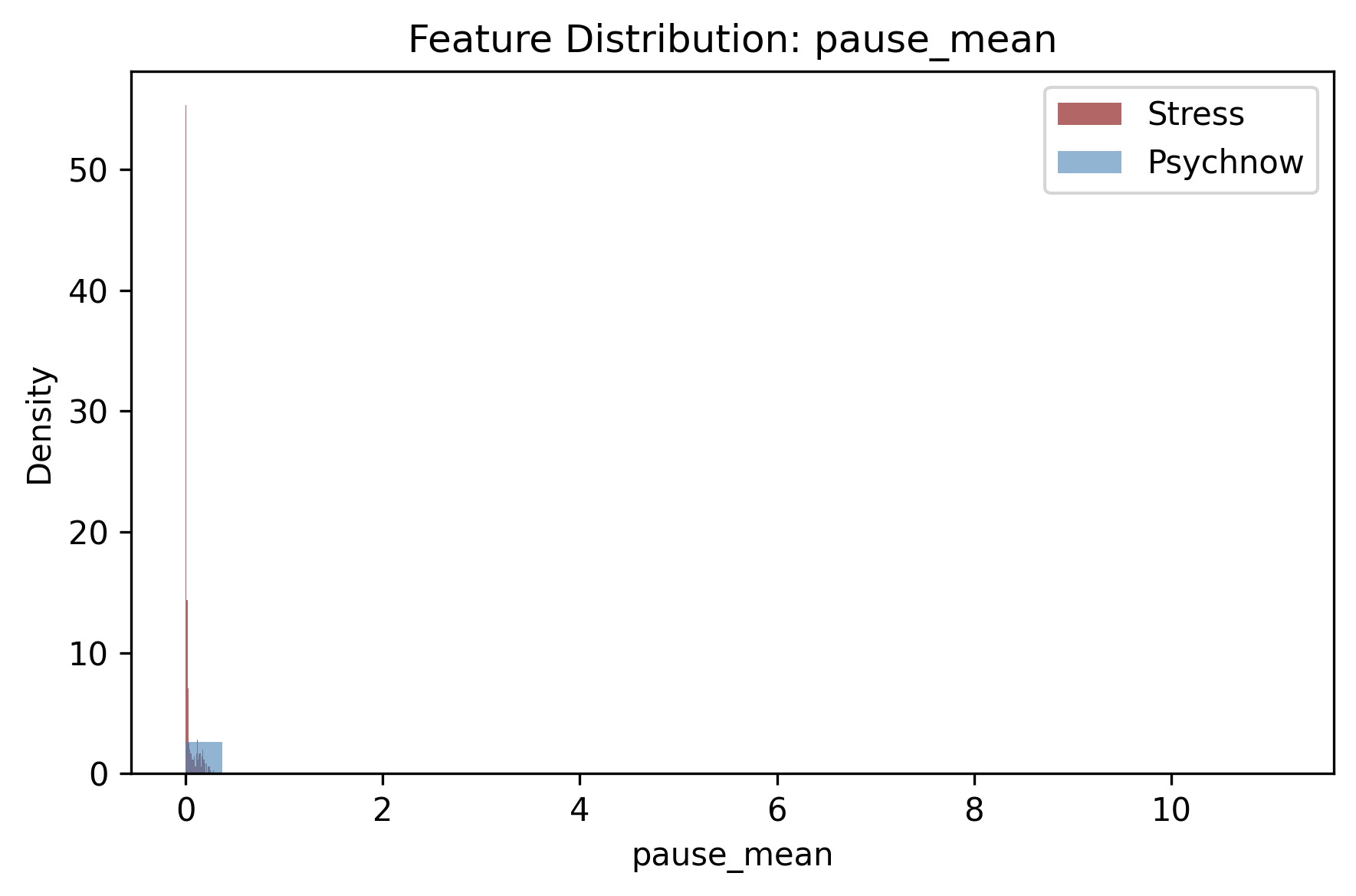} &
        \includegraphics[width=0.09\textwidth]{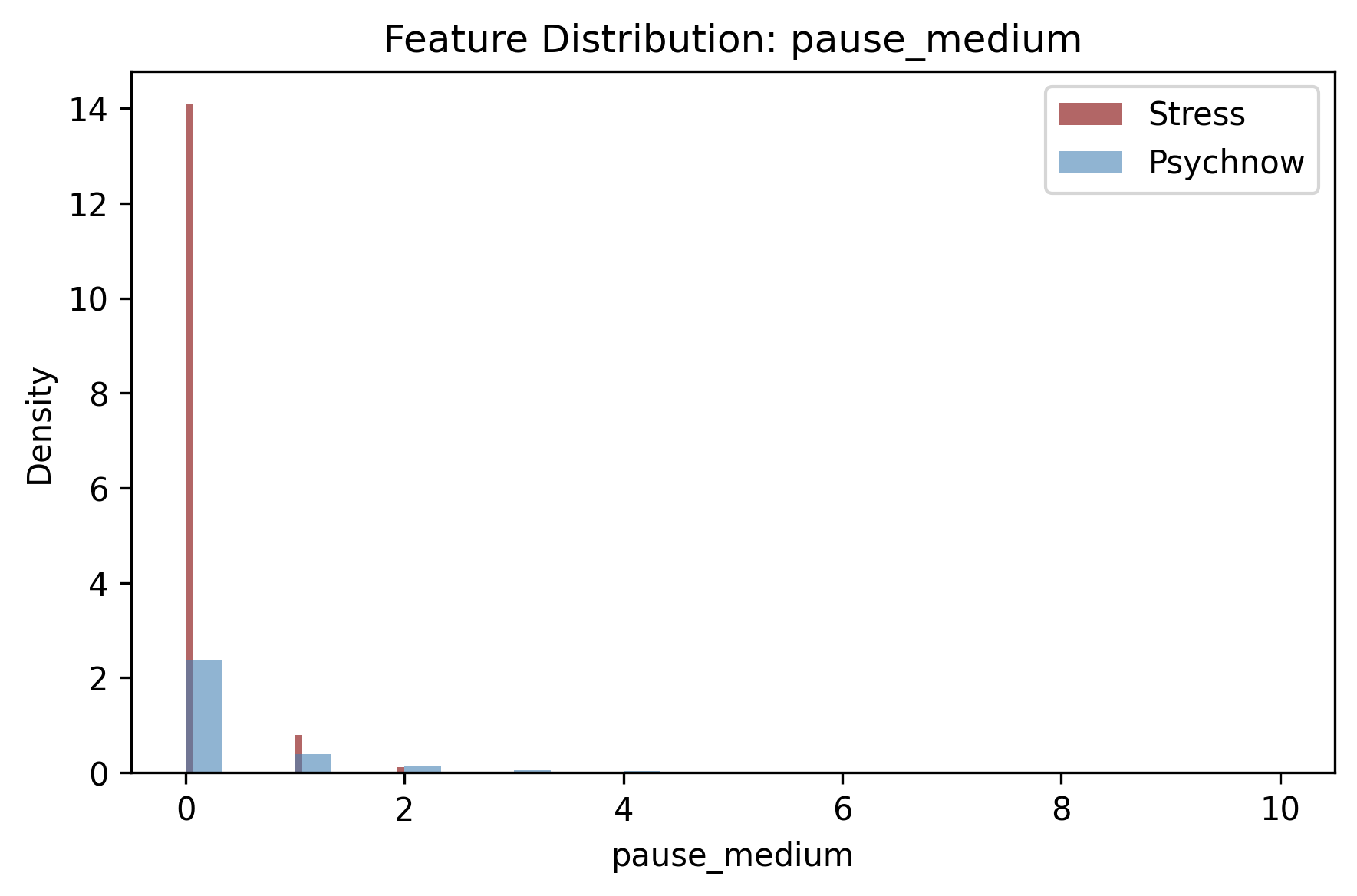} &
        \includegraphics[width=0.09\textwidth]{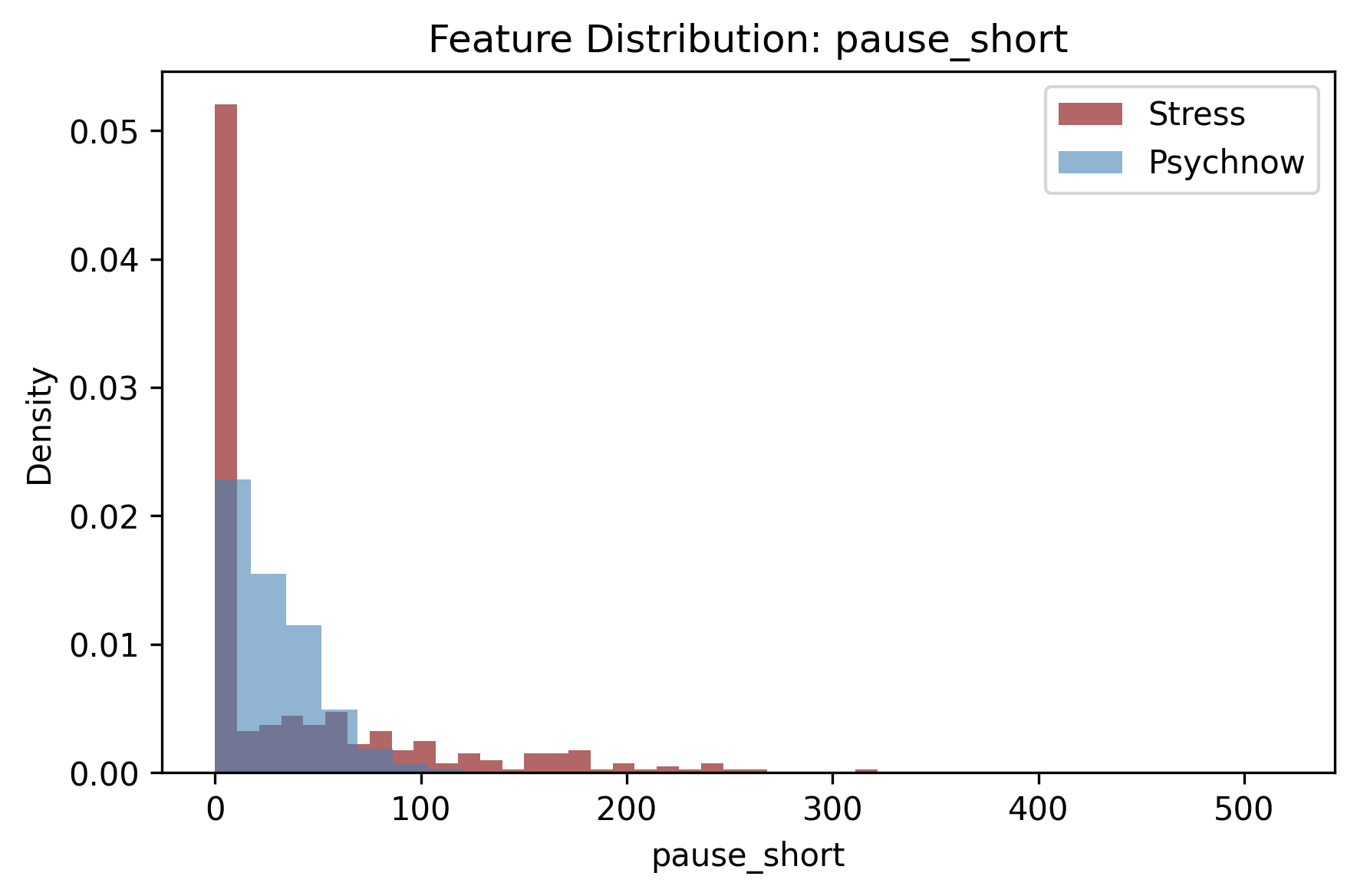} &
        \includegraphics[width=0.09\textwidth]{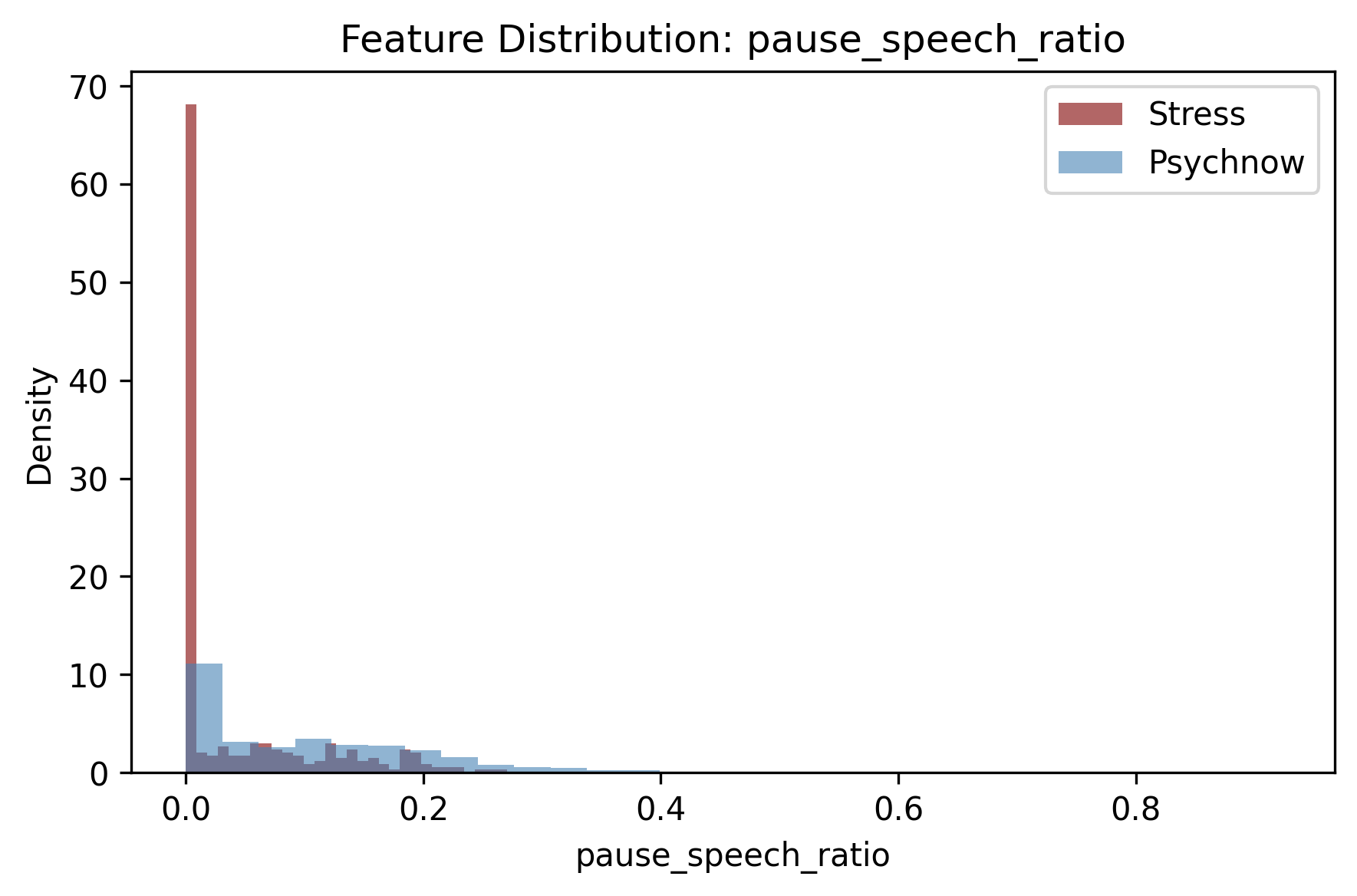} &
        \includegraphics[width=0.09\textwidth]{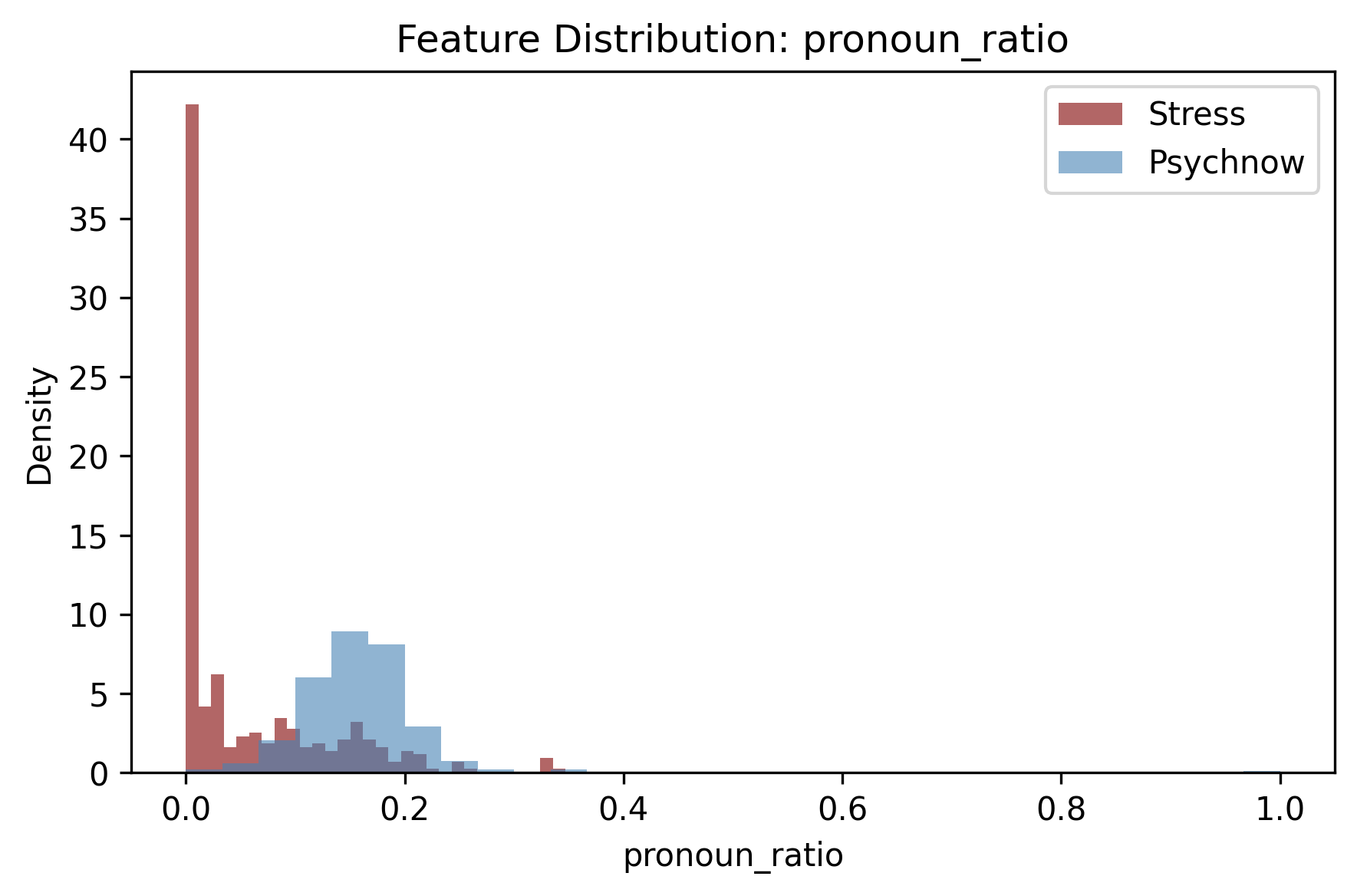} &
        \includegraphics[width=0.09\textwidth]{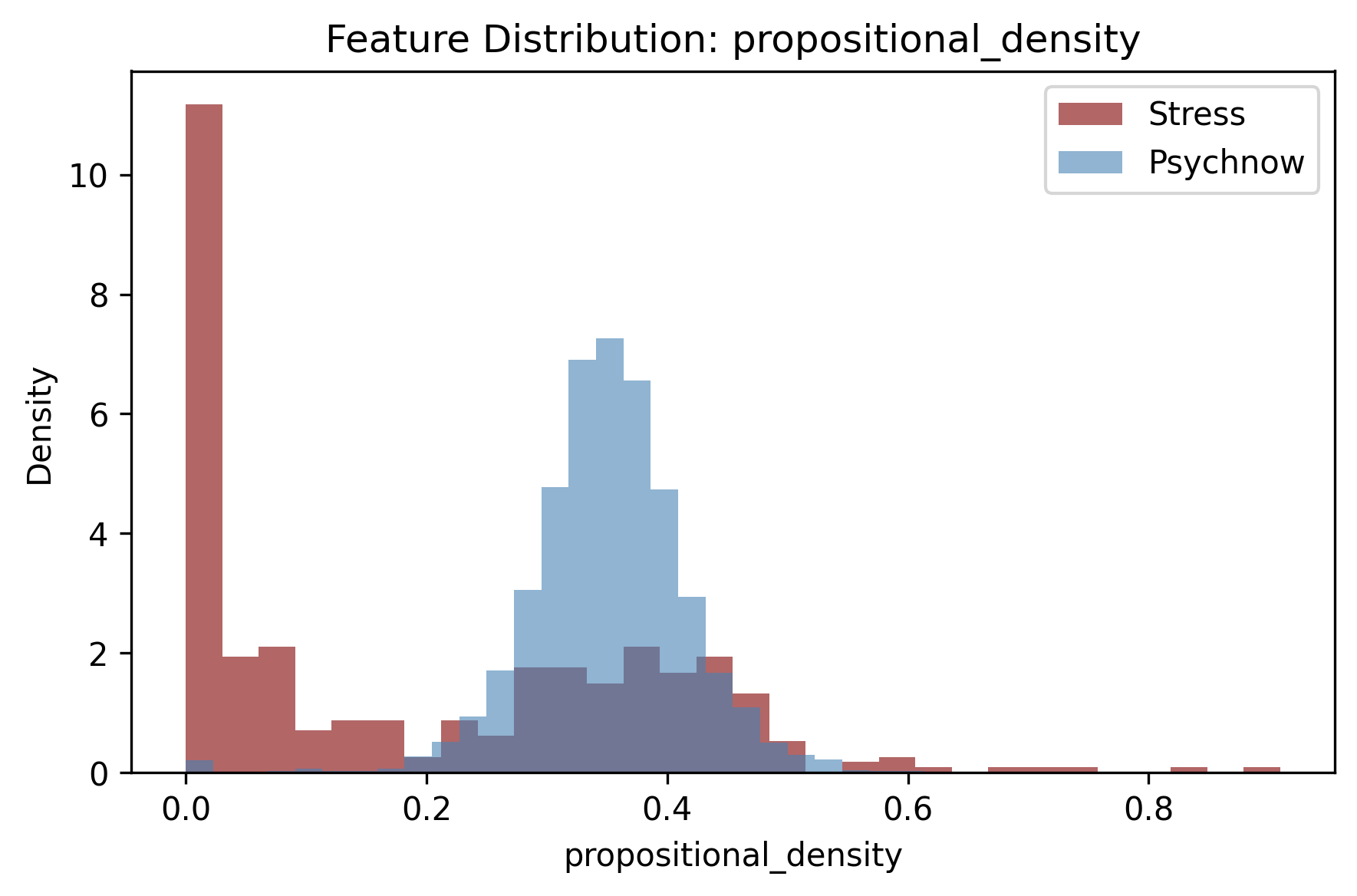} &
        \includegraphics[width=0.09\textwidth]{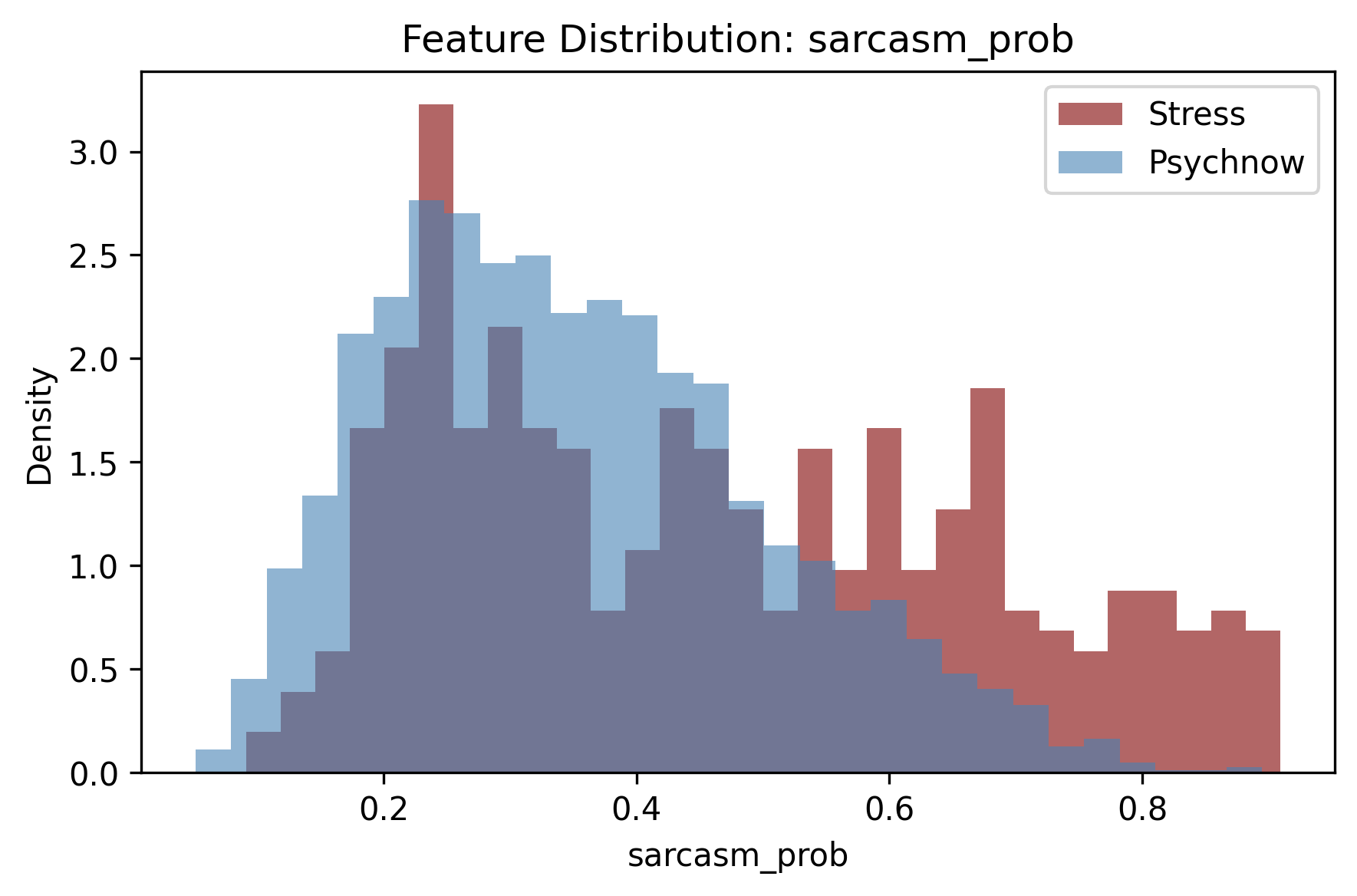} &
        \includegraphics[width=0.09\textwidth]{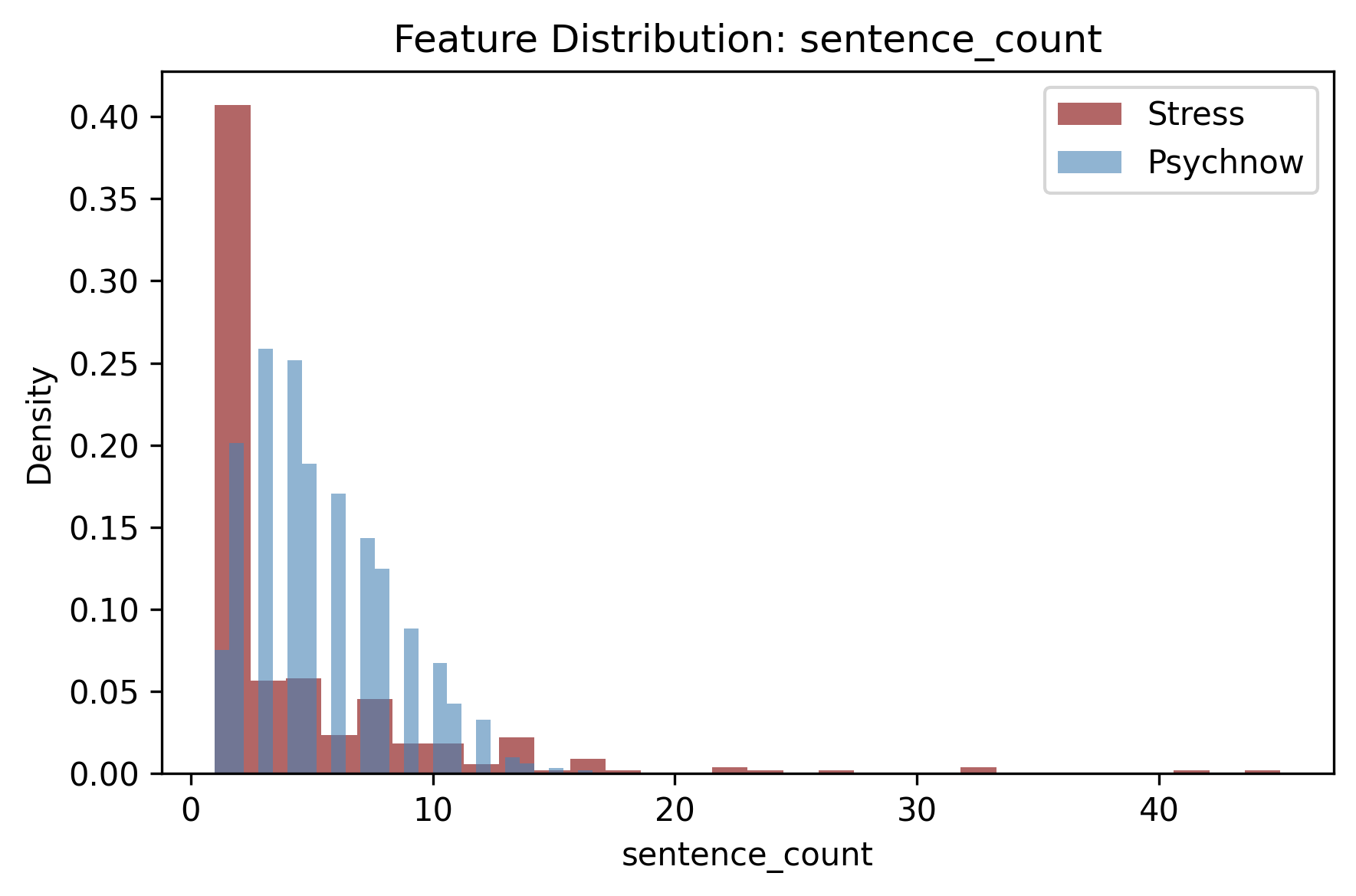} &
        \includegraphics[width=0.09\textwidth]{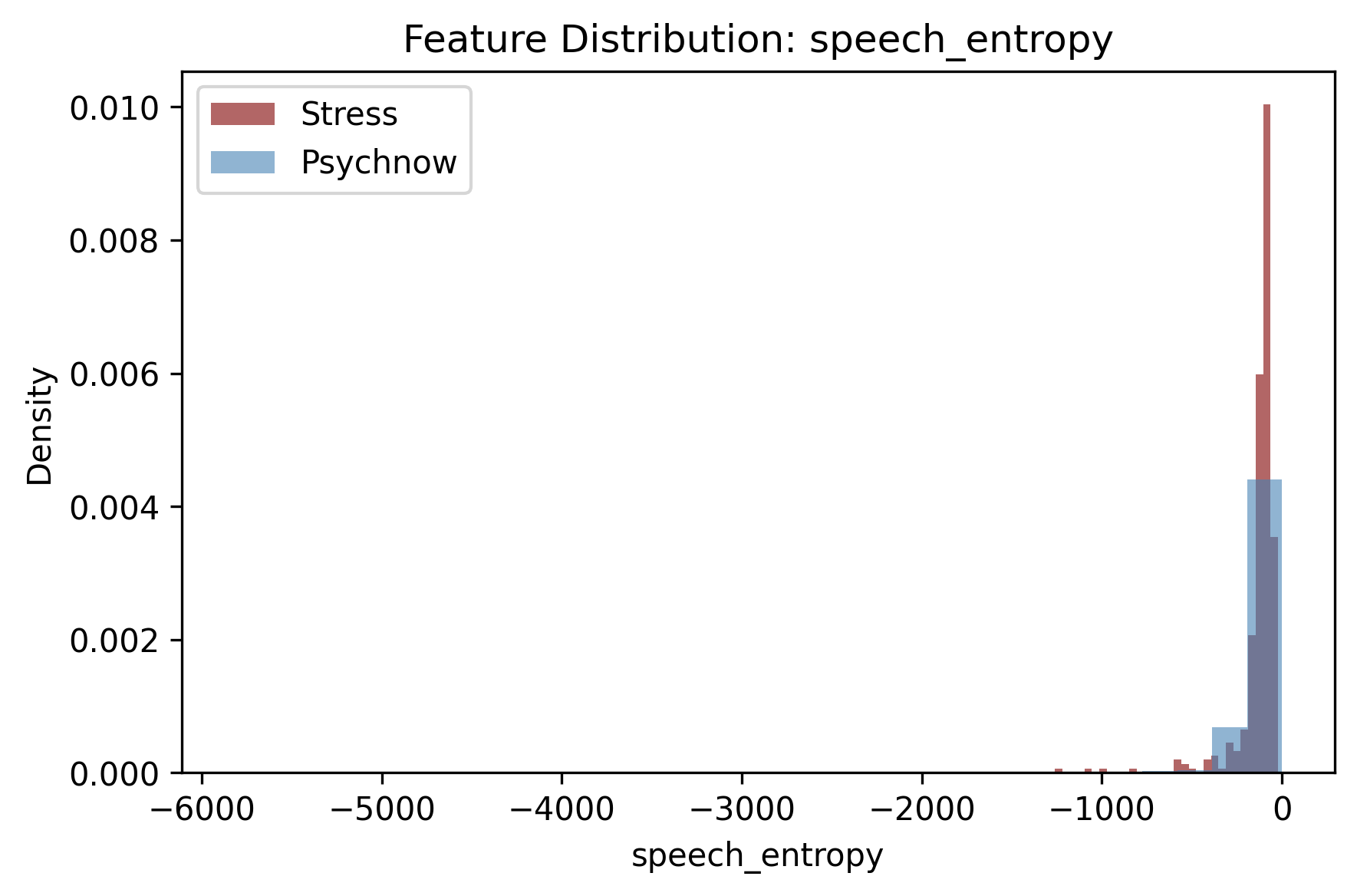} \\[-2pt]

        \includegraphics[width=0.09\textwidth]{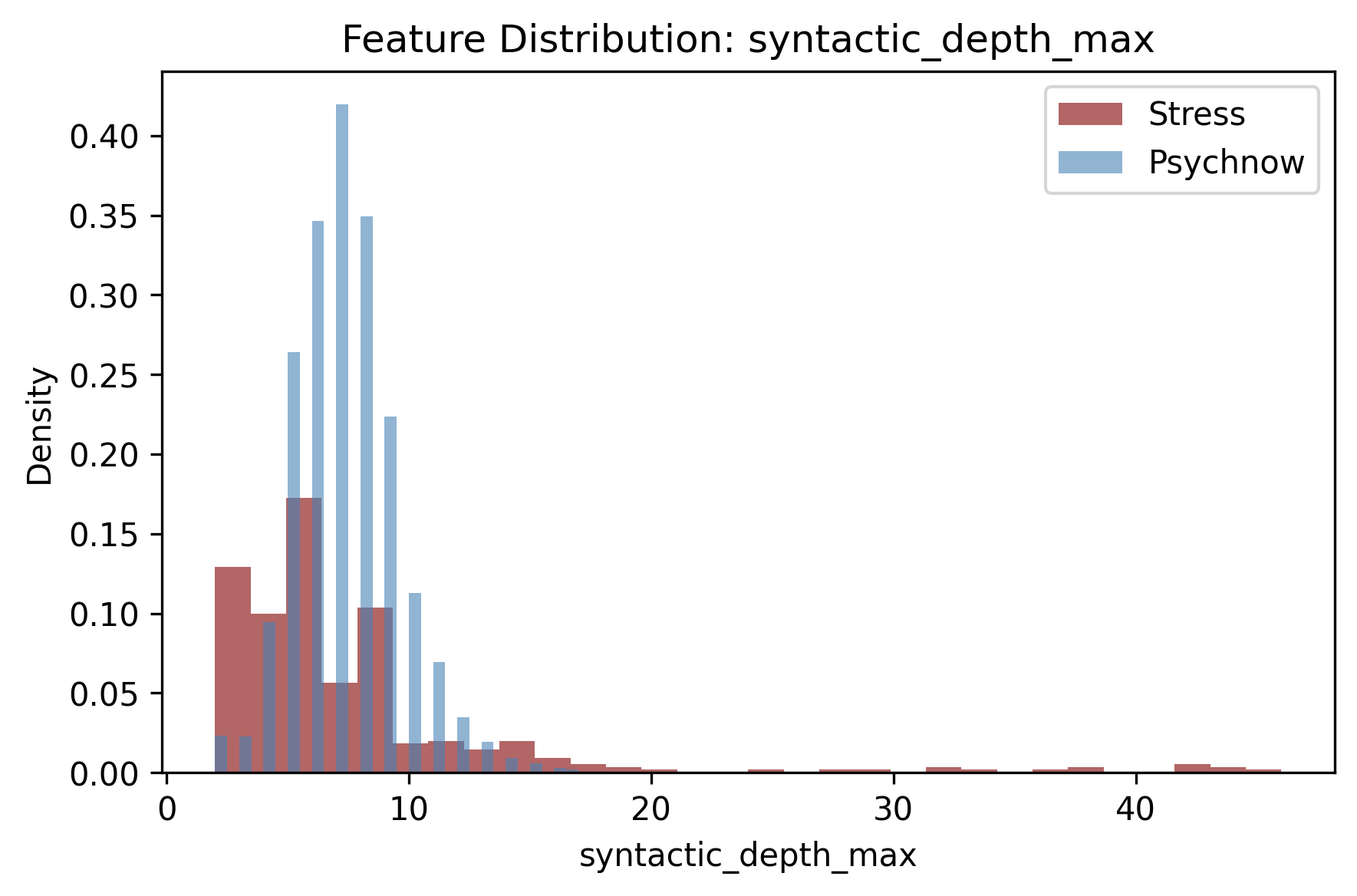} &
        \includegraphics[width=0.09\textwidth]{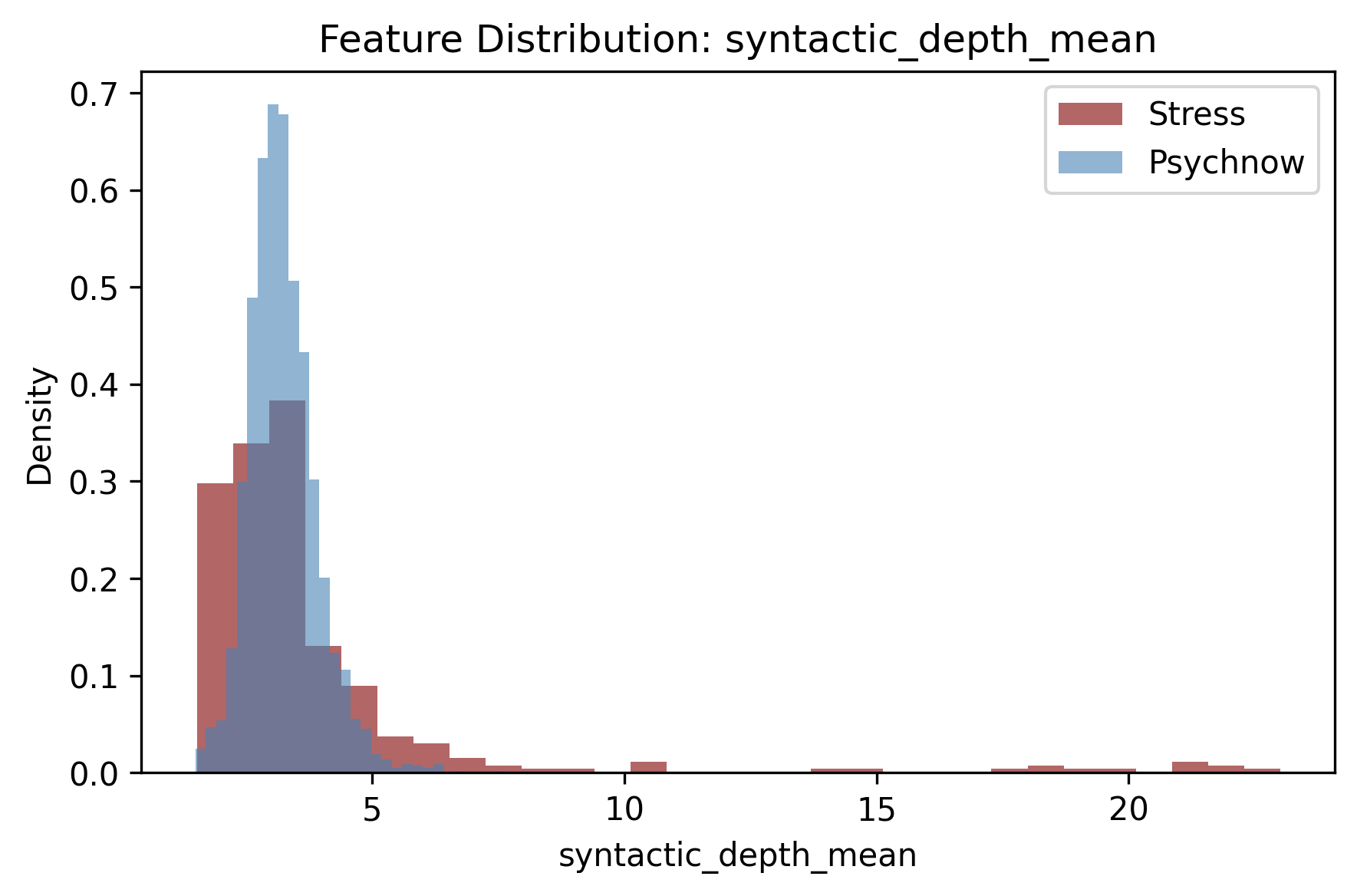} &
        \includegraphics[width=0.09\textwidth]{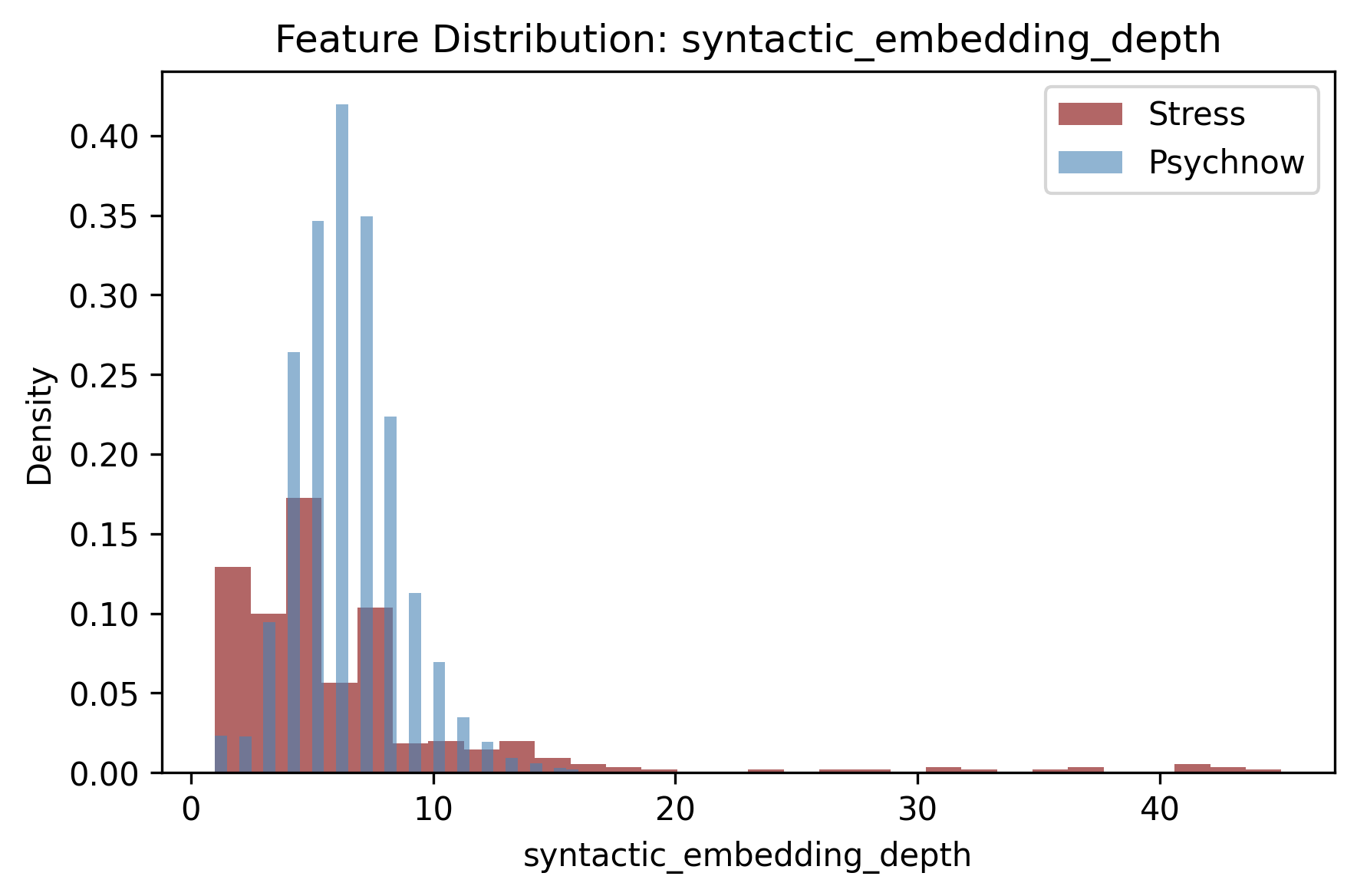} &
        \includegraphics[width=0.09\textwidth]{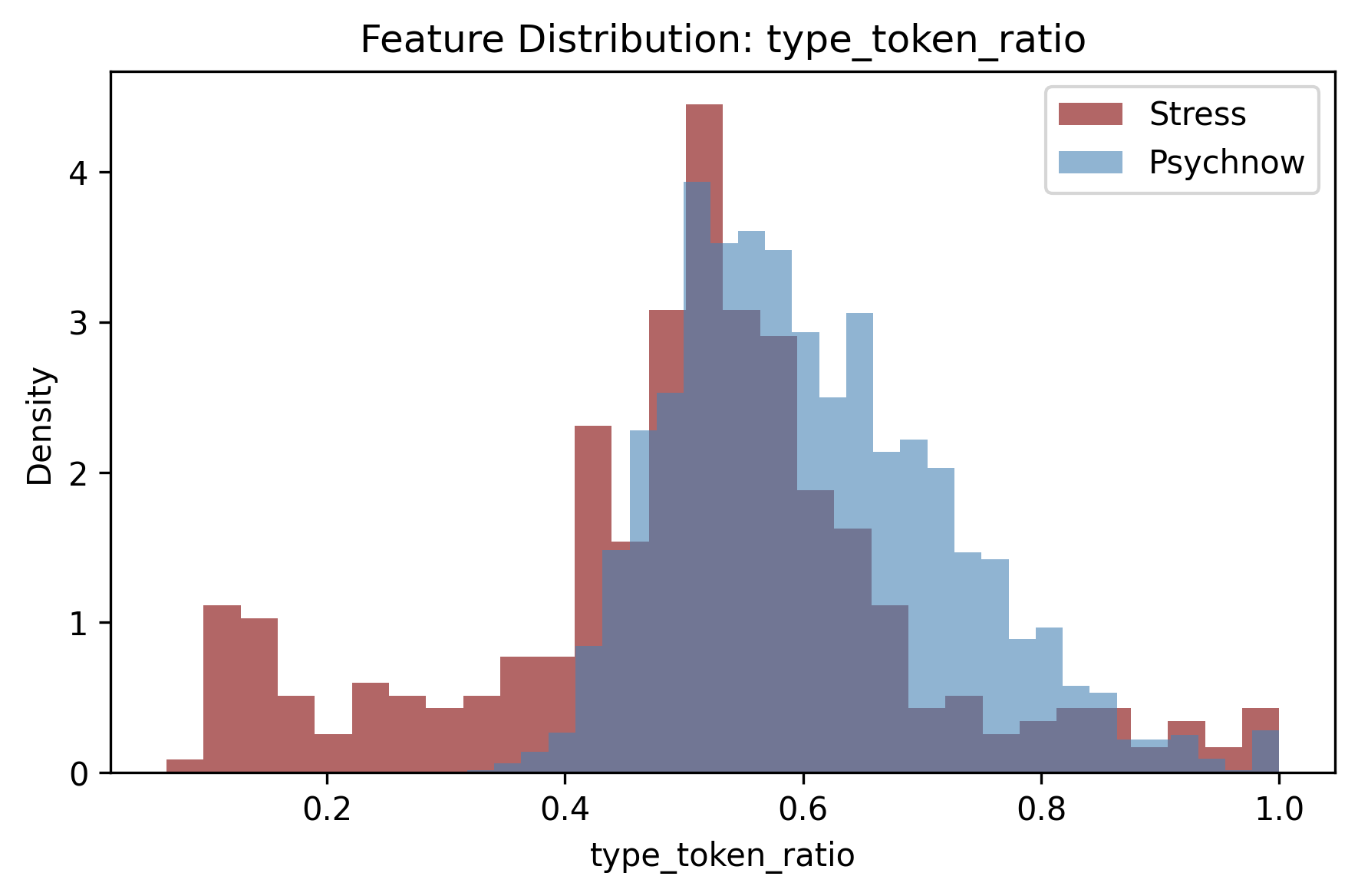} &
        \includegraphics[width=0.09\textwidth]{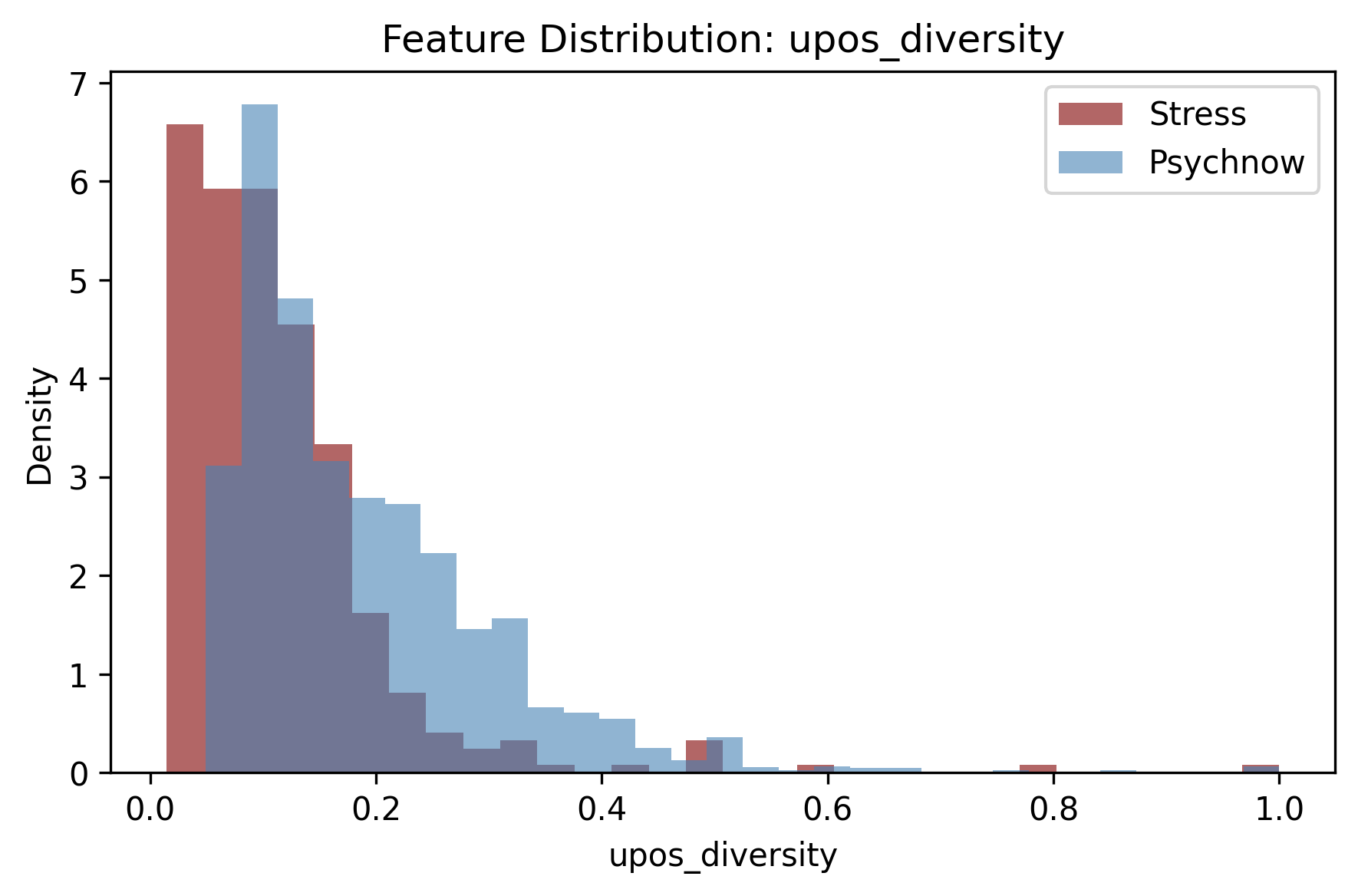} &
        \includegraphics[width=0.09\textwidth]{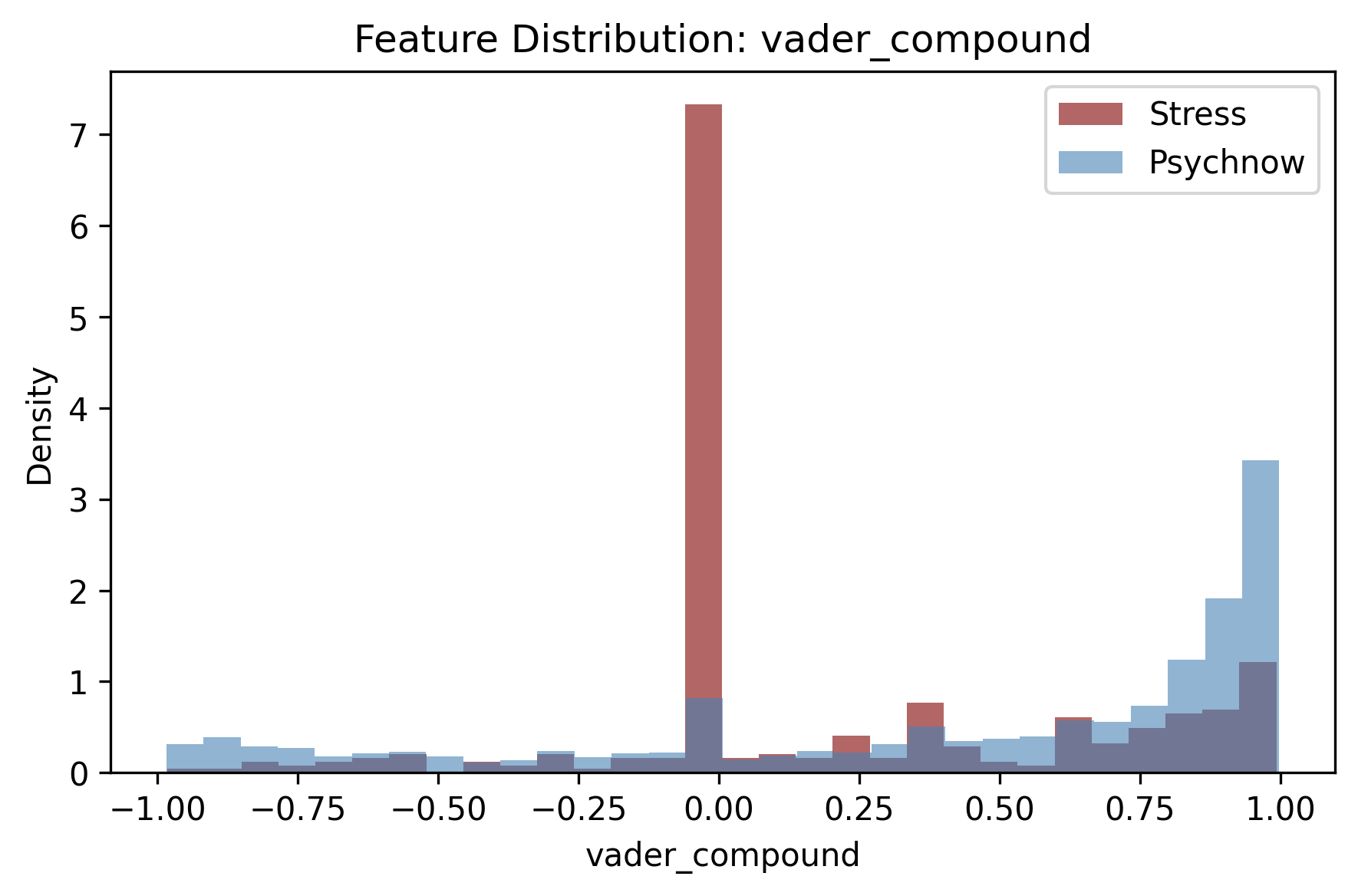} &
        \includegraphics[width=0.09\textwidth]{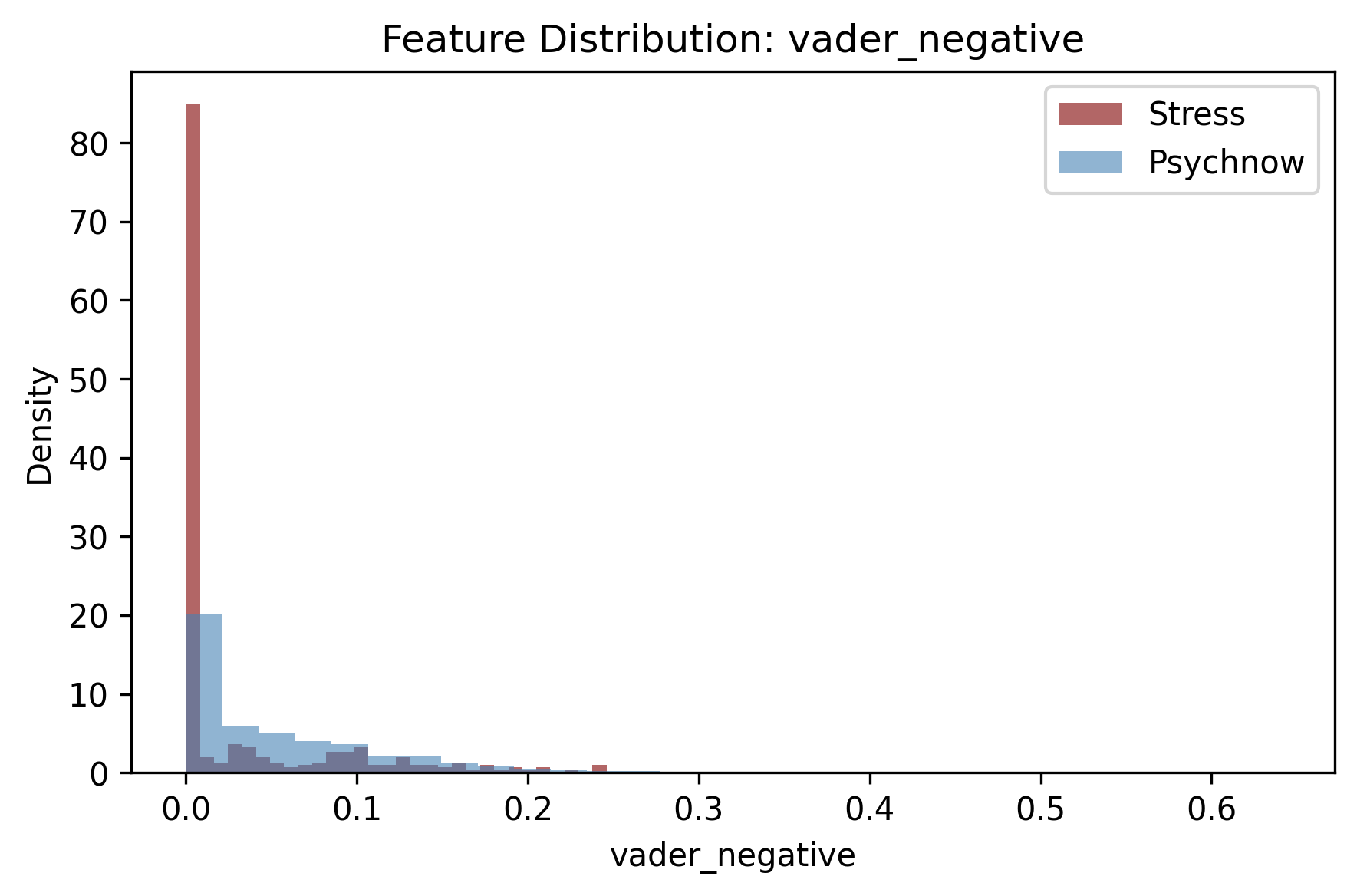} &
        \includegraphics[width=0.09\textwidth]{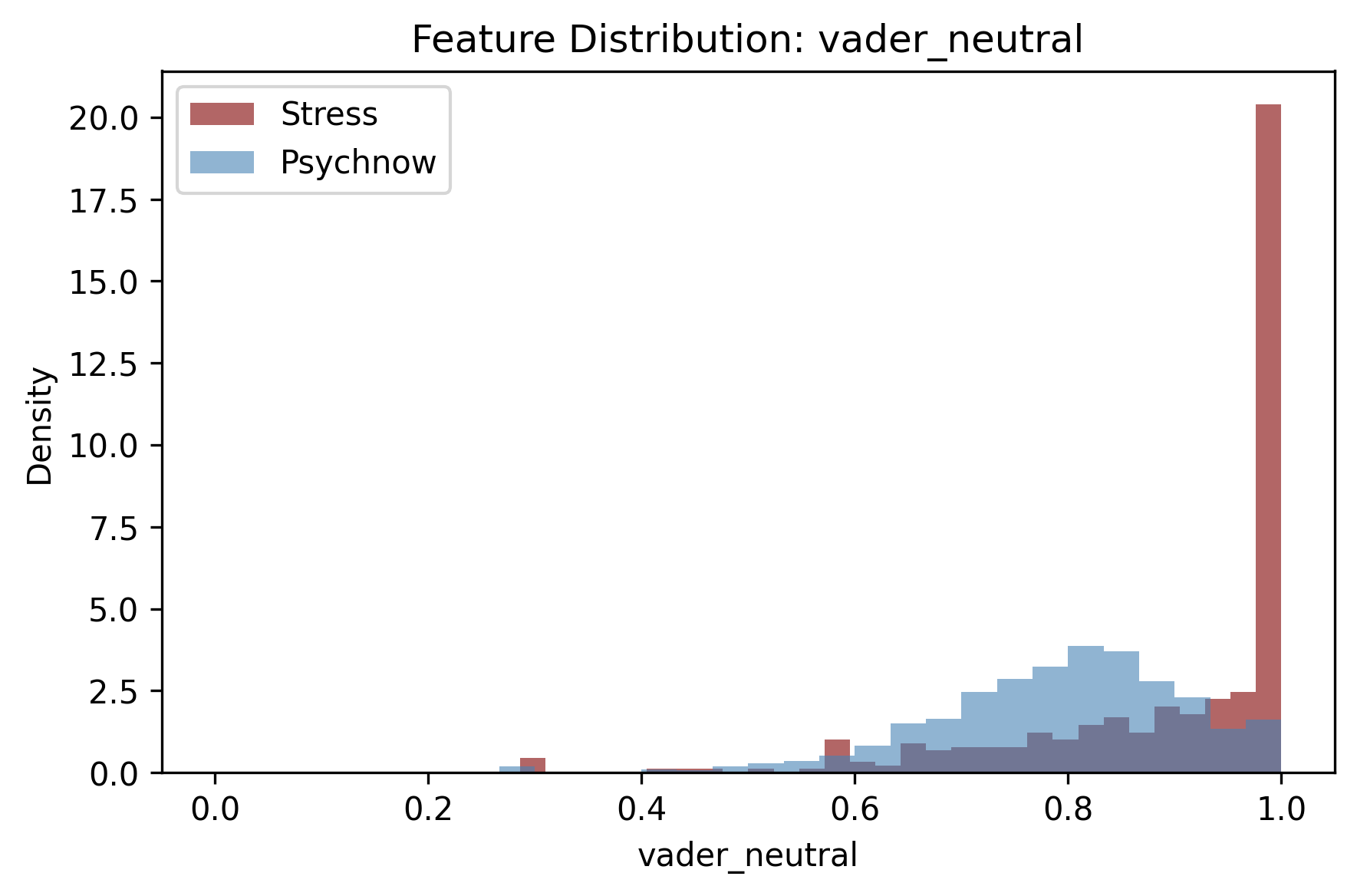} &
        \includegraphics[width=0.09\textwidth]{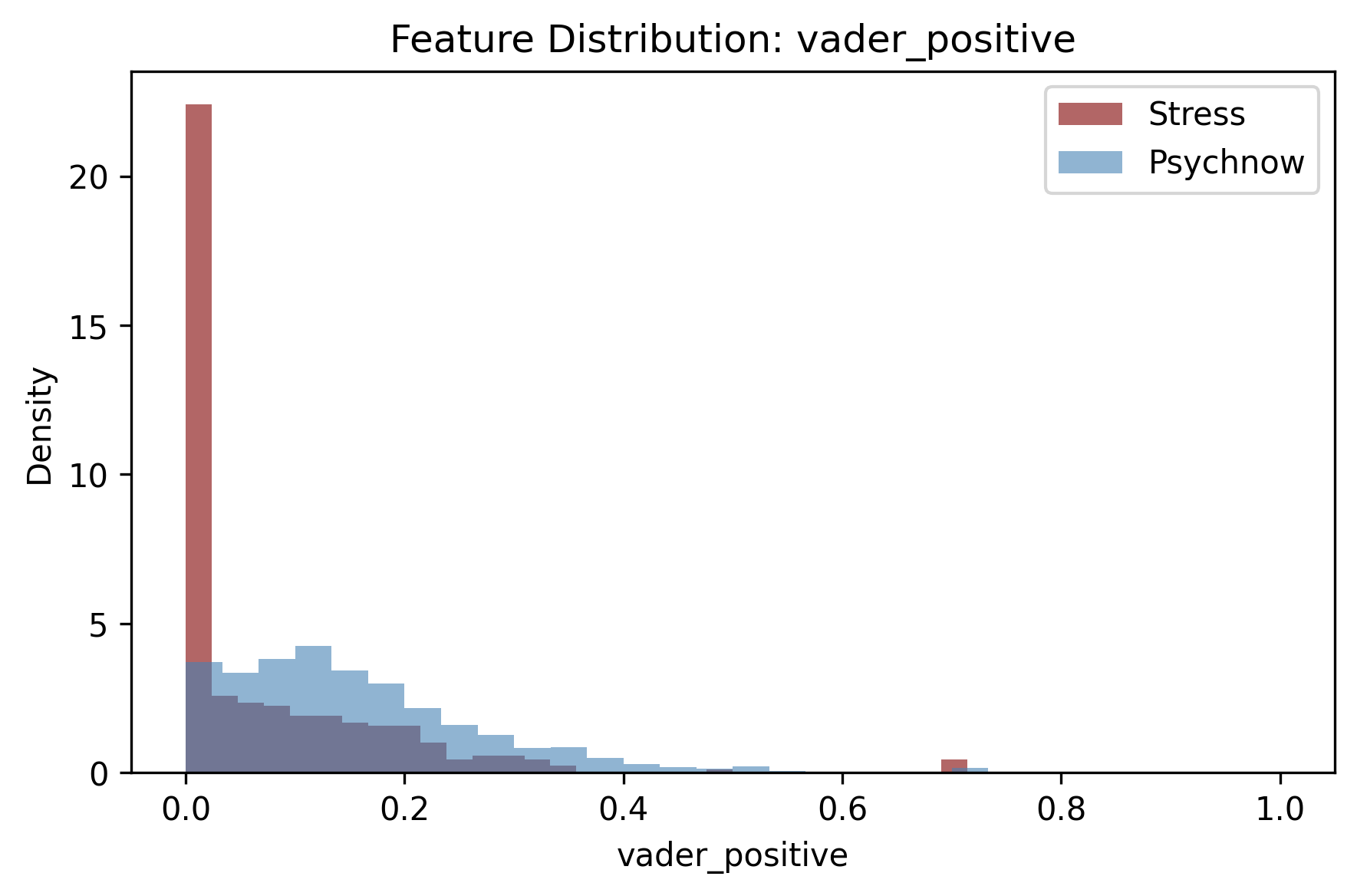} &
        \includegraphics[width=0.09\textwidth]{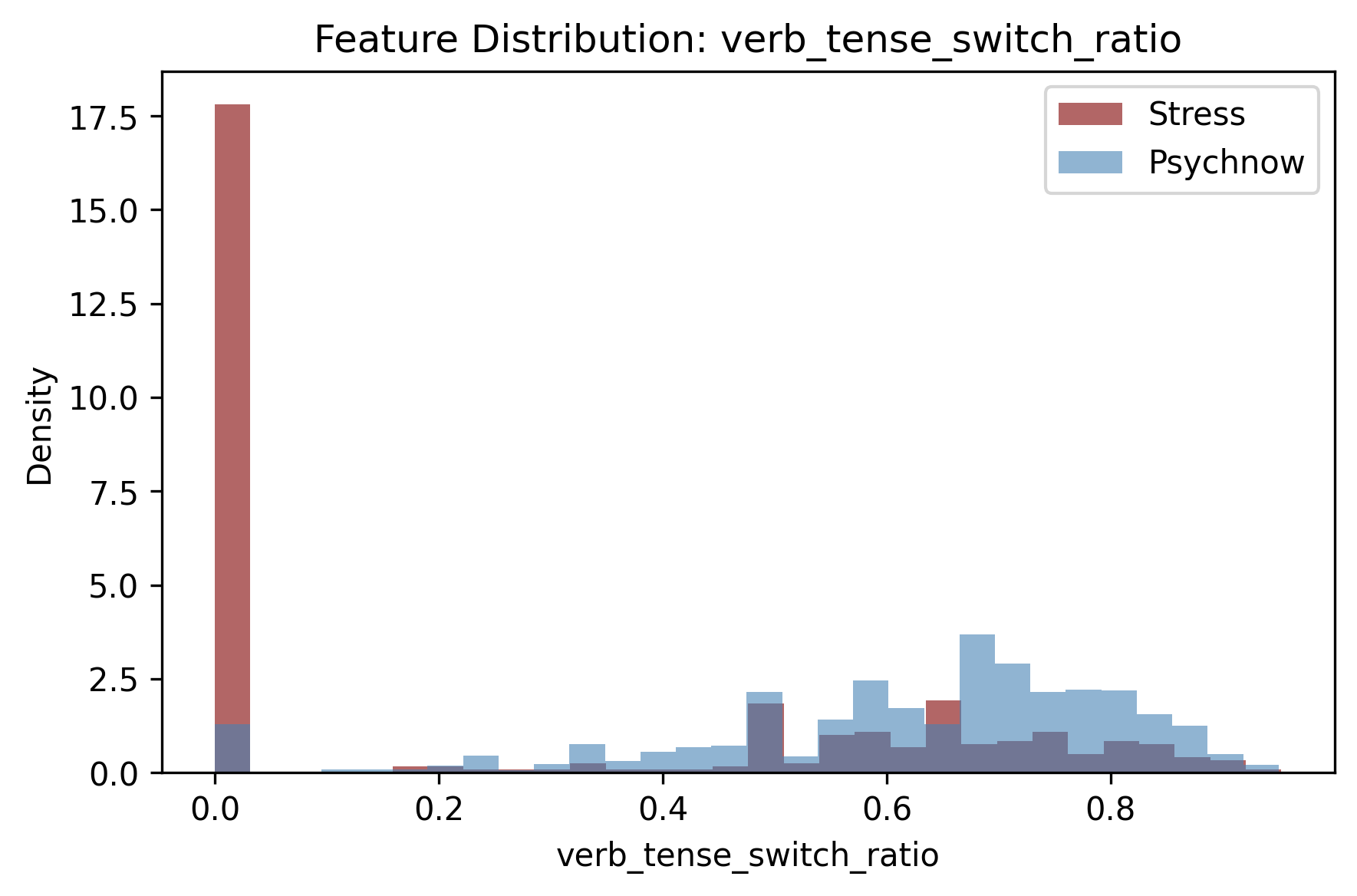} \\[-2pt]

        \includegraphics[width=0.09\textwidth]{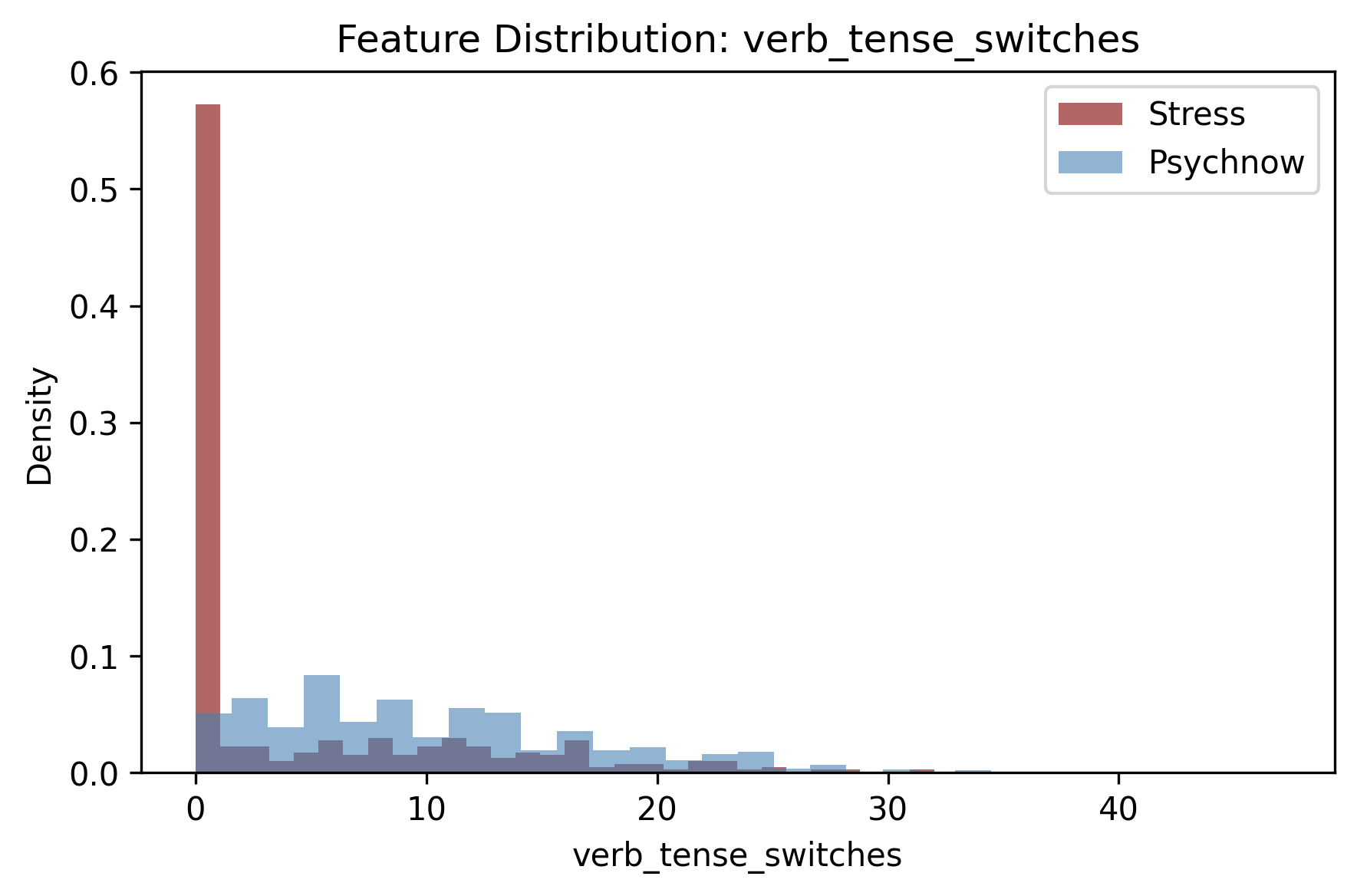} &
        \includegraphics[width=0.09\textwidth]{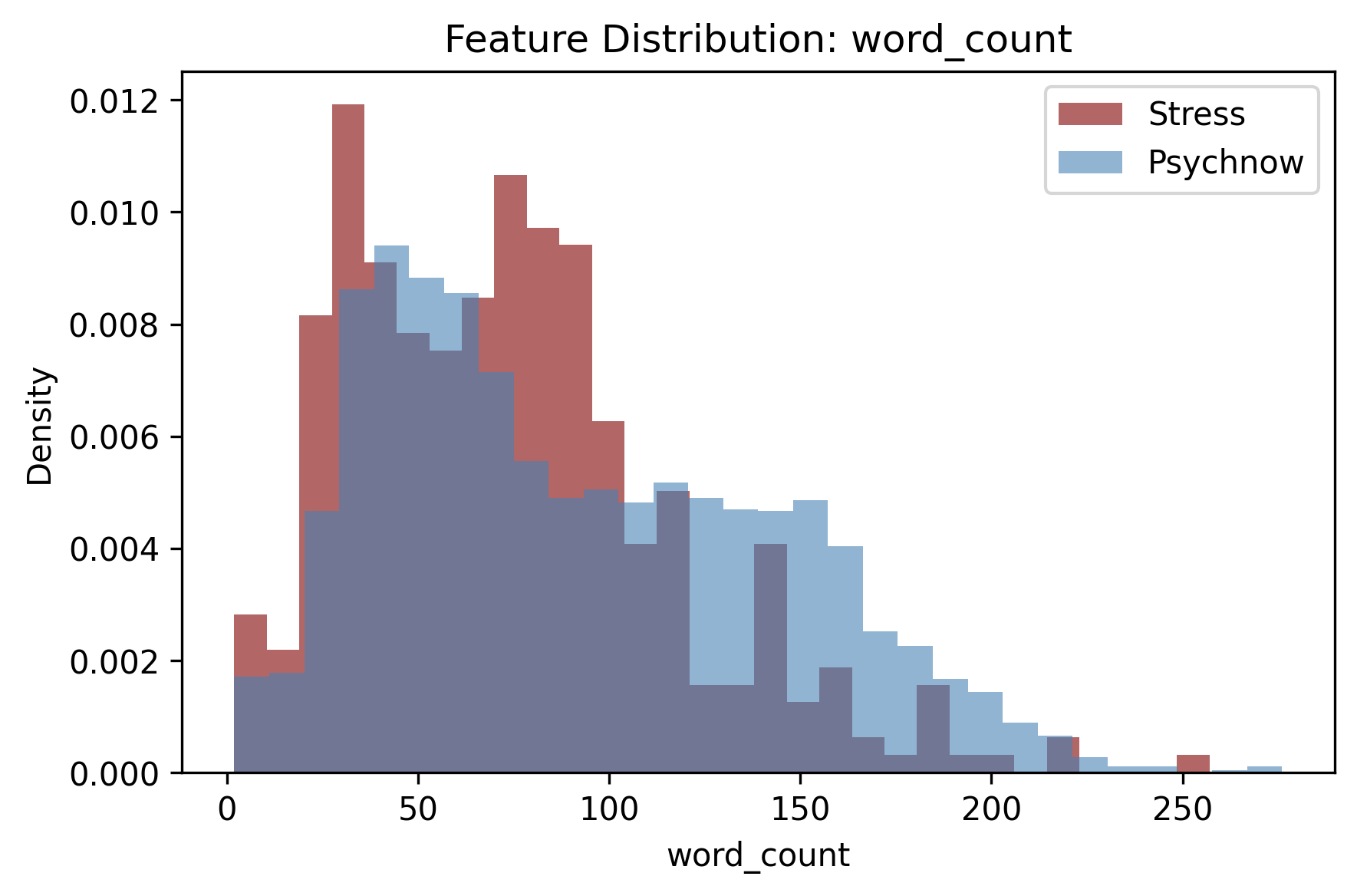}
    \end{tabular}

    \caption{Comparison of feature value distributions between the \textit{Stress} (maroon) and \textsc{Real} (steelblue) datasets across 82 extracted features.}
    \label{fig:feature_distributions_all}
\end{figure*}

To compare the feature distributions between the \textit{Stress} and \textsc{Real} datasets, 
we visualize the normalized histograms for each extracted feature (see Figure~\ref{fig:feature_distributions_all}). 
Red denotes the Stress dataset, while blue corresponds to \textsc{Real}. Overall, the majority of features display similar distributional patterns between the \textit{Stress} and \textsc{Real} datasets. 
This consistency supports the validity and reliability of our feature extraction procedure, suggesting that the computed features 
capture comparable underlying characteristics across recording conditions.

\section{Partial Dependence Plots}

Partial Dependence Plots (PDPs) illustrate the marginal effect of individual features on model predictions while averaging over all other variables. By doing so, they reveal whether a given feature has a positive, negative, or non-linear influence on the target outcome. PDPs are particularly informative for complex, non-linear models such as ensemble or deep learning architectures, as they provide interpretable insight into learned feature-outcome relationships.

\begin{figure}[p]
  \centering
  \resizebox{0.45\textwidth}{0.45\textheight}{%
    \begin{tabular}{ccc}
      \includegraphics[width=0.32\textwidth]{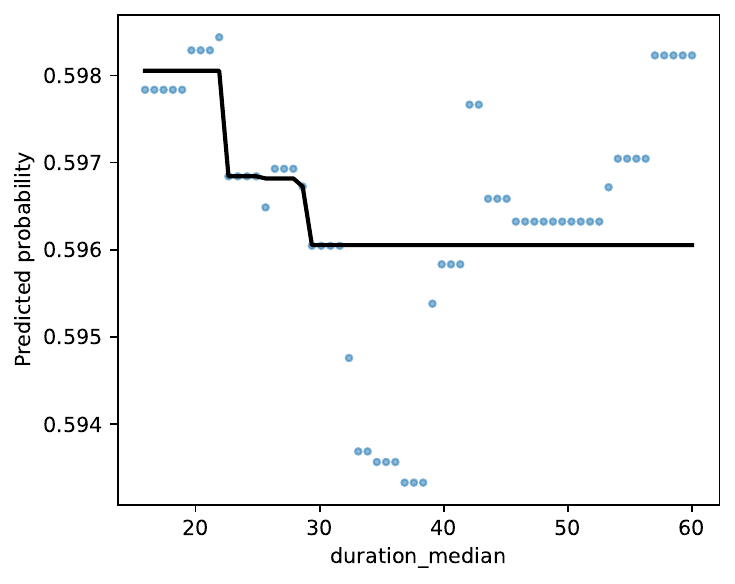} &
      \includegraphics[width=0.32\textwidth]{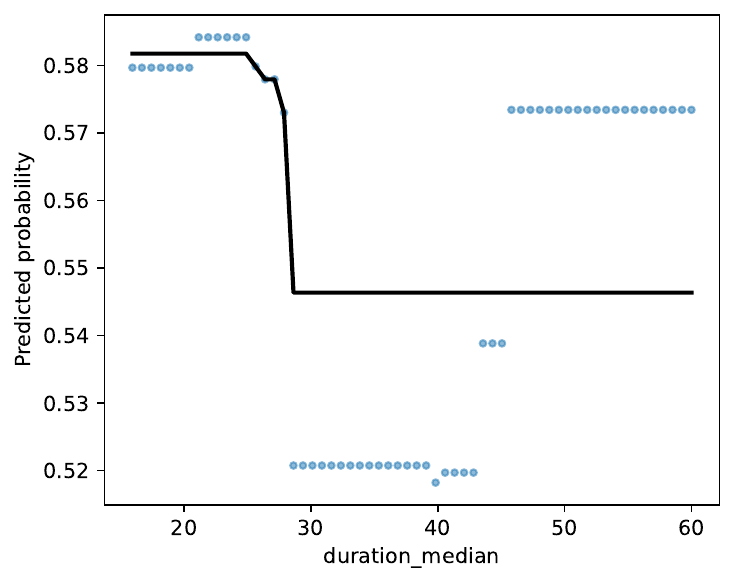} &
      \includegraphics[width=0.32\textwidth]{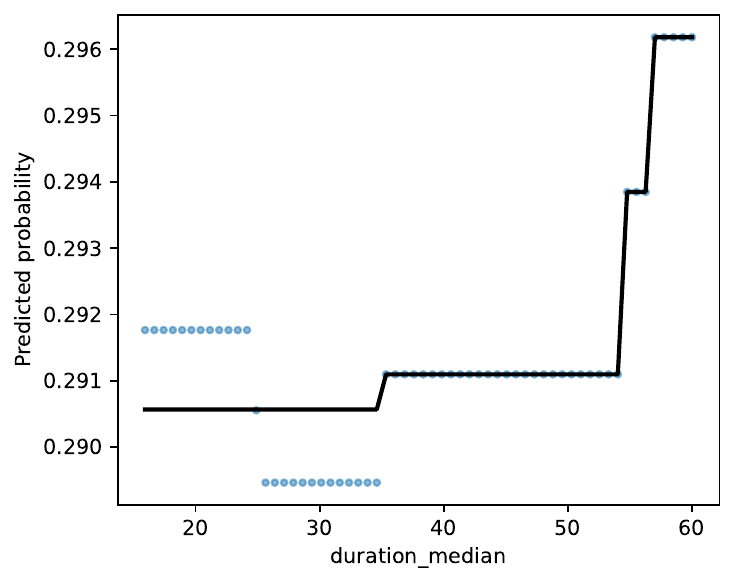} \\

      \includegraphics[width=0.32\textwidth]{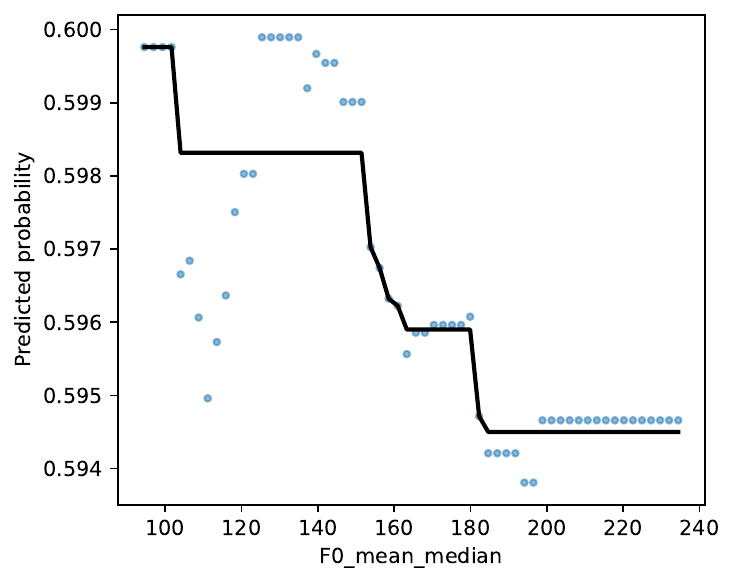} &
      \includegraphics[width=0.32\textwidth]{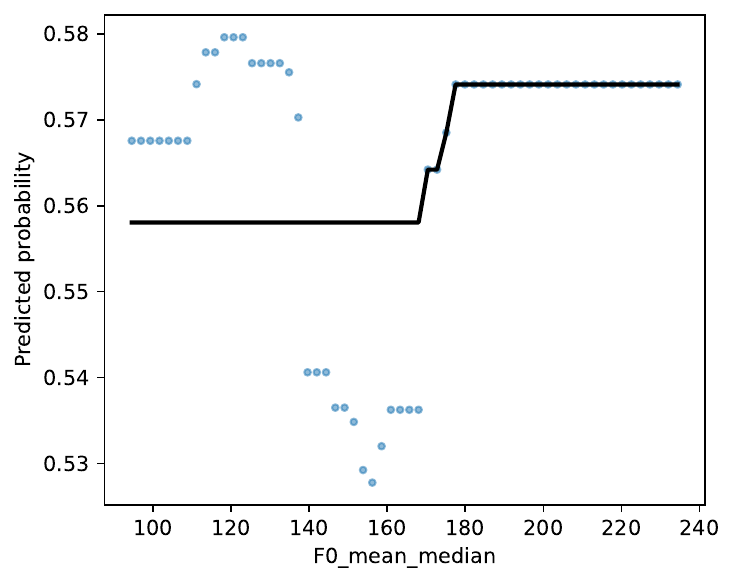} &
      \includegraphics[width=0.32\textwidth]{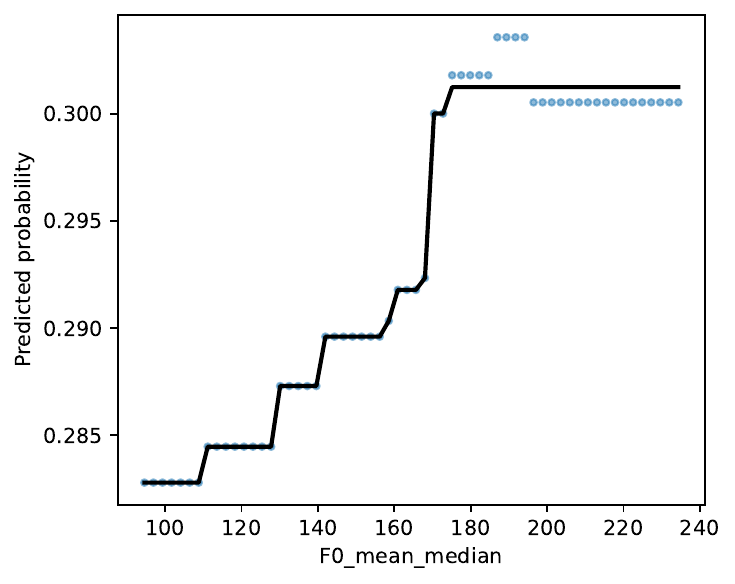} \\

      \includegraphics[width=0.32\textwidth]{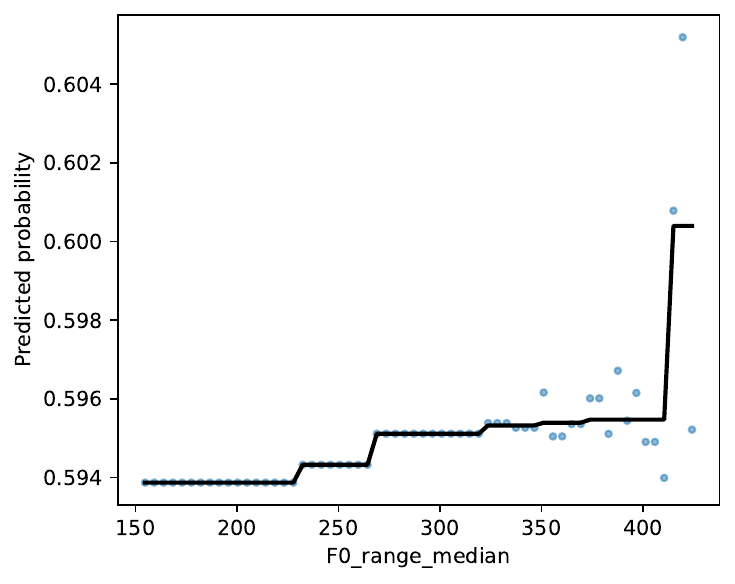} &
      \includegraphics[width=0.32\textwidth]{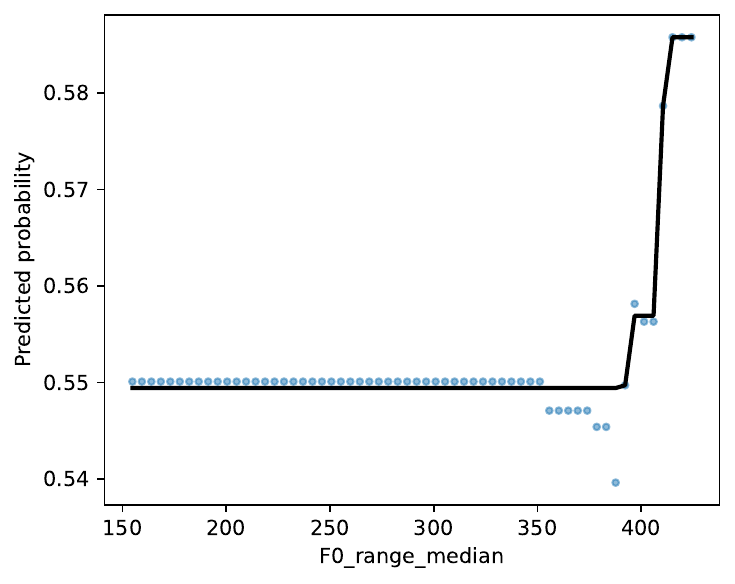} &
      \includegraphics[width=0.32\textwidth]{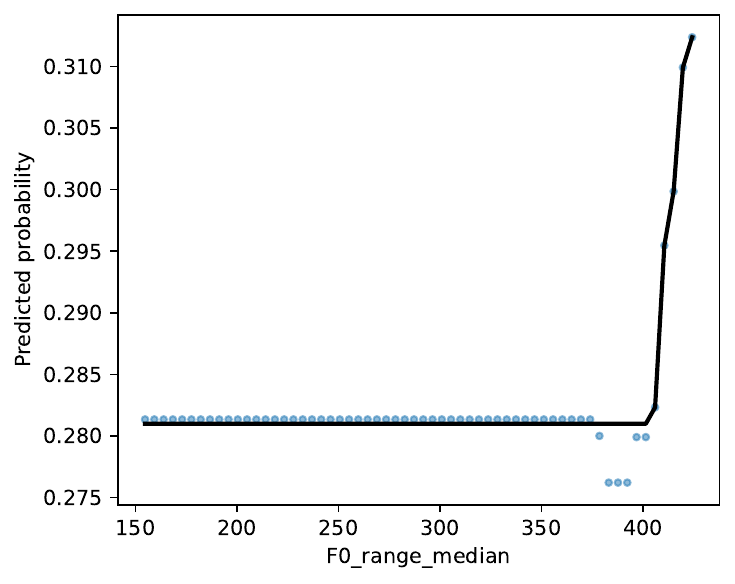} \\

      \includegraphics[width=0.32\textwidth]{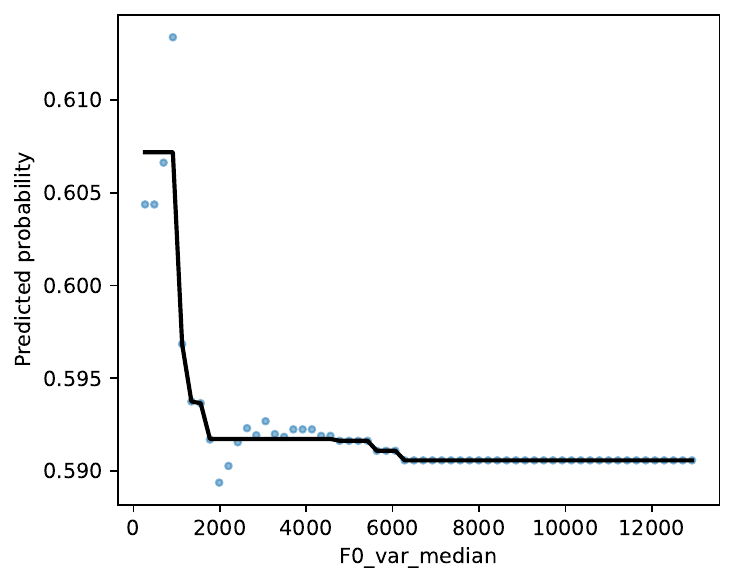} &
      \includegraphics[width=0.32\textwidth]{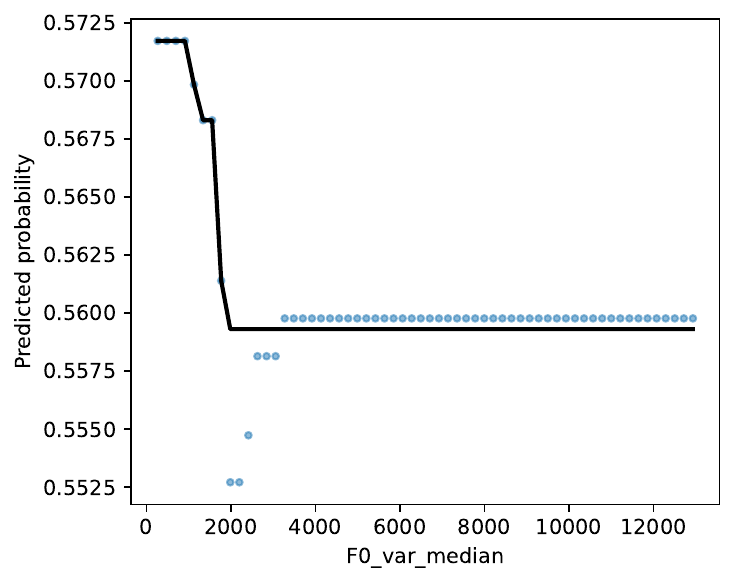} &
      \includegraphics[width=0.32\textwidth]{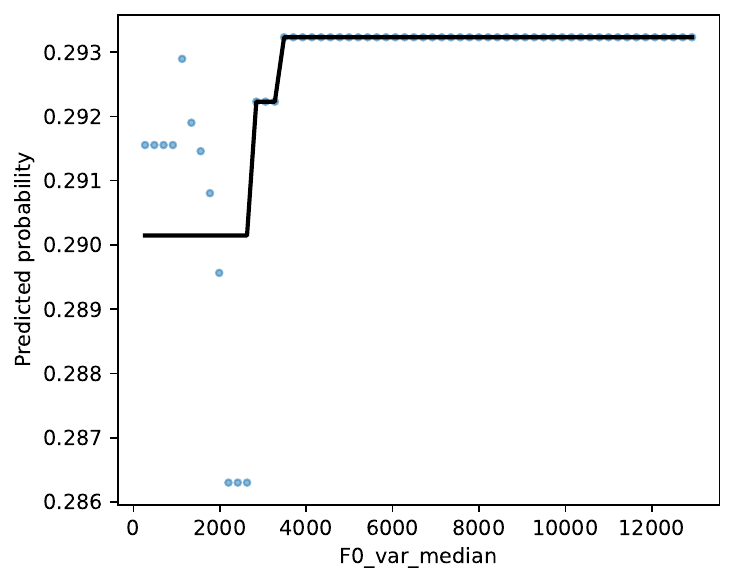} \\

      \includegraphics[width=0.32\textwidth]{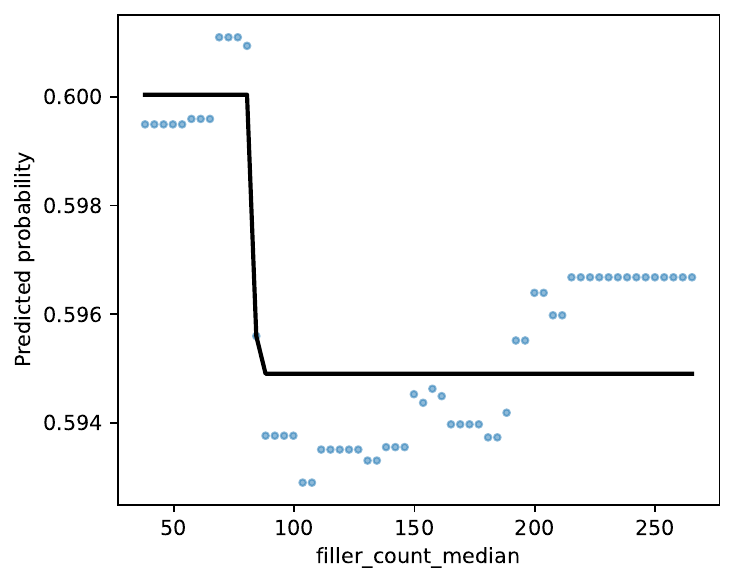} &
      \includegraphics[width=0.32\textwidth]{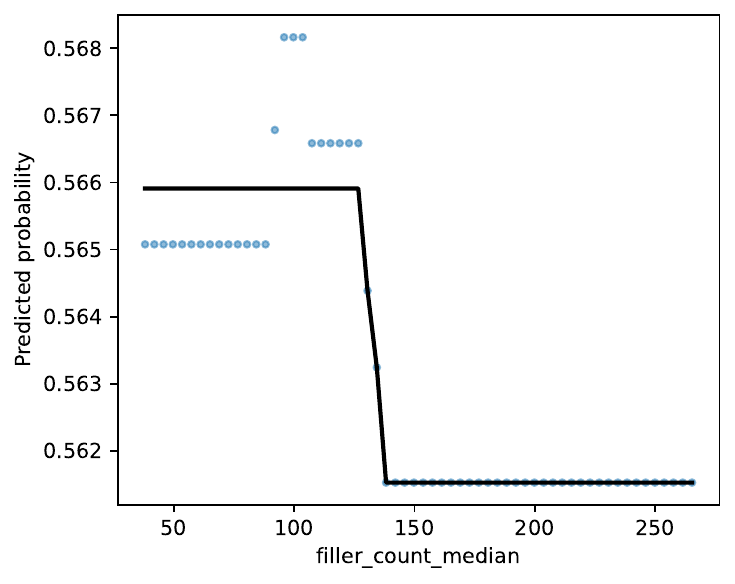} &
      \includegraphics[width=0.32\textwidth]{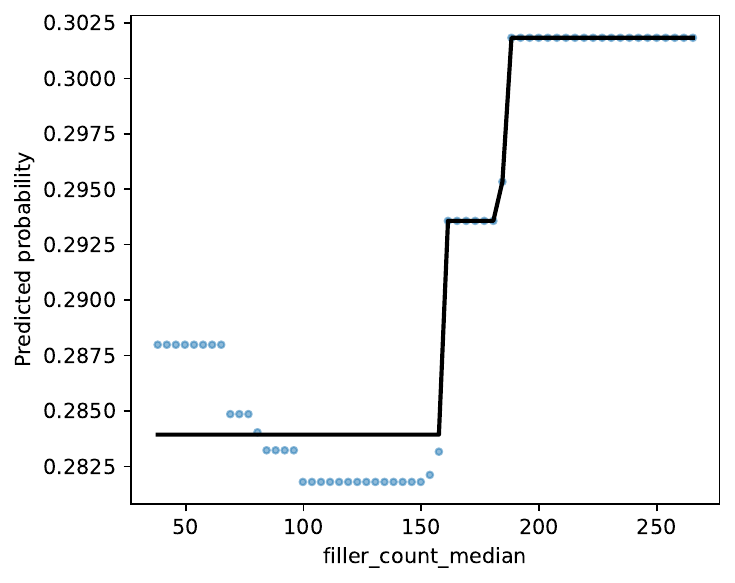} \\

      \includegraphics[width=0.32\textwidth]{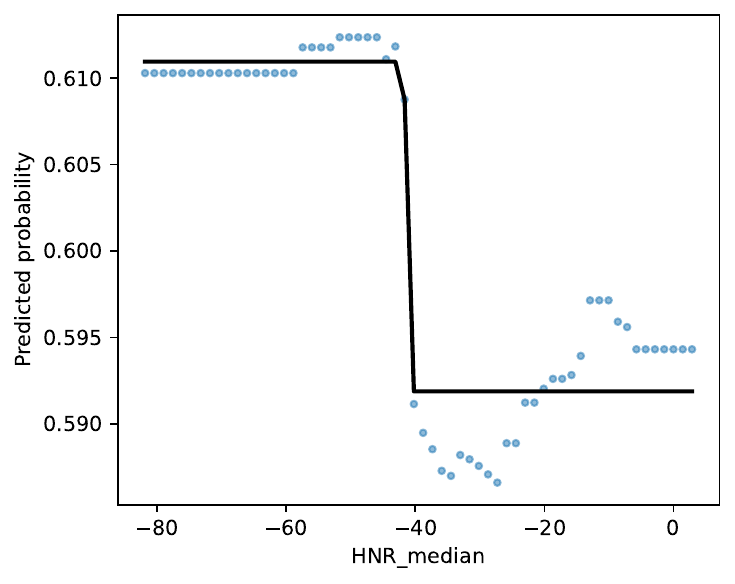} &
      \includegraphics[width=0.32\textwidth]{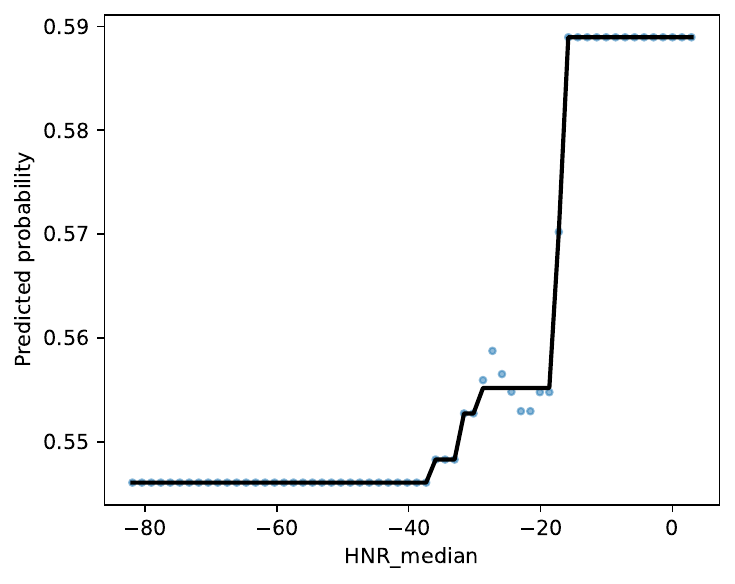} &
      \includegraphics[width=0.32\textwidth]{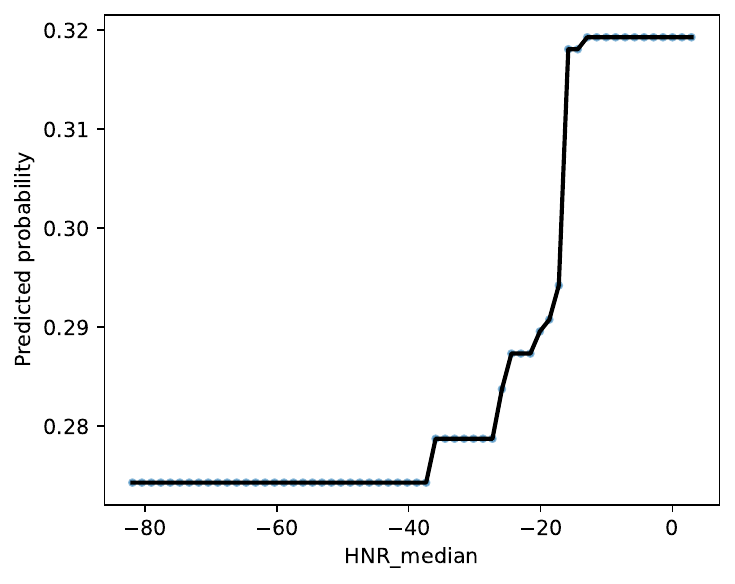} \\

      \includegraphics[width=0.32\textwidth]{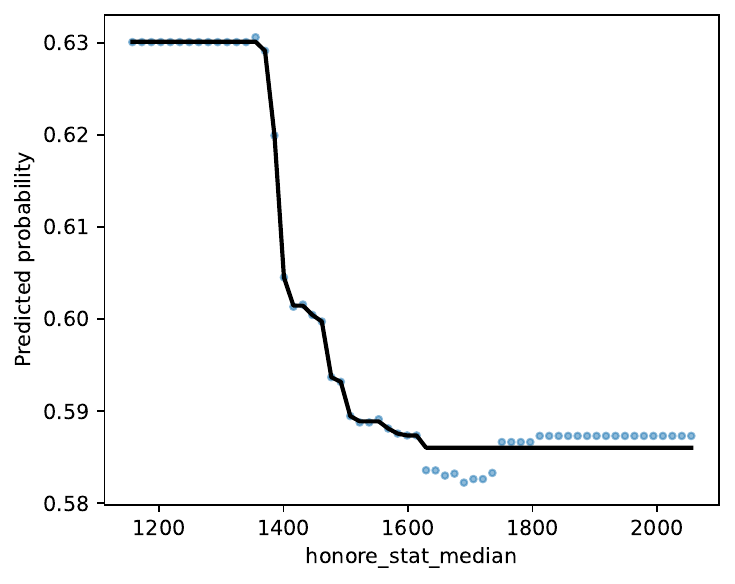} &
      \includegraphics[width=0.32\textwidth]{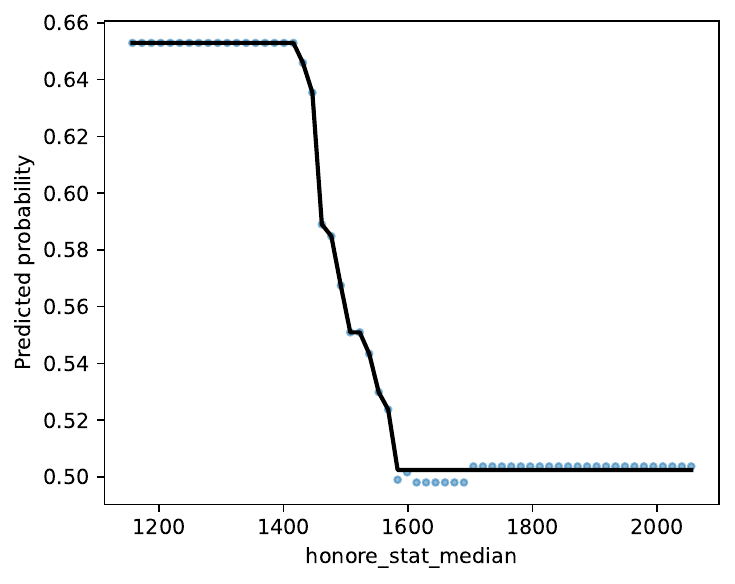} &
      \includegraphics[width=0.32\textwidth]{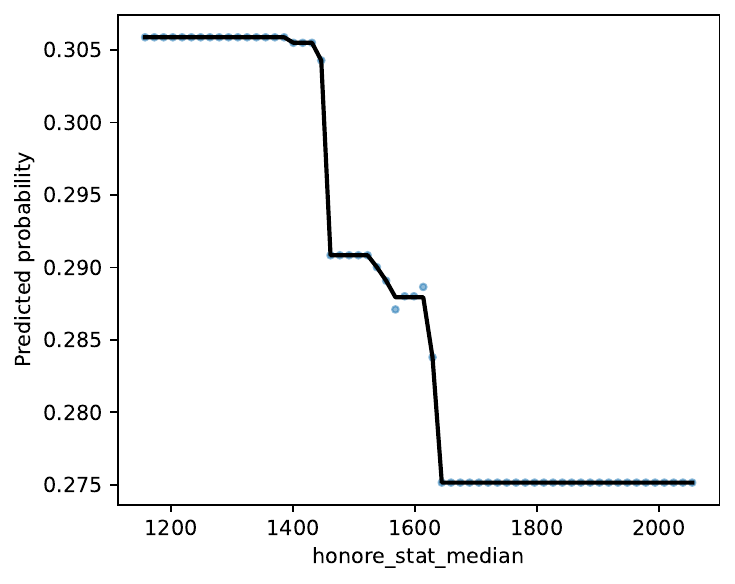} \\

      \includegraphics[width=0.32\textwidth]{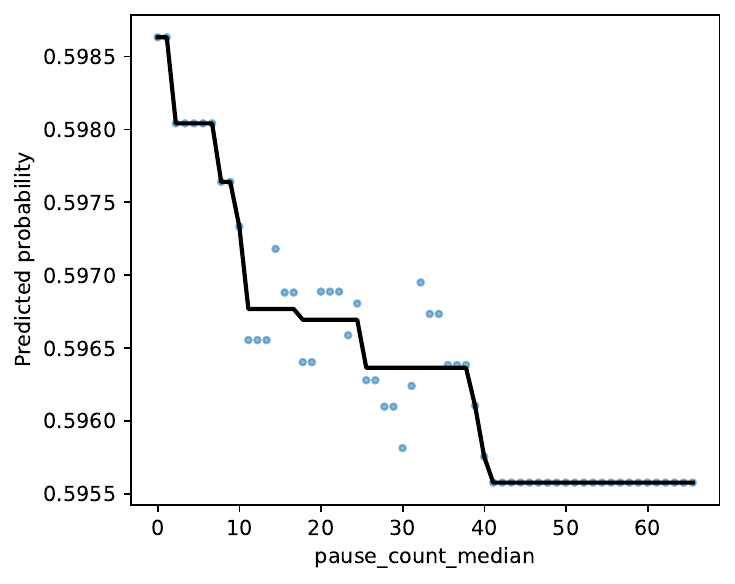} &
      \includegraphics[width=0.32\textwidth]{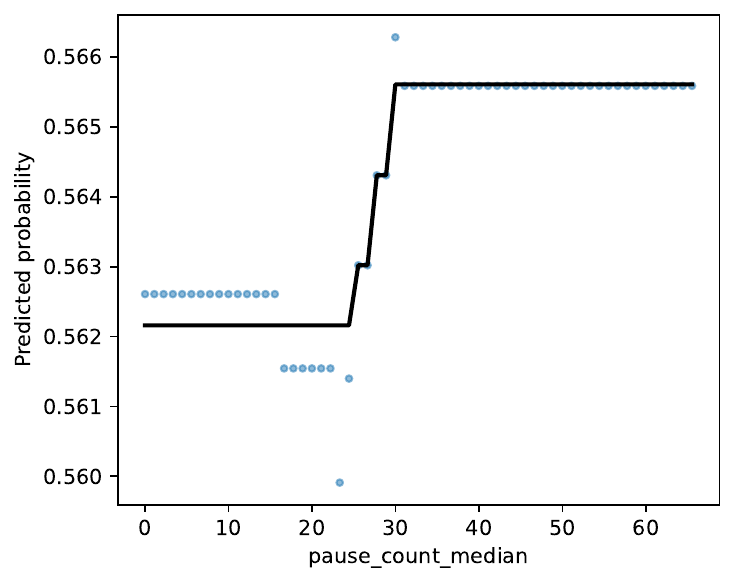} &
      \includegraphics[width=0.32\textwidth]{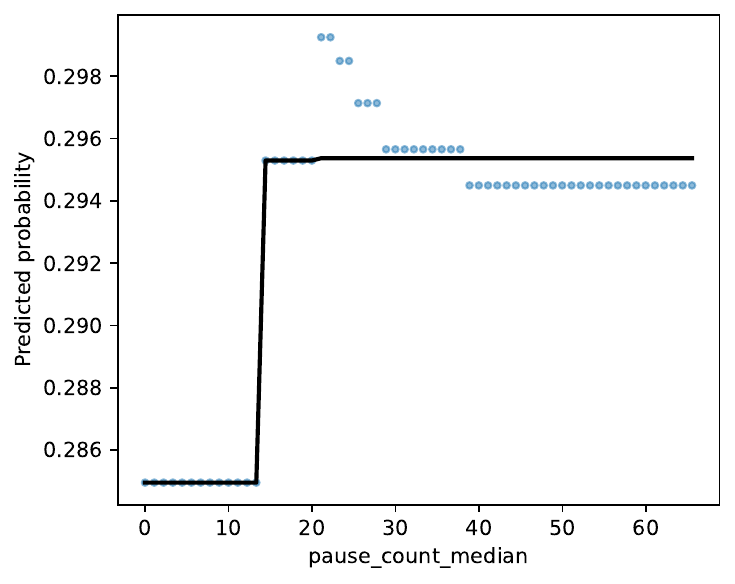} \\

      \includegraphics[width=0.32\textwidth]{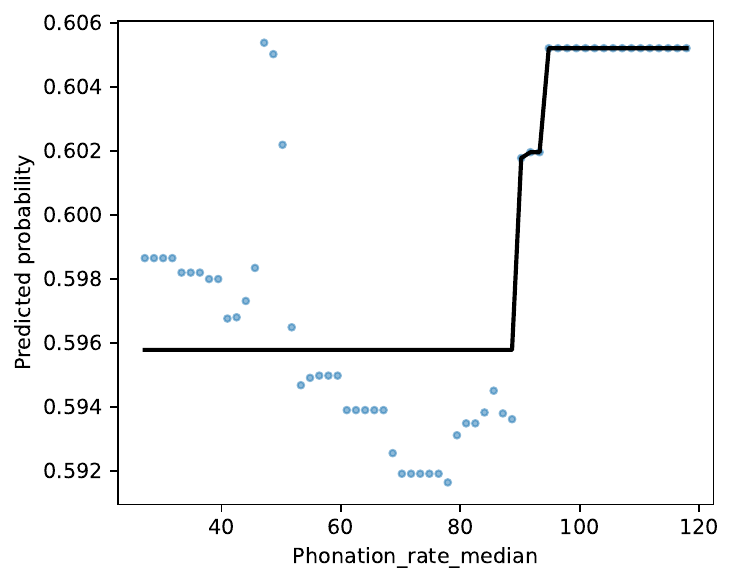} &
      \includegraphics[width=0.32\textwidth]{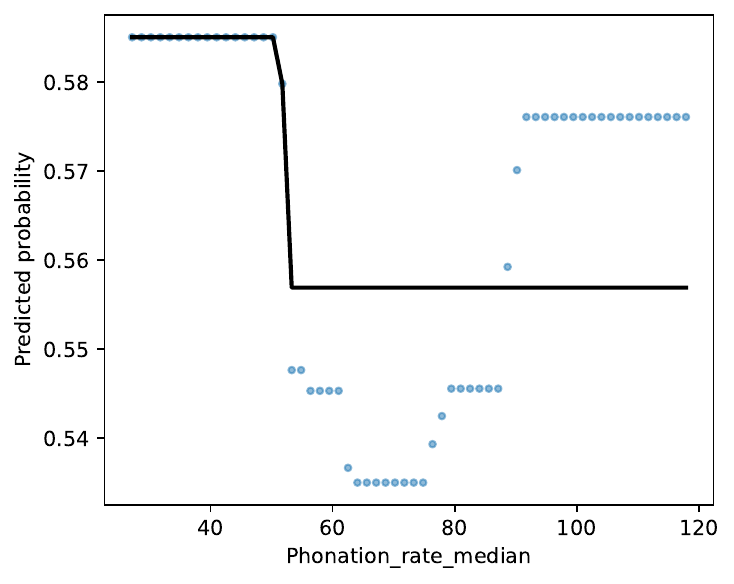} &
      \includegraphics[width=0.32\textwidth]{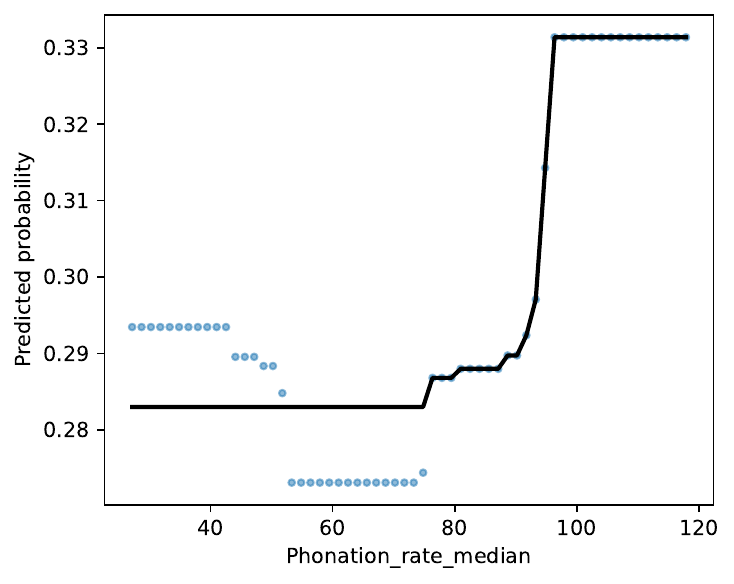} \\

      \includegraphics[width=0.32\textwidth]{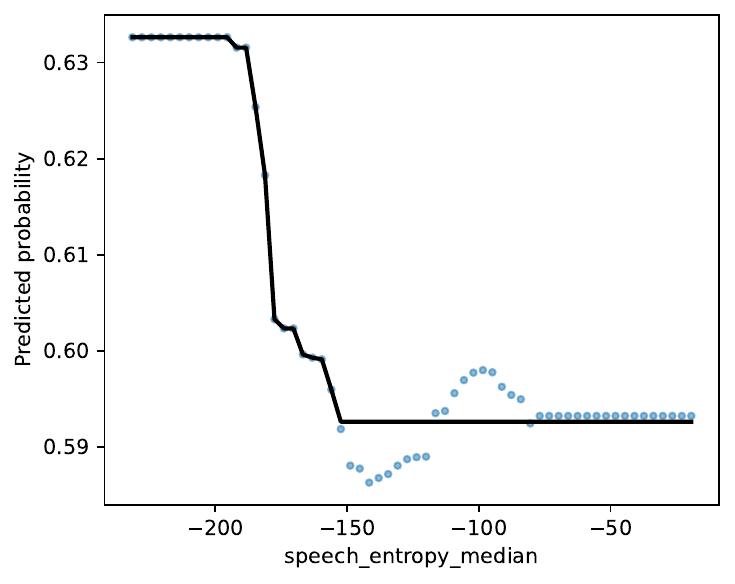} &
      \includegraphics[width=0.32\textwidth]{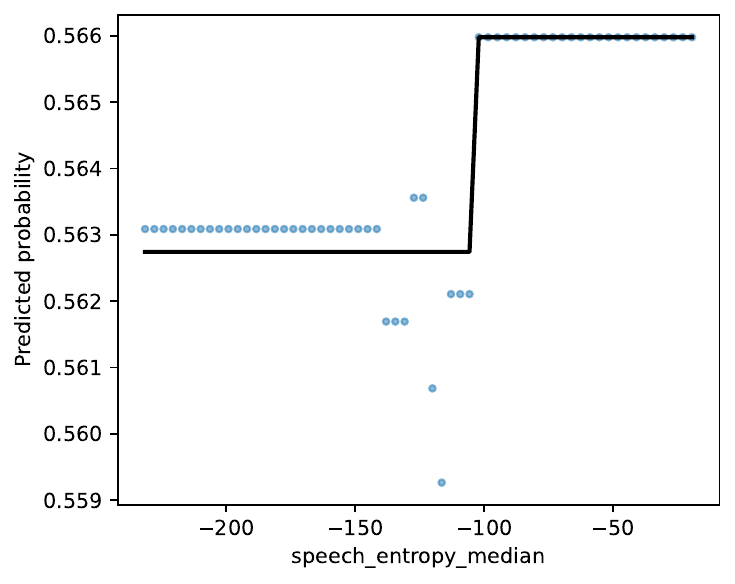} &
      \includegraphics[width=0.32\textwidth]{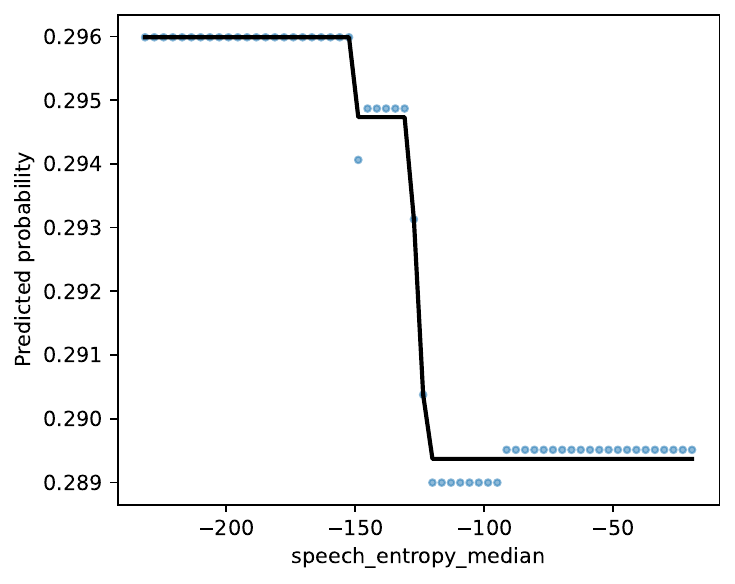} \\
    \end{tabular}
  }
    \caption{Partial Dependence Plots for speech-derived features across the \textsc{Real} model predictions for ASRS (first column), GAD-7 (second column), and PHQ-9 (third column).}
  \label{fig:pdp_all}
\end{figure}

Figure~\ref{fig:pdp_all} presents the PDPs for a range of speech-derived features across the \textit{Real} dataset models (ASRS, GAD-7, and PHQ-9). These plots highlight how prosodic, lexical, and semantic attributes contribute to predicting mental health symptom severity. 

\begin{figure}[!t]
  \centering
  \setlength{\tabcolsep}{2pt} 
  \renewcommand{\arraystretch}{1.0}
  \begin{tabular}{ccc}
    \includegraphics[width=0.32\columnwidth]{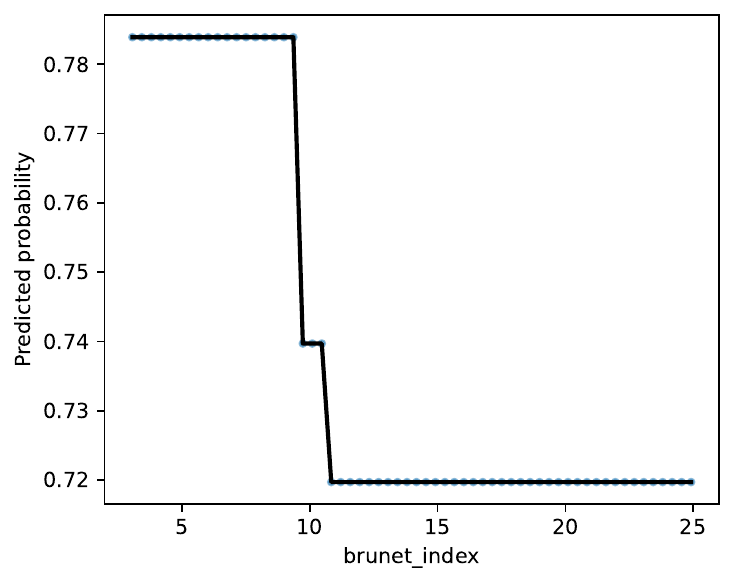} &
    \includegraphics[width=0.32\columnwidth]{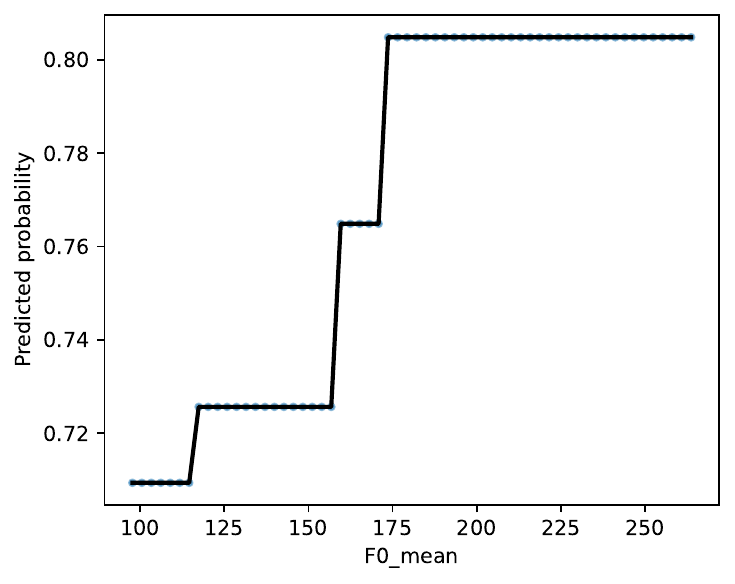} &
    \includegraphics[width=0.32\columnwidth]{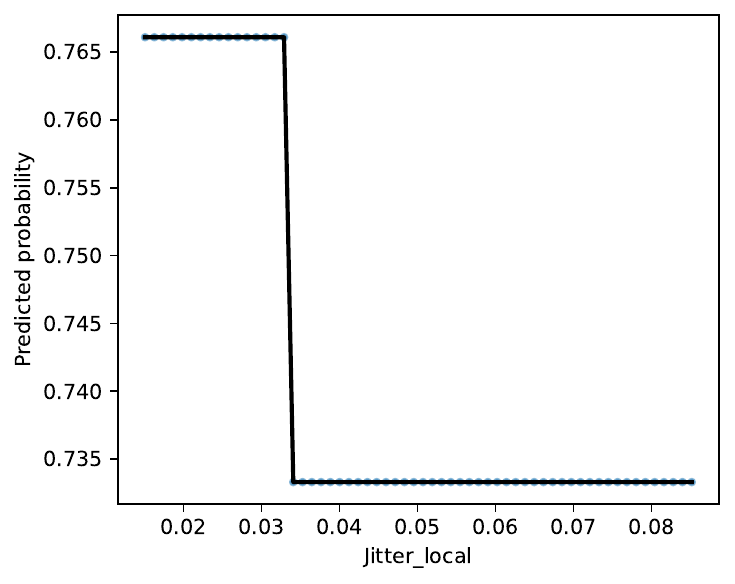} \\[2pt]

    \includegraphics[width=0.32\columnwidth]{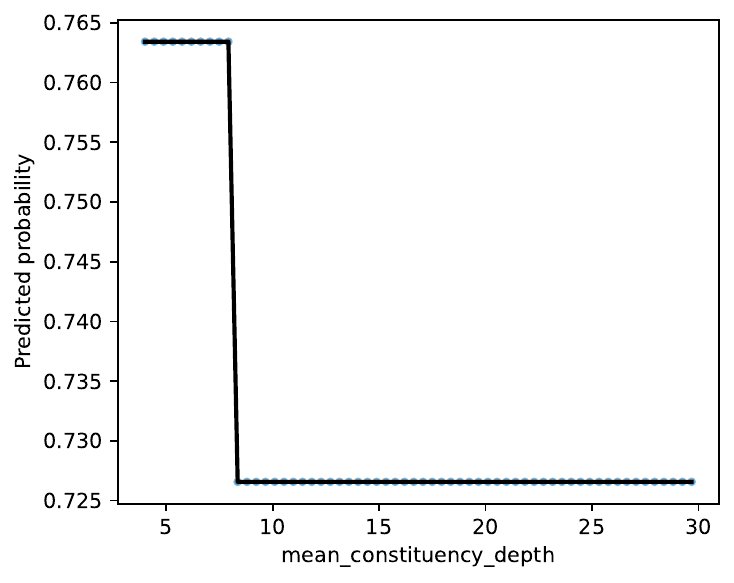} &
    \includegraphics[width=0.32\columnwidth]{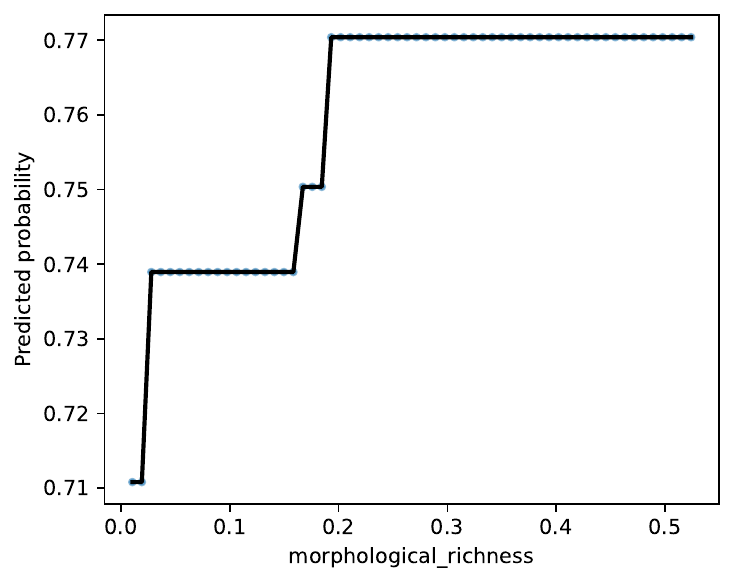} &
    \includegraphics[width=0.32\columnwidth]{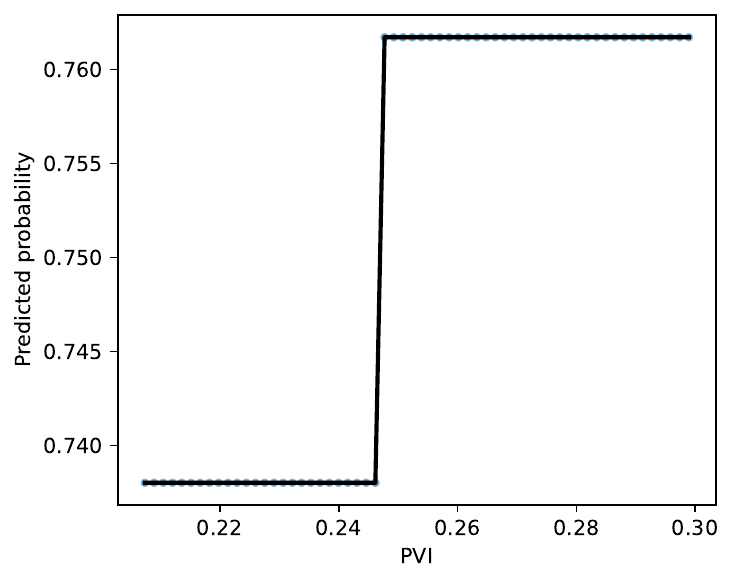} \\[2pt]

    \includegraphics[width=0.32\columnwidth]{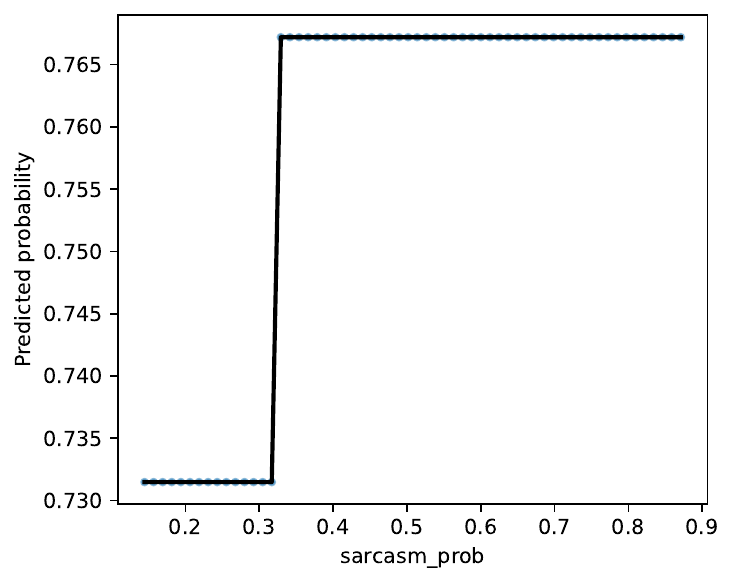} &
    \includegraphics[width=0.32\columnwidth]{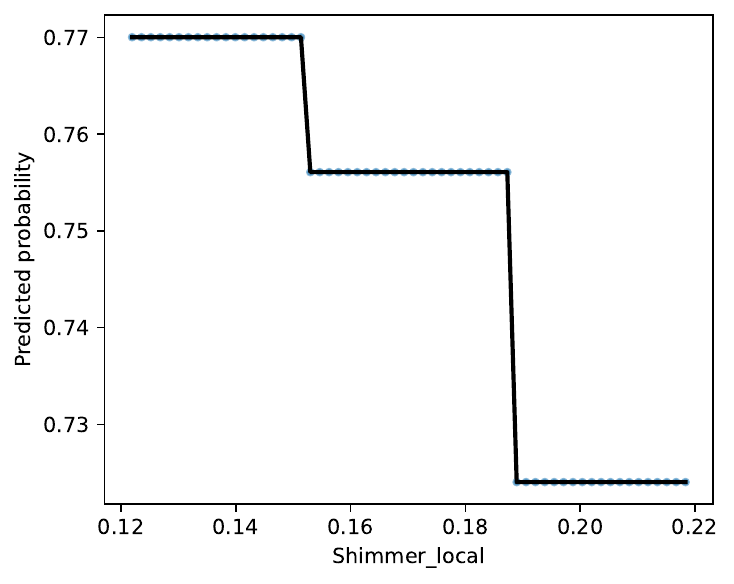} &
    \includegraphics[width=0.32\columnwidth]{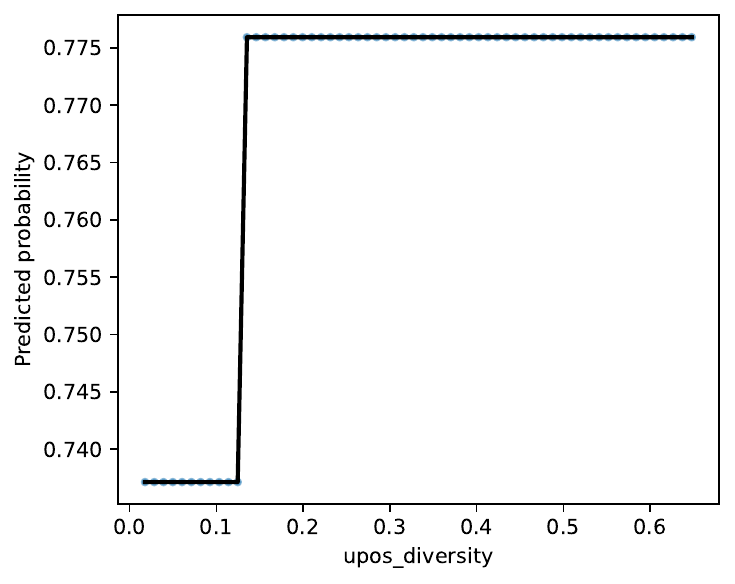} \\[2pt]

    \includegraphics[width=0.32\columnwidth]{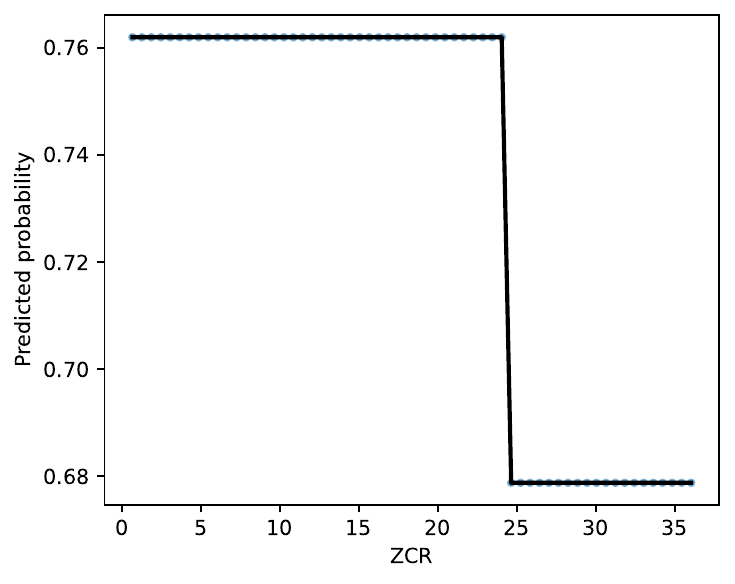} & & \\
  \end{tabular}
  \caption{
    Partial Dependence Plots for key linguistic and acoustic features in the \textsc{StressID} model. 
    Each curve shows the monotonic relationship between the feature value and model prediction.
  }
  \label{fig:pdp_single_model}
\end{figure}

Figure~\ref{fig:pdp_single_model} focuses on the \textsc{StressID} model, illustrating the partial dependencies for key linguistic and acoustic predictors. The monotonic and threshold-based trends observed, such as the influence of speech rate, pause duration, and pitch variability, suggest interpretable and physiologically plausible mappings between vocal behavior and stress-related outcomes. 

The Partial Dependence Plots for the \textsc{DAIC-WOZ} dataset (Figure~\ref{fig:pdp_single_model_daic}) show that increased pause duration and reduced pitch variability are associated with higher depression probability. Elevated jitter and shimmer further indicate reduced phonatory stability in depressed speech. Linguistic features such as shorter word length and lower lexical diversity exhibit monotonic trends, reflecting cognitive and expressive slowing.

For the \textsc{EATD} corpus, PDPs (Figure~\ref{fig:eatd_pdp_all}) reveal strong non-linear effects, with higher articulation rate and wider pitch range associated with lower depression likelihood. Increased sad emotion activation sharply increases predicted depression probability. Linguistic simplification beyond threshold levels further contributes to elevated risk.

In the \textsc{Androids Corpus}, PDPs (Figure~\ref{fig:pdp_single_model_androids}) indicate that reduced vocal intensity, increased shimmer, and longer pauses correspond to higher depression probability. Emotional polarity features show strong monotonic relationships, with higher sadness and lower positive sentiment increasing risk. Disrupted discourse coherence further amplifies depression predictions across tasks.

\section{Feature Descriptions}\label{sec:feature}

To ensure interpretability and transparency in our modeling pipeline, we provide a detailed description of all extracted acoustic and linguistic features used in our analyses, see Table~\ref{tab:features_mathsafe}. Each feature is annotated with its extraction method, type, subtype, and psycholinguistic interpretation. These features encompass prosodic, voice quality, lexical, syntactic, semantic, and psycholinguistic dimensions, thereby capturing both low-level acoustic cues and higher-level cognitive-linguistic markers of mental state.

\begin{table*}[htbp]
\vspace{-0.35em}
\centering
\scriptsize
\setlength{\tabcolsep}{2pt}
\renewcommand{\arraystretch}{0.60}
\begin{tabular}{>{\raggedright\arraybackslash}p{3.5cm} >{\raggedright\arraybackslash}p{2.1cm} >{\raggedright\arraybackslash}p{2.6cm} >{\raggedright\arraybackslash}p{7cm}}
\toprule
\textbf{Feature} & \textbf{Type} & \textbf{Subtype} & \textbf{Calculation / Definition} \\
\midrule
ZCR & Acoustic & Prosodic & Zero-crossing rate of the normalized voiced waveform. \\
F0\_mean & Acoustic & Prosodic & Mean pitch (Hz) from Praat pitch analysis. \\
F0\_range & Acoustic & Prosodic & Maximum pitch minus minimum pitch. \\
F0\_var & Acoustic & Prosodic & Variance of pitch values across voiced frames. \\
F0\_std & Acoustic & Prosodic & Standard deviation of F0 values. \\
Intensity\_mean & Acoustic & Prosodic & Mean amplitude of the waveform. \\
Intensity\_std & Acoustic & Prosodic & Standard deviation of intensity values. \\
Jitter\_local & Acoustic & Voice Quality &Mean absolute F0 period difference. \\
Shimmer\_local & Acoustic & Voice Quality & Mean absolute amplitude difference. \\
HNR & Acoustic & Voice Quality & Harmonics-to-noise ratio (Praat Harmonicity). \\
PVI & Acoustic & Voice Quality & Pairwise variability index across voiced intervals. \\
filler\_count & Linguistic & Voice Quality & Count of filler tokens such as "um", "uh", "erm". \\
duration & Acoustic & Prosodic & Total length of the audio file (seconds). \\
Phonation\_rate & Acoustic & Prosodic & Number of voiced frames divided by total duration. \\
pause\_count & Acoustic & Prosodic & Count of silent segments below amplitude threshold. \\
pause\_short & Acoustic & Prosodic & Pauses $<1$ s. \\
pause\_medium & Acoustic & Prosodic & Pauses $1{-}2$ s. \\
pause\_long & Acoustic & Prosodic & Pauses $>2$ s. \\
pause\_mean & Acoustic & Prosodic & Mean duration of all pauses (s). \\
pause\_speech\_ratio & Acoustic & Prosodic & Total pause time divided by total duration. \\
articulation\_rate & Acoustic & Prosodic & Voiced frames / (duration $-$ total pause time). \\
speech\_entropy & Acoustic & Prosodic & Shannon entropy of voiced amplitudes. \\
emotion\_neu & Acoustic & Psycholinguistic & Neutral emotion probability (HuBERT model). \\
emotion\_hap & Acoustic & Psycholinguistic & Happiness probability (HuBERT model). \\
emotion\_ang & Acoustic & Psycholinguistic & Anger probability (HuBERT model). \\
emotion\_sad & Acoustic & Psycholinguistic & Sadness probability (HuBERT model). \\
word\_count & Linguistic & Lexical & Total number of words in transcript. \\
sentence\_count & Linguistic & Lexical & Total number of sentences in transcript. \\
type\_token\_ratio & Linguistic & Lexical & Ratio of unique words to total words. \\
MATTR & Linguistic & Lexical & Moving-average type-token ratio (window=50). \\
brunet\_index & Linguistic & Lexical & $N^{(V^{-0.165})}$ (lexical richness). \\
honore\_stat & Linguistic & Lexical & $100 \log N / (1 - H/V)$ (lexical richness measure). \\
lexical\_density & Linguistic & Lexical & Ratio of content words to total words. \\
idea\_density & Linguistic & Lexical & Ratio of idea-bearing words to total words. \\
content\_function\_ratio & Linguistic & Lexical & Ratio of content to function words. \\
pronoun\_ratio & Linguistic & Lexical & Pronouns divided by total words. \\
Tense\_Past & Linguistic & Lexical & Count of past tense verbs. \\
Tense\_Pres & Linguistic & Lexical & Count of present tense verbs. \\
Voice\_Pass & Linguistic & Lexical & Count of passive voice verbs. \\
Number\_Plur & Linguistic & Lexical & Count of plural nouns or verbs. \\
lemma\_ttr & Linguistic & Lexical & Unique lemmas / total lemmas. \\
upos\_diversity & Linguistic & Lexical & Unique UPOS tags / total tokens. \\
morphological\_richness & Linguistic & Lexical & Unique morph features / total words. \\
propositional\_density & Linguistic & Lexical & Ratio of verbs, adj, adv, prep, conj to total words. \\
mean\_sentence\_length & Linguistic & Syntactic & Mean words per sentence. \\
mean\_clause\_length & Linguistic & Syntactic & Mean number of tokens per clause. \\
syntactic\_depth\_mean & Linguistic & Syntactic & Mean dependency tree depth. \\
syntactic\_depth\_max & Linguistic & Syntactic & Maximum dependency tree depth. \\
clause\_ratio & Linguistic & Syntactic & Clauses per sentence. \\
verb\_tense\_switches & Linguistic & Syntactic & Count of verb tense changes. \\
verb\_tense\_switch\_ratio & Linguistic & Syntactic & Tense switches / total verbs. \\
syntactic\_embedding\_depth & Linguistic & Syntactic & Maximum dependency embedding depth. \\
passive\_voice\_ratio & Linguistic & Syntactic & Passive voice verbs / total verbs. \\
graph\_nodes & Linguistic & Syntactic & Number of unique nodes in the speech graph. \\
graph\_edges & Linguistic & Syntactic & Number of total edges in the speech graph. \\
graph\_repeated\_edges & Linguistic & Syntactic & Count of edges appearing multiple times. \\
graph\_largest\_scc & Linguistic & Syntactic & Size of largest strongly connected component. \\
graph\_density & Linguistic & Syntactic & Ratio of actual to maximum possible edges. \\
graph\_loops\_L1 & Linguistic & Syntactic & Number of self-loops (1-node cycles). \\
graph\_loops\_L2 & Linguistic & Syntactic & Number of 2-node cycles. \\
graph\_loops\_L3 & Linguistic & Syntactic & Number of 3-node cycles. \\
graph\_avg\_total\_degree & Linguistic & Syntactic & Average in+out degree across nodes. \\
graph\_diameter & Linguistic & Syntactic & Longest shortest path in undirected graph. \\
graph\_avg\_shortest\_path & Linguistic & Syntactic & Mean shortest path length in graph. \\
nodes\_per\_word & Linguistic & Syntactic & graph\_nodes / total\_words. \\
edges\_per\_word & Linguistic & Syntactic & graph\_edges / total\_words. \\
atd\_per\_word & Linguistic & Syntactic & graph\_avg\_total\_degree / total\_words. \\
parallel\_edges\_per\_word & Linguistic & Syntactic & graph\_repeated\_edges / total\_words. \\
loops\_L1\_per\_word & Linguistic & Syntactic & graph\_loops\_L1 / total\_words. \\
loops\_L2\_per\_word & Linguistic & Syntactic & graph\_loops\_L2 / total\_words. \\
loops\_L3\_per\_word & Linguistic & Syntactic & graph\_loops\_L3 / total\_words. \\
mean\_constituency\_depth & Linguistic & Syntactic & Mean depth of Stanza constituency tree. \\
max\_constituency\_depth & Linguistic & Syntactic & Maximum depth of Stanza constituency tree. \\
first\_order\_coherence & Linguistic & Semantic &Mean cosine simmilarity between sentences.\\
second\_order\_coherence & Linguistic & Semantic & Average similarity between sentences separated by one. \\
discourse\_cohesion & Linguistic & Semantic & Mean token overlap between consecutive sentences. \\
sentence\_repetition\_ratio & Linguistic & Semantic & Exact repeated sentences / total sentences. \\
vader\_negative & Linguistic & Psycholinguistic & VADER negative sentiment proportion. \\
vader\_neutral & Linguistic & Psycholinguistic & VADER neutral sentiment proportion. \\
vader\_positive & Linguistic & Psycholinguistic & VADER positive sentiment proportion. \\
vader\_compound & Linguistic & Psycholinguistic & Composite sentiment score in range $[-1,1]$. \\
sarcasm\_prob & Bimodal & Psycholinguistic & Probability of sarcasm (BERT + Wav2Vec classifier). \\
\bottomrule
\end{tabular}
\caption{
List of 82 acoustic and linguistic features with their computational definitions and grouping by type and subtype.
}
\label{tab:features_mathsafe}
\end{table*}

\begin{figure}[!h]
  \centering
  \setlength{\tabcolsep}{2pt}
  \renewcommand{\arraystretch}{1.0}
  \begin{tabular}{ccc}
    \includegraphics[width=0.32\columnwidth]{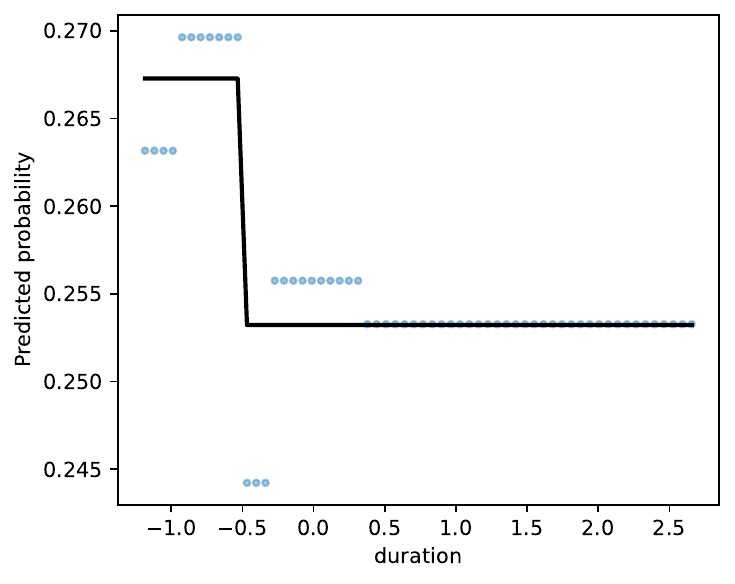} &
    \includegraphics[width=0.32\columnwidth]{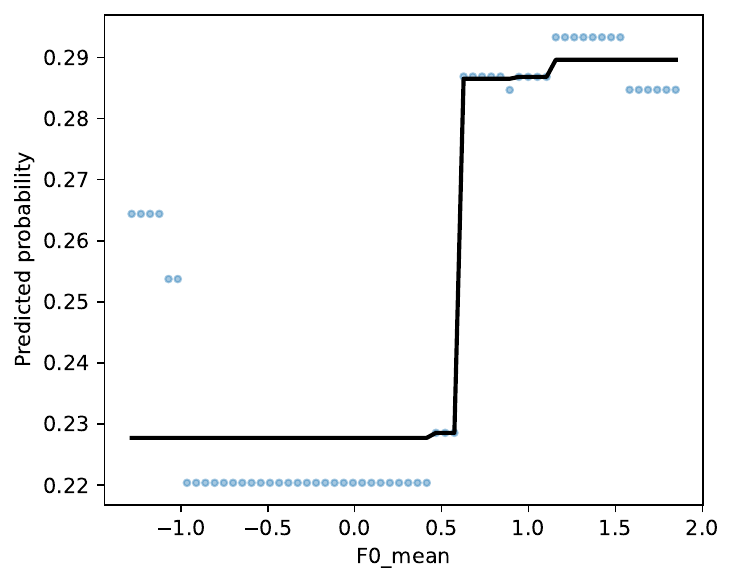} &
    \includegraphics[width=0.32\columnwidth]{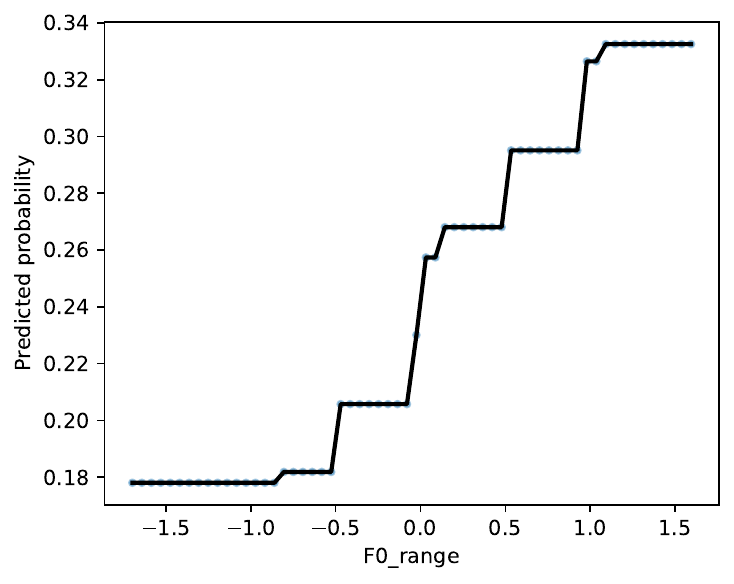} \\[2pt]
    \includegraphics[width=0.32\columnwidth]{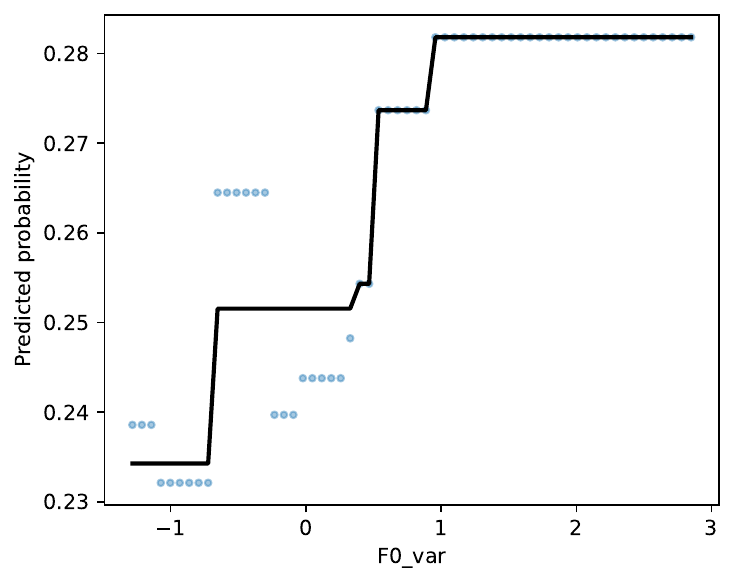} &
    \includegraphics[width=0.32\columnwidth]{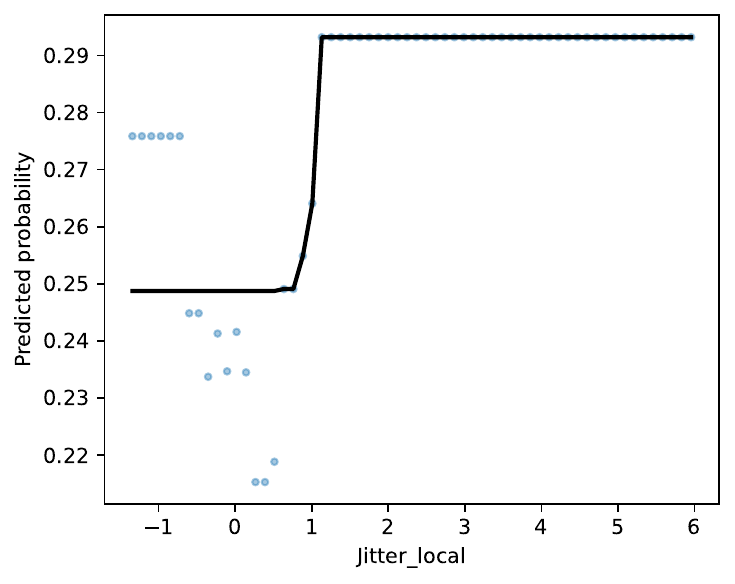} &
    \includegraphics[width=0.32\columnwidth]{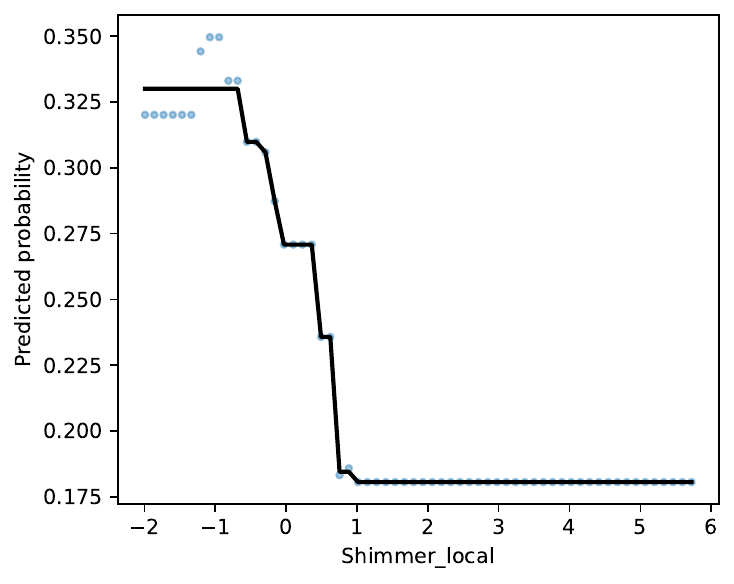} \\[2pt]
    \includegraphics[width=0.32\columnwidth]{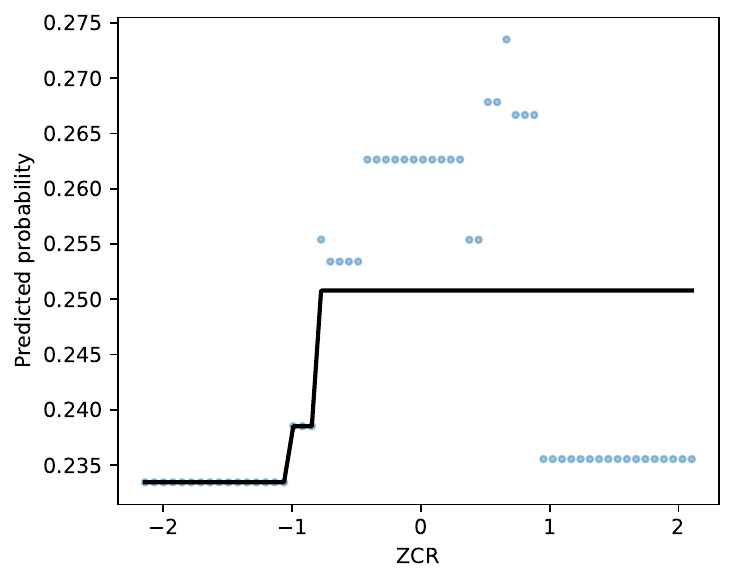} & & \\
  \end{tabular}
  \caption{
    Partial Dependence Plots for key acoustic features in the \textsc{DAIC-WOZ} model.
  }
  \label{fig:pdp_single_model_daic}
\end{figure}

\begin{figure}[!h]
  \centering
  \setlength{\tabcolsep}{2pt}
  \renewcommand{\arraystretch}{1.0}
  \begin{tabular}{ccc}
    \includegraphics[width=0.32\columnwidth]{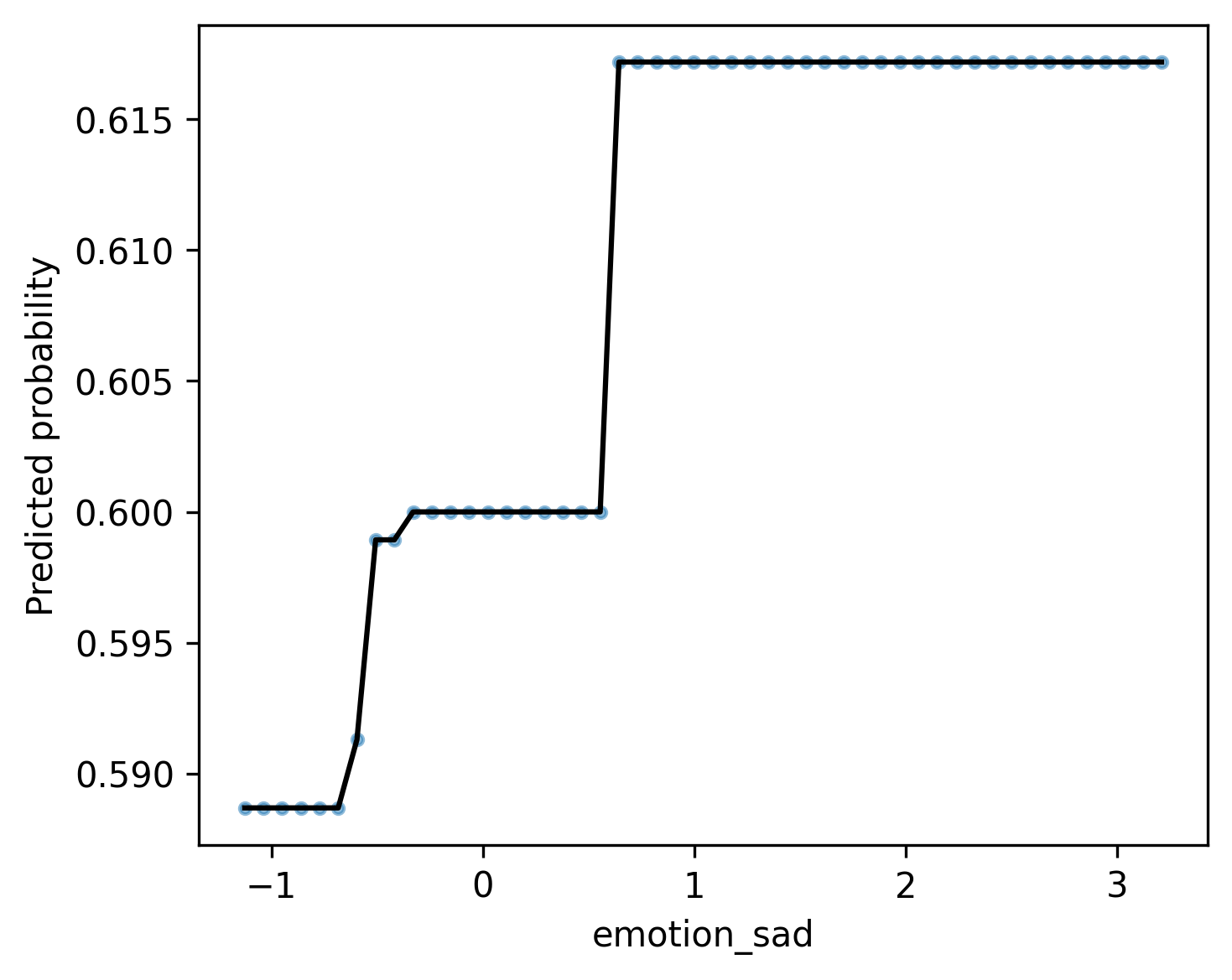} &
    \includegraphics[width=0.32\columnwidth]{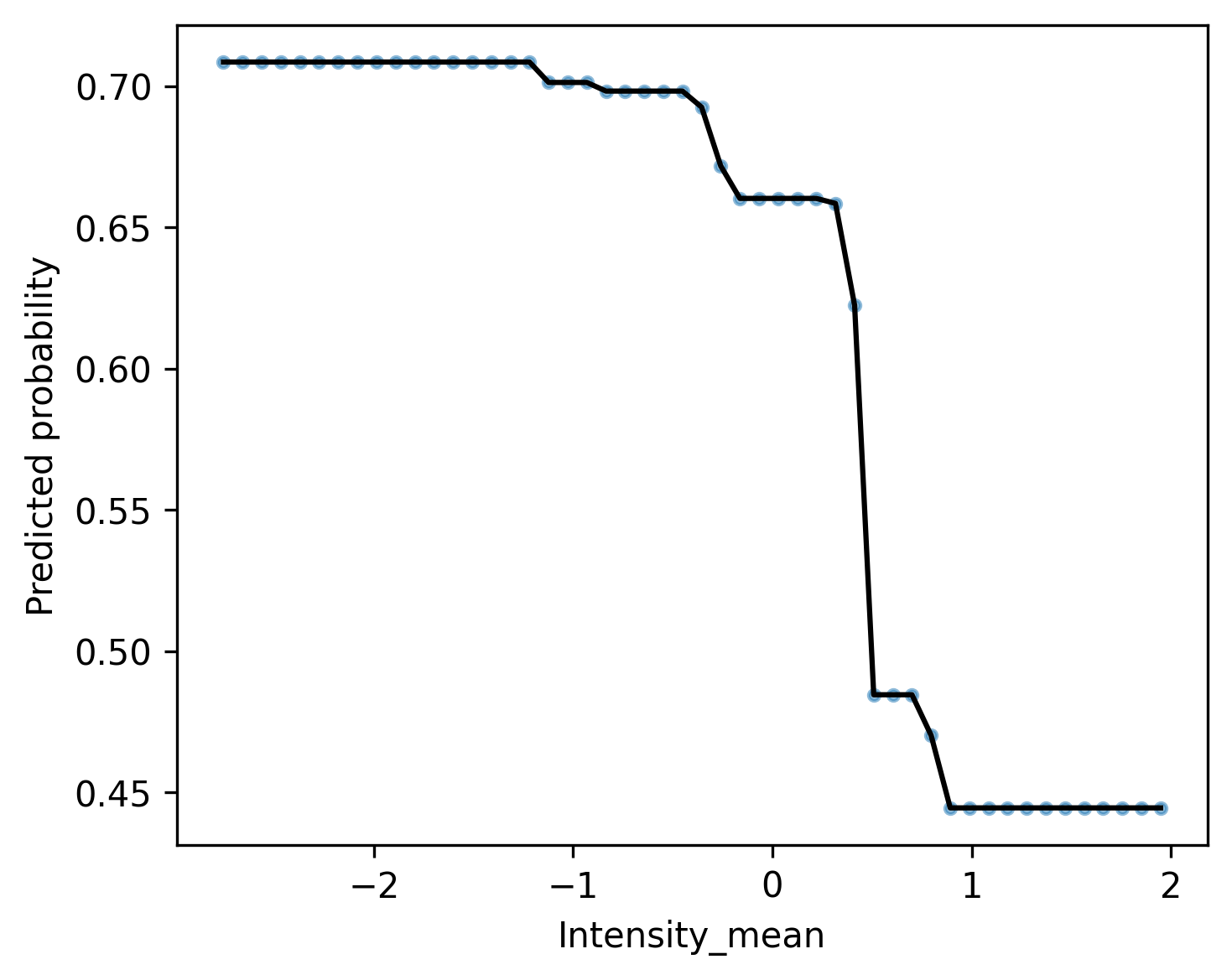} &
    \includegraphics[width=0.32\columnwidth]{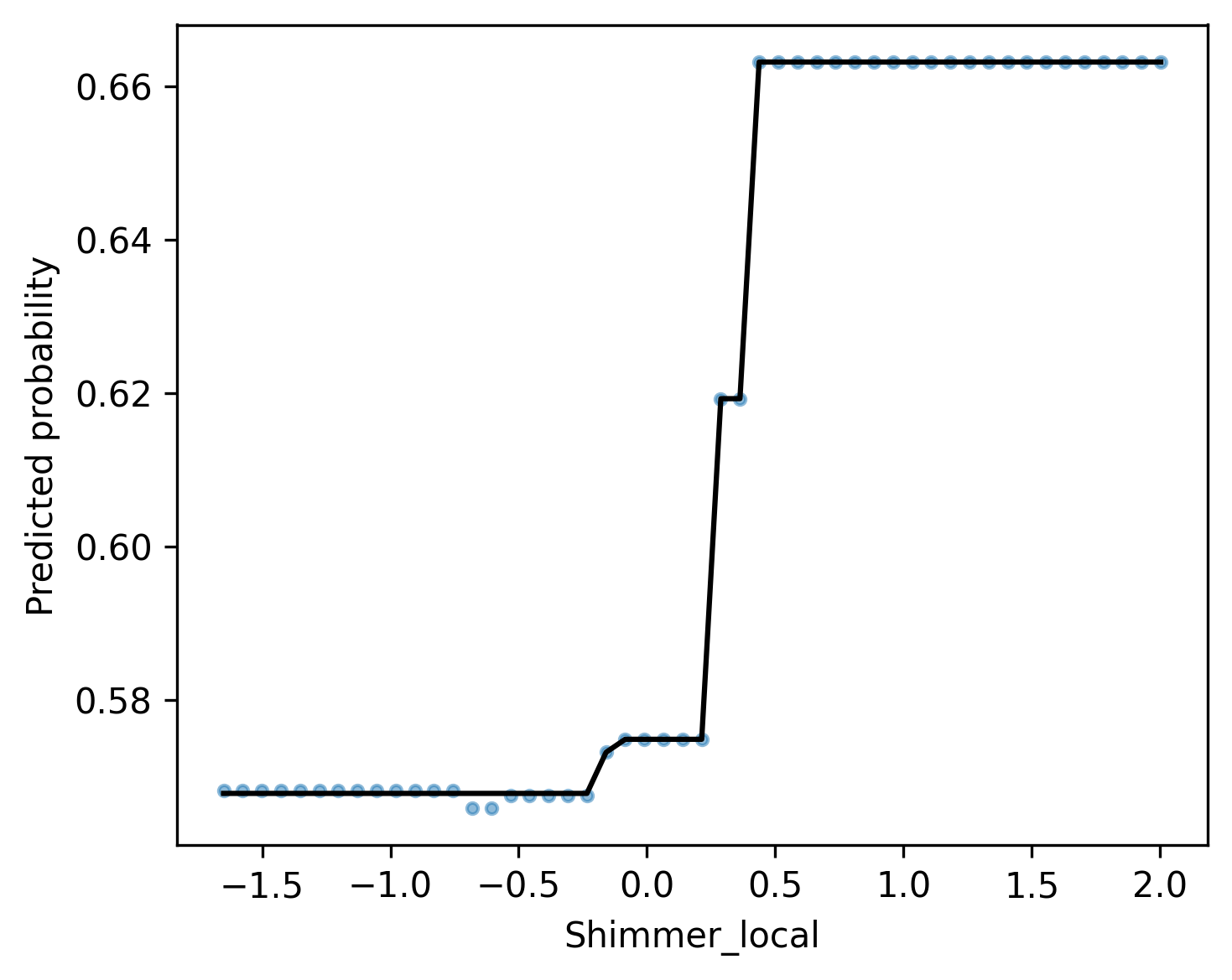} \\[2pt]

    \includegraphics[width=0.32\columnwidth]{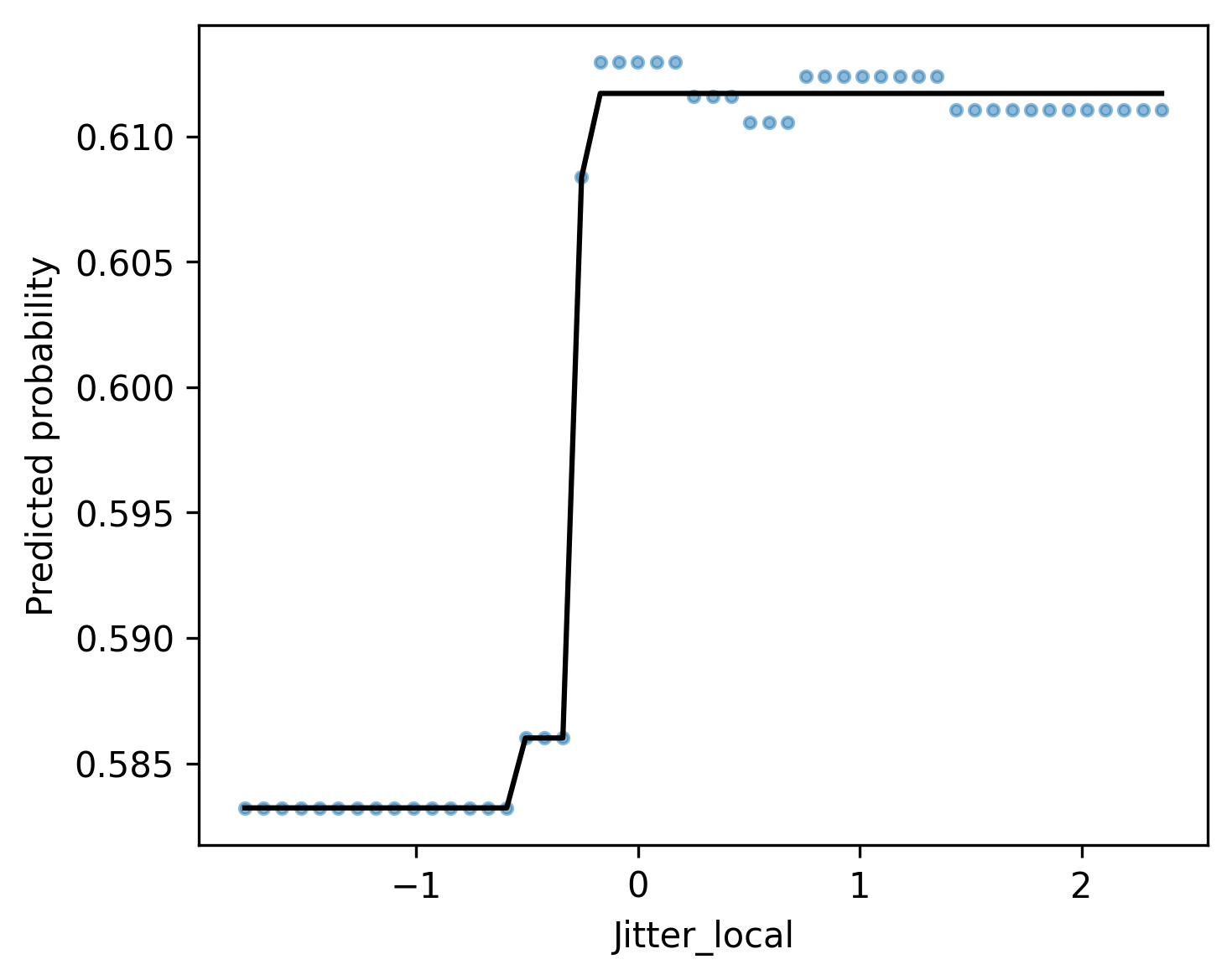} &
    \includegraphics[width=0.32\columnwidth]{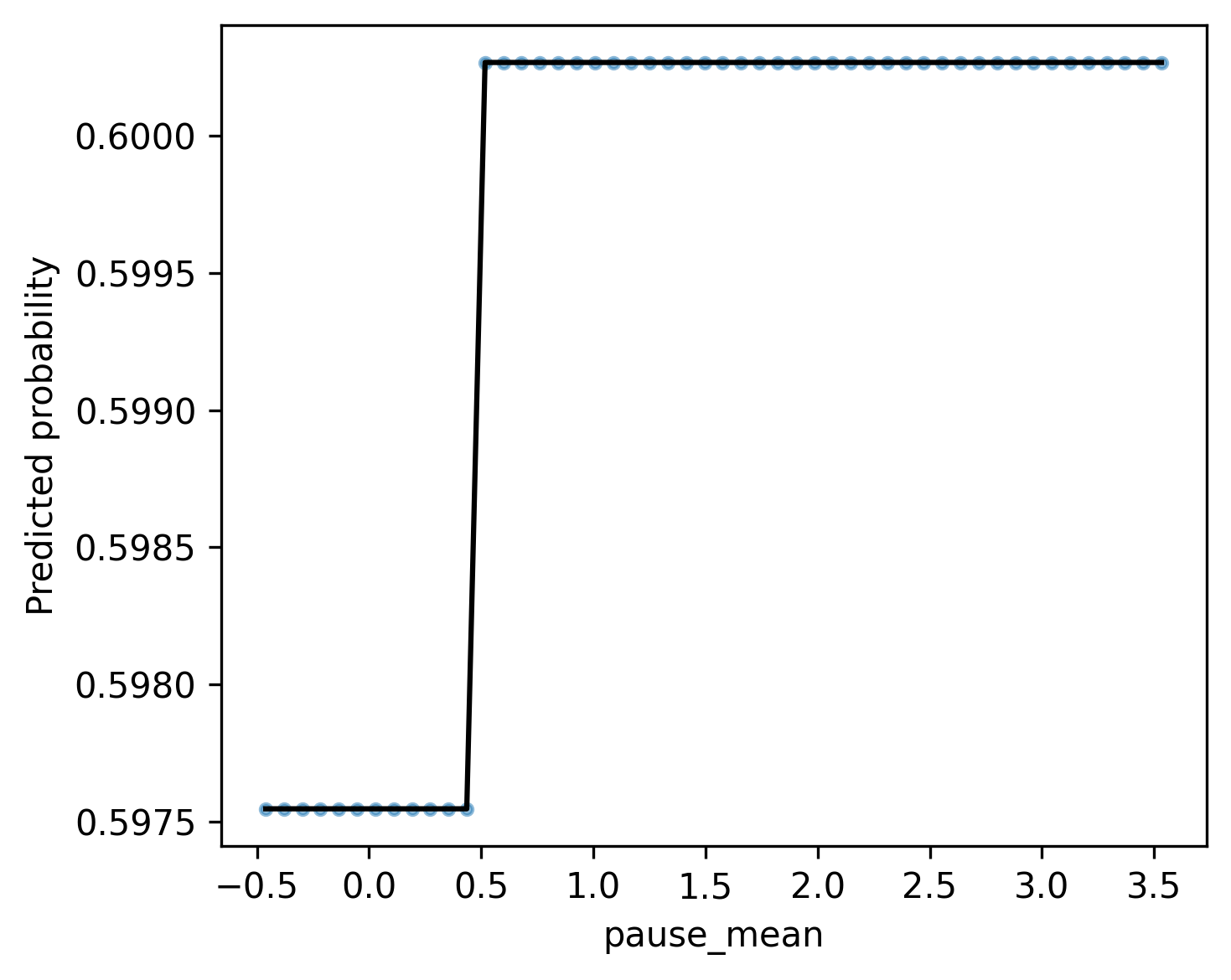} &
    \includegraphics[width=0.32\columnwidth]{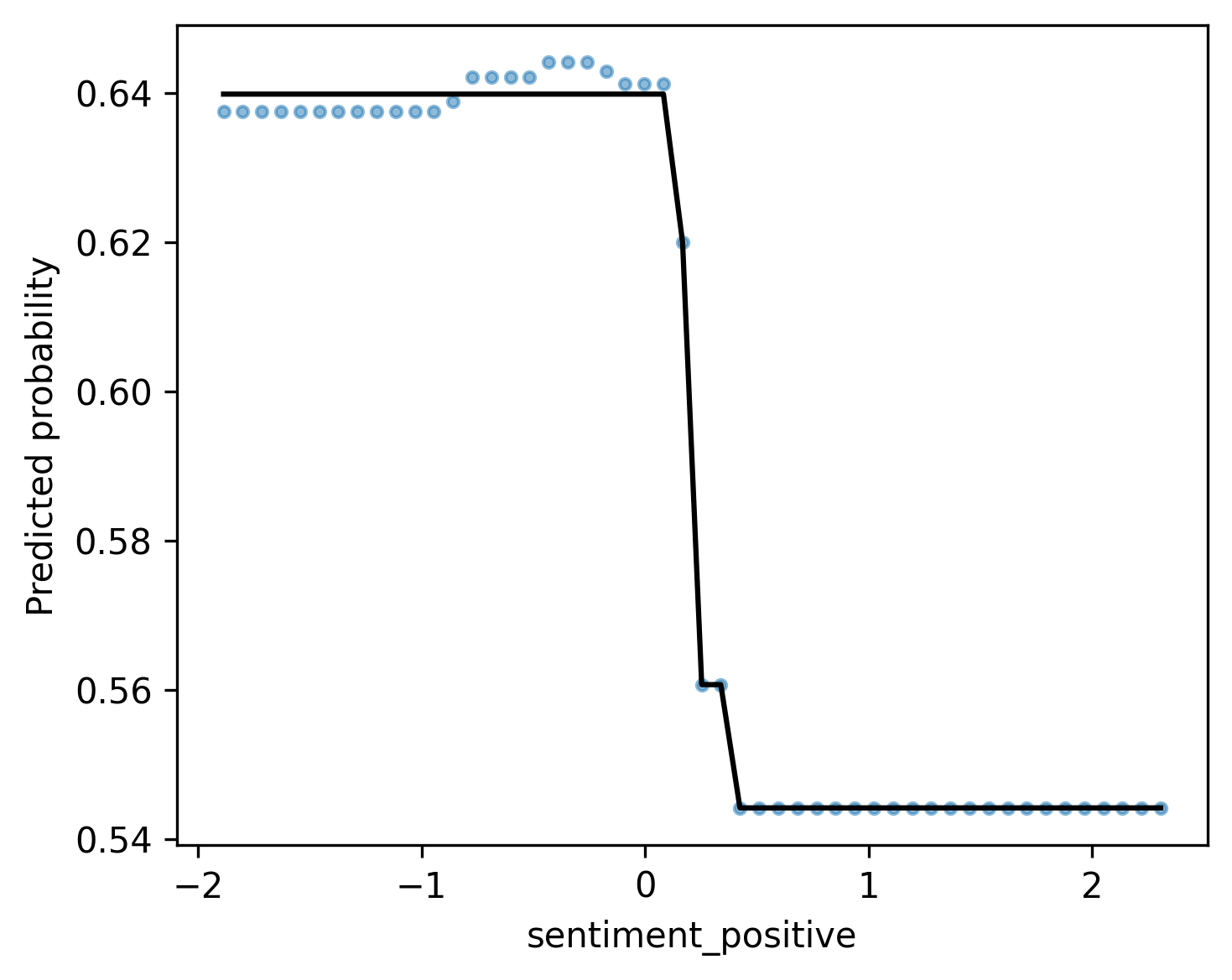} \\
  \end{tabular}
  \caption{
    Partial Dependence Plots for key acoustic, emotional, and linguistic features in the \textsc{Androids} model.
  }
  \label{fig:pdp_single_model_androids}
\end{figure}

\begin{figure}[!h]
  \centering
  \setlength{\tabcolsep}{2pt}
  \renewcommand{\arraystretch}{1.0}
  \begin{tabular}{ccc}
    \includegraphics[width=0.32\columnwidth]{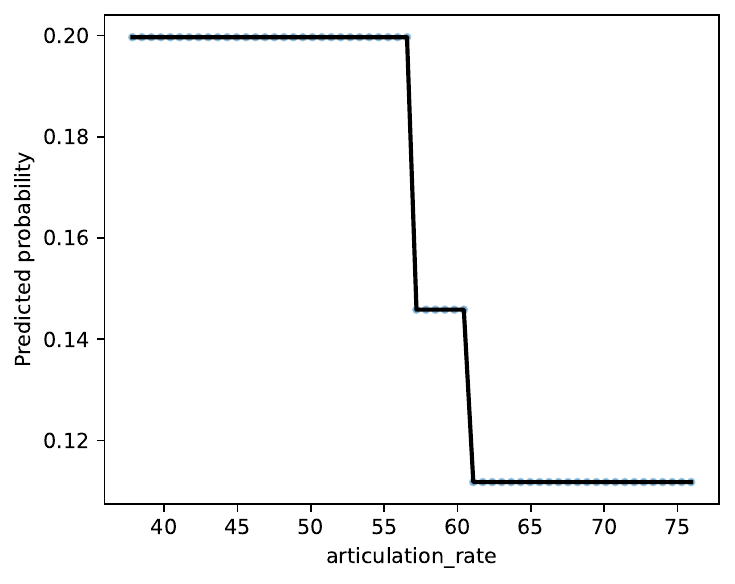} &
    \includegraphics[width=0.32\columnwidth]{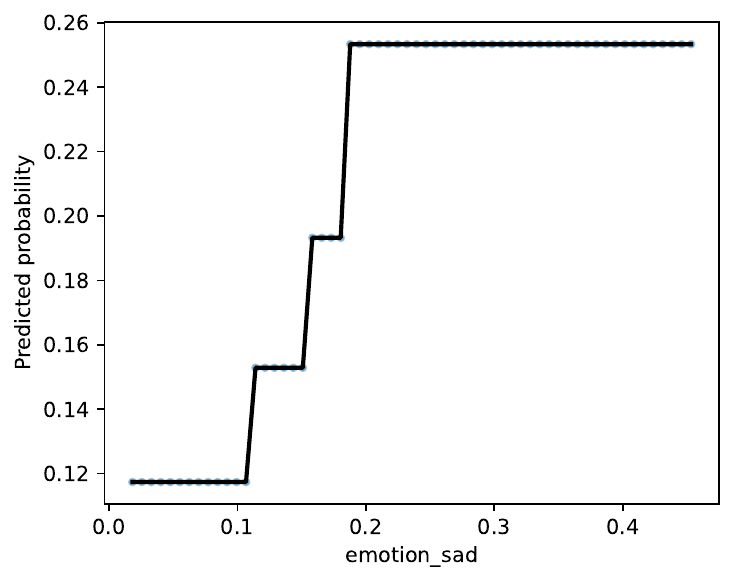} &
    \includegraphics[width=0.32\columnwidth]{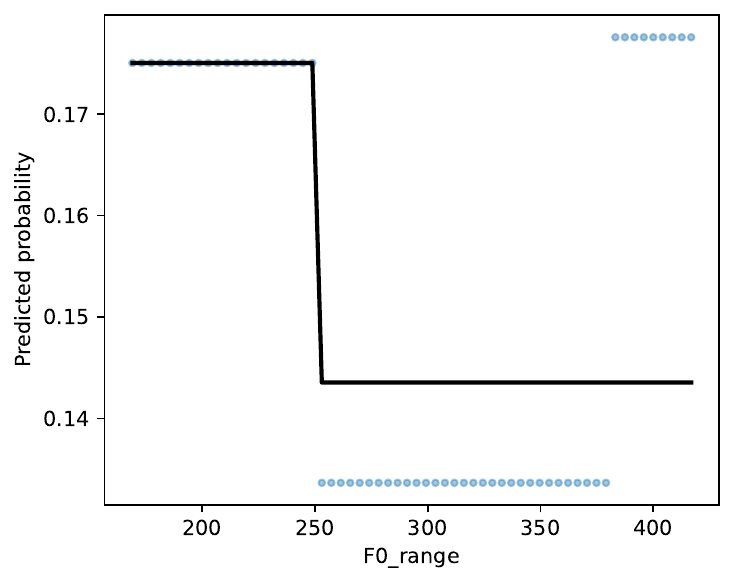} \\[2pt]

    \includegraphics[width=0.32\columnwidth]{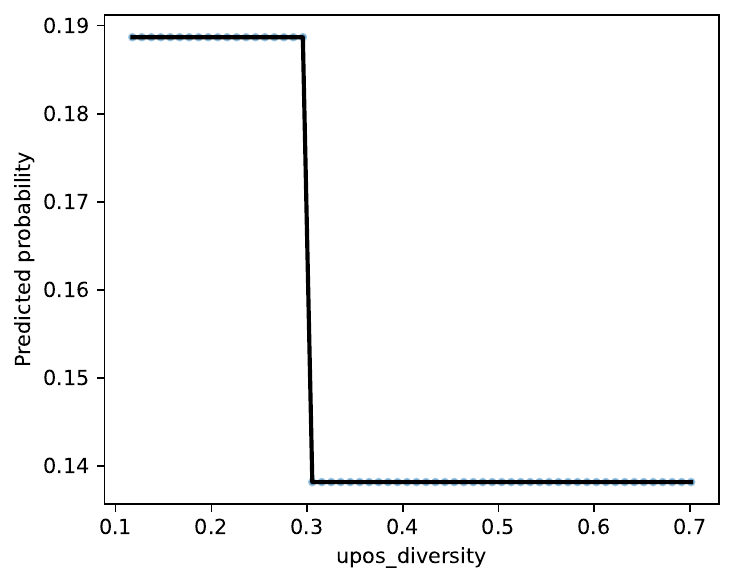} &
    \includegraphics[width=0.32\columnwidth]{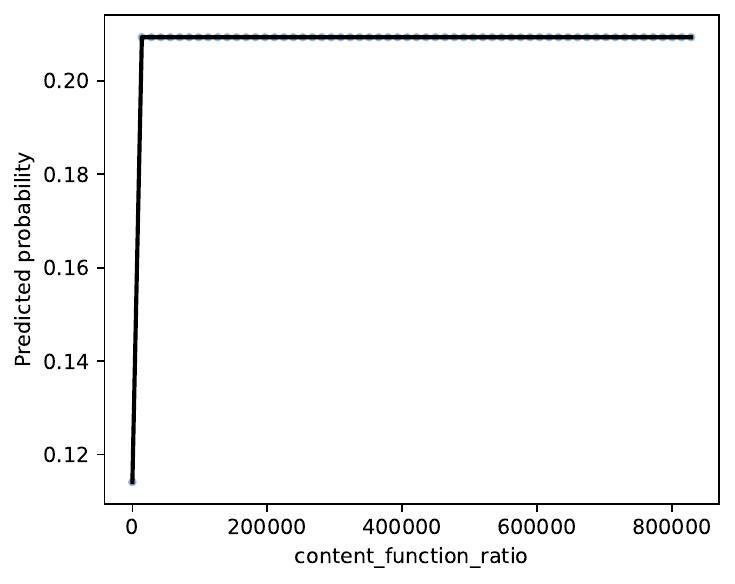} &
    \includegraphics[width=0.32\columnwidth]{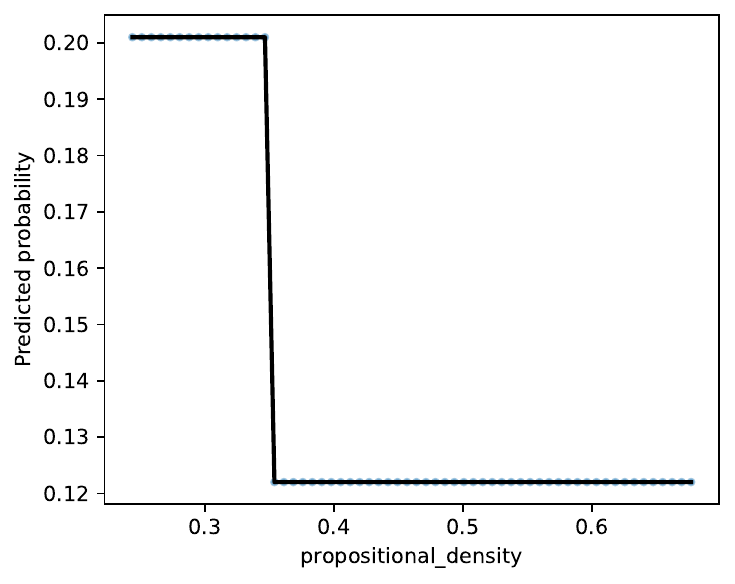} \\[2pt]

  \end{tabular}
  \caption{
    Partial Dependence Plots for key acoustic and linguistic features in the \textsc{EATD} model.
  }
  \label{fig:eatd_pdp_all}
\end{figure}

\section{Feature Correlation}

\begin{figure}[t]
    \centering
    \includegraphics[width=0.6\linewidth]{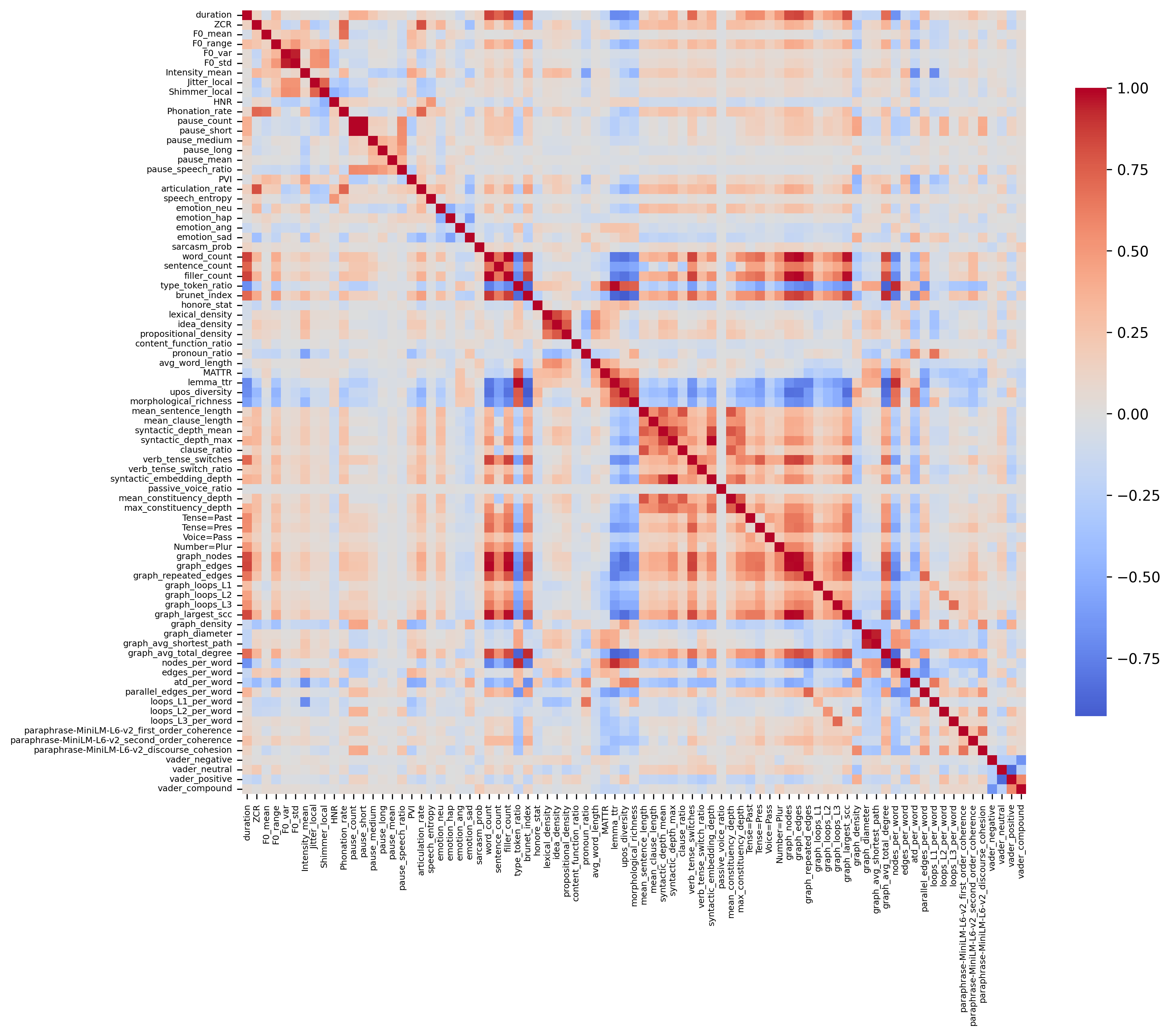}
    
    \vspace{0.1cm}
    \includegraphics[width=0.6\linewidth]{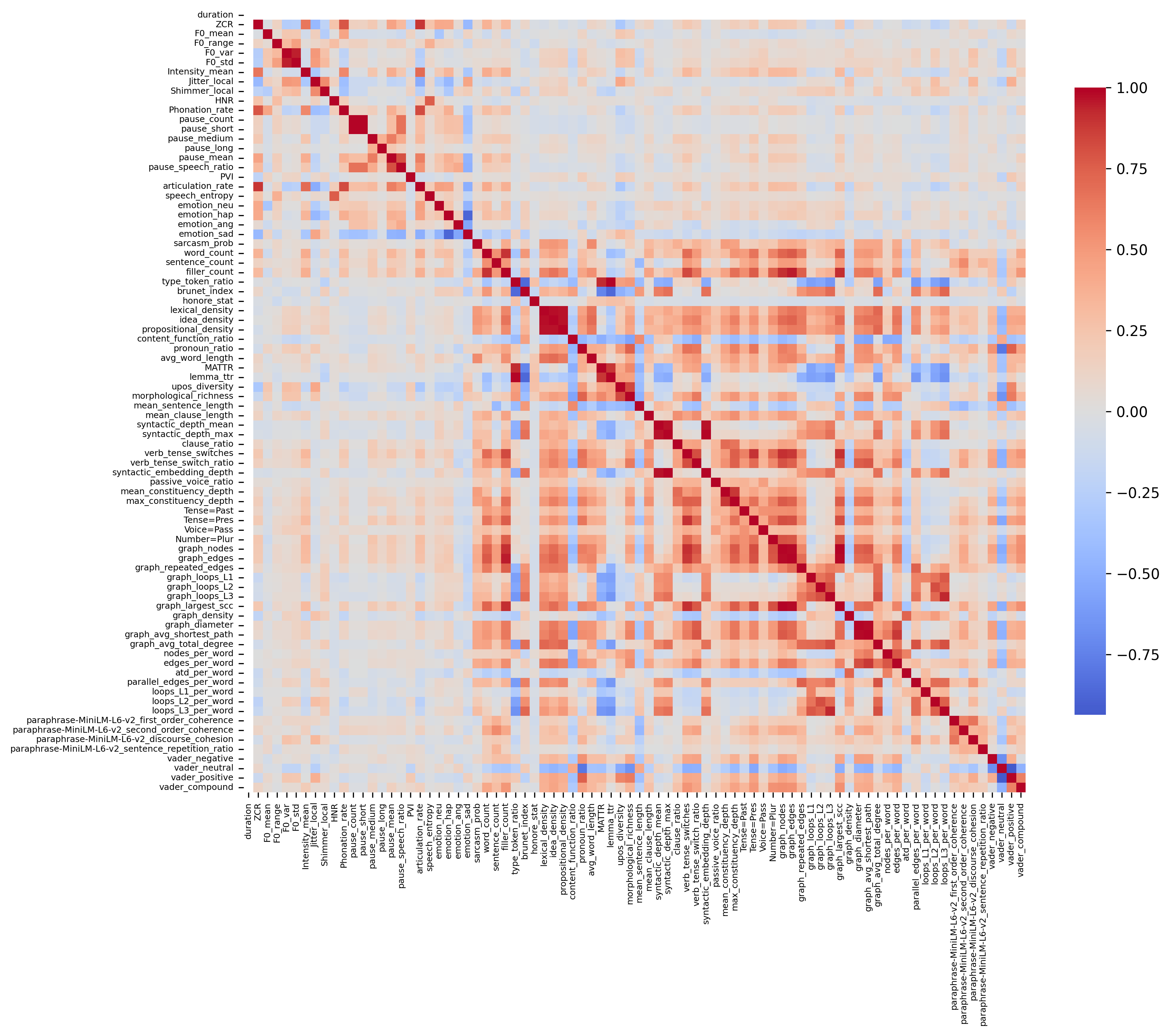}
    
    \vspace{0.1cm}
    \includegraphics[width=0.7\linewidth]{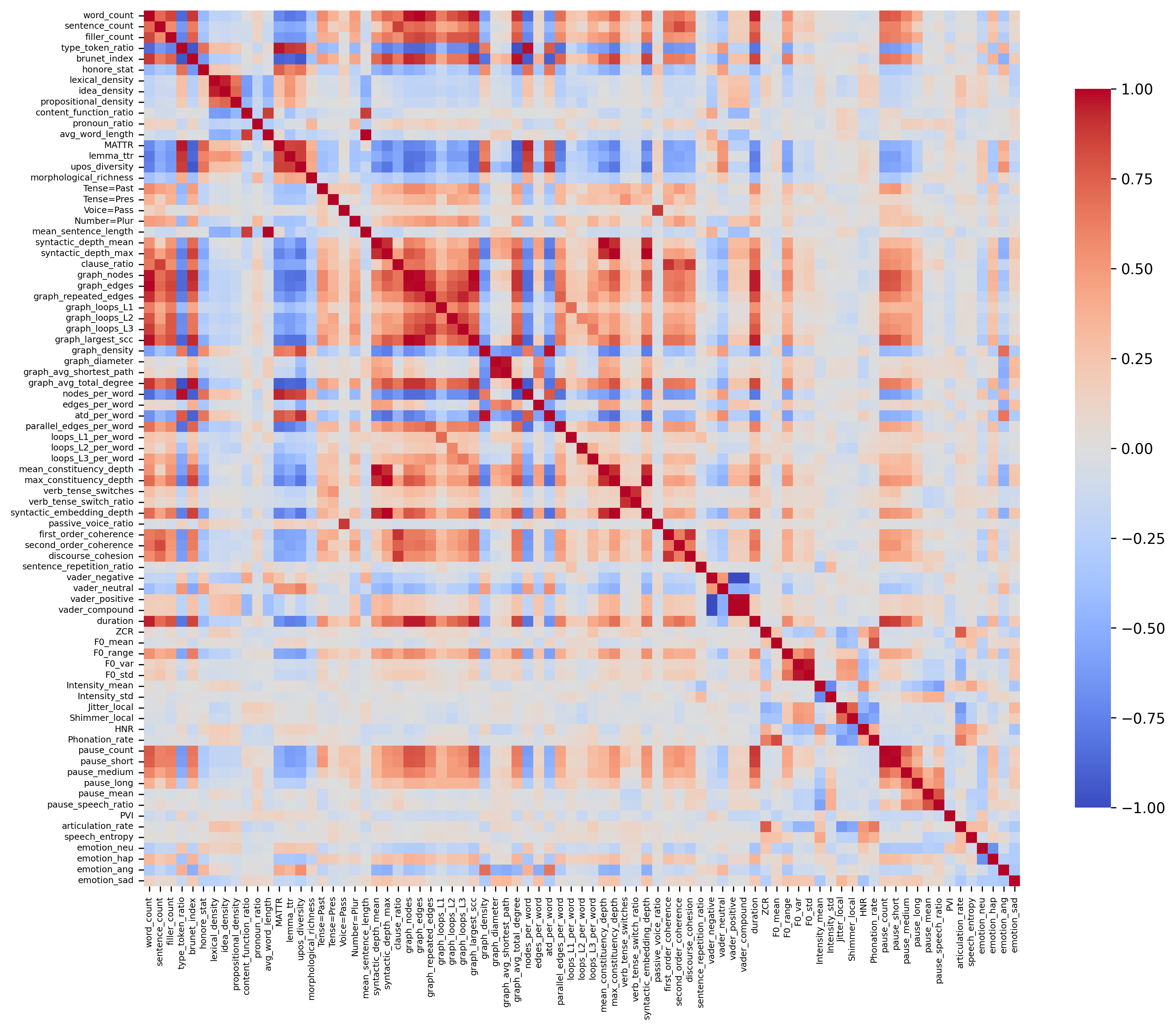}
    
    \vspace{0.1cm}
    \includegraphics[width=0.6\linewidth]{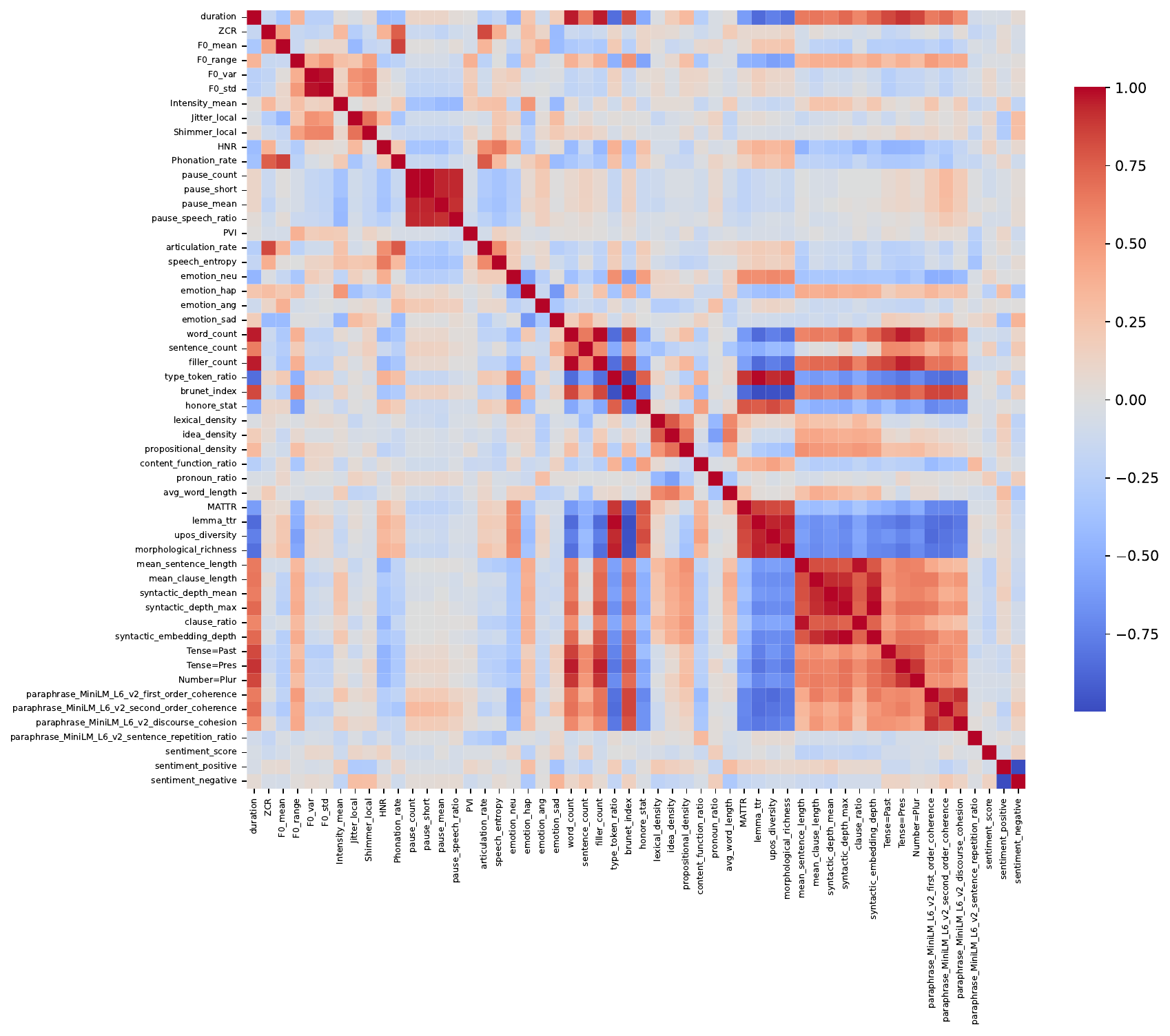}
    
    \vspace{0.1cm}
    \includegraphics[width=0.6\linewidth]{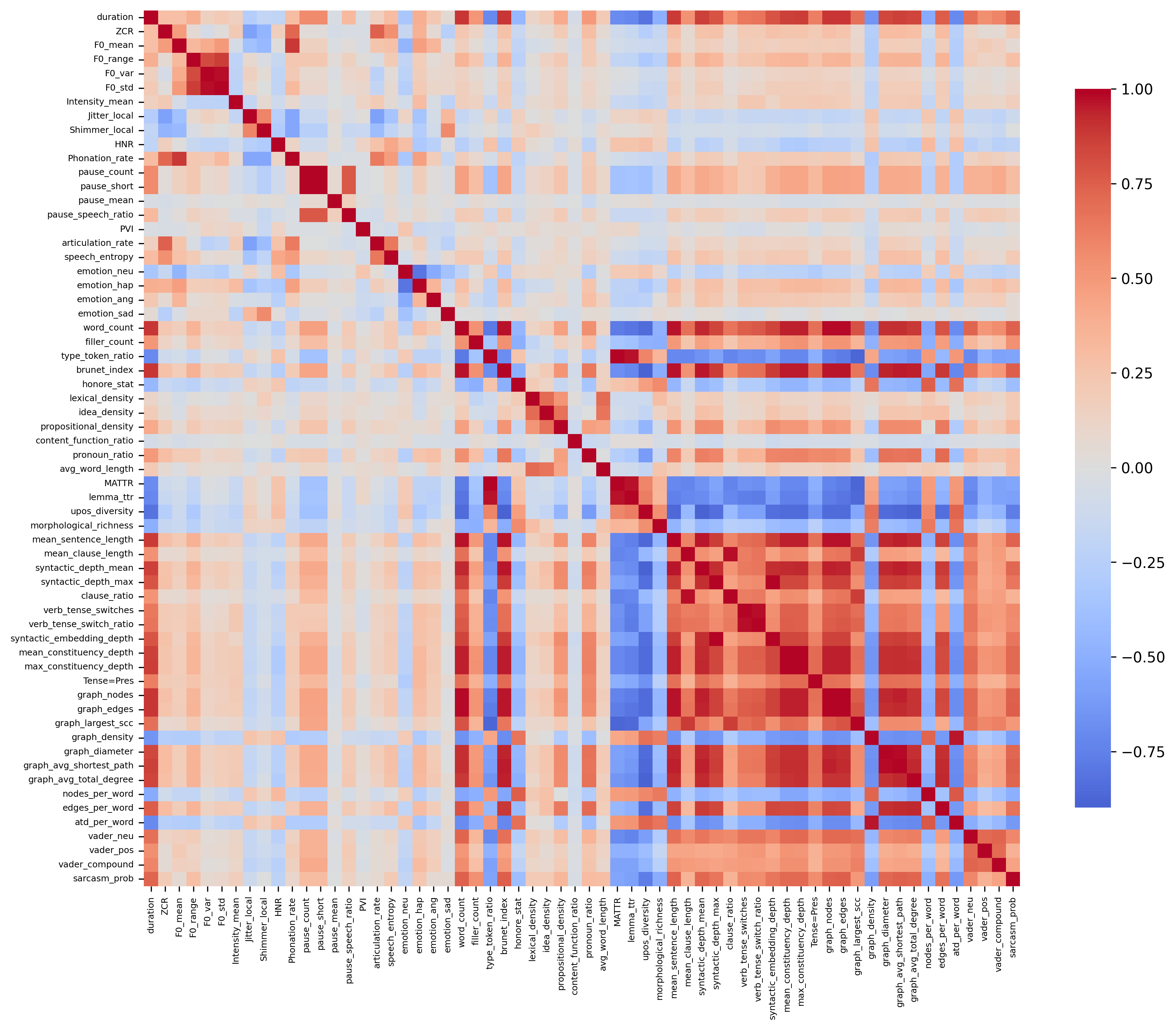}

    \caption{
    Correlation matrices of the extracted acoustic and linguistic features across datasets.
    From top to bottom, panels correspond to \textsc{Real}, \textsc{StressID}, \textsc{EATD}, \textsc{Androids}, and \textsc{DAIC-WOZ}, respectively.
    Each figure shows pairwise linear correlations among features.
    }
    
    \label{fig:correlation_all}
\end{figure}

Figure~\ref{fig:correlation_all} presents correlation matrices of the extracted acoustic and linguistic features across all evaluated datasets. The matrices illustrate pairwise linear relationships among features, revealing structured intra-group correlations and dataset-specific differences in feature dependence. Variations in correlation strength across datasets reflect differences in recording conditions, task design, and population characteristics, and highlight potential feature redundancy addressed by the modeling approach.

\section{Additional Results and Cross-Dataset Analysis}\label{sec:additional}

This section presents additional results obtained using the proposed analytical framework across multiple clinical and real-world speech datasets, including \textsc{StressID}, \textsc{DAIC-WOZ}, \textsc{Androids}, \textsc{EATD}, and the \textsc{Real} dataset. We report both interpretable model explanations and statistical group comparisons to provide complementary perspectives on feature relevance. Figures illustrate feature importance rankings derived from XGBoost built-in importance, LIME, and SHAP, while accompanying tables summarize statistically significant differences between clinical and non-clinical groups based on two-sample $t$-tests. Across datasets and mental health conditions, certain patterns can be observed, suggesting a potential role for prosodic variability, emotional expression, lexical richness, and graph-based linguistic structure as indicators of stress, depression, anxiety, and attentional symptoms.

\begin{table}[!h]
\centering
\tiny
\setlength{\tabcolsep}{2.5pt}
\renewcommand{\arraystretch}{1.1}
\begin{tabular}{lccc}
\toprule
\textbf{Feature} & \textbf{Non-Depressed} & \textbf{Depressed} & \textbf{$p$-value} \\
\midrule
sarcasm\_prob         & 0.465 & 0.431 & 2.44$\times$10$^{-3}$ \\
pause\_short          & 0.357 & 0.768 & 4.13$\times$10$^{-3}$ \\
pause\_speech\_ratio  & 0.002 & 0.006 & 6.67$\times$10$^{-3}$ \\
avg\_word\_length     & 3.540 & 3.490 & 5.28$\times$10$^{-2}$ \\
\bottomrule
\end{tabular}
\label{tab:depression_features}
\caption{
Mean feature values for Non-Depressed and Depressed groups (\textsc{DAIC-WOZ}), 
with corresponding $p$-values from two-sample $t$-tests ($p<0.05$). 
Features were standardized (z-scored) before analysis.
}
\end{table}

\begin{figure*}[!h]
    \centering
    \begin{subfigure}[b]{0.32\linewidth}
        \includegraphics[width=\linewidth]{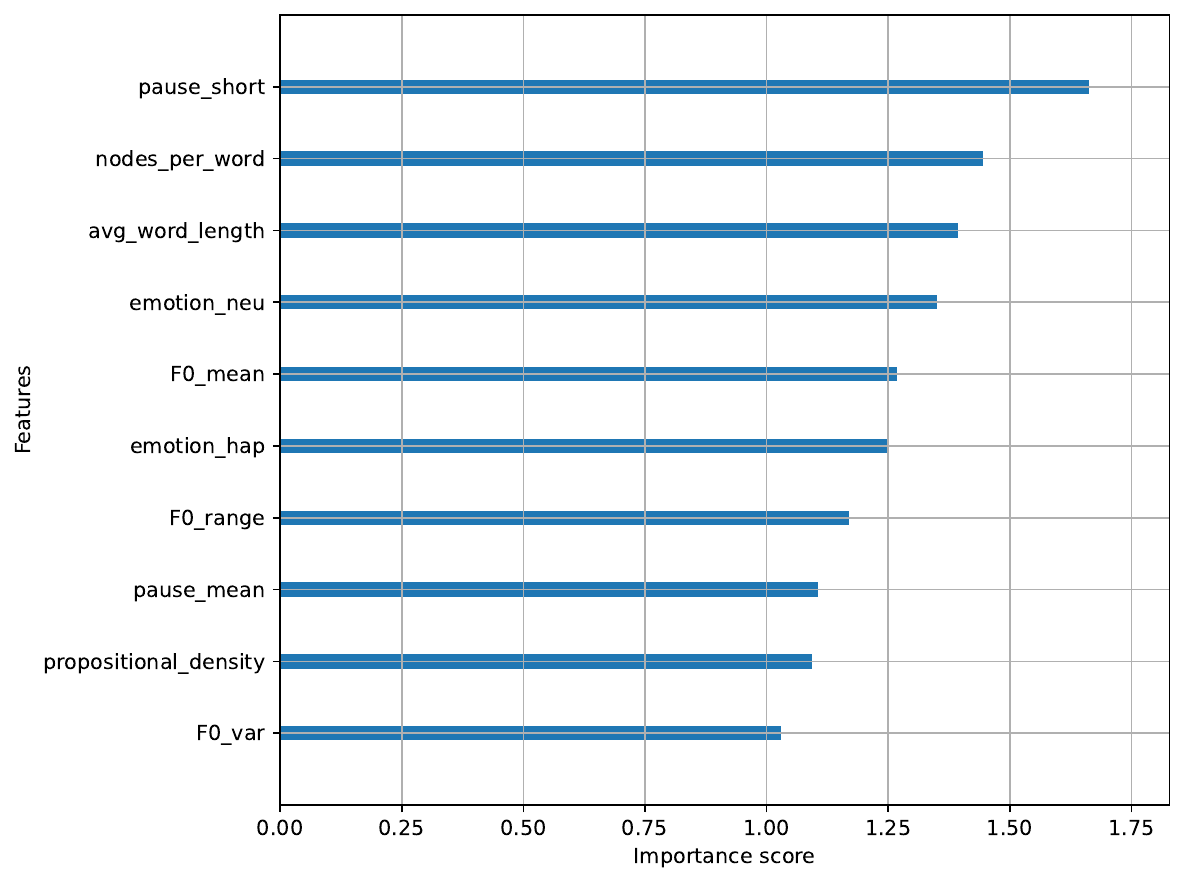}
        \caption{\textsc{DAIC-WOZ} -- XGBoost}
        \label{fig:daic_xgb}
    \end{subfigure}
    \hfill
    \begin{subfigure}[b]{0.32\linewidth}
        \includegraphics[width=\linewidth]{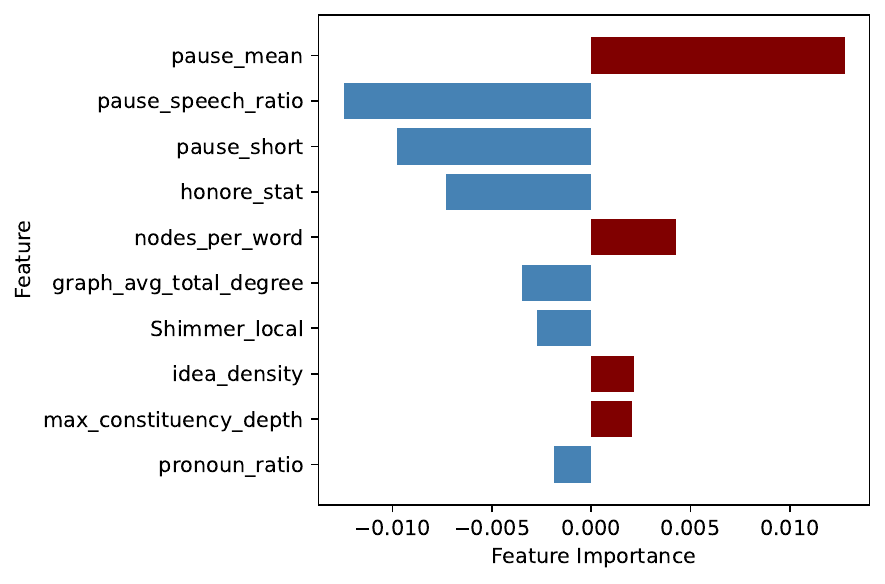}
        \caption{\textsc{DAIC-WOZ} -- LIME}
        \label{fig:daic_lime}
    \end{subfigure}
    \hfill
    \begin{subfigure}[b]{0.32\linewidth}
        \includegraphics[width=\linewidth]{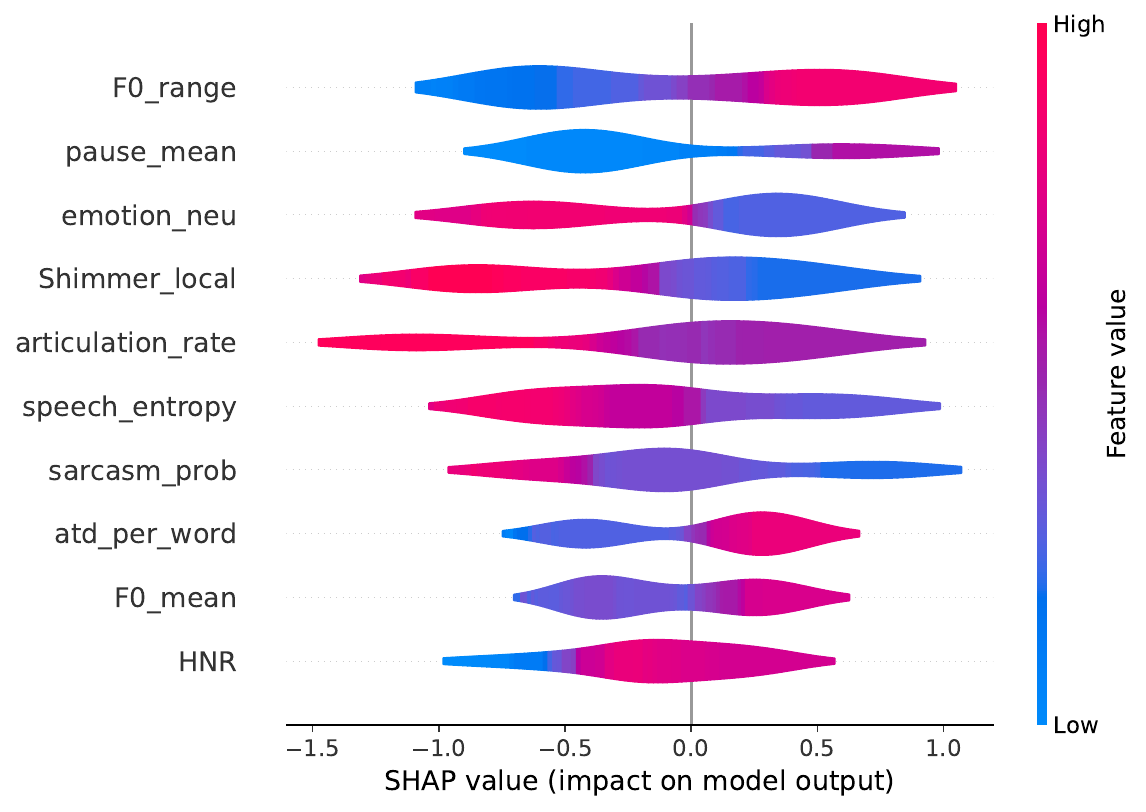}
        \caption{\textsc{DAIC-WOZ} -- SHAP}
        \label{fig:daic_shap}
    \end{subfigure}
    \caption{
        Top predictive features for the \textsc{DAIC-WOZ} dataset derived from acoustic and linguistic descriptors. 
    }
    \label{fig:daic_feature_comparison}
\end{figure*}

\begin{table}[!h]
\centering
\tiny
\setlength{\tabcolsep}{2.5pt}
\renewcommand{\arraystretch}{1.1}
\begin{tabular}{lccc}
\toprule
\textbf{Feature} & \textbf{Non-Depressed} & \textbf{Depressed} & \textbf{$p$-value} \\
\midrule
emotion\_sad & 0.099 & 0.182 & 2.83$\times$10$^{-5}$ \\
Intensity\_mean & 58.360 & 54.726 & 2.73$\times$10$^{-5}$ \\
sentiment\_negative & 0.519 & 0.693 & 6.49$\times$10$^{-5}$ \\
sentiment\_positive & 0.469 & 0.300 & 9.91$\times$10$^{-5}$ \\
second\_order\_coherence & 0.213 & 0.313 & 1.08$\times$10$^{-3}$ \\
Jitter\_local & 0.021 & 0.024 & 1.26$\times$10$^{-2}$ \\
emotion\_neu & 0.199 & 0.156 & 1.96$\times$10$^{-2}$ \\
Shimmer\_local & 0.110 & 0.122 & 1.84$\times$10$^{-2}$ \\
brunet\_index & 8.852 & 9.740 & 1.70$\times$10$^{-2}$ \\
duration & 54.971 & 72.298 & 3.56$\times$10$^{-2}$ \\
Phonation\_rate & 108.486 & 94.456 & 3.73$\times$10$^{-2}$ \\
pause\_count & 1.074 & 3.594 & 4.45$\times$10$^{-2}$ \\
pause\_short & 1.074 & 3.594 & 4.45$\times$10$^{-2}$ \\
pause\_mean & 0.0013 & 0.0038 & 3.63$\times$10$^{-2}$ \\
pause\_speech\_ratio & 0.00035 & 0.00119 & 4.03$\times$10$^{-2}$ \\
type\_token\_ratio & 0.698 & 0.646 & 2.68$\times$10$^{-2}$ \\
lemma\_ttr & 0.628 & 0.575 & 4.64$\times$10$^{-2}$ \\
sentence\_count & 5.212 & 6.536 & 3.19$\times$10$^{-2}$ \\
first\_order\_coherence & 0.309 & 0.375 & 3.94$\times$10$^{-2}$ \\
upos\_diversity & 0.276 & 0.210 & 2.44$\times$10$^{-2}$ \\
emotion\_hap & 0.662 & 0.611 & 4.94$\times$10$^{-2}$ \\
MATTR & 0.796 & 0.775 & 5.32$\times$10$^{-2}$ \\
\bottomrule
\end{tabular}
\label{tab:depression_features_2}
\caption{
Mean acoustic and linguistic feature values for Non-Depressed and Depressed participants in the \textsc{Androids} corpus, with corresponding $p$-values from two-sample $t$-tests ($p<0.05$ shown).}
\end{table}

\begin{figure*}[!h]
    \centering
    \begin{subfigure}[b]{0.32\linewidth}
        \includegraphics[width=\linewidth]{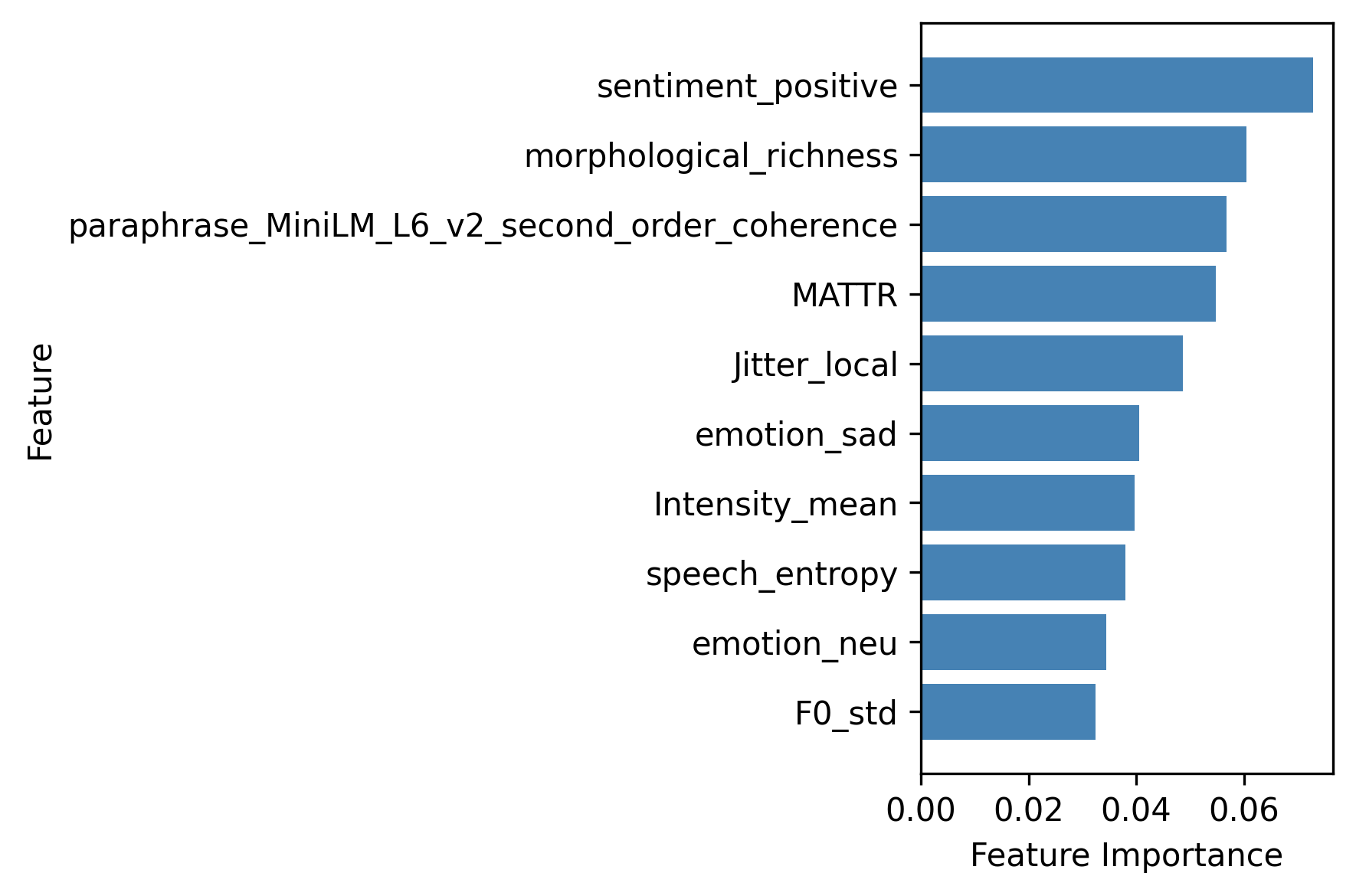}
        \caption{Androids -- XGBoost}
        \label{fig:androids_xgb}
    \end{subfigure}
    \hfill
    \begin{subfigure}[b]{0.32\linewidth}
        \includegraphics[width=\linewidth]{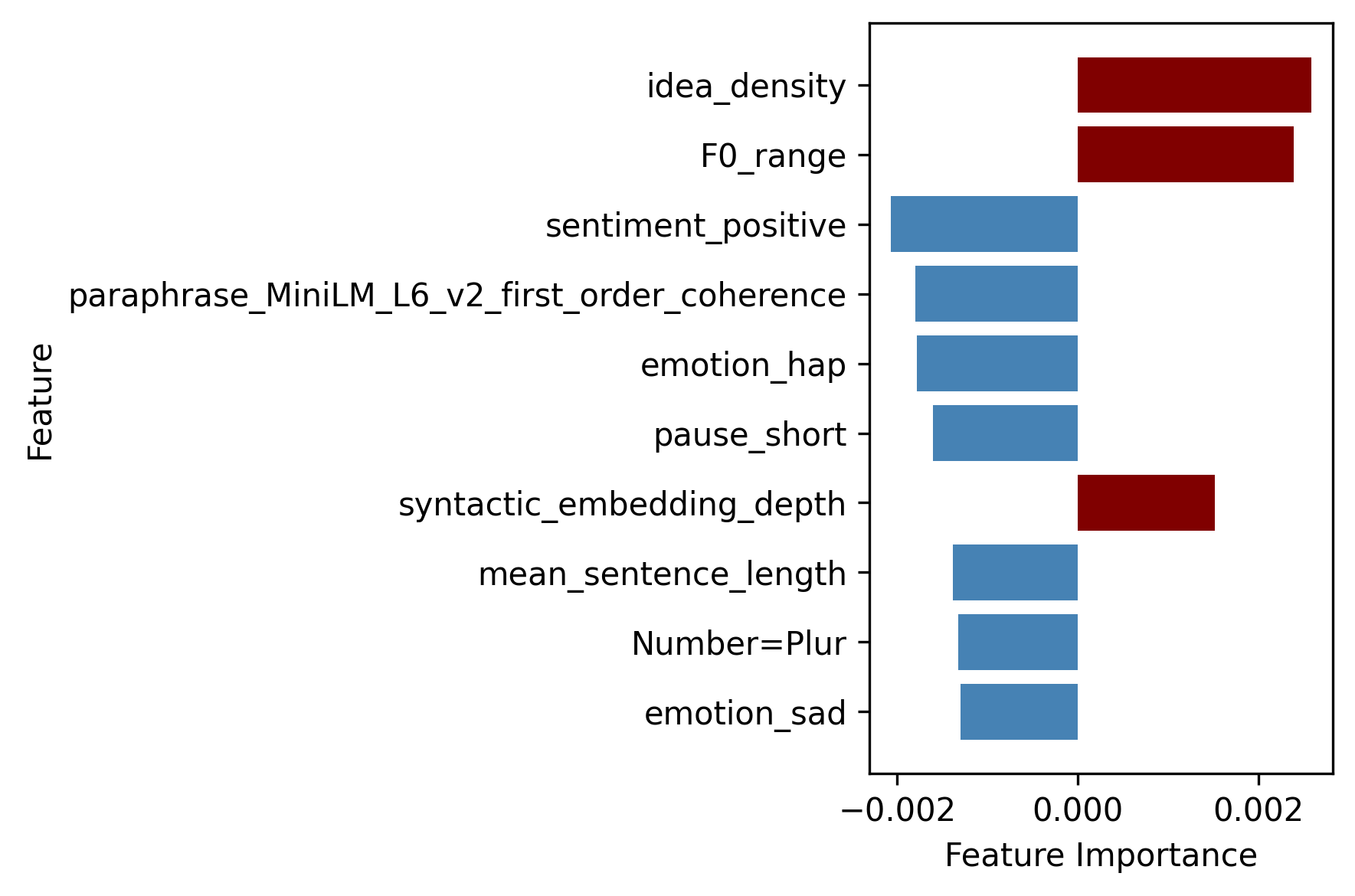}
        \caption{Androids -- LIME}
        \label{fig:androids_lime}
    \end{subfigure}
    \hfill
    \begin{subfigure}[b]{0.32\linewidth}
        \includegraphics[width=\linewidth]{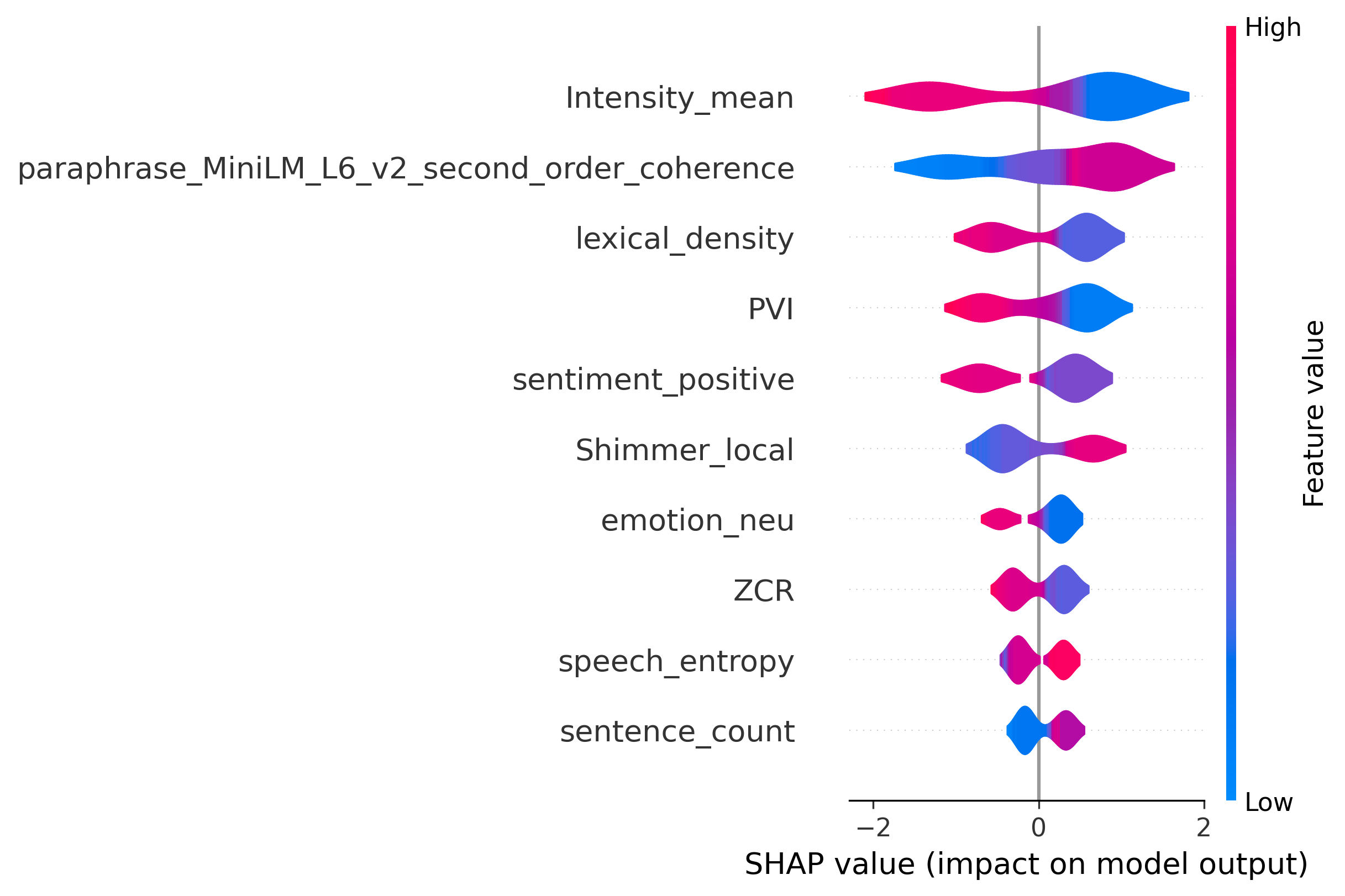}
        \caption{Androids -- SHAP}
        \label{fig:androids_shap}
    \end{subfigure}
    \caption{
        Top predictive features for the \textsc{Androids Corpus} dataset derived from acoustic and linguistic descriptors.
    }
    \label{fig:androids_feature_comparison}
\end{figure*}

\begin{table}[!h]
\centering
\tiny
\setlength{\tabcolsep}{3pt}
\renewcommand{\arraystretch}{1.1}
\begin{tabular}{lccc}
\toprule
\textbf{Feature} & \textbf{Non-Depressed} & \textbf{Depressed} & \textbf{$p$-value} \\
\midrule
emotion\_neu & 0.112 & -0.493 & 8.58$\times$10$^{-4}$ \\
emotion\_sad & -0.128 & 0.562 & 1.49$\times$10$^{-2}$ \\
F0\_mean & -0.102 & 0.448 & 1.86$\times$10$^{-2}$ \\
passive\_voice\_ratio & 0.046 & -0.201 & 2.82$\times$10$^{-2}$ \\
pause\_count & 0.052 & -0.230 & 4.78$\times$10$^{-2}$ \\
\bottomrule
\end{tabular}
\label{tab:eatd_depression_features}
\caption{
Mean acoustic and linguistic feature values for Non-Depressed and Depressed participants in the \textsc{EATD} corpus, with corresponding $p$-values from two-sample $t$-tests ($p<0.05$ shown).}
\end{table}

\begin{figure*}[!h]
    \centering
    \begin{subfigure}[b]{0.32\linewidth}
        \includegraphics[width=\linewidth]{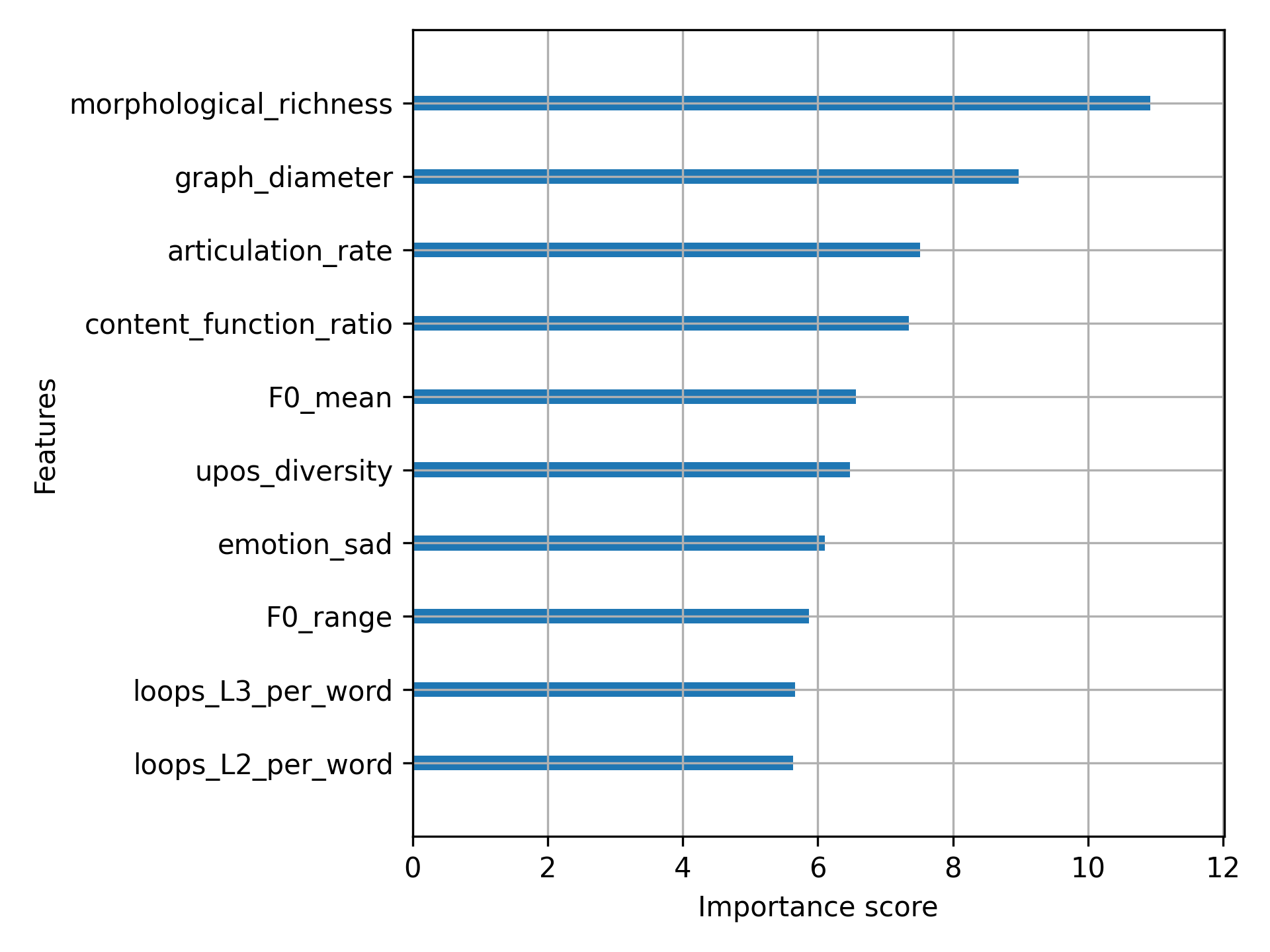}
        \caption{\textsc{EATD} -- XGBoost}
        \label{fig:eatd_xgb}
    \end{subfigure}
    \hfill
    \begin{subfigure}[b]{0.32\linewidth}
        \includegraphics[width=\linewidth]{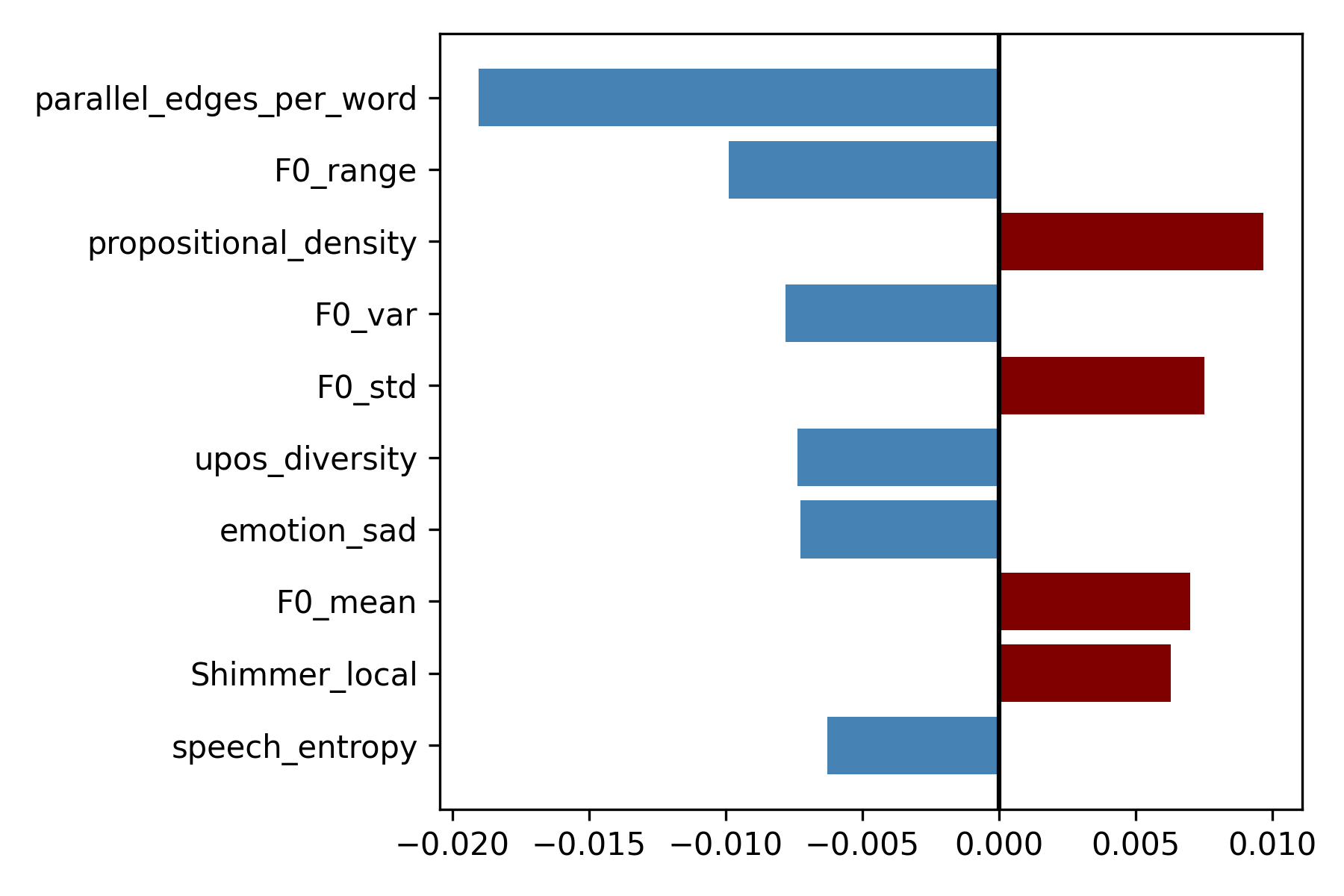}
        \caption{\textsc{EATD} -- LIME}
        \label{fig:eatd_lime}
    \end{subfigure}
    \hfill
    \begin{subfigure}[b]{0.32\linewidth}
        \includegraphics[width=\linewidth]{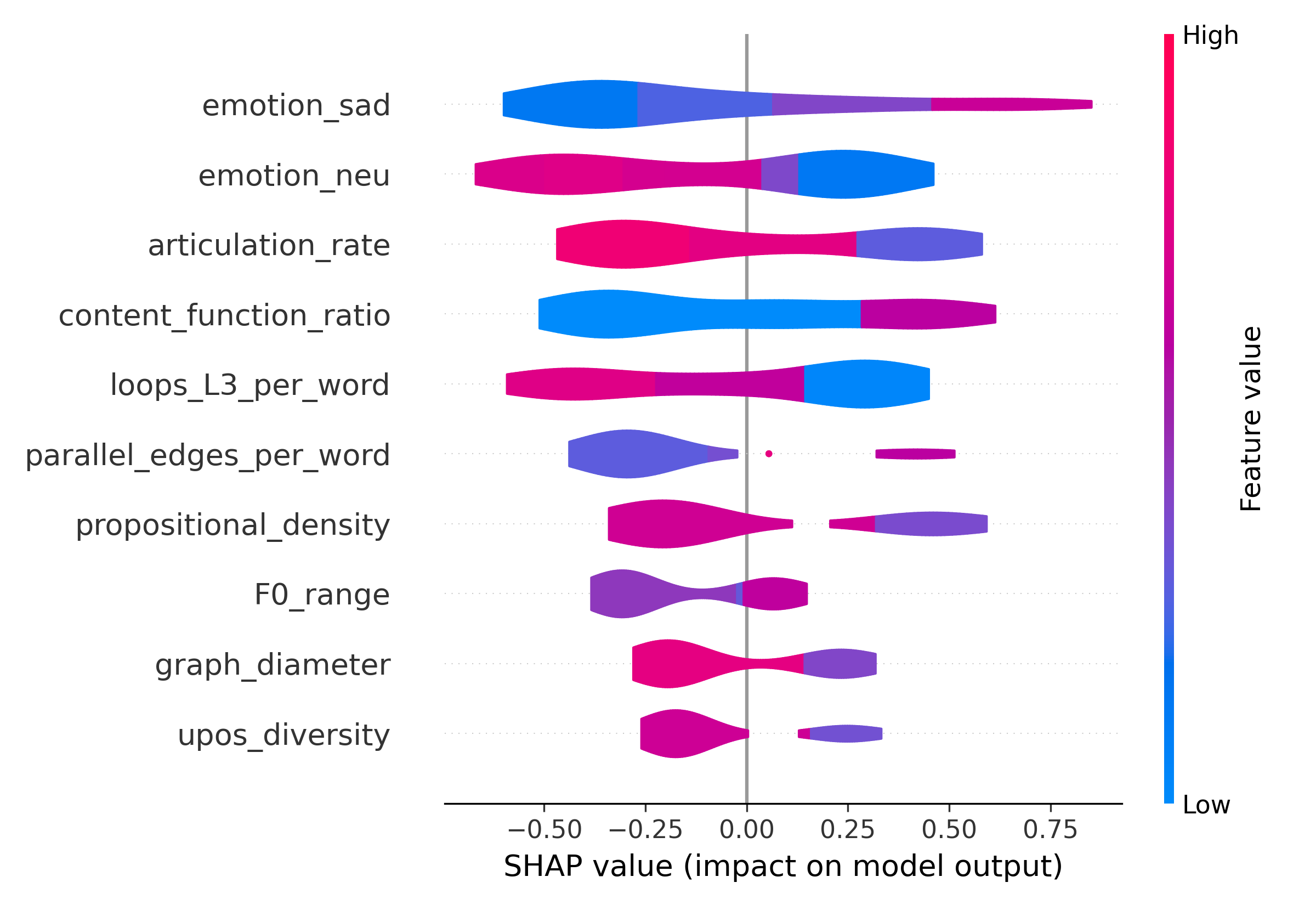}
        \caption{\textsc{EATD} -- SHAP}
        \label{fig:eatd_shap}
    \end{subfigure}
    \caption{
        Top predictive features for the \textsc{EATD} dataset derived from acoustic and linguistic descriptors. 
    }
    \label{fig:eatd_feature_comparison}
\end{figure*}

\begin{table}[!h]

\centering
\tiny
\setlength{\tabcolsep}{2.5pt} 
\renewcommand{\arraystretch}{1.1} 
\begin{tabular}{lccc}
\toprule
\textbf{Feature} & \textbf{Non-Depressed} & \textbf{Depressed} & \textbf{$p$-value} \\
\midrule
vader\_negative          & 0.030 & 0.050 & 1.22$\times$10$^{-4}$ \\
vader\_compound          & 0.652 & 0.497 & 4.98$\times$10$^{-3}$ \\
content\_function\_ratio & 1.333 & 1.273 & 9.93$\times$10$^{-3}$ \\
loops\_L1\_per\_word     & 0.003 & 0.001 & 1.26$\times$10$^{-2}$ \\
pause\_medium            & 0.295 & 0.102 & 1.91$\times$10$^{-2}$ \\
graph\_loops\_L1         & 0.302 & 0.148 & 1.77$\times$10$^{-2}$ \\
MATTR                    & 0.693 & 0.704 & 4.78$\times$10$^{-2}$ \\
emotion\_neu             & 0.198 & 0.169 & 4.26$\times$10$^{-2}$ \\
\bottomrule
\end{tabular}
\label{tab:phq9_medians_pvalues_2}
\caption{
Mean feature values for Non-Depression and Depression groups (PHQ-9) from the \textsc{Real} dataset, 
with corresponding $p$-values from two-sample $t$-tests ($p<0.05$).
}
\end{table}

\begin{figure*}[!h]
    \centering
    \begin{subfigure}[b]{0.32\linewidth}
        \includegraphics[width=\linewidth]{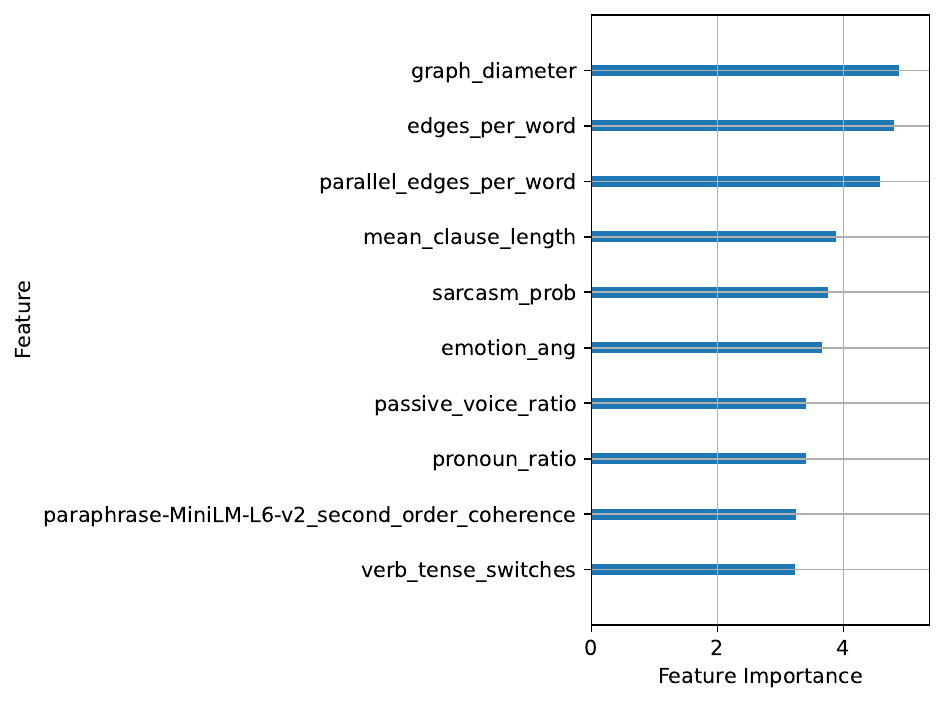}
        \caption{ASRS -- XGBoost}
        \label{fig:asrs_xgb}
    \end{subfigure}
    \hfill
    \begin{subfigure}[b]{0.32\linewidth}
        \includegraphics[width=\linewidth]{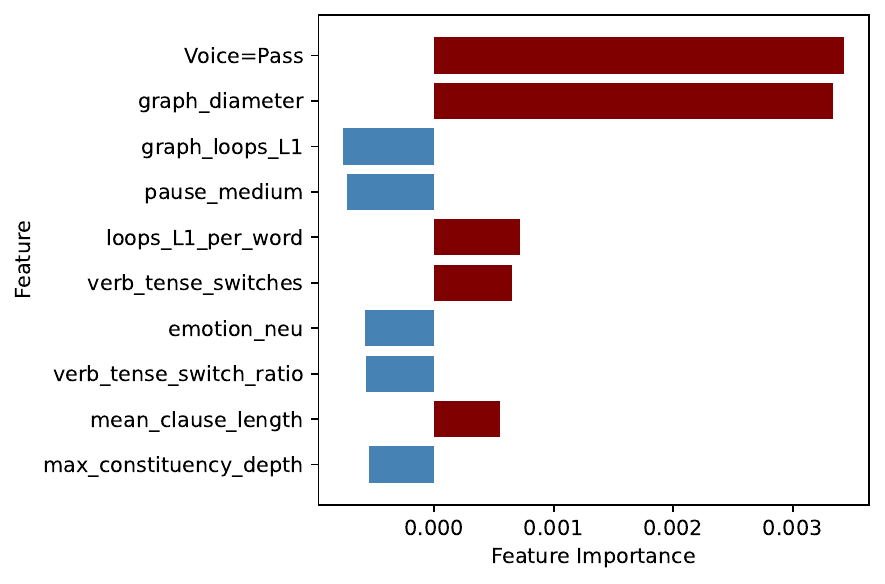}
        \caption{ASRS -- LIME}
        \label{fig:asrs_lime}
    \end{subfigure}
    \hfill
    \begin{subfigure}[b]{0.32\linewidth}
        \includegraphics[width=\linewidth]{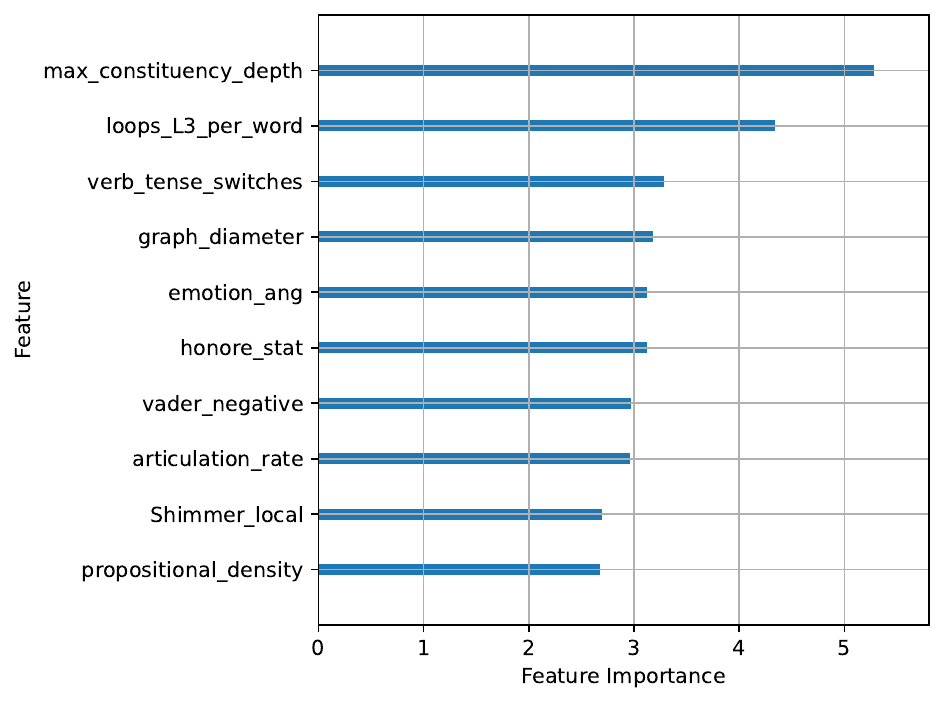}
        \caption{GAD-7 -- XGBoost}
        \label{fig:gad7_xgb}
    \end{subfigure}

    \vspace{0.5em}

    \begin{subfigure}[b]{0.32\linewidth}
        \includegraphics[width=\linewidth]{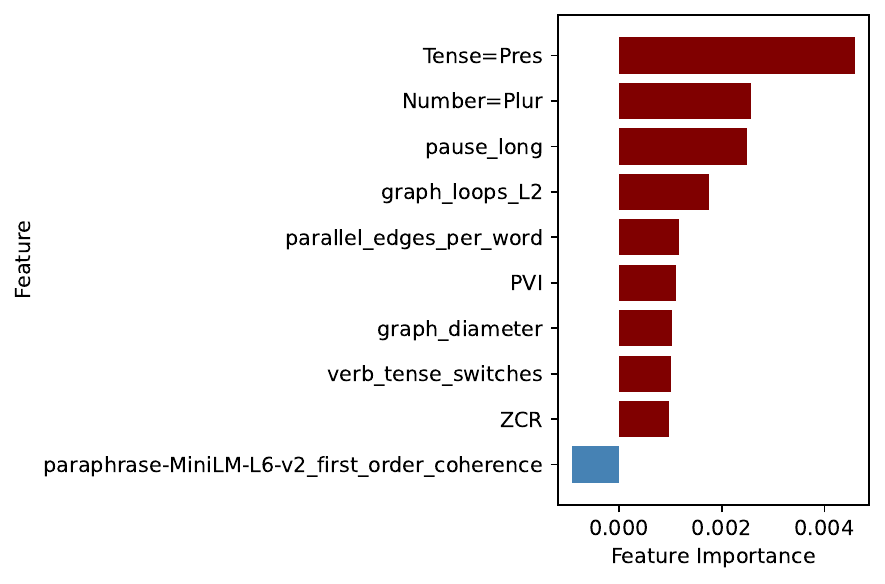}
        \caption{GAD-7 -- LIME}
        \label{fig:gad7_lime}
    \end{subfigure}
    \hfill
    \begin{subfigure}[b]{0.32\linewidth}
        \includegraphics[width=\linewidth]{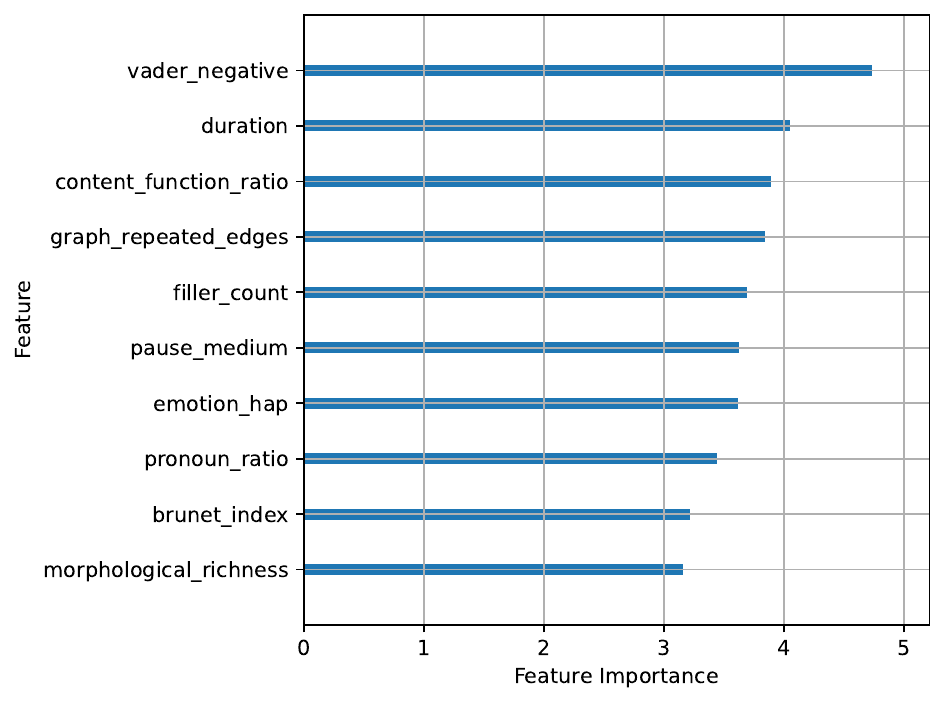}
        \caption{PHQ-9 -- XGBoost}
        \label{fig:phq9_xgb}
    \end{subfigure}
    \hfill
    \begin{subfigure}[b]{0.32\linewidth}
        \includegraphics[width=\linewidth]{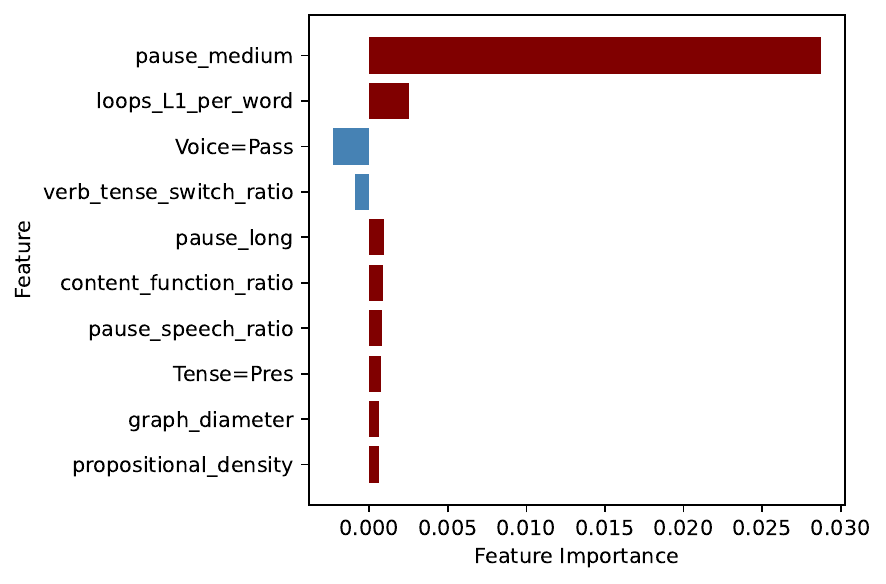}
        \caption{PHQ-9 -- LIME}
        \label{fig:phq9_lime}
    \end{subfigure}

    \caption{
        Top predictive features from the \textsc{Real} dataset for ASRS, GAD-7, and PHQ-9 classification tasks
        using XGBoost built-in importance and LIME.
    }
    \label{fig:feature_comparison_second}
\end{figure*}

\end{document}